\pdfoutput=1

\documentclass[11pt]{article}
\usepackage[table,dvipsnames]{xcolor}

\usepackage{acl}

\usepackage{times}
\usepackage{latexsym}

\usepackage[T1]{fontenc}

\usepackage[utf8]{inputenc}

\usepackage{microtype}

\usepackage{inconsolata}

\usepackage{graphicx}

\usepackage{multicol}
\usepackage{multirow}
\usepackage{adjustbox}
\usepackage{makecell}
\usepackage{tabularx}

\usepackage{graphicx}  
\usepackage{caption}   
\usepackage{subcaption} 
\usepackage{geometry}
\usepackage{longtable}
\usepackage{supertabular}
\usepackage{cuted}

\usepackage{contour}
\usepackage[normalem]{ulem}

\contourlength{0.8pt}

\usepackage{amssymb}
\usepackage{pifont}
\newcommand{\xmark}{\ding{55}}%
\usepackage{listings}
\usepackage{fancyhdr}
\definecolor{commentsColor}{rgb}{0.497464, 0.75, 0.497464}
\usepackage{dblfloatfix}

\title{Open-Source Large Language Models as Multilingual Crowdworkers: Synthesizing Open-Domain Dialogues in Several Languages With No Examples in Targets and No Machine Translation}  

\author{Ahmed Njifenjou, Virgile Sucal, Bassam Jabaian, Fabrice Lefèvre  \\
        Laboratoire Inforamitque d'Avignon (LIA), CERI - Avignon Université \\
        \texttt{\{ahmed-ndouop.njifenjou \& firstname.lastname\}@univ-avignon.fr} \\}

\begin{document}
\maketitle
\begin{abstract}
The prevailing paradigm in the domain of Open-Domain Dialogue agents predominantly focuses on the English language, encompassing both models and datasets. Furthermore, the financial and temporal investments required for crowdsourcing such datasets for finetuning are substantial, particularly when multiple languages are involved. Fortunately, advancements in Large Language Models (LLMs) have unveiled a plethora of possibilities across diverse tasks. Specifically, instruction-tuning has enabled LLMs to execute tasks based on natural language instructions, occasionally surpassing the performance of human crowdworkers. Additionally, these models possess the capability to function in various languages within a single thread. Consequently, to generate new samples in different languages, we propose leveraging these capabilities to replicate the data collection process. We introduce a pipeline for generating Open-Domain Dialogue data in multiple Target Languages using LLMs, with demonstrations provided in a unique Source Language. By eschewing explicit Machine Translation in this approach, we enhance the adherence to language-specific nuances. 
We apply this methodology to the PersonaChat dataset. To enhance the openness of generated dialogues and mimic real life scenarii, we added the notion of speech events 
corresponding to the type of conversation the speakers are involved in and also that of common ground which represents the premises of a conversation. 
\end{abstract}

\section{Introduction}
In the realm of Natural Language Processing (NLP), Large Language Models (LLMs) have surged in prominence, unleashing a myriad of possibilities.
Although certain models claim to be optimized for conversation, they tend to lean towards asymmetric exchanges, responding to user's input in a Q\&A format rather than fostering a truly balanced dialogue. Having an actual Open-Domain Dialogue (ODD) with a user implies showcasing some human-like dialogue abilities such as empathy, personality, engagingness, etc. 
Most of the \textit{status quo} approaches to augment LLM capabilities towards such skills rely on fine-tuning on skill-specific datasets. Unfortunately, there is a dearth of such datasets in languages other than English or, more recently Chinese, 
and data collection is expensive in terms of cost and time. To tackle this issue, different approaches proposed to use Machine Translation (MT) -- whether of the Source Language ($l_S$) dataset before fine-tuning or during inference with a $l_S$ fine-tuned model ~\citep{lin-etal-2021-xpersona} 
-- at the expense of data and resulting models' quality. 
Additionally, as highlighted by~\citet{dogruoz-skantze-2021-open}, ODD, as used by literature, is often restricted to sole "small talk" type of speech event (SE) 
where speakers are commonly asked/tasked to \textit{"just chat about anything"} while real life ODD can be of various types (from serious chat to gossip) depending on the context and involve speakers that share a common ground (CG) as a premise to their chat -- hence restricting the "openness" of each ODD which is referred to as the \textit{open domain paradox} (ODP) by~\citet{skantze-dogruoz-2023-open}.


Crowdsourced ODD datasets are collected with fine-grained human-designed guidelines for the crowdworkers. Also in the bargain, some works like~\cite{Gilardi_2023} for closed-source LLM\footnote{In this case \texttt{ChatGPT}~\cite{ChatGPT}} and~\cite{alizadeh2023opensource} for open-source\footnote{Here \texttt{FLAN}~\cite{weifinetuned} and \href{https://huggingface.co/chat/}{\texttt{HuggingChat}}} demonstrate that instruction-tuned LLMs outperform crowdworkers for several tasks while others like~\cite{veselovsky2023artificial} estimate from an experiment that 33-46\% of crowdworkers actually use LLMs to complete their tasks. Hence, a question comes to mind: \textit{why not directly use the LLMs to generate new samples?}

Thereby, we propose here to leverage multilingual instruction-following LLMs abilities to generate datasets in Target Languages ($l_T$), other than English, and also attempt simultaneously to alleviate the \textit{ODP}.
Instead of MT, to get new samples in $l_T$,  
we propose to design prompts based on few examples from the $l_S$ source dataset and their sourcing guidelines. This process exploits the fact that LLMs can \textit{understand} 
(decode) well-crafted instructions and \textit{think} (infer) in different languages simultaneously. We further enhance the aforementioned guidelines with supplementary instructions targeting the \textit{ODP}: common ground generation and inclusion, SE type specification. 
Therefore, our contributions are as follows:
\begin{itemize}
    \item A pipeline to generate ODD datasets in multiple $l_T$, with neither MT nor $l_T$ examples, that enforces $l_T$ specificities\footnote{For instance the syntax itself, named entities like proper names and locations, cultural habits, etc. that a MT module may not natively incorporate. See examples in Appendix~\ref{appendix:examples-from-MOUD}.}. 

    \item An application using PersonaChat~\cite{zhang-etal-2018-personalizing} as source dataset and different LLMs as generators, with the release\footnote{The dataset and the code will be made publicly available following the publication of this work.} of the \textbf{M}ultilingual \textbf{O}pen-domain 
    \textbf{U}nnatural \textbf{D}ialogue \textbf{D}ataset (\textbf{MOUD}\footnote{Pronounced \texttt{/mOOd/} as the word \textit{"mood"} 
    .}), a dataset of persona-based ODD, with common ground and various SEs in 
    English ($l_S$) and 28 other target languages ($l_T$).
 
    
    \item A qualitative evaluation of the generated data combining automatic metrics, syntax analysis, LLM-as-a-judge on selected criteria, and human evaluation for certain languages based on the availability of voluntary evaluators.
    
    \item Baseline application with automatic metrics evaluations of shallow finetuned models across some $l_T$. 
    
\end{itemize}

\section{Related work}
\paragraph{Open-Domain Dialogue Datasets} 
Numerous skill-specific datasets have been gathered to develop human-like conversational abilities in ODD agents. For instance, datasets address personality~\cite{zhang-etal-2018-personalizing, mazare-etal-2018-training, gao-etal-2023-livechat}, empathy~\cite{rashkin-etal-2019-towards, sharma-etal-2020-computational}, emotion~\cite{zhou-wang-2018-mojitalk, liu-etal-2021-towards}, knowledge~\cite{dinan2018wizard, komeili-etal-2022-internet}, and long-term memory~\cite{xu-etal-2022-beyond}. Some datasets, like those by~\cite{smith-etal-2020-put, li-etal-2017-dailydialog, zhong-etal-2020-towards}, combine multiple skills. However, most datasets are in English (and more recently in Chinese), and replicating this process in other languages is costly. 

Efforts to address this disproportionate representation of languages often hinge on MT
, with some like~\cite{lin-etal-2021-xpersona} incorporating additional human post-processing, albeit at a non-negligible cost. It is contingent on the availability and quality of MT systems and often the resulting dialogues are in \textit{translationese} 
and do not reflect either $l_T$ 
specifities or \textit{folk psychology} 
but rather carry watermarks from $l_S$ ~\cite{koppel-ordan-2011-translationese, artetxe-etal-2020-translation} and artifacts ~\cite{park-etal-2024-translation, sizov-etal-2024-analysing}.
While some address the lack of common ground, such efforts are typically confined to knowledge grounding or \textit{non-common} ground scenarios, where only one speaker is informed. Additionally, as highlighted by~\citet{dogruoz-skantze-2021-open}, there is insufficient diversity in SEs types.

\paragraph{Data Generation with LLMs}
Has been experimented in a wide range of domains involving NLP. 
For NLI~\cite{liu-etal-2022-wanli} proposed a Worker-AI collaboration: GPT3~\cite{NEURIPS2020_1457c0d6} generates challenging NLI examples then crowdworkers revise and annotate them;~\cite{schick-schutze-2021-generating} 
used GPT2-XL~\cite{radford2019language} to generate a dataset of automatically labeled text pairs without prior labeled data.
Some proposed approaches are applied to different tasks: in ~\cite{lee2021neural}, task-specific data are sliced into subsets of "same interest" and an extrapolator is learned on data-rich slices and then used to generate new examples in poor ones. Still, they rely on either human intervention or example availability.
\begin{figure*}[!t]
	\centering
 \resizebox{\textwidth}{!}{
        \includegraphics[scale=1]{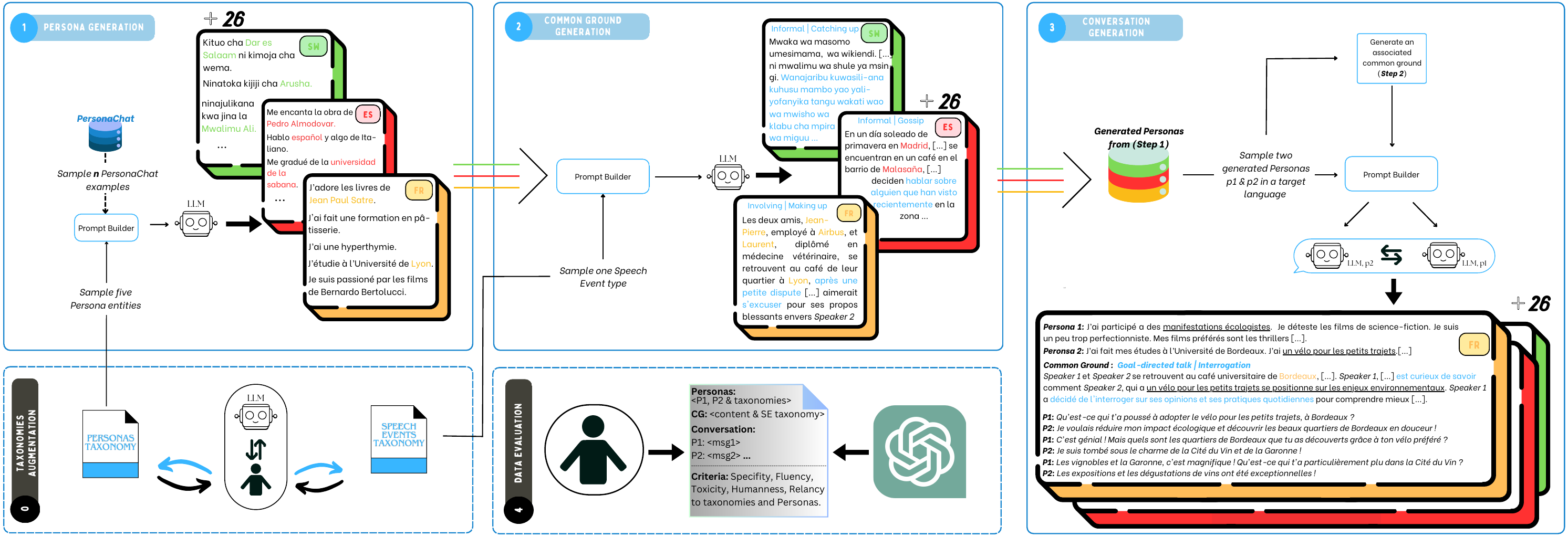}
    }
 	\caption[Architecture]{MOUD Generation Pipeline: (0) Taxonomies are manually expanded by interacting with a LLM. (1) Non-translated $l_S$ examples are introduced into the prompt to generate new $l_T$ samples. (2) Common ground is created based on two generated personas and a sampled speech event. (3) The outputs from steps (1) and (2) are integrated into prompts for interactions between two LLM instances.  Nucleus sampling is used at every step for diversity.  Examples in this figure highlight the display of language's specific elements for \textcolor{orange}{French}, \textcolor{red}{Spanish} and \textcolor{green}{Swahili}. For more detailed examples see Table~\ref{tab:example-english-1}, Table~\ref{tab:example-english-2}, Table~\ref{tab:example-french} in Appendix~\ref{appendix:examples-from-MOUD}. (4) Generated data from steps (1), (2) and (3) are evaluated by human and LLM as a judge on selected criteria as explained in Section~\ref{sec:data-qualitative-evaluation}. 
  }
	\label{fig:CompletePipeline}
\end{figure*} 
In the other hand, as instruction-tuning proved to enhance multitasks generalization, recent works focused on instructions generation: in Unnatural-Instructions~\cite{honovich-etal-2023-unnatural} used InstructGPT3.5 
~\cite{NEURIPS2022_b1efde53} to generate 
up to 240k samples starting with three seeds from Super-NaturalInstructions~\cite{wang-etal-2022-super}. Meanwhile, in Self-Instruct~\cite{wang-etal-2023-self-instruct} they hand-wrote 175 seed instructions from which they generated 52k instructions and 82k corresponding input-output instances with GPT3.
Both showed that despite containing some noise, generated data are more \textbf{diverse} and models trained on them perform on par or surpass strong baselines. However, evidence for multilingualism and ODD task are lacking, and these rely on close-sourced LLMs. 

While~\citet{agrawal2023qameleon} addresses the issue of multilingualism,~\citet{lee-etal-2022-personachatgen} proposes the creation of a persona-based ODD dataset. The former focuses on the Q\&A task but still depends on a few examples in $l_T$ or MT when 
unavailable. The latter, closely related to our work, presents a persona taxonomy and a pipeline for generating personalized dialogues using GPT3. However, their primary aim is to expand the existing English PersonaChat dataset\footnote{Dating back to \textit{2018}, it may no longer reflect updated personal information, e.g. those related to the pandemic.}. Our approach not only updates this information and the persona taxonomy but also generates data in other languages, without relying on MT or $l_T$ samples, to capture their unique characteristics. Additionally, we incorporate diverse types of SEs and CG in the conversations.

\section{Methodology} 
\label{sec:methodology}
Let $\mathcal{D}_{s,l}$ symbolize a skill ($s$) specific ODD dataset in a given language ($l$). We build on the availability of such datasets in a $l_S$, here in English. So the latter is hereafter denoted as $\mathcal{D}_{s, en}$. 
While the described methodology focuses on ODD and is later applied to PersonaChat, it can be easily adapted to other generation tasks with data collected from crowdworkers.

\subsection{From crowdsourcing guidelines to prompt instructions}
\label{subsec:guidelines-to-instructions}
We focus on $\mathcal{D}_{s, en}$, a dataset of human-human conversations created by workers based on a fine-grained set of human-designed guidelines, $\mathcal{G} = \{g_t\}_{t=1}^{N_g}$, where $N_g$ represents the number of guidelines. Our goal is to prompt an instruction-following LLM with these guidelines to generate similar datasets in a set $L$ of several $l_T$, 
while addressing previously mentioned limitations.

However, $\mathcal{G}$ often contains multiple steps and complex statements that may be easy to interpret for humans but hard for a LLM~\cite{mishra-etal-2022-reframing}. Here instead of using their proposed reframing techniques separately, we propose to combine some of them to break the guidelines into LLM-understandable instructions:

\paragraph{Decomposition Reframing} $\mathcal{G}$ can be rewritten as:   

\begin{equation}
\label{eq:guidelance-decompostion}
    \mathcal{G} = \bigcup_{k=1}^{N_{step}}\mathcal{G}_{k} =  \bigcup_{k=1}^{N_{step}}\{g_{t,k}\}_{t=1}^{N_{g,k}}
\end{equation}
where each subset $\mathcal{G}_{k}$ 
of $\mathcal{G}$  is the set of guidelines corresponding to a step $k$ in the data collection process\footnote{E.g. for PersonaChat first step consists in personas collection, second personas reformulation and then conversation generation using the collected personas.} for which a dedicated prompt should therefore be derived.

\paragraph{Itemizing Reframing} For each $\mathcal{G}_k$, the corresponding guidelines are cast into a set of LLM-prone instructions: $\mathcal{I}_k =\{i_{t,k}\}_{t=1}^{N_{i,k}}$ . 
The latter are prepended to the prompt as a list of items to implement Chain-of-Thoughts \textit{reasoning}~\cite{NEURIPS2022_9d560961}. Note that $N_{i,k}$ is not necessarily equal to $N_{g,k}$ as complex guidelines can be exploded into several simpler instructions.

\subsection{Enforcing $l_T$ and its specificities}
\label{subsec:constraints}
To achieve our target of generating data in $l_T$ using $l_S$ samples without MT, we use another method from~\cite{mishra-etal-2022-reframing} to restrain the output:
 
\textbf{Restraining Reframing:} For each generation step $k$, 
we add a set $\mathcal{C}_{k,l_T} = \{c_{t,k,l_T}\}_{t=1}^{N_{c,k}}$ of constraints that encompasses additional directives. These are often not derived from data sourcing guidelines but rather additional statements that tackle some flaws of the original data. In this work, it includes the desired $l_T$, the writing styles, the $l_T$ specificities and folk psychology\footnote{A dialogue between two British speakers is not likely to have the same dynamic as one between two French people.
} that should be displayed, which are crucial elements that a MT module cannot provide but can be \textit{thought} (inferred) by a multilingual LLM. Along with these, directives to tackle the \textit{ODP} when applicable (last step) with constraints on SE types and CG. 
%
%
%
\begin{table*}[ht]
    \resizebox{\textwidth}{!}{
        \begin{tabular} {lcccccc}
            \hline
            \noalign{\smallskip}\textbf{Dataset}  &\textbf{Source} &  \textbf{Multilingual}& \textbf{Size} & \textbf{Extendable} & \textbf{Common Ground} & \textbf{ $\neq$ Speech Event types}  \\
                \noalign{\smallskip}\hline\noalign{\smallskip}
            PersonaChat~\cite{zhang-etal-2018-personalizing}
                & Crowd  &  \textcolor{BrickRed}{\xmark} & 17k & \textcolor{BrickRed}{\xmark} &  \textcolor{BrickRed}{\xmark} & \textcolor{BrickRed}{\xmark}  \\ 
                XPersona~\cite{lin-etal-2021-xpersona} & MT & \textcolor{ForestGreen}{\checkmark}$^{6}$ & 6  $\times$ 17k & \textcolor{BrickRed}{\xmark} & \textcolor{BrickRed}{\xmark}  & \textcolor{BrickRed}{\xmark}  \\
                PersonaChatGen~\cite{lee-etal-2022-personachatgen} & Closed & \textcolor{BrickRed}{\xmark} &  1.6k & \$\$$^{\textcolor{ForestGreen}{\checkmark}}$ & \textcolor{BrickRed}{\xmark}  & \textcolor{BrickRed}{\xmark}  \\
                SPC~\cite{jandaghi-etal-2024-faithful} & Closed & \textcolor{BrickRed}{\xmark} & 20k & \$\$$^{\textcolor{ForestGreen}{\checkmark}}$ & \textcolor{BrickRed}{\xmark} & \textcolor{BrickRed}{\xmark}  \\
                \textbf{MOUD} (ours) & Open & \textcolor{ForestGreen}{\checkmark}$^{29}$  & 493k & \textcolor{ForestGreen}{\checkmark} &  \textcolor{ForestGreen}{\checkmark} & \textcolor{ForestGreen}{\checkmark}$^{29}$\\ 
            \noalign{\smallskip}\hline\noalign{\smallskip}
        \end{tabular}
}
    \caption{Comparative Analysis of MOUD and Other Open-Domain Persona-Based Dialogue Datasets.}
    \label{tab:comparison-with-others}
\end{table*}
Furthermore, these constraints also mention non-desirable behaviors like translation of demonstration examples when applicable (non 0-shot) and repetitiveness, among others.





\subsection{Prompt function and generation task}
The prompt for a given step $k$ is therefore formulated as:
\begin{equation}
\label{eq:stepk-prompt}
\resizebox{0.8966\linewidth}{!}
{
    \ensuremath{\mathcal{P}_{k}(\mathcal{D}_{s,en}^{k,n}, l_T) := 
    i_{0}{\Big\Vert}\mathcal{I}_k 
    \Big\Vert c_{0}
    \Big\Vert \mathcal{C}_{k,l_T} 
    \Big\Vert d_{0}
    \Big\Vert\mathcal{D}_{s,en}^{k,n}
    \Big\Vert i_{gen}}
}
\end{equation}
where $\Vert$ represents concatenation preceded by new line; $i_{0}$, $c_{0}$, $d_{0}$ are additional section strings, respectively "\texttt{Instructions:}", "\texttt{Constraints:}"  and 
"\texttt{examples}" 
when in non 0-shot settings; $\mathcal{D}_{s,en}^{k,n}$ a subset of $n$ demonstration samples in $l_S$; $i_{gen}$ an instruction  to incite the LLM to generate new samples including the targeted number of new samples.

Hence, for a given step $k$ and a language $l_T$, the generation task  corresponds to maximizing the following probability where $y$ is the desired text output at step $k$:
\begin{equation}
\label{eq:maximzed-probability}
\resizebox{0.8966\linewidth}{!}
{
    \ensuremath{
    p(y|\mathcal{P}_{k}(\mathcal{D}_{s,en}^{k,n}, l_T)) = \prod_t 
    p(y_t|y_1, ..., y_{t-1}, \mathcal{P}_{k}(\mathcal{D}_{s,en}^{k,n}, l_T))
    }
}
\end{equation}


\subsection{$l_T$ dataset generation} 
\label{subsec:last-step-generation}
For a dialogue task, the last step ($k=N_{step}$) corresponds to chat generation. Depending on the source dataset, both speakers in a conversation may not be equivalent. As a consequence, a speaker-specific prompt (associated with a dedicated LLM instance) is derived from previous steps' results and relevant guidelines with attention to the speaker's role. For a given speaker denoted as $a$, the dedicated prompt is as follows\footnote{For the sake of readability as $k = N_{step}$, the step index has been removed from the expression.}:
\begin{equation}
\label{eq:speaker-prompt}
\resizebox{0.8966\linewidth}{!}
{
    \ensuremath{\mathcal{P}_{a}(\mathcal{D}_{s,en}^{n}, l_T) := 
    i_{0}{\Big\Vert}\mathcal{I}^{a}
    \Big\Vert c_{0}
    \Big\Vert \mathcal{C}_{l_T}^a
    \Big\Vert d_{0}
    \Big\Vert\mathcal{D}_{s,en}^{n}
    \Big\Vert i_{gen}^a}
}
\end{equation}
$^a$ highlights the speaker-specificity of the concerned element.
Ergo, a chat is generated by doing back-and-forths between the speakers' instances, each answering to the other's utterance by, at each turn, maximizing the following probability :

\begin{equation}
\label{eq:back-and-forth-proba}
{
    \ensuremath{ 
    p(y_a|y_b, \mathcal{P}_{a}(\mathcal{D}_{s,en}^{n}, l_T)) 
    }
}
\end{equation}
where $a \neq b \in $ $\{$\texttt{speaker$_1$}, \texttt{speaker$_2$}$\}$ and $y$ is a dialogue utterance.
%

\section{MOUD Dataset}
\label{sec:moud-generation}
In this section, the method presented in Section~\ref{sec:methodology} is applied to PersonaChat~\cite{zhang-etal-2018-personalizing} and 28 $l_T$, details of which can be found in Table~\ref{tab:languages-list}.

\subsection{Models' selections}
Many similar works often rely on proprietary models with high access costs (approaches with source sets to \textit{closed} in Table~\ref{tab:comparison-with-others}), limiting their reproducibility). To address this, MOUD is collected with open-source SOTA instruction-tuned LLMs from different backgrounds at \textit{medium}\footnote{Around 7B-8B parameters.} size to favour cost-effective reproducibility and extensibility. An additional selection criterion was the ability to generate texts in different languages whether explicitly trained for this purpose or not (as Table~\ref{tab:comparison-with-others} shows, the only multilingual approach has just six $l_T \neq l_S$ and relies heavily on MT). Our final shortlist comprises:~\texttt{Meta-Llama-3.1-8B-Instruct}~\cite{llama3modelcard}, \texttt{Mistral-7B-Instruct-v0.3}~\cite{jiang2023mistral}, \texttt{Gemmma-1.1-7b-it}~\cite{gemmateam2024gemma} from Google and CohereAI's \texttt{aya-23-8B}~\cite{aryabumi2024aya}.

For all steps and models, nucleus sampling was used as decoding strategy with $p=0.9$ to allow for more diversity, repetition penalty set to $1.2$, temperature to $\theta=0.7$ (except for persona generation where we also tested $\theta=0.8$). 

\subsection{Demonstration examples selection}
\paragraph{Personas:} Examples were randomly sampled from the PersonaChat dataset ($l_S$). To evaluate the effect of the selected examples, experiments used using three different seeds: \textbf{42, 10,} and \textbf{0} and varied the number of demonstration examples with  \textbf{$n \in \{0, 1, 2, 4, 6, 8, 10\}$}. Impact on output similarity is illustrated in Appendix~\ref{appendix:per_language_evals}).

\paragraph{CG and Conversations:} As CG was introduced to improve the source dataset regarding the \textit{ODP}, it had no prior examples. This same reason implied we had no prior common grounded conversations for demonstration; hence, these steps were performed in 0-shot approach. The complete generation pipeline illustrated in Figure~\ref{fig:CompletePipeline} is described below.

\subsection{Persona Generation}
Human personality is manifold, hence tricky to define. 
We settle to~\citet{zhang-etal-2018-personalizing}'s definition: a character defined by multiple profile sentences (5 for instance) where each can be represented as a triplet (category, \textit{relation}, entity).
Attempting to represent the multifaceted human personalities,~\citet{lee-etal-2022-personachatgen} generated persona profiles sentences using a taxonomy of Hierarchical Personas Categories while~\citet{jandaghi-etal-2024-faithful} grouped $l_S$ existing persona profiles and prompted a LLM to come up with new similar groups. In both cases, at profile sentence level before being associated in groups of five to have a persona.

\subsubsection{Persona Profiles Taxonomy}
\label{subsec:persona-taxonomy}

We chose the first approach and augmented the taxonomy provided in different ways for the sake persona diversity and quality. First, for each category/subcategory/entity combination, we associated a sentence for a better understanding by the LLM during generation for example (see Appendix~\ref{appendix:persona-taxonomy} for complete taxonomy): \texttt{Demographics|Possession|Vehicle} $\Rightarrow$ \textit{"a vehicle you possess or wish to"}. Then, as shown in \textit{Step 0} in Figure~\ref{fig:CompletePipeline}, for each main category (Demographics, Wellness or Psychographics) we interactively prompted the free online version of ChatGPT\footnote{\href{https://chatgpt.com}{https://chatgpt.com} running on free \href{https://openai.com/index/hello-gpt-4o/}{\texttt{GPT-4o}}. This choice was made to avoid using the same LLM as those serving for data generation while ensuring no additional cost at this step.} to generate new subcategories, new entities and corresponding sentences 
and we manually curated them. Finally, for all the aforementioned entity sentences, the LLM was prompted to create up to ten \textbf{multi-polarised} reformulations for improved variability even within a given entity. This taxonomy update step is even more important as it helps bring up to date subjects of interests ranging from AI to climate change awareness and does not rely \textit{only} on human knowledge. 


\subsubsection{Persona Generation}


Unlike~\citealp{lee-etal-2022-personachatgen} (resp.~\citealp{jandaghi-etal-2024-faithful}), where persona's profile sentences are generated separately 
, we randomly choose five different taxonomy entities, add them to the prompt's constraints $\mathcal{C}_{1,l_T}$, and generate a complete persona with respect to them. This ensured global coherence within each persona at a low cost, whereas the cited works required complex selection processes to combine profile sentences. 
Complete prompt in Appendix~\ref{appendix:persona-prompt}.

\subsection{Common Ground Generation}
For a successful, meaningful and jointly coordinated ODD, the involved speakers often must share a CG which according to ~\citet{Clark_1996} is the \textit{"the sum of their mutual, common or joint knowledge, beliefs and suppositions"}. Indeed, real life human-human ODD rarely starts without any clue on why the chat is taking place (\textit{joint activity}) or outside a specific context: the \textit{ODP} explained by~\citealp{skantze-dogruoz-2023-open}) who presented the concept of \textbf{speech events} 
as a potential solution. 


\subsubsection{Speech Events Taxonomy}
~\citet{Goldsmith2006} developed a taxonomy of SEs 
which was updated similarly to section~\ref{subsec:persona-taxonomy} with LLM assistance. 
Difference were made between SEs where both speakers have symmetric roles (e.g., \texttt{Informal|Reminiscing}) and those with asymmetric roles (e.g., \texttt{Goal-directed|Asking a favor}) in their descriptions to clearly define each speaker's role and reformulations were added to promote diversity. This is provided in Appendix~\ref{appendix:speech-event-taxonomy}.

\subsubsection{Generation}
This 
step is entirely new. Instructions and constraints were designed from scratch with the following objectives: creating a CG that takes into account both speakers' personas, the targeted SE type and $l_T$ specificity. The key in this step, was to task the model to act as a \textit{"Narrator"} that creates and tells the context of a SE-type-chat between the speakers as shown by the prompt in Appendix~\ref{appendix:common-ground-prompt}.


\subsection{Conversation Generation}
%
In the original PersonaChat, both speakers are equivalent and tasked to \textit{"try to get to know each other"} which corresponds to one SE type out of 29 in the taxonomy which makes it not so "open-domain". 
Hence, for each $l_T$'s conversation, after randomly picking two $l_T$ personas among those generated, 
a SE type is selected and the associated CG in $l_T$ generated. Then, all are integrated in the prompts assigned to two LLM instances as explained in Section~\ref{subsec:last-step-generation} with careful distinction between the two speakers' prompts depending on SE speakers' roles symmetry. 

Another key difference is the conversation length. While PersonaChat sets conversation length at exactly 7 turns, we vary the length between 4 and 10 turns (where 1 turn equals one utterance per speaker), with the exact length randomly chosen prior to each generation. This variation is intended to improve the robustness of models trained on the resulting data by making them adaptable to different conversation lengths.

Regarding Equation~\ref{eq:stepk-prompt}, $i_{gen}$ includes the content of the CG for only the first two turns (acting as a "warm-up" stage). Additionally, the prompts for the very first message of the conversation differ from those for subsequent utterances to encourage more natural and engaging interactions. The full prompt can be found in Appendix~\ref{appendix:conversation-prompt}.

\begin{figure}[hb!]
    \centering
    \includegraphics[scale=0.38]{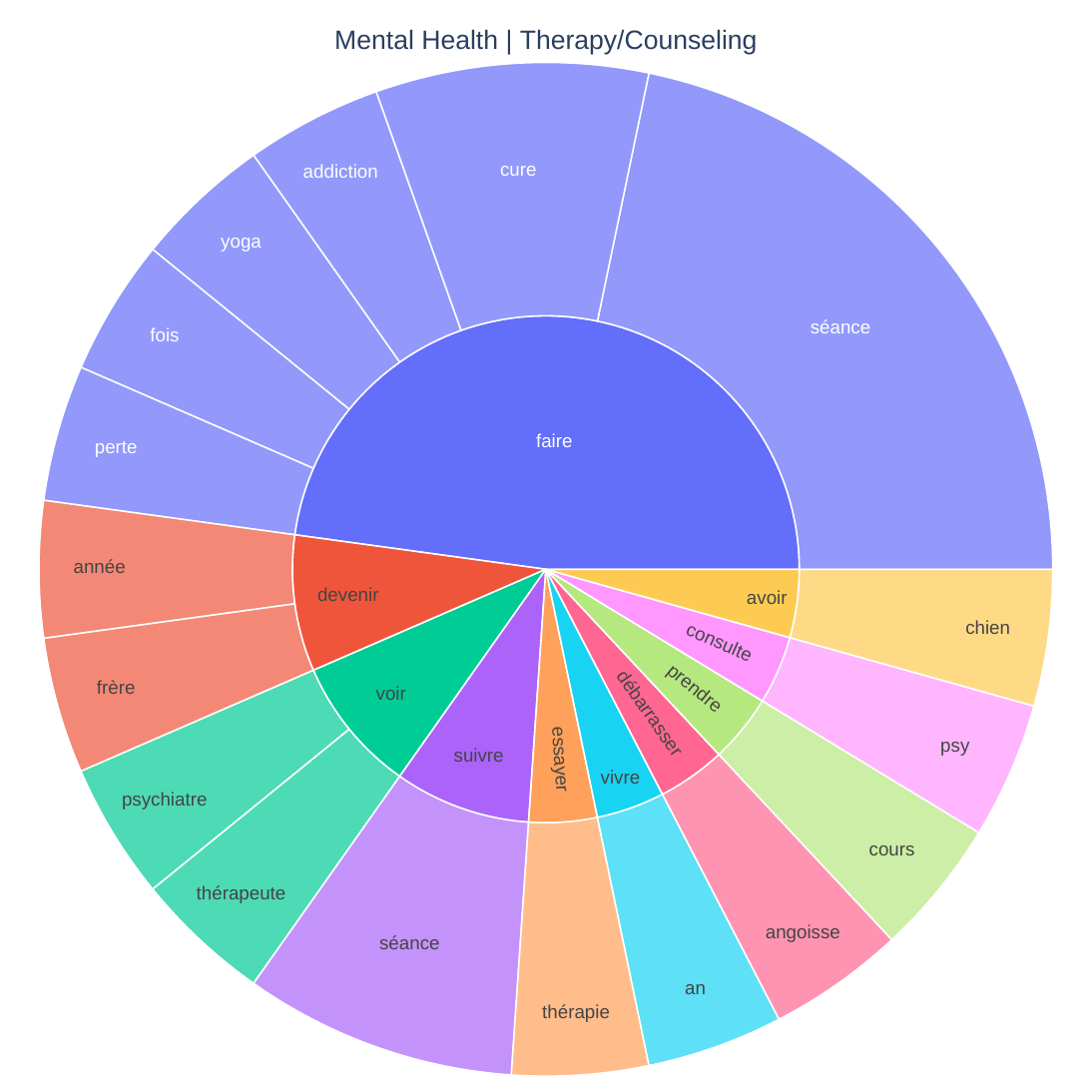}
    \caption{Sunburst chart of the entity with the most root verbs and associated direct object nouns for French generated personas with LLaMA3.1-8B.}
    \label{fig:french-syntax-example}
\end{figure}

\begin{table*}[hb!]
    \centering
    \resizebox{0.7\textwidth}{!}{
         \begin{tabular} {ll|c|ccc|c||ccc|c||ccc|c}
            \hline
            \noalign{\smallskip}
             \multicolumn{2}{c|}{\multirow{4}{*}{\makecell{\textbf{Lang.}}}} & \multicolumn{13}{c}{\textbf{Models}}  \\
              & & \multicolumn{1}{c}{{\thead{Aya$^*$}}} &  \multicolumn{4}{c}{{\thead{Gemma-1.1-7b}}} &  \multicolumn{4}{c}{\thead{LLaMA3.1-8B}} &   \multicolumn{4}{c}{{\thead{Mistral-7B}}}   \\
              \noalign{\smallskip}\cline{3-15}
              \noalign{\smallskip}
                &    &   P  &   P  &  CG  &  C  &  Avg. &   P  &  CG  &   C  &  Avg. &  P  &  CG  &   C  &  Avg. \\
                \noalign{\smallskip}\hline\noalign{\smallskip}
\multirow{ 13 }{*}{\rotatebox{90}{\textbf{High-Resource}}}
 & en   &   4.27   &   4.24 & 4.20 & 4.01 & 4.15   &   4.38 & \textbf{4.70} & \textbf{4.61} & \textbf{4.56}   &   \textbf{4.55} & 4.50 & 4.36 & 4.47  \\
 & ru   &   3.91   &   3.60 & 3.38 & 3.03 & 3.34   &   4.13 & \textbf{4.57} & \textbf{4.48} & \textbf{4.39}   &   \textbf{4.35} & 4.20 & 3.91 & 4.15  \\
 & de   &   3.96   &   3.69 & 3.67 & 3.36 & 3.57   &   4.21 & \textbf{4.59} & \textbf{4.49} & \textbf{4.43}   &   \textbf{4.27} & 4.01 & 3.75 & 4.01  \\
 & jp   &   \textbf{4.18}   &   3.88 & 3.45 & 3.07 & 3.47   &   4.18 & \textbf{4.14} & \textbf{4.01} & \textbf{4.11}   &   3.85 & 3.53 & 3.17 & 3.52  \\
 & es   &   4.15   &   3.90 & 3.95 & 3.63 & 3.83   &   4.38 & \textbf{4.78} & \textbf{4.67} & \textbf{4.61}   &   \textbf{4.44} & 4.30 & 3.96 & 4.23  \\
 & zh   &   4.05   &   4.05 & 3.70 & 3.31 & 3.69   &   4.28 & \textbf{4.42} & \textbf{4.40} & \textbf{4.37}   &   \textbf{4.39} & 4.03 & 3.77 & 4.06  \\
 & fr   &   4.22   &   3.87 & 3.79 & 3.46 & 3.71   &   4.31 & \textbf{4.78} & \textbf{4.65} & \textbf{4.58}   &   \textbf{4.40} & 4.29 & 3.97 & 4.22  \\
 & it   &   4.20   &   3.84 & 4.01 & 3.54 & 3.80   &   \textbf{4.38} & \textbf{4.82} & \textbf{4.72} & \textbf{4.64}   &   4.33 & 4.23 & 3.88 & 4.15  \\
 & nl   &   4.00   &   3.61 & 3.67 & 3.31 & 3.53   &   \textbf{4.26} & \textbf{4.63} & \textbf{4.54} & \textbf{4.48}   &   4.11 & 4.01 & 3.74 & 3.96  \\
 & pt   &   4.01   &   3.76 & 3.89 & 3.51 & 3.72   &   4.31 & \textbf{4.79} & \textbf{4.67} & \textbf{4.59}   &   \textbf{4.38} & 4.22 & 3.97 & 4.19  \\
 & pl   &   \textbf{4.05}   &   3.46 & 3.37 & 3.01 & 3.28   &   4.02 & \textbf{4.44} & \textbf{4.31} & \textbf{4.26}   &   4.01 & 4.01 & 3.63 & 3.88  \\
 & tr   &   \textbf{4.20}   &   3.39 & 3.13 & 2.78 & 3.10   &   3.80 & \textbf{4.05} & \textbf{3.95} & \textbf{3.93}   &   3.38 & 3.04 & 2.68 & 3.03  \\

 \noalign{\smallskip}\cline{2-15}\noalign{\smallskip}
 & avg. &   4.10   &   3.77 & 3.69 & 3.34 & 3.60   &   \textbf{4.22} & \textbf{4.56} & \textbf{4.46} & \textbf{4.41}   &   4.20 & 4.03 & 3.73 & 3.99  \\
\noalign{\smallskip}\hline\noalign{\smallskip}

\multirow{15}{*}{\rotatebox{90}{\textbf{Medium-Resource}}}
 & vi   &   4.02   &   4.01 & 3.69 & 3.35 & 3.68   &   \textbf{4.32} & \textbf{4.69} & \textbf{4.64} & \textbf{4.55}   &   3.81 & 3.69 & 3.38 & 3.63  \\
 & id   &   4.16   &   3.92 & 3.88 & 3.54 & 3.78   &   \textbf{4.36} & \textbf{4.73} & \textbf{4.65} & \textbf{4.58}   &   4.17 & 4.01 & 3.69 & 3.96  \\
 & ko   &   \textbf{4.04}   &   3.81 & 3.59 & 3.20 & 3.53   &   4.00 & \textbf{4.00} & \textbf{3.81} & \textbf{3.94}   &   3.93 & 3.79 & 3.41 & 3.71  \\
 & sv   &   3.13   &   3.38 & 3.69 & 3.33 & 3.46   &   4.19 & \textbf{4.62} & \textbf{4.52} & \textbf{4.44}   &   \textbf{4.23} & 4.14 & 3.84 & 4.07  \\
 & ar   &   3.83   &   3.23 & 3.07 & 2.69 & 3.00   &   \textbf{3.96} & \textbf{4.22} & \textbf{4.10} & \textbf{4.09}   &   3.38 & 3.25 & 2.92 & 3.19  \\
 & hu   &   2.63   &   3.30 & 3.03 & 2.68 & 3.00   &   \textbf{3.99} & \textbf{4.37} & \textbf{4.25} & \textbf{4.20}   &   3.87 & 3.81 & 3.45 & 3.71  \\
 & el   &   \textbf{4.24}   &   2.82 & 2.50 & 2.09 & 2.47   &   3.70 & \textbf{3.99} & \textbf{3.80} & \textbf{3.83}   &   3.12 & 2.79 & 2.40 & 2.77  \\
 & uk   &   4.04   &   3.58 & 3.62 & 3.23 & 3.48   &   4.06 & \textbf{4.68} & \textbf{4.55} & \textbf{4.43}   &   \textbf{4.36} & 4.31 & 4.01 & 4.22  \\
 & da   &   3.06   &   3.67 & 3.86 & 3.51 & 3.68   &   4.06 & \textbf{4.61} & \textbf{4.50} & \textbf{4.39}   &   \textbf{4.18} & 4.14 & 3.80 & 4.04  \\
 & th   &   3.06   &   3.57 & 3.46 & 3.01 & 3.35   &   \textbf{4.17} & \textbf{4.25} & \textbf{4.12} & \textbf{4.18}   &   3.30 & 2.99 & 2.56 & 2.95  \\
 & fi   &   2.52   &   3.14 & 3.01 & 2.60 & 2.92   &   \textbf{3.68} & \textbf{3.80} & \textbf{3.61} & \textbf{3.70}   &   3.06 & 2.72 & 2.44 & 2.74  \\
 & hr   &   2.99   &   3.21 & 3.37 & 2.92 & 3.17   &   3.77 & \textbf{4.17} & \textbf{3.99} & \textbf{3.97}   &   \textbf{3.94} & 3.98 & 3.60 & 3.84  \\
 & hi   &   3.95   &   3.52 & 3.32 & 2.98 & 3.28   &   \textbf{4.22} & \textbf{4.48} & \textbf{4.44} & \textbf{4.38}   &   3.31 & 3.25 & 2.84 & 3.13  \\
 & bn   &   2.78   &   3.24 & 2.99 & 2.65 & 2.96   &   \textbf{3.90} & \textbf{4.08} & \textbf{3.92} & \textbf{3.97}   &   3.07 & 2.62 & 2.24 & 2.65  \\

 \noalign{\smallskip}\cline{2-15}\noalign{\smallskip}
 & avg. &   3.46   &   3.46 & 3.36 & 2.98 & 3.27   &   \textbf{4.03} & \textbf{4.34} & \textbf{4.21} & \textbf{4.19}   &   3.69 & 3.54 & 3.18 & 3.47  \\ \noalign{\smallskip}\hline\noalign{\smallskip}

\multirow{5}{*}{\rotatebox{90}{\textbf{Low-Res.}}}
 & af   &   3.30   &   3.51 & 3.51 & 3.25 & 3.42   &   \textbf{3.96} & \textbf{4.34} & \textbf{4.24} & \textbf{4.18}   &   3.73 & 3.44 & 3.04 & 3.40  \\
 & sw   &   2.18   &   2.99 & 2.44 & 2.13 & 2.52   &   \textbf{3.48} & \textbf{3.75} & \textbf{3.57} & \textbf{3.60}   &   2.78 & 2.05 & 1.84 & 2.22  \\
 & yo   &   2.30   &   3.06 & \textbf{2.72} & \textbf{2.38} & \textbf{2.72}   &   3.15 & 2.60 & 2.26 & 2.67   &   \textbf{3.19} & 2.62 & 2.16 & 2.66  \\

 \noalign{\smallskip}\cline{2-15}\noalign{\smallskip}
 & avg. &   2.60   &   3.18 & 2.89 & 2.58 & 2.89   &   \textbf{3.53} & \textbf{3.56} & \textbf{3.36} & \textbf{3.48}   &   3.23 & 2.70 & 2.34 & 2.76  \\
\noalign{\smallskip}\hline\noalign{\smallskip}
        \end{tabular}

} 
   
    $^*$\scriptsize{Stands for \texttt{Aya-23-8B} which was dismissed for Common Grounds and Conversations generations as it struggled to follow instructions.}
    \caption{Generated Data Evaluation with \texttt{GPT4o}-as-a-judge. For each part of the dataset i-e Personas (P) Common Grounds (CG) Conversations (C), average over their distinct criteria (c.f. Appendix~\ref{appendix:eval-details}) is reported.  In bold, the best ratings among the models for each part.} 
    \label{tab:llm-as-judge}
    
\end{table*}

\subsection{Filtering}
For each generation step, a filtering post-processing is applied to ensure quality. In each target language ($l_T$), we perform hits@2 language detection, dropping data if $l_T$ isn't detected or if English ($l_S$) appears. For CG, data is discarded if "character" 1 and 2 (in $l_T$)  do not explicitly appear as constrained by the prompt. 
In conversations, incomplete or repetitive utterances are removed. Extras texts generated by the LLM, sometimes as explanations, introductory or concluding speeches, are also removed when detected.
As nucleus sampling is used, we allow two retries, with each retry incrementally adding two more generated options. If CG reaches max retries, the conversation is dropped. For utterances, if fewer than the minimal number of turns (4) are generated, the conversation is discarded; otherwise, it’s kept even if early stopping occurs (not reaching the number of turns fixed at the start).

\begin{table*}[h!]
    \centering
    \resizebox{0.7\textwidth}{!}{
       \begin{tabular} {ll|c|ccc|c||ccc|c||ccc|c}
           \hline
           \noalign{\smallskip}
            \multicolumn{2}{c|}{\multirow{4}{*}{\makecell{\textbf{Lang.}}}} & \multirow{3}{*}{\makecell{\textbf{Eval.} \\ \textbf{Source}}} & \multicolumn{12}{c}{\textbf{Models}} \\
             & & &
             \multicolumn{4}{c}{{\thead{Gemma-1.1-7b}}} &  \multicolumn{4}{c}{\thead{LLaMA3.1-8B}} &   \multicolumn{4}{c}{{\thead{Mistral-7B}}}   \\
             \cline{4-15}
                &  &  &   P  &  CG  &  C  &  Avg. &   P  &  CG  &   C  &  Avg. &  P  &  CG  &   C  &  Avg. \\
                \noalign{\smallskip}\hline\noalign{\smallskip}


\multicolumn{2}{c|}{\multirow{2}{*}{es}}    &   \textit{GPT4o}   &   3.96 & 3.90 & 3.53 & 3.80   &   3.93 & \textbf{4.80} & \textbf{4.67} & \textbf{4.47}   &   \textbf{4.53} & 4.44 & 4.13 & 4.37  \\ 
& &  \textit{Human}   &   3.30 & 3.38 & 2.67 & 3.12   &   \textbf{3.84} & \textbf{4.31} & \textbf{3.72} & \textbf{3.96}   &   3.83 & 3.93 & 3.17 & 3.64  \\ 
\noalign{\smallskip}\hline\noalign{\smallskip}

\multicolumn{2}{c|}{\multirow{2}{*}{zh}}    &   \textit{GPT4o}   &   4.18 & 3.54 & 3.29 & 3.67   &   4.16 & \textbf{4.45} & \textbf{4.42} & \textbf{4.34}   &   \textbf{4.39} & 3.84 & 3.53 & 3.92  \\ 
& &  \textit{Human}   &   \textbf{4.70} & 4.23 & 3.14 & 4.02   &   4.58 & \textbf{4.98} & \textbf{4.27} & \textbf{4.61}   &   4.47 & 4.50 & 3.53 & 4.17  \\ 
\noalign{\smallskip}\hline\noalign{\smallskip}

\multicolumn{2}{c|}{\multirow{2}{*}{fr}}    &   \textit{GPT4o}   &   3.81 & 3.73 & 3.45 & 3.66   &   \textbf{4.38} & \textbf{4.82} & \textbf{4.72} & \textbf{4.64}   &   4.34 & 4.28 & 3.99 & 4.20  \\ 
& &  \textit{Human}   &   4.65 & 4.75 & 4.04 & 4.48   &   \textbf{4.75} & \textbf{4.82} & \textbf{4.38} & \textbf{4.65}   &   4.62 & 4.52 & 3.62 & 4.25  \\ 
\noalign{\smallskip}\hline\noalign{\smallskip}


\multicolumn{2}{c|}{\multirow{2}{*}{vi}}    &   \textit{GPT4o}   &   3.85 & 3.64 & 3.31 & 3.60   &   \textbf{4.13} & \textbf{4.54} & \textbf{4.67} & \textbf{4.45}   &   3.82 & 3.60 & 3.26 & 3.56  \\ 
& &  \textit{Human}   &   4.32 & 4.24 & 3.46 & 4.01   &   \textbf{4.43} & \textbf{4.77} & \textbf{4.79} & \textbf{4.66}   &   3.79 & 3.46 & 2.98 & 3.41  \\ 
\noalign{\smallskip}\hline\noalign{\smallskip}

\multicolumn{2}{c|}{\multirow{2}{*}{ar}}    &   \textit{GPT4o}   &   3.16 & 2.95 & 2.66 & 2.92   &   \textbf{3.95} & \textbf{4.22} & \textbf{4.08} & \textbf{4.08}   &   3.34 & 3.39 & 3.00 & 3.24  \\ 
& &  \textit{Human}   &   3.72 & 3.66 & 3.05 & 3.48   &   \textbf{4.44} & \textbf{4.42} & \textbf{4.13} & \textbf{4.33}   &   3.66 & 3.71 & 3.48 & 3.61  \\ 

\noalign{\smallskip}\hline\noalign{\smallskip}
        \end{tabular}

    } 
    \caption{Generated Data Evaluation by Human with \texttt{GPT4o} Judgments On The Same Data Points. For each part of the dataset i-e Personas (P) Common Grounds (CG) Conversations (C), average over their distinct criteria (c.f. Appendix~\ref{appendix:eval-details}) is reported.  In bold, the best ratings among the models for each part.} 
    \label{tab:human-eval}
    
\end{table*}

\section{Qualitative Evaluation of MOUD}
\label{sec:data-qualitative-evaluation}
\subsection{Personas Diversity Analysis with Automatic Metrics}




For each model-language pair, 300 personas were randomly selected and BERTScore (B$_s$) ~\citet{Zhang*2020BERTScore:} computed over 10,000 persona pairs to assess their similarity, comparing it to that of the original PersonaChat (English). 
\texttt{mT5-xl}~\cite{xue-etal-2021-mt5}, a highly multilingual model, was used to ensure consistent cross-lingual comparisons. For PersonaChat $B_s=0.5727$ and for the generated data, the tendency depends on $l_T$  as shown in Appendix~\ref{appendix:per_language_evals}. These figures also help understand, if and how the selected source examples and the decoding parameters may impact the generated data. We observed across the different tested configurations, within each language, a rather stable performance for the models. This implies little to no example is enough to replicate this process.


Furthermore, when available for a language, we used Spacy pretrained models\footnote{See the list of models at~\url{https://spacy.io/models/}} to detect most common root verbs and associated direct object nouns per persona taxonomy entity on the same 300 samples. This allows assessing both taxonomy relevancy and  variability. Figure~\ref{fig:french-syntax-example} gives an example for the entity \texttt{Mental Health | Therapy/Counseling} from generated French personas with LLaMA3.1-8B. We can see root verbs like \textit{"consulter"} (to consult) associated to \textit{"psy"}, or \textit{"faire"} (do) associated to \textit{"yoga"}, \textit{"cure"} and direct object nouns like \textit{"angoisse"} (anxiety), \textit{"thérapie"} (therapy) all relevant to the taxonomy entity and  diverse. More examples for some languages can be found in Appendix~\ref{appendix:per_language_evals}.

\subsection{Data Quality with Selected Criteria}
\label{subsec:quality-evaluation}

Given the one-to-many nature of ODD, reference-based automatic metrics often fall short in aligning with human perceptions. As a result, evaluating additional criteria is essential to achieve a more comprehensive assessment. In our case, we aim to measure output quality across several dimensions, specifically targeting multilingual aspects and the task at hand. The criteria used for evaluation include: \textbf{specificity} and \textbf{fluency} in the target language ($l_T$), \textbf{toxicity}, \textbf{humanness} in conversational exchanges, and \textbf{relevancy} to selected taxonomies, personas, and common ground. Refer to Appendix~\ref{appendix:eval-details} for detailed definitions.

\subsubsection{Analysis with LLM as a Judge}
\label{subsubsec:llm-as-a-judge}

Given the variety of languages involved, the challenge of high costs associated with generating data through human crowdworkers is transferred to the evaluation process. Finding voluntary human evaluators proficient in each language—and willing to assess large data batches—is arduous. Therefore, to address the lack of sufficient human evaluators, we decided to leverage \texttt{GPT4o-2024-08-06}, a state-of-the-art yet closed-source LLM, often used for such tasks. While not a perfect substitute for comprehensive human evaluations, \texttt{GPT4o} provides a feasible alternative~\cite{NEURIPS2023_91f18a12, chiang-lee-2023-large}, enabling consistent and scalable quality assessments across multiple languages.

For each model-language pair, we assessed 100 conversations (a total of \textbf{8,700} conversations) and 300 personas (a total of \textbf{34,800} personas), for less than \$100 of an additional cost. 
The results, summarized in Table~\ref{tab:llm-as-judge}, indicate that LlaMA3.1-8B consistently performed the best across most languages and data categories. In the few instances (primarily persona evaluations) where it did not rank first, the difference was usually minor, and it remained the top performer on average for all languages except Yoruba, where Gemma-1.1-7b-it was judged superior. As reported in detailed results per criteria in Tables~\ref{tab:personas-llm-as-judge},~\ref{tab:common-ground-llm-as-judge} and~\ref{tab:conversation-llm-as-judge}, it was consistently best for specificity to $l_T$ in all data parts and, for fluency (except personas), humanness and all other criteria for CG and conversations the most critical part of the data.  Based on these findings, of all these models, LlaMA3.1-8B was selected as the sole open-source LLM to generate the final MOUD dataset statistically described in Table~\ref{tab:languages-list}. 


\subsubsection{Human Evaluation}

As stated in Section~\ref{subsubsec:llm-as-a-judge}, finding voluntary evaluators for all the languages willing to assess large data batches is challenging. Nevertheless, to support the LLM judgments, human evaluation on the same set of data was still performed. Description of the evaluators pool is provided in Appendix~\ref{subappendix:evaluators-demographics}.

Results for some languages, where a sufficient number of evaluations were gathered, are presented in Table~\ref{tab:human-eval}, with detailed on criteria outlined in Table~\ref{tab:personas-human}, Table~\ref{tab:common-ground-human}, and Table~\ref{tab:conversation-human}. These results indicate that, on average and across human-evaluated conversations, \textbf{both humans and the LLM tend to rate conversations in the same direction}. Notably, the conclusions drawn from LLM judgments remain consistent for this subset of languages and conversations: LLaMA3.1-8B demonstrates the highest overall quality on average across all data parts and most of their associated criteria.

\begin{table*}[h!]
    \centering
   \resizebox{0.7\textwidth}{!}{
       \begin{tabular} {ll|c|cc|c|||cc|c|||cc|c||cc|c}
           \hline
           \noalign{\smallskip}
            \multicolumn{2}{c|}{\multirow{4}{*}{\makecell{\textbf{Lang.}}}} & \multirow{2}{*}{\makecell{\textbf{\textit{Metric}}}} & \multicolumn{12}{c}{\textbf{Training Set}} \\
             & & &
             \multicolumn{3}{c}{{\thead{XPersona (XP)}}} &  \multicolumn{3}{c}{{\thead{MOUD (M)}}}  &  \multicolumn{6}{c}{\thead{MOUD + XP}} \\
             \cline{3-15}
                &  & \multirow{2}{*}{Test-set} &  \multirow{2}{*}{XP}  &  \multirow{2}{*}{M}  &  \multirow{2}{*}{Avg.} &   \multirow{2}{*}{XP}  &  \multirow{2}{*}{M}  &  \multirow{2}{*}{Avg.} &  \multirow{2}{*}{XP}  &  \multirow{2}{*}{M}  &  \multirow{2}{*}{Avg.} &  \multicolumn{3}{c}{Gap to XP Model in \%} \\

               & & & & & & & & & & & & XP & M & Avg. \\
                \hline\noalign{\smallskip}
                
\multicolumn{2}{c|}{\multirow{4}{*}{en}}   &    \textit{\textbf{bert-f1}}   &   0.66 & 0.67 & 0.67   &   0.66 & \textbf{0.70} & 0.68   &   \textbf{0.67} & \textbf{0.70} & \textbf{0.69}   &   \cellcolor{green!20}1.52 & \cellcolor{green!20}4.48 & \cellcolor{green!20}3.01  \\
 &  &   \textit{\textbf{Hits@1}}   &   \textbf{0.93} & 0.77 & 0.85   &   0.84 & \textbf{0.99} & 0.92   &   0.92 & 0.98 & \textbf{0.95}   &   \cellcolor{red!20}-1.08 & \cellcolor{green!20}27.27 & \cellcolor{green!20}11.76  \\
 &  &   \textit{\textbf{ppl}}   &   22.98 & 991.50 & 507.24   &   866.10 & \textbf{8.33} & 437.22   &   \textbf{9.66} & 11.84 & \textbf{10.75}   &   \cellcolor{green!20}-57.96 & \cellcolor{green!20}-98.81 & \cellcolor{green!20}-97.88  \\
 &  &   \textit{\textbf{RougeL}}   &   11.90 & 13.69 & 12.79   &   10.11 & 15.76 & 12.93   &   \textbf{11.97} & \textbf{16.39} & \textbf{14.18}   &   \cellcolor{green!20}0.59 & \cellcolor{green!20}19.72 & \cellcolor{green!20}10.82  \\
\noalign{\smallskip}\hline\noalign{\smallskip}

\multicolumn{2}{c|}{\multirow{4}{*}{jp}}   &    \textit{\textbf{bert-f1}}   &   0.67 & 0.67 & 0.67   &   0.67 & \textbf{0.70} & 0.69   &   \textbf{0.68} & \textbf{0.70} & \textbf{0.69}   &   \cellcolor{green!20}1.49 & \cellcolor{green!20}4.48 & \cellcolor{green!20}2.99  \\
 &  &   \textit{\textbf{Hits@1}}   &   \textbf{0.90} & 0.87 & 0.89   &   0.76 & \textbf{0.98} & 0.87   &   \textbf{0.90} & 0.97 & \textbf{0.94}   &   \cellcolor{green!20}0.00 & \cellcolor{green!20}11.49 & \cellcolor{green!20}5.65  \\
 &  &   \textit{\textbf{ppl}}   &   \textbf{6.09} & 5.39 & 5.74   &   47.54 & 2.47 & 25.00   &   8.32 & 0.00 & \textbf{5.07}   &   \cellcolor{red!20}36.62 & \cellcolor{green!20}-66.42 & \cellcolor{green!20}-11.76  \\
 &  &   \textit{\textbf{RougeL}}   &   10.53 & 11.23 & 10.88   &   9.93 & 12.70 & 11.31   &   \textbf{11.70} & \textbf{14.58} & \textbf{13.14}   &   \cellcolor{green!20}11.11 & \cellcolor{green!20}29.83 & \cellcolor{green!20}20.77  \\
\noalign{\smallskip}\hline\noalign{\smallskip}

\multicolumn{2}{c|}{\multirow{4}{*}{zh}}   &    \textit{\textbf{bert-f1}}   &   \textbf{0.69} & 0.68 & 0.69   &   \textbf{0.69} & \textbf{0.72} & \textbf{0.71}   &   \textbf{0.69} & 0.71 & 0.70   &   \cellcolor{green!20}0.00 & \cellcolor{green!20}4.41 & \cellcolor{green!20}2.19  \\
 &  &   \textit{\textbf{Hits@1}}   &   \textbf{0.91} & 0.80 & 0.85   &   0.79 & \textbf{0.99} & 0.89   &   \textbf{0.91} & \textbf{0.99} & \textbf{0.95}   &   \cellcolor{green!20}0.00 & \cellcolor{green!20}23.75 & \cellcolor{green!20}11.11  \\
 &  &   \textit{\textbf{ppl}}   &   \textbf{11.68} & 56.24 & 33.96   &   122.00 & 0.00 & 66.54   &   14.35 & 13.21 & \textbf{13.78}   &   \cellcolor{red!20}22.86 & \cellcolor{green!20}-76.51 & \cellcolor{green!20}-59.42  \\
 &  &   \textit{\textbf{RougeL}}   &   \textbf{15.15} & 14.26 & 14.71   &   14.33 & 0.00 & 16.77   &   15.07 & 18.92 & \textbf{17.00}   &   \cellcolor{red!20}-0.53 & \cellcolor{green!20}32.68 & \cellcolor{green!20}15.57  \\
\noalign{\smallskip}\hline\noalign{\smallskip}

\multicolumn{2}{c|}{\multirow{4}{*}{fr}}   &    \textit{\textbf{bert-f1}}   &   \textbf{0.67} & 0.68 & 0.68   &   0.65 & \textbf{0.70} & 0.68   &   \textbf{0.67} & \textbf{0.70} & \textbf{0.69}   &   \cellcolor{green!20}0.00 & \cellcolor{green!20}2.94 & \cellcolor{green!20}1.48  \\
 &  &   \textit{\textbf{Hits@1}}   &   \textbf{0.92} & 0.88 & 0.90   &   0.78 & \textbf{0.99} & 0.89   &   0.90 & \textbf{0.99} & \textbf{0.95}   &   \cellcolor{red!20}-2.17 & \cellcolor{green!20}12.50 & \cellcolor{green!20}5.00  \\
 &  &   \textit{\textbf{ppl}}   &   \textbf{7.04} & 96.40 & 51.72   &   136.70 & 7.14 & 71.92   &   7.17 & \textbf{3.66} & \textbf{5.42}   &   \cellcolor{red!20}1.85 & \cellcolor{green!20}-96.20 & \cellcolor{green!20}-89.53  \\
 &  &   \textit{\textbf{RougeL}}   &   11.59 & 12.41 & 12.00   &   9.54 & 14.07 & 11.80   &   \textbf{12.21} & \textbf{15.88} & \textbf{14.05}   &   \cellcolor{green!20}5.35 & \cellcolor{green!20}27.96 & \cellcolor{green!20}17.04  \\
\noalign{\smallskip}\hline\noalign{\smallskip}

\multicolumn{2}{c|}{\multirow{4}{*}{it}}   &    \textit{\textbf{bert-f1}}   &   \textbf{0.66} & 0.66 & 0.66   &   0.65 & 0.67 & 0.66   &   \textbf{0.66} & \textbf{0.69} & \textbf{0.68}   &   \cellcolor{green!20}0.00 & \cellcolor{green!20}4.55 & \cellcolor{green!20}2.27  \\
 &  &   \textit{\textbf{Hits@1}}   &   0.89 & 0.82 & 0.85   &   0.75 & \textbf{0.99} & 0.87   &   \textbf{0.90} & \textbf{0.99} & \textbf{0.95}   &   \cellcolor{green!20}1.12 & \cellcolor{green!20}20.73 & \cellcolor{green!20}10.53  \\
 &  &   \textit{\textbf{ppl}}   &   18.77 & 14.24 & 16.50   &   126.90 & 5.04 & 65.97   &   \textbf{5.75} & \textbf{4.98} & \textbf{5.37}   &   \cellcolor{green!20}-69.37 & \cellcolor{green!20}-65.03 & \cellcolor{green!20}-67.49  \\
 &  &   \textit{\textbf{RougeL}}   &   8.96 & 10.35 & 9.66   &   7.75 & 10.32 & 9.04   &   \textbf{9.10} & \textbf{12.96} & \textbf{11.03}   &   \cellcolor{green!20}1.56 & \cellcolor{green!20}25.22 & \cellcolor{green!20}14.24  \\
\noalign{\smallskip}\hline\noalign{\smallskip}

\multicolumn{2}{c|}{\multirow{4}{*}{id}}   &    \textit{\textbf{bert-f1}}   &   \textbf{0.70} & 0.70 & 0.70   &   \textbf{0.70} & \textbf{0.73} & \textbf{0.72}   &   \textbf{0.70} & \textbf{0.73} & \textbf{0.72}   &   \cellcolor{green!20}0.00 & \cellcolor{green!20}4.29 & \cellcolor{green!20}2.14  \\
 &  &   \textit{\textbf{Hits@1}}   &   \textbf{0.89} & 0.91 & 0.90   &   0.80 & \textbf{0.99} & 0.90   &   \textbf{0.89} & \textbf{0.99} & \textbf{0.94}   &   \cellcolor{green!20}0.00 & \cellcolor{green!20}8.79 & \cellcolor{green!20}4.44  \\
 &  &   \textit{\textbf{ppl}}   &   \textbf{42.36} & 74.25 & 58.30   &   250.60 & \textbf{6.81} & 128.70   &   48.40 & 9.02 & \textbf{28.71}   &   \cellcolor{red!20}14.26 & \cellcolor{green!20}-87.85 & \cellcolor{green!20}-50.76  \\
 &  &   \textit{\textbf{RougeL}}   &   12.95 & 13.39 & 13.17   &   11.54 & \textbf{19.63} & 15.58   &   \textbf{13.13} & 19.14 & \textbf{16.14}   &   \cellcolor{green!20}1.39 & \cellcolor{green!20}42.94 & \cellcolor{green!20}22.51  \\
\noalign{\smallskip}\hline\noalign{\smallskip}

\multicolumn{2}{c|}{\multirow{4}{*}{ko}}   &    \textit{\textbf{bert-f1}}   &   0.59 & 0.57 & 0.58   &   \textbf{0.66} & \textbf{0.67} & \textbf{0.67}   &   0.63 & 0.60 & 0.61   &   \cellcolor{green!20}6.78 & \cellcolor{green!20}5.26 & \cellcolor{green!20}6.03  \\
 &  &   \textit{\textbf{Hits@1}}   &   \textbf{0.85} & 0.90 & 0.88   &   0.76 & \textbf{0.96} & 0.86   &   \textbf{0.85} & \textbf{0.96} & \textbf{0.91}   &   \cellcolor{green!20}0.00 & \cellcolor{green!20}6.67 & \cellcolor{green!20}3.43  \\
 &  &   \textit{\textbf{ppl}}   &   \textbf{4.18} & 6.26 & 5.22   &   6.86 & \textbf{2.26} & 4.56   &   5.05 & 2.56 & \textbf{3.81}   &   \cellcolor{red!20}20.81 & \cellcolor{green!20}-59.11 & \cellcolor{green!20}-27.11  \\
 &  &   \textit{\textbf{RougeL}}   &   3.89 & 4.15 & 4.02   &   \textbf{6.92} & \textbf{9.23} & \textbf{8.08}   &   6.23 & 4.76 & 5.50   &   \cellcolor{green!20}60.15 & \cellcolor{green!20}14.70 & \cellcolor{green!20}36.69  \\
\noalign{\smallskip}\hline\noalign{\smallskip}
        \end{tabular}

} 
    \caption{Automatic Evaluation of Finetuned BLOOM on MOUD with and without CG (MOUD-CG/M-CG) and XPersona in different Languages. In \textbf{bold} are the best average scores per metric across resulting models. \colorbox{green!20}{Green cells} represent the gain in \% over XPersona trained models while \colorbox{red!20}{red cells} what has been lost.} 
    \label{tab:models-automatic-evaluations}
    
\end{table*}

\section{Baseline Experiments with MOUD}

We conduct our experiments using the smallest variant of BLOOM~\cite{workshop2023bloom176bparameteropenaccessmultilingual}, the 560M parameter model\footnote{\url{https://huggingface.co/bigscience/bloom-560m}}. The model is fine-tuned on a multitask objective, detailed in Appendix~\ref{appendix:fine-tuning-details}, and evaluated using automatic metrics, including \textbf{BertScore (Bert-F1, $\uparrow$)}, \textbf{Hits@1} ($\uparrow$), \textbf{Perplexity (PPL, $\downarrow$)}, and \textbf{Rouge-L} ($\uparrow$), with further explanations provided in Appendix~\ref{appendix:models-evaluation-details}.  
Given the one-to-many nature of ODD, automatic metrics may not fully capture conversational quality. However, they still offer valuable insights into performance.

As shown in Table~\ref{tab:models-automatic-evaluations}, models trained on MOUD often achieve better average performance across most metrics with some exceptions. However, similar to models trained on XPersona, they exhibit significantly higher perplexity on other dataset test set. This highlights the distinct nature of MOUD compared to PersonaChat and XPersona, reinforcing its value as a complementary resource.
Notably, when training on a combination of both datasets—maintaining the same total size as the XPersona training set by balancing the samples equally (50\% from each) and shuffling them during training—we observe substantial improvements across languages and metrics compared to models trained solely on XPersona. While a few exceptions exist where the performance drop is minimal, the overall trend highlights the complementary contribution of MOUD to existing datasets like XPersona. This further underscores its potential for enhancing multilingual conversational models and suggests promising directions for future research, particularly with specialized architectures tailored to its unique characteristics.

\section{Conclusion}  
In this study, we addressed two key dimensions of \textbf{Openness} in Open-Domain Dialogue: cultural openness, achieved through multilingualism and $l_T$ specificity, and \textit{ODP}, which we enhanced by integrating CG with a diverse range of SE types in the generated data.  
We evaluated four medium-sized, open-source LLMs, with LLaMA3.1-8B-Instruct consistently outperforming the others across multiple criteria according to both human and LLM assessments. It excelled not only in taxonomy relevance—particularly in effectively incorporating SEs within CG—but also in $l_T$ specificity, fluency, and overall humanness. This led to its selection as the model for generating the final MOUD dataset, an \textbf{O3}DD dataset, where \textbf{O3} represents \textbf{O}pen in language and culture, \textbf{O}pen in Speech-Event diversity, and \textbf{O}pen-Domain dialogue.  

Baseline automatic evaluations on shallow fine-tuned models highlight MOUD’s potential for advancing multilingual ODD systems. Models trained on MOUD exhibit distinct characteristics compared to those trained on XPersona, reinforcing its complementary value. Furthermore, models trained on a combination of both datasets—while maintaining the same overall training size—demonstrate improved performance over XPersona-trained models. This suggests that MOUD not only enhances diversity in dialogue modeling but also holds promise for further improvements, particularly with more specialized model's architectures.





\section{Limitations}

Since our evaluations on all the languages are performed using the LLM-as-Judge process, it may not be as relevant as evaluations performed by humans.
However, due to the high cost of human evaluations, we did not collect enough results for all the languages. Yet we report  results for the languages with a decent amount of evaluations across the models in Table~\ref{tab:personas-human}, Table~\ref{tab:common-ground-human} and Table~\ref{tab:conversation-human}.
Furthermore, the overall pipeline depend on the availability of rather high quality open-source multilingual instruction-tuned LLMs. And even assuming the existence of such models, the compute resource still comes at some costs, preventing some research from being replicated or augmented.



\section*{Acknowledgments}

This work was supported by the $\mu$DialBot project funded by the French National Research Agency (\textit{Agence Nationale de Recherche, ANR}) under the grant \texttt{ANR-20-CE33-0008} and benefited from computational resources provided by the Jean Zay supercomputer under the dossier \texttt{AD011013966R1}. We also extend our gratitude the evaluators who volunteered during the evaluation process of the generated data. 

\bibliography{custom}

\appendix

\section{Statistics}  

MOUD consist in open-source-LLM-based generated ODD in \textbf{29} languages (the list is provided in Table~\ref{tab:languages-list}) ranging from high-resource to very low resource.  %
As such, to the best of our knowledge, there is no ODD dataset with such a range of languages as shown by comparison with similar ODD datasets or approaches shown in Table~\ref{tab:comparison-with-others}.


\begin{table}[h]
    \centering
    \resizebox{0.5\textwidth}{!}{
        \begin{tabular} {ll|ccc|c|c|c|c}
            \hline
            \noalign{\smallskip}\multicolumn{2}{c|}{\textbf{Languages}} & \textbf{Code} & \textbf{Pop. (M)} & \textbf{\% in CC} & \textbf{\#Dial.\footnotemark} & \textbf{\#Utt.} & \textbf{Avg. \#Utt.} & \textbf{Avg \#words}\\
                \noalign{\smallskip}\hline\noalign{\smallskip}
\multirow{ 12 }{*}{\rotatebox{90}{\textbf{High-Resource}}}
 & English & en & 1132 & 44.5 & 21296 & 298699 & 14.03 & 18.19  \\
 & Russian & ru & 258 & 5.95 & 18799 & 262490 & 13.96 & 13.75  \\
 & German & de & 135 & 5.26 & 18353 & 257331 & 14.02 & 17.88  \\
 & Japanese & jp & 126 & 5.16 & 22738 & 318267 & 14.00 & 14.18  \\
 & Spanish & es & 595 & 4.59 & 18984 & 265315 & 13.98 & 18.19  \\
 & Chinese & zh & 1100 & 4.42 & 23811 & 333020 & 13.99 & 13.00  \\
 & French & fr & 321 & 4.31 & 18596 & 259554 & 13.96 & 20.84  \\
 & Italian & it & 85 & 2.61 & 17867 & 249600 & 13.97 & 18.44  \\
 & Dutch & nl & 28 & 1.91 & 20030 & 280712 & 14.01 & 17.86  \\
 & Portuguese & pt & 274 & 1.95 & 18966 & 264364 & 13.94 & 17.16  \\
 & Polish & pl & 50 & 1.76 & 13650 & 191263 & 14.01 & 13.48  \\
 & Turkish & tr & 88 & 1.06 & 20890 & 292683 & 14.01 & 10.75  \\
\noalign{\smallskip}\hline\noalign{\smallskip}

\multirow{14}{*}{\rotatebox{90}{\textbf{Medium-Resource}}}
 & Vietnamese & vi & 86 & 0.98 & 13325 & 187144 & 14.04 & 15.16  \\
 & Indonesian & id & 199 & 0.92 & 21519 & 300147 & 13.95 & 16.93  \\
 & Korean & ko & 82 & 0.69 & 18438 & 257939 & 13.99 & 8.52  \\
 & Swedish & sv & 10 & 0.65 & 13149 & 183980 & 13.99 & 17.46  \\
 & Arabic & ar & 375 & 0.62 & 19692 & 275816 & 14.01 & 11.75  \\
 & Hungarian & hu & 13 & 0.58 & 12103 & 169207 & 13.98 & 12.99  \\
 & Greek & el & 12 & 0.56 & 14051 & 196935 & 14.02 & 12.93  \\
 & Ukrainian & uk & 41 & 0.54 & 12896 & 181270 & 14.06 & 12.17  \\
 & Danish & da & 6 & 0.43 & 11983 & 167252 & 13.96 & 18.56  \\
 & Thai & th & 70 & 0.41 & 12827 & 180116 & 14.04 & 11.12  \\
 & Finnish & fi & 6 & 0.36 & 11105 & 156546 & 14.10 & 11.21  \\
 & Croatian & hr & 5.6 & 0.21 & 11511 & 161547 & 14.03 & 15.02  \\
 & Hindi & hi & 600 & 0.19 & 35905 & 502468 & 13.99 & 29.86  \\
\noalign{\smallskip}\hline\noalign{\smallskip}

\multirow{3}{*}{\rotatebox{90}{\textbf{Low-R.}}}
 & Bengali & bn & 270 & 0.11 & 21505 & 300271 & 13.96 & 25.75  \\
 & Afrikaans & af & 17 & 0.009 & 11247 & 157212 & 13.98 & 19.18  \\
 & Swahili & sw & 200 & 0.008 & 10167 & 142400 & 14.01 & 16.43  \\
 & Yoruba & yo & 45 & 0.0008 & 8182 & 113990 & 13.93 & 26.62  \\\noalign{\smallskip}\hline\noalign{\smallskip}
\end{tabular}
    }
    \caption{Detailed list of languages included and their number of conversations in the current version of MOUD. Their order and groups are determined by their percentage in Common Crawl with High-resource being $\geq 1 \%$, Medium-resource $\geq 0.1\%$ and Low-resource for the rest.}
    \label{tab:languages-list}
\end{table}

\footnotetext{Given access to an open source LLM, one can repeat the generation process described in Section~\ref{sec:moud-generation} and illustrated in Figure~\ref{fig:CompletePipeline} to generate additional samples.}

\section{Details on Quality Evaluation}
\label{appendix:eval-details}
Both LLM-as-a-judge and human evaluation were performed on the same criteria rated from 1 to 5 with the following descriptions. The texts in \textcolor{blue}{blue} correspond to what was added when prompting \texttt{GPT4o} to perform the evaluations.

\noindent\textbf{For Personas:}

\noindent\textcolor{blue}{\#\#\# Input: Personas and Taxonomies}

\noindent\textcolor{blue}{(id: <persona\_id>)}

\noindent<profile\_1> \textcolor{blue}{(Taxonomy:} <category|entity>\textcolor{blue}{)} 

...

\noindent<profile\_5> \textcolor{blue}{(Taxonomy:} <category|entity>\textcolor{blue}{)} 

\noindent\textcolor{blue}{(id: <persona2\_id>)}

...

\noindent\textcolor{blue}{(id: <persona6\_id>)}

\noindent<profile\_1> \textcolor{blue}{(Taxonomy:} <category|entity>\textcolor{blue}{)} 

...

\noindent<profile\_5> \textcolor{blue}{(Taxonomy:} <category|entity>\textcolor{blue}{)} 

\noindent\textbf{Specificity}: How much are the persona's sentences specific to \{language\}, in terms of entities provided like names, cities, culture, activities and folk psychology in general?

\noindent1: Not at all
2: A little
3: Somewhat
4: Quite a bit 
5: A lot

\noindent\textbf{Fluency}: Judge the language quality of the persona's sentences. How is the language skills of the provider of this persona?

\noindent1: Terrible
2: Bad
3: Decent
4: Good
5: Very good

\noindent\textbf{Taxonomy relevancy}: How relevant is each persona's sentence to the taxonomy provided? Are all the personas sentences coherent together (no contradictory facts among the sentences in the same persona)? \textcolor{blue}{Evaluate both}.

\noindent1: Not at all
2: A little
3: Somewhat
4: Quite a bit 
5: A lot

\noindent\textbf{Toxicity}: How would you rate this personas is in terms of toxicity. Is there any harmful or offending content in the personas sentences? How much is it toxic ?

\noindent1: Extremely
2: Quite
3: A little
4: Not really
5: Harmless



\noindent\textcolor{blue}{\#\#\# Output: Return your evaluation in a dictionary with each persona id as key and a dictionary with your evaluations as value and do not explain:}


\vspace{10pt}
\noindent\textbf{For Common Grounds and Conversations:}

\noindent\textcolor{blue}{\#\#\# Input: Conversations}

\noindent\textcolor{blue}{(id: <conversation\_id>)}

\noindent\textcolor{blue}{\# Personas:}

\noindent Speaker 1:

\noindent<profile\_sentence\_1>

...

\noindent<profile\_sentence\_5>

\noindent Speaker 2:

\noindent<profile\_sentence\_1>

...

\noindent<profile\_sentence\_5>

\noindent\textcolor{blue}{\# Common Ground:} <speech\_event | taxonomy> 

\noindent<complete\_common\_ground\_text\_content>

\noindent\textcolor{blue}{\# Dialogue:}

\noindent Speaker 1: <message1>

\noindent Speaker 2: <message2>

\noindent Speaker 1: <message3>

\noindent Speaker 2: <message4>

...

\noindent\textcolor{blue}{(id: <conversation2\_id>)}

...

\noindent \textcolor{blue}{\#\#\# Evaluation:}

\noindent\textcolor{blue}{\# Common ground evaluation:}

\noindent \textbf{Specificity}: How much is the common ground specific to {language}, in terms of entities provided like names, cities, culture, activities and folk psychology in general?

\noindent1: Not at all
2: A little
3: Somewhat
4: Quite a bit 
5: A lot

\noindent \textbf{Fluency}: Judge the language quality of the provided common ground, is it plausible? How is the language skills of the provider of this common ground?

\noindent 1: Terrible
2: Bad
3: Decent
4: Good
5: Very good

\noindent \textbf{Personas relevancy}: Is the common ground coherent with both speakers’ personas? Is it a context/joint activity that is likely to happen between the speakers?

\noindent1: Not at all
2: A little
3: Somewhat
4: Quite a bit 
5: A lot

\noindent\textbf{Speech event type relevancy}: Does the common ground take into account the type of talk provided in taxonomy above? How much would it allow that type of talk to happen between the speakers?

\noindent1: Not at all
2: A little
3: Somewhat
4: Quite a bit 
5: A lot

\noindent\textbf{Toxicity}: How would you rate this common ground in terms of toxicity. Is there any harmful or offending content in the personas sentences? How much is it toxic ?

\noindent1: Extremely
2: Quite
3: A little
4: Not really
5: Harmless

\noindent\textcolor{blue}{\# Dialogue evaluation:}

\noindent \textbf{Common ground relevancy}:  How consistent and faithful is the conversation to the common ground context provided and is the associated type of talk displayed in the conversation?

\noindent1: Not at all
2: A little
3: Somewhat
4: Quite a bit 
5: A lot

\noindent \textbf{Specificity}: How much is the conversation specific to the \{language\}, in terms of entity provided like names, cities, culture, and folk psychology in general?

\noindent1: Not at all
2: A little
3: Somewhat
4: Quite a bit 
5: A lot

\noindent \textbf{Humanness}: Do you think this conversation is from a model or human?

\noindent1:Definitely a model
2: Probably a model
3: Can be both but more human
4: Probably a human
5: Definitely a human

\noindent \textbf{Fluency}: Judge the language quality of the speakers in this conversation. Is what is said plausible? How would you rate their skills in \{language\}?

\noindent1: Terrible
2: Bad
3: Decent
4: Good
5: Very good

\noindent \textbf{Toxicity}: How would you rate this conversation is in terms of toxicity (harmful or offending content display)? How much is it toxic ?

\noindent1: Extremely
2: Quite
3: A little
4: Not really
5: Harmless

\noindent \textbf{Personas relevancy:} How consistent and faithful (no contradictory elements) is the conversation to the speakers' personas provided?

\noindent1: Not at all
2: A little
3: Somewhat
4: Quite a bit 
5: A lot

\noindent\textcolor{blue}{\#\#\# Output: Return your evaluation in a dictionary with each conversation id as key and two dictionaries for your "common\_ground" and "dialogue" evaluations and do not explain:}

For both evaluation batches, a typical OpenAI API system prompt was also added when sending request: \textcolor{blue}{"You are a smart evaluator, native \{language\} speaker, tasked to evaluate the quality of \{language\} \{data\_type\} on different aspects. You carefully read the criteria before giving your rating from 1 (worst) to 5 (best). The evaluated \{data\_type\} are in \{language\}, ensure you carefully pay attention to all details before making your rating decisions from grammar to content."} ~where \{language\} is replaced by the corresponding language and \{data\_type\} either  "personas" or "open domain conversations" depending on the part of the data we assessed.

In the detailed results tables below, the correspondence between each criterion and its abbreviation is as follows:  Specificity (S), Fluency (F), and Toxicity (Tx) appear in all tables; Relevance to Taxonomy for Personas (TR); Relevance to Speech Event Taxonomy for Common Grounds (T);  Relevance to Personas (P) is provided for Common Grounds and Conversations; Relevance to Common Ground and Taxonomy (CGT) and Humanness (H) are assessed specifically for Conversations.


\section{Human Evaluation}
\label{subsubsec:human-evaluation}


To support the LLM judgment we also performed human evaluation on the same set of data. As stated in Section~\ref{subsubsec:llm-as-a-judge}, finding voluntary evaluators for all the languages willing to assess large data batches is challenging. Nevertheless we gathered a decent amount of evaluations for some languages: \textbf{87} for Arabic (ar), \textbf{50} for French (fr), \textbf{34} for Spanish (es), \textbf{25} for Chinese (zh), and \textbf{23} for Vietnamese (vi).

\begin{table}[h]
        \centering
        \resizebox{\columnwidth}{!}{
            \begin{tabular} {l|c c c c c|c}
                    \hline\noalign{\smallskip}
                    \multirow{2}{*}{\textbf{Models}}     & \multicolumn{5}{c|}{\textbf{Languages}} & \multirow{2}{*}{\textbf{Total}} \\
                                               & es & zh & fr &  vi & ar &  \\
                    \noalign{\smallskip}\hline\noalign{\smallskip}
                    gemma-1.1-7b-it            & 12 & 7  & 17 &  9  & 31 & 76  \\
                    llama-3.1-8b-instruct      & 7  & 8  & 17 &  7  & 36 & 75  \\
                    mistral-7b-instruct-v0.3   & 15 & 10 & 16 &  7  & 20 & 68  \\
                    \noalign{\smallskip}\hline\noalign{\smallskip}
                    \textbf{Total}            & 34 & 25 & 50 &  23 & 87 & 219 \\
                    \noalign{\smallskip}\hline
            \end{tabular}
        }
        \caption{Number of Human Evaluated Conversations and CGs ($\times$ 2 for Personas counts) per Language and Model}
        \label{tab:lm-conv-stats}
\end{table}



We describe our pool of evaluators in Appendix~\ref{subappendix:evaluators-demographics} below and present the results compared to LLM judgments in Appendix~\ref{subappendix:human-eval-results}.

\subsection{Evaluators' Demographics Description}
\label{subappendix:evaluators-demographics}
Evaluators were \textbf{voluntary} participants recruited via various online channels (mailing lists, LinkedIn, direct contact, etc.). Participants were prompted to enroll only for \textbf{their native} language(s), even if fluent in others. Below is a demographic summary of our evaluators pool, based on a survey completed upon their first login on the evaluation platform presented in Appendix~\ref{appendix:evaluation-platform}. A total of \textbf{30} evaluators with 97\% at PhD or Grad education level, mostly (87\%) employed or student; 53\% directly contacted by us, 70\% female and 67\% aged between 18 and 29 (given education level more middle to late 20s).

\begin{table}[h]
        \centering
        \resizebox{0.3\textwidth}{!}{
                \begin{tabular} {l|c c c c c|c}
                    \hline\noalign{\smallskip}
                    \multirow{2}{*}{\textbf{Age}}     & \multicolumn{5}{c|}{\textbf{Languages}} & \multirow{2}{*}{\textbf{Total}} \\
                                   & es & zh & fr &  vi & ar &  \\
                    \noalign{\smallskip}\hline\noalign{\smallskip}
                        Under 18    & 0  & 0  & 0  & 0  & 1  & 1   \\
                        18 - 29     & 0  & 2  & 4  & 2  & 12 & 20  \\
                        30 - 49     & 2  & 1  & 0  & 0  & 1  & 4   \\
                        50 +        & 2  & 0  & 2  & 0  & 0  & 4   \\
                        Other       & 1  & 0  & 0  & 0  & 0  & 1   \\
                        \noalign{\smallskip}\hline\noalign{\smallskip}
                        \textbf{Total}       & 5  & 3  & 6  & 2  & 14 & 30  \\
                        \noalign{\smallskip}\hline
                \end{tabular}
        }
        \caption{Human Evaluators Age}
        \label{tab:age-user-stats}
\end{table}


\begin{table}[h]
        \centering
        \resizebox{0.3\textwidth}{!}{
                \begin{tabular} {l|c c c c c|c}
                    \hline\noalign{\smallskip}
                    \multirow{2}{*}{\textbf{Gender}} & \multicolumn{5}{c|}{\textbf{Languages}} & \multirow{2}{*}{\textbf{Total}} \\
                               & es & zh & fr &  vi & ar &  \\
                    \noalign{\smallskip}\hline\noalign{\smallskip}
                        Female & 2  & 1  & 4  & 0  & 14 & 21  \\
                        Male   & 2  & 2  & 1  & 2  & 0  & 7  \\
                        Other  & 1  & 0  & 1  & 0  & 0  & 2   \\
                        \noalign{\smallskip}\hline\noalign{\smallskip}
                        \textbf{Total}  & 5  & 3  & 6  & 2  & 14 & 30  \\
                        \noalign{\smallskip}\hline
                \end{tabular}
        }
        \caption{Human Evaluators Gender}
        \label{tab:gender-user-stats}
\end{table}


\begin{table}[h]
        \centering
        \resizebox{0.3\textwidth}{!}{
                \begin{tabular} {l|c c c c c|c}
                    \hline\noalign{\smallskip}
                    \multirow{2}{*}{\makecell{\textbf{Education} \\ \textbf{Level}}} & \multicolumn{5}{c|}{\textbf{Languages}} & \multirow{2}{*}{\textbf{Total}} \\
                                       & es & zh & fr &  vi & ar &  \\
                    \noalign{\smallskip}\hline\noalign{\smallskip}
                        Grad           & 1  & 2  & 6  & 1  & 13 & 23  \\
                        PhD            & 3  & 1  & 0  & 1  & 1  & 6   \\
                        Other          & 1  & 0  & 0  & 0  & 0  & 1   \\
                        \noalign{\smallskip}\hline\noalign{\smallskip}
                        \textbf{Total} & 5  & 3  & 6  & 2  & 14 & 30  \\
                        \noalign{\smallskip}\hline
                \end{tabular}
        }
        \caption{Human Evaluators Education Level}
        \label{tab:education-user-stats}
\end{table}


\begin{table}[h!]
        \centering
        \resizebox{0.3\textwidth}{!}{
                \begin{tabular} {l|c c c c c|c}
                    \hline\noalign{\smallskip}
                    \multirow{2}{*}{\makecell{\textbf{Employment} \\ \textbf{Status}}} & \multicolumn{5}{c|}{\textbf{Languages}} & \multirow{2}{*}{\textbf{Total}} \\
                                        & es & zh & fr &  vi & ar &  \\
                    \noalign{\smallskip}\hline\noalign{\smallskip}
                        Employed        & 3  & 0  & 1  & 1  & 1  & 6   \\
                        Unemployed      & 0  & 0  & 2  & 0  & 0  & 2   \\
                        Student         & 1  & 3  & 3  & 1  & 12 & 20  \\
                        Other           & 1  & 0  & 0  & 0  & 1  & 2   \\
                        \noalign{\smallskip}\hline\noalign{\smallskip}
                        \textbf{Total}  & 5  & 3  & 6  & 2  & 14 & 30  \\
                        \noalign{\smallskip}\hline
                \end{tabular}
        }
        \caption{Human Evaluators Employment Status}
        \label{tab:employment-user-stats}
\end{table}


\begin{table}[h!]
        \centering
        \resizebox{0.3\textwidth}{!}{
                \begin{tabular} {l|c c c c c|c}
                    \hline\noalign{\smallskip}
                    \multirow{2}{*}{\makecell{\textbf{Recruiting} \\ \textbf{Channel}}} & \multicolumn{5}{c|}{\textbf{Languages}} & \multirow{2}{*}{\textbf{Total}} \\
                                        & es & zh & fr &  vi & ar &  \\
                    \noalign{\smallskip}\hline\noalign{\smallskip}
                        Authors         & 4  & 1  & 4  & 1  & 6  & 16  \\
                        LinkedIn        & 0  & 2  & 0  & 0  & 2  & 4   \\
                        Mailing         & 0  & 0  & 0  & 1  & 0  & 1   \\
                        Referral        & 0  & 0  & 2  & 0  & 6  & 8   \\
                        Other           & 1  & 0  & 0  & 0  & 0  & 1   \\
                        \noalign{\smallskip}\hline\noalign{\smallskip}
                        \textbf{Total}  & 5  & 3  & 6  & 2  & 14 & 30  \\
                        \noalign{\smallskip}\hline
                \end{tabular}
        }
        \caption{Human Evaluators Recruiting Channel}
        \label{tab:find-user-stats}
\end{table}

\subsection{LLM to Human Correlation and Inter-Annotators Agreement (IAA)}

\begin{table}[h]
\centering
    \resizebox{0.4\textwidth}{!}{
        \begin{tabular} {c|cc|cc|cc}
            \hline
            \noalign{\smallskip}
            \multirow{3}{*}{\makecell{\rotatebox{-90}{\textbf{Criteria}}}} & \multicolumn{6}{c}{\textbf{Measures}} \\
            \noalign{\smallskip}\cline{2-7}
            \noalign{\smallskip}
            & \multicolumn{2}{c|}{{\thead{Pearson}}} & \multicolumn{2}{c|}{{\thead{Spearman}}} & \multicolumn{2}{c}{{\thead{Kendall}}} \\
            \noalign{\smallskip}\cline{2-7}
            \noalign{\smallskip}
            & $r$ & $p_r$ & $\rho$ & $p_{\rho}$ & $\tau$ & $p_{\tau}$  \\
            \noalign{\smallskip}\hline
            \noalign{\smallskip}
            \textbf{S} & 0.205 & 5.40e-06 & 0.227 & 4.90e-07 & 0.200 & 7.10e-07 \\
            \textbf{F} & 0.330 & 1.00e-13 & 0.297 & 2.70e-11 & 0.254 & 5.50e-11 \\
            \textbf{TR} & 0.395 & 1.80e-19 & 0.366 & 1.10e-16 & 0.317 & 1.50e-16 \\
            \textbf{Tx} & 0.066 & 1.50e-01 & 0.047 & 3.00e-01 & 0.047 & 3.00e-01 \\
            \noalign{\smallskip}\hline
        \end{tabular}
    }
    \caption{Correlation for Personas between Human Annotations and LLM Judgments}
    \label{tab:human-llm-personas-corr}
\end{table}

\begin{table}[h]
\centering
    \resizebox{0.4\textwidth}{!}{
        \begin{tabular} {c|cc|cc|cc}
            \hline
            \noalign{\smallskip}
            \multirow{3}{*}{\makecell{\rotatebox{-90}{\textbf{Criteria}}}} & \multicolumn{6}{c}{\textbf{Measures}} \\
            \noalign{\smallskip}\cline{2-7}
            \noalign{\smallskip}
            & \multicolumn{2}{c|}{{\thead{Pearson}}} & \multicolumn{2}{c|}{{\thead{Spearman}}} & \multicolumn{2}{c}{{\thead{Kendall}}} \\
            \noalign{\smallskip}\cline{2-7}
            \noalign{\smallskip}
            & $r$ & $p_r$ & $\rho$ & $p_{\rho}$ & $\tau$ & $p_{\tau}$  \\
            \noalign{\smallskip}\hline
            \noalign{\smallskip}
            \textbf{S} & 0.381 & 9.40e-10 & 0.391 & 3.20e-10 & 0.340 & 7.10e-10 \\
            \textbf{F} & 0.219 & 6.20e-04 & 0.241 & 1.60e-04 & 0.206 & 2.00e-04 \\
            \textbf{Tx$^*$} & \textit{NaN} & \textit{NaN} & \textit{NaN} & \textit{NaN} & \textit{NaN} & \textit{NaN} \\
            \textbf{P} & 0.163 & 1.10e-02 & 0.139 & 3.10e-02 & 0.120 & 3.10e-02 \\
            \textbf{T} & 0.276 & 1.30e-05 & 0.271 & 2.00e-05 & 0.243 & 1.70e-05 \\
            \noalign{\smallskip}\hline
        \end{tabular}
    }
    \caption{Correlation for Common Grounds between Human Annotations and LLM Judgments}
    \label{tab:human-llm-cg-corr}
\end{table}

\begin{table}[h!]
\centering
    \resizebox{0.4\textwidth}{!}{
        \begin{tabular} {c|cc|cc|cc}
            \hline
            \noalign{\smallskip}
            \multirow{3}{*}{\makecell{\rotatebox{-90}{\textbf{Criteria}}}} & \multicolumn{6}{c}{\textbf{Measures}} \\
            \noalign{\smallskip}\cline{2-7}
            \noalign{\smallskip}
            & \multicolumn{2}{c|}{{\thead{Pearson}}} & \multicolumn{2}{c|}{{\thead{Spearman}}} & \multicolumn{2}{c}{{\thead{Kendall}}} \\
            \noalign{\smallskip}\cline{2-7}
            \noalign{\smallskip}
            & $r$ & $p_r$ & $\rho$ & $p_{\rho}$ & $\tau$ & $p_{\tau}$  \\
            \noalign{\smallskip}\hline
            \noalign{\smallskip}
            \textbf{S} & 0.438 & 1.10e-12 & 0.459 & 6.20e-14 & 0.392 & 5.30e-13 \\
            \textbf{F} & 0.245 & 1.20e-04 & 0.250 & 8.50e-05 & 0.209 & 9.90e-05 \\
            \textbf{H} & 0.354 & 1.60e-08 & 0.351 & 2.10e-08 & 0.295 & 4.50e-08 \\
            \textbf{Tx$^*$} & \textit{NaN} & \textit{NaN} & \textit{NaN} & \textit{NaN} & \textit{NaN} & \textit{NaN} \\
            \textbf{P} & 0.276 & 1.30e-05 & 0.256 & 5.90e-05 & 0.217 & 6.10e-05 \\
            \textbf{CGT} & 0.212 & 9.40e-04 & 0.235 & 2.40e-04 & 0.197 & 2.80e+00 \\
            \noalign{\smallskip}\hline
        \end{tabular}
    }
    \caption{Correlation for Conversations between Human Annotations and LLM Judgments}
    \label{tab:human-llm-conversations-corr}
\end{table}

Low to moderate correlations are observed, yet all are \textbf{highly statistically significant}.

$^*$ The toxicity correlations in Table~\ref{tab:human-llm-cg-corr} and Table~\ref{tab:human-llm-conversations-corr} are reported as \textit{NaN} because the LLM consistently rated the toxicity of CG and Conversations as \textbf{5} (not toxic at all) across all human-evaluated conversations. This consistent score resulted in a standard deviation of zero, making correlation computation impossible. This observation aligns with human evaluators’ average toxicity ratings, which were similarly high: Tx $= 4.80$ with $\sigma = 0.59$ for CG and Tx $= 4.70$ with $\sigma = 0.74$ for Conversations.
Furthermore, the toxicity correlation for Personas in Table~\ref{tab:human-llm-personas-corr} appear to be the lowest and least significant. However, when looking into the average toxicity scores, they further confirm a general agreement on the absence of toxicity. Human evaluators rated Personas at Tx $= 4.83$ with $\sigma = 0.49$, while the LLM rated them at Tx $= 4.98$ with $\sigma = 0.19$.

Overall, these results indicate a shared assessment between human evaluators and the LLM, reinforcing the conclusion that the generated data is predominantly perceived as non-toxic.

\begin{table}[h!]
    \centering
    \resizebox{0.4\textwidth}{!}{
         \begin{tabular} {ll|ccc|ccc}
            \hline
            \noalign{\smallskip}
             \multicolumn{2}{c|}{\multirow{4}{*}{\makecell{\textbf{Lang.}}}} & \multicolumn{6}{c}{\textbf{Scales}} \\
              & &  \multicolumn{3}{c}{{\thead{5 ratings: 1,2,3,4,5}}} &  \multicolumn{3}{c}{\thead{Grouped: (1,2); (3,4) \& (5,)}}  \\
              \noalign{\smallskip}\cline{3-8}
              \noalign{\smallskip}
                &    &  P & CG & C &   P & CG & C  \\
                \noalign{\smallskip}\hline\noalign{\smallskip}
\multicolumn{2}{c|}{es}   &   0.171  & 0.110 & 0.119   &   0.317  & 0.147 & 0.202  \\
\multicolumn{2}{c|}{zh}   &   -0.049 & 0.110 & 0.053   &   -0.087 & 0.100 & 0.171  \\
\multicolumn{2}{c|}{fr}   &   0.005  & 0.031 & 0.113   &   0.012  & 0.051 & 0.174  \\
\multicolumn{2}{c|}{vi}   &   0.146  & 0.237 & 0.281   &   0.268  & 0.294 & 0.459  \\
\multicolumn{2}{c|}{ar}   &   0.209  & 0.216 & 0.185   &   0.310  & 0.330 & 0.319  \\
            
            \noalign{\smallskip}\hline\noalign{\smallskip}
        \end{tabular}
    }
    \caption{Cohen's $\kappa$ Inter-Annotator Agreement}
    \label{tab:human-llm-iaa}
\end{table}

The $\kappa$ values are relatively low but improve slightly when scores are grouped, as shown in the Table~\ref{tab:human-llm-iaa}. This grouping represents broader categories, such as bad, decent, and excellent, which help smooth minor differences between evaluators.

\subsection{Human Evaluation Detailed Results}
\label{subappendix:human-eval-results}
Despite the low $\kappa$ values presented in Table~\ref{tab:human-llm-iaa} (which can be attributed to some of its inherent limitations, such as its tendency to decrease with an increasing number of classes,~\citealp{10.1093/ptj/85.3.257}), \textbf{both humans and LLMs tend to rate conversations in the same direction} as shown by Table~\ref{tab:personas-human}, Table~\ref{tab:common-ground-human} and Table~\ref{tab:conversation-human}, which is the most critical aspect of alignment. This is supported by works such as~\citep{amidei-etal-2018-rethinking}, which argue that high IAA is not always desirable, and (\citet{chiang-lee-2023-large}, \citet{iskender-etal-2021-reliability}), which highlight that even between human experts, values can be low — \textit{a fortiori} when comparing humans to LLMs.

\section{Prompts Templates}
\label{appendix:prompts}

\subsection{Personas Generation}
\label{appendix:persona-prompt}

\begin{lstlisting}
### Instructions:
The aim is to create new examples similar to those 
provided bellow with respect to the following: 
1. Generate a character (persona) description using 
five short sentences as profile.
2. The profile SHOULD BE natural and descriptive.
3. The profile SHOULD BE a short sentence. Mostly
using the first person. For example: "I'm not a fan 
of something", "My preferred stuff is something".
4. The profile SHOULD contain typical topics of 
human interest that the described speaker can bring 
up in a conversation

### Constraints:
1. Each sentence in the persona should be in {lang}.
2. Generate persona that are coherent with the fact 
that it describe a {lang}-speaking person in terms 
of locations, names, culture etc.
3. Each sentence should be short with a maximum of 
15 words.
4. DO NOT TRANSLATE PROVIDED EXAMPLES NOR THE ONES 
YOU GENERATE.
5. DO NOT REPEAT A PATTERN, EACH NEW EXAMPLE 
SHOULD BE UNIQUE. BE CREATIVE.

Below are examples of the type of character descrip-
tions you should create:

Example <1>
<example_1_profile_sentence_1>
...
<example_1_profile_sentence_5>
...
Example <n>
<example_n_profile_sentence_1>
...
<example_n_profile_sentence_5>

Generate new {num_requested} varied examples respec-
ting the following taxonomy:
Example 1
- a sentence on <persona_taxonomy_entity_sentence_1>
...
- a sentence on <persona_taxonomy_entity_sentence_5>
\end{lstlisting}

Where $n$ is the number of demonstration examples fixed before generation, if $n=0$ all the part on examples is removed; \{lang\} is replaced by the target language at hand. 

\subsection{Common-grounds Generation}
\label{appendix:common-ground-prompt}

Here, we tasked the LLM to act as a narrator which tells the context of an ODD between two characters associated with two randomly selected personas. A speech event type is randomly selected and associated to the conversation and input in the prompt in place of \{speech\_event\}. \{language\} corresponds to the target language, \{category\} is the category of the speech event w.r.t the taxonomy (e.g. Informal/Superficial Talk) and \{speech\_event\_sentence$|$description\} are selected from the augmented taxonomy for the LLM to better understand the type of speech event. And finally, it is  forced to include a translation of the word "character" in the target languages,  \{translation\_of\_character\_in\_target\}, followed by 1 and 2  to clearly  specify the role of both speakers in the resulting conversation.

\begin{lstlisting}
### Input: Below are the personas of the only two 
characters that will conduct the conversations. 
Take it into account in the common-ground:

Character 1 persona:
<character_1_profile_sentence_1>
...
<character_1_profile_sentence_5>

Character 2 persona:
<character_2_profile_sentence_1>
...
<character_2_profile_sentence_5>

### Instructions:
You are a Narrator fluent in {language} that ex-
plains the context of a discussion between two 
charcaters described by their personas in Input. 
The context in {language} may include a topic, 
a situation, a subject to talk about, an object 
of interest and maybe environment description. 
The context should allow for an open-domain dia-
logue where {speech_event_sentence}

### Constraints:
1. The context and topics should be coherent with 
the personas in Input and suitable for an {category} 
talk especially {speech_event} i-e {speech_event
_description}
2. The context should be in {language} coherent with
the fact that the resulting conversation will be 
performed by {language}-speaking persons in terms
of locations, names, culture, folk psychology etc.
3. The context should be coherent with characters 
personas in Input.
4. Do not repeat the characters personas in Input 
instead create a context that is likely to happen 
between them.
5. Do not add or infer other characters than those 
described in the Input.
6. Adding names is restricted unless mentioned in
the characters personas in the Input.
7. The context is a short paragraph that ALWAYS 
mention "{translation_of_character_in_target} 1" 
and "{translation_of_character_in_target}  2"  
and the purpose of their chat.
8. Do not translate the context you provide.
9. The proposed context should be encapsulated in
a very short paragraph. 
10. Remember you are the narrator do not do the 
conversation between the characters, only return 
the context.

### Narrator:
\end{lstlisting}

\subsection{Conversations Generation}
\label{appendix:conversation-prompt}

Again, \{language\} is replaced by the target language, the type of speech event expected to be displayed in the conversation is specified in \{speech\_event\_type\} with its full taxonomy in \{speech\_event\_taxonomy\} and a sentence describing the role of the current speaker-LLM instance in the conversation, especially for asymmetric type of talk; this is provided in \{speech\_event\_sentence\_description\_with\_speakers \_role\}. Please refer to Appendix~\ref{appendix:speech-event-taxonomy} for details on speech event taxonomy, description and sentences. The common ground is provided in \{common\_ground\}, the speaker-LLM instance is reminded its role in the CG with  wit the translation of "character" provided as the CG is in target language. The number of total turns and the current turn number are also provided to help the speakers instance to have a conversation that should last accordingly. 

\begin{lstlisting}
### Instructions
You are a fluent {language} speaker. You do not mix
{language} with any other language when speaking as 
{language} is your native language. You read the 
prompt carefully and pay close attention to your 
character, your role in the conversation, its con-
text and the level of details required. You make 
sure you give factual and precise responses using 
correct grammar in {language}. 
You role play as the character described in the fol-
lowing lines. You always speak with short and simple
answers in {language}.


### Constraints:
1. You SHALL ALWAYS respond in {language}.
2. Your response should be coherent with the fact
you are a {language}-speaking person in terms of lo-
cations, names, culture, folk psychology etc.
3. You shall be creative.
4. You avoid copying 'Your Persona Information' 
exactly in your response. Use them creatively.
5. Your response should be a SHORT sentence with 
less than 15 words coherent with your persona and
the context provided below.
6. Always stay true to your character provided in
'Your Persona Information' below.
7. You should try as much as possible to have a 
{speech_event_type} talk especially {speech_event
_taxonomy} i-e a converation where {speech_event_
sentence_description_with_speakers_role}

YOUR Persona Information: how you describe Your-
self, Not the User!
<character_profile_sentence_1>
...
<character_profile_sentence_5>


The underlying CONTEXT of this discussion is:
{common_ground}. You are character ({translation_of
_character_in_target}) {1_or_2}.

[Complete the following conversation expected to 
last {num_turns} and you are at turn {current_turn}. 
Take this into account to respond with a SHORT and
PRECISE message in {language} as your character des
cribed above would. Do not repeat previous messages,
instead keep the conversation flow:  
# % if first  message % 
Start the conversation with a SHORT sentence in 
{language}:]

<formatted_chat_with_model_template>
\end{lstlisting}

\begin{lstlisting}
# for Gemma-1.1-7b-it, the chat was embedded in the 
# prompt as follows
Persona: <message1>
User: <message2>
Persona: <message3>
User: <message4>
Persona: 
\end{lstlisting}

\section{XPersona LLM judgments on the criteria}

XPersona~\cite{lin-etal-2021-xpersona} is one of the approaches to addressing multilingualism in persona-based ODD. It leverages MT to translate PersonaChat into 6 languages other than English, with additional revisions: rule-based (rules defined by human based on observations on a subset of the data) for the training set and human-based for the test and validation sets. We conducted a quality analysis of this dataset using LLM as a judge. Where applicable, the assessment utilized the same criteria as for MOUD, excluding taxonomy relevance and all common-ground-related metrics.

The results indicate that XPersona consistently received lower ratings compared to MOUD for the given language, with particularly low scores in \textbf{Specificity} and \textbf{Humaneness}. These aspects are crucial for fostering better multilingualism and cultural inclusivity, which are more effectively addressed by our proposed approach.

\begin{table}[h]
   \centering
   \resizebox{0.3\textwidth}{!}{
       \begin{tabular} {ll|c|ccc|c}
           \hline
           \noalign{\smallskip}
            \multicolumn{2}{c|}{\multirow{2}{*}{\makecell{\textbf{Lang.}}}} & \multirow{2}{*}{\makecell{\textbf{Revision} \\ \textbf{Type}}} & \multicolumn{4}{c}{\textbf{Criteria}} \\
             \cline{4-7}
              &  &  & S & F & Tx & Avg.\\
\noalign{\smallskip}\hline\noalign{\smallskip}
\hline\noalign{\smallskip}

\multicolumn{2}{c|}{en}   &   \textit{PersonaChat} & \textcolor{red}{\textbf{2.87}} & 3.93 & 4.95 & 3.91  \\
\noalign{\smallskip}\hline\noalign{\smallskip}
\hline\noalign{\smallskip}

\multicolumn{2}{c|}{\multirow{2}{*}{jp}}   &   \textit{Human} & 1.86 & 4.33 & 4.96 & 3.71  \\
& & \textit{Rules Based} & 1.92 & 4.50 & 4.94 & 3.79  \\
\noalign{\smallskip}\hline\noalign{\smallskip}

\multicolumn{2}{c|}{\multirow{2}{*}{zh}}   &   \textit{Human} & 1.36 & 4.58 & 4.97 & 3.64  \\
& & \textit{Rules Based} & 1.35 & 4.49 & 4.97 & 3.60  \\
\noalign{\smallskip}\hline\noalign{\smallskip}

\multicolumn{2}{c|}{\multirow{2}{*}{fr}}   &   \textit{Human} & 2.18 & 4.19 & 4.94 & 3.77  \\
& & \textit{Rules Based} & 2.10 & 4.06 & 4.97 & 3.71  \\
\noalign{\smallskip}\hline\noalign{\smallskip}

\multicolumn{2}{c|}{\multirow{2}{*}{it}}   &   \textit{Human} & 2.01 & 4.11 & 4.98 & 3.70  \\
& & \textit{Rules Based} & 2.05 & 4.19 & 4.95 & 3.73  \\
\noalign{\smallskip}\hline\noalign{\smallskip}

\multicolumn{2}{c|}{\multirow{2}{*}{id}}   &   \textit{Human} & 1.40 & 3.75 & 4.97 & 3.37  \\
& & \textit{Rules Based} & 1.45 & 3.65 & 4.94 & 3.35  \\
\noalign{\smallskip}\hline\noalign{\smallskip}

\multicolumn{2}{c|}{\multirow{2}{*}{ko}}   &   \textit{Human} & 1.86 & 4.44 & 4.92 & 3.74  \\
& & \textit{Rules Based} & 1.78 & 4.35 & 4.91 & 3.68  \\
\noalign{\smallskip}\hline\noalign{\smallskip}
\hline\noalign{\smallskip}

\multicolumn{2}{c|}{\multirow{2}{*}{Avg.}}   &  \textit{Human} & \textcolor{red}{\textbf{1.78}} & 4.23 & 4.96 & 3.66  \\
& &  \textit{Rule Based} & \textcolor{red}{\textbf{1.78}} & 4.21 & 4.94 & 3.64  \\

\noalign{\smallskip}\hline\noalign{\smallskip}

\end{tabular}
   }
   \caption{Detailed LLM Judgments of XPersona Conversations. Average over the evaluated conversations for each language is reported.}
  \label{tab:xpersona-personas-llm}
\end{table}

\begin{table}[h]
   \centering
   \resizebox{0.39\textwidth}{!}{
       \begin{tabular} {ll|c|ccccc|c}
           \hline
           \noalign{\smallskip}
            \multicolumn{2}{c|}{\multirow{2}{*}{\makecell{\textbf{Lang.}}}} & \multirow{2}{*}{\makecell{\textbf{Revision} \\ \textbf{Type}}} & \multicolumn{4}{c}{\textbf{Criteria}} \\
             \cline{4-9}
              &  &  & S & F & H & Tx & PR & Avg. \\
\noalign{\smallskip}\hline\noalign{\smallskip}
\hline\noalign{\smallskip}

\multicolumn{2}{c|}{en} & \textit{PersonaChat}  &  \textcolor{red}{\textbf{2.88}} & 3.6 & \textcolor{red}{\textbf{2.74}} & 4.9 & 4.39 & 3.70  \\
\noalign{\smallskip}\hline\noalign{\smallskip}
\hline\noalign{\smallskip}

\multicolumn{2}{c|}{\multirow{2}{*}{jp}} & \textit{Human}  &  1.97 & 3.57 & 2.54 & 4.88 & 4.06 & 3.40  \\
& & \textit{Rules Based}  &  2.15 & 3.06 & 2.21 & 4.93 & 3.78 & 3.23  \\
\noalign{\smallskip}\hline\noalign{\smallskip}

\multicolumn{2}{c|}{\multirow{2}{*}{zh}} & \textit{Human}  &  2.22 & 3.58 & 2.53 & 4.88 & 4.09 & 3.46  \\
& & \textit{Rules Based}  &  2.05 & 3.15 & 2.37 & 4.9 & 3.89 & 3.27  \\
\noalign{\smallskip}\hline\noalign{\smallskip}

\multicolumn{2}{c|}{\multirow{2}{*}{fr}} & \textit{Human}  &  2.18 & 3.11 & 2.34 & 4.88 & 4.12 & 3.33  \\
& & \textit{Rules Based}  &  1.94 & 3.07 & 2.39 & 4.89 & 4.02 & 3.26  \\
\noalign{\smallskip}\hline\noalign{\smallskip}

\multicolumn{2}{c|}{\multirow{2}{*}{it}} & \textit{Human}  &  2.27 & 3.32 & 2.42 & 4.95 & 4.21 & 3.43  \\
& & \textit{Rules Based}  &  2.07 & 2.99 & 2.18 & 4.88 & 3.94 & 3.21  \\
\noalign{\smallskip}\hline\noalign{\smallskip}

\multicolumn{2}{c|}{\multirow{2}{*}{id}} & \textit{Human}  &  2.04 & 3.57 & 2.68 & 4.92 & 4.25 & 3.49  \\
& & \textit{Rules Based}  &  2.07 & 3.36 & 2.57 & 4.95 & 4.16 & 3.42  \\
\noalign{\smallskip}\hline\noalign{\smallskip}

\multicolumn{2}{c|}{\multirow{2}{*}{ko}} & \textit{Human}  &  2.29 & 3.59 & 2.36 & 4.86 & 4.01 & 3.42  \\
& & \textit{Rules Based}  &  2.09 & 3.25 & 2.26 & 4.93 & 3.77 & 3.26  \\
\noalign{\smallskip}\hline\noalign{\smallskip}
\hline\noalign{\smallskip}

\multicolumn{2}{c|}{Avg.} & \textit{Human}  &  \textcolor{red}{\textbf{2.16}} & 3.46 & \textcolor{red}{\textbf{2.48}} & 4.89 & 4.12 & 3.42  \\
& & \textit{Rules Based}  &  \textcolor{red}{\textbf{2.06}} & 3.15 & \textcolor{red}{\textbf{2.33}} & 4.91 & 3.93 & 3.28  \\

\noalign{\smallskip}\hline\noalign{\smallskip}

\end{tabular}
   }
   \caption{Detailed LLM Judgments of XPersona Personas. Average over the evaluated personas for each language is reported.}
  \label{tab:xpersona-personas-llm}
\end{table}

\section{Details on Baselines Experiments with MOUD}

\subsection{Datasets}

We utilize the MOUD dataset as outlined in Table~\ref{tab:lm-conv-stats}, restricting our selection to languages present in the XPersona dataset. This allows for a direct comparison between MOUD-based models performance and those based on an existing related dataset. For each language, we retain 1,000 conversations for the validation set and another 1,000 for the test set. Detailed statistics for the training and evaluation splits for both datasets are provided in Table~\ref{tab:training-datasets}.

\begin{table}[h!]
    \centering
   \resizebox{\columnwidth}{!}{
       \begin{tabular} {ll|c|c|c||c|c}
           \hline
           \noalign{\smallskip}
            \multicolumn{2}{c|}{\multirow{2}{*}{\makecell{\textbf{Lang.}}}} & \multirow{2}{*}{\makecell{\textbf{Split}}} & 
            \multicolumn{2}{c||}{{\thead{XPersona}}} &  \multicolumn{2}{c}{\thead{MOUD}} \\
             \cline{4-7}
            &  &  &  \textbf{\#Dialogues}  &  \textbf{\#Utterances}  &  \textbf{\#Dialogues}  &  \textbf{\#Utterances}  \\
                \hline\noalign{\smallskip}
                
\multicolumn{2}{c|}{\multirow{3}{*}{en}}    &  
\textit{\textbf{Train}}             &       16878 & 248244       &          19296 & 270552   \\ 
& &  \textit{\textbf{Valid.}}       &        1000 & 14632        &           1000 & 14110    \\ 
& &  \textit{\textbf{Test}}         &        1000 & 15602        &          1000 & 14032    \\ 
\noalign{\smallskip}\hline\noalign{\smallskip}

\multicolumn{2}{c|}{\multirow{3}{*}{jp}}    &  
\textit{\textbf{Train}}             &     16878 & 248244      &           20738 & 289892   \\ 
& &  \textit{\textbf{Valid.}}       &       275 & 4278         &            1000 & 14228    \\ 
& &  \textit{\textbf{Test}}         &       275 & 4322         &            1000 & 14130    \\ 
\noalign{\smallskip}\hline\noalign{\smallskip}

\multicolumn{2}{c|}{\multirow{3}{*}{zh}}   &   
\textit{\textbf{Train}}             &     16878 & 248244      &          21811 & 305156  \\ 
& &  \textit{\textbf{Valid.}}       &       222 & 3440        &           1000 & 13812   \\ 
& &  \textit{\textbf{Test}}         &       222 & 3458        &           1000 & 14020   \\ 
\noalign{\smallskip}\hline\noalign{\smallskip}

\multicolumn{2}{c|}{\multirow{3}{*}{fr}}   &   
\textit{\textbf{Train}}             &     16878 & 248244         &          16596 & 231508   \\ 
& &  \textit{\textbf{Valid.}}       &       248 & 3868           &           1000 & 13942    \\ 
& &  \textit{\textbf{Test}}         &       249 & 3900           &           1000 & 14102    \\ 
\noalign{\smallskip}\hline\noalign{\smallskip}

\multicolumn{2}{c|}{\multirow{3}{*}{it}}   &   
\textit{\textbf{Train}}             &     16878 & 248244       &           15867 & 221824  \\ 
& &  \textit{\textbf{Valid.}}       &       140 & 2160         &            1000 & 13758   \\ 
& &  \textit{\textbf{Test}}         &       140 & 2192         &            1000 & 14016   \\ 
\noalign{\smallskip}\hline\noalign{\smallskip}

\multicolumn{2}{c|}{\multirow{3}{*}{id}}  &   
\textit{\textbf{Train}}             &     16878 & 248244          &           11325 & 159098   \\ 
& &  \textit{\textbf{Valid.}}       &       484 & 7562            &            1000 & 13958   \\ 
& &  \textit{\textbf{Test}}         &       484 & 7540            &            1000 & 14088   \\ 
\noalign{\smallskip}\hline\noalign{\smallskip}

\multicolumn{2}{c|}{\multirow{3}{*}{ko}}    &   
\textit{\textbf{Train}}             &     16878 & 248244         &          16438 & 229970   \\ 
& &  \textit{\textbf{Valid.}}       &       299 & 4684           &           1000 & 14018   \\ 
& &  \textit{\textbf{Test}}         &       300 & 4678           &           1000 & 13916   \\ 
\noalign{\smallskip}\hline\noalign{\smallskip}
        \end{tabular}

    } 
    \caption{Detailed Statistics of the Training Data} 
    \label{tab:training-datasets}
    
\end{table}

\subsection{Fine-Tuning Approach}
\label{appendix:fine-tuning-details}
\paragraph{Multitask Learning Setup} 
We fine-tuned \texttt{BLOOM-560M}\footnote{\url{https://huggingface.co/bigscience/bloom-560m}} on the two tasks of the PersonaChat~\cite{zhang-etal-2018-personalizing} dataset: (1) Next Utterance Generation using a Causal Language Modeling (CLM) head and (2) Next Utterance Classification using a Multi-Choice Classification (MC) head. 

Following the architecture proposed by~\citet{DBLP:journals/corr/abs-1901-08149}, which is not specifically designed for MOUD constraints such as common ground, speech-event variations, access to both speakers' personas (PersonaChat only provides the \textit{second} speaker’s persona, as does XPersona), and language-specific considerations, we establish baseline metrics using a simple model fine-tuned on this dataset. Future improvements are expected with alternative backbone models and dedicated architectures, facilitated by the dataset's release and community contributions.

\paragraph{Hyperparameter Configuration} 
The model was fine-tuned with a total \textbf{batch size of 32}, where each sequence block consists of a concatenation of persona, common ground (if applicable), dialogue history, and the reply. Each block contains \textbf{num\_candidates = 4} sequences: one with the golden response for CLM loss computation and three distractors for the MC head to learn correct response selection.

Training was performed for \textbf{1 epoch} using the AdamW optimizer with a linearly decayed learning rate of \textbf{6.25e-5}, \textbf{$\beta_1$ = 0.9}, \textbf{$\beta_2$ = 0.999}, an L2 weight decay of \textbf{0.01}, and a weighting factor of \textbf{2} for the CLM loss in the final objective function: 
\[
loss = 2 \times clm\_loss + mc\_loss.
\]  
During training, model performance was evaluated on the validation set every \textbf{10\%} of an epoch, with an evaluation delay of \textbf{2,000} steps. The best model checkpoint was selected based on perplexity on the validation set. Training times, including these validation intervals, vary by hardware: approximately \textbf{12 hours} on a single V100 GPU and under \textbf{5 hours} on an A100 GPU.

\subsection{Evaluation}
\label{appendix:models-evaluation-details}

\subsubsection{Metrics}
Evaluating ODD models remains challenging due to the inherent subjectivity of responses. While automatic metrics provide some indication of performance, they may not fully capture conversational quality. We employ the following metrics:  
\begin{itemize}
    \item \textbf{BERTScore-F1}: Measures semantic similarity between model outputs and references using contextual embeddings.
    \item \textbf{Hits@1}: A ranking-based metric specifically designed for PersonaChat, assessing whether the correct response is ranked highest in a set with \textbf{num\_candidates - 1} dummies.
\end{itemize} 

\newgeometry{top=0.5cm, bottom=1.5cm, left=2.5cm, right=2.5cm}
\begin{itemize}
    \item \textbf{Perplexity (PPL)}: Estimates the fluency of generated text, with lower values indicating better coherence.
    \item \textbf{Rouge-L}: Captures n-gram overlap, serving as an indicator of lexical similarity.
\end{itemize}

\subsubsection{Sampling-Based Decoding Strategy}
Text generation for Rouge-L and BertScore was performed using a single-beam search with sampling, applying a temperature of \textbf{0.7} and a nucleus sampling threshold of \textbf{0.9}. To reduce redundancy, a repetition penalty of \textbf{1.2} and an \textbf{n-gram constraint of size 4} were enforced. The model was configured to generate up to \textbf{250 new tokens}, with a minimum of \textbf{1 new token}. 

Prioritizing stochasticity over deterministic ranking, the absence of multiple beams enhances output diversity while maintaining coherence. As suggested by~\citet{DBLP:journals/corr/abs-1901-08149}, this approach may better align with human conversational experience, though we did not explicitly evaluate this aspect. Additionally, omitting top-k filtering allows the model to sample from a broader range of token probabilities, fostering more varied yet contextually relevant responses.

\begin{table*}[h]
    \resizebox{\textwidth}{!}{
\begin{tabular} {ll|cccc|c||cccc|c||cccc|c||cccc|c}
            \hline
            \noalign{\smallskip}
             \multicolumn{2}{c|}{\multirow{4}{*}{\makecell{\textbf{Lang.}}}} & \multicolumn{20}{c}{\textbf{Models}} \\
              & & \multicolumn{5}{c}{{\thead{Aya-23-8B}}} &  \multicolumn{5}{c}{{\thead{Gemma-1.1-7b}}} &  \multicolumn{5}{c}{\thead{LLaMA3.1-8B}} &   \multicolumn{5}{c}{{\thead{Mistral-7B}}}   \\
              \noalign{\smallskip}\cline{3-22}
              \noalign{\smallskip}
               &    &  S & F & TR & Tx & Avg &   S & F & TR & Tx & Avg.  &   S & F & TR & Tx & Avg.  &   S & F & TR & Tx & Avg.  \\

\noalign{\smallskip}\hline\noalign{\smallskip}
\multirow{ 13 }{*}{\rotatebox{90}{\textbf{High-Resource}}}
& en   &   2.96 & 4.79 & 4.33 & 4.99 & 4.27   &   3.23 & 4.87 & 3.87 & 4.99 & 4.24   &   \textbf{3.40} & 4.40 & 4.73 & 4.99 & 4.38   &   3.35 & \textbf{4.90} & \textbf{4.94} & \textbf{5.00} & \textbf{4.55}  \\
 & ru   &   2.40 & \textbf{4.74} & 3.53 & 4.98 & 3.91   &   2.60 & 3.76 & 3.05 & \textbf{4.99} & 3.60   &   \textbf{3.27} & 4.56 & 3.72 & 4.98 & 4.13   &   3.16 & 4.70 & \textbf{4.60} & 4.95 & \textbf{4.35}  \\
 & de   &   2.57 & \textbf{4.75} & 3.56 & 4.97 & 3.96   &   2.74 & 3.87 & 3.18 & \textbf{4.99} & 3.69   &   \textbf{3.45} & 4.43 & 3.98 & 4.98 & 4.21   &   3.04 & 4.42 & \textbf{4.64} & 4.98 & \textbf{4.27}  \\
 & jp   &   2.81 & \textbf{4.84} & \textbf{4.06} & 4.99 & \textbf{4.18}   &   2.66 & 4.12 & 3.74 & \textbf{5.00} & 3.88   &   \textbf{3.16} & 4.68 & 3.89 & 4.98 & 4.18   &   2.74 & 3.71 & 3.99 & 4.97 & 3.85  \\
 & es   &   2.54 & \textbf{4.93} & 4.16 & 4.98 & 4.15   &   2.81 & 4.53 & 3.28 & \textbf{4.99} & 3.90   &   \textbf{3.57} & 4.74 & 4.23 & 4.98 & 4.38   &   3.27 & 4.77 & \textbf{4.72} & 4.99 & \textbf{4.44}  \\
\end{tabular}
    } 
    \caption{Detailed Personas Evaluation with \texttt{GPT4o}-as-a-judge. Average over the \textit{300} evaluated personas for each model-language pair is reported. In bold, is the best rating among the models for each criterion.}
   \label{tab:personas-llm-as-judge}
\end{table*}

\begin{table*}[h]
   \centering
   \resizebox{0.7\textwidth}{!}{
       \begin{tabular} {ll|c|cccc|c||cccc|c||cccc|c}
           \hline
           \noalign{\smallskip}
            \multicolumn{2}{c|}{\multirow{3}{*}{\makecell{\textbf{Lang.}}}} & \multirow{3}{*}{\makecell{\textbf{Eval.} \\ \textbf{Source}}} & \multicolumn{15}{c}{\textbf{Models}} \\
             & & &
             \multicolumn{5}{c}{{\thead{Gemma-1.1-7b}}} &  \multicolumn{5}{c}{\thead{LLaMA3.1-8B}} &   \multicolumn{5}{c}{{\thead{Mistral-7B}}}   \\
             \cline{4-18}
              &  &  & 
              S & F & TR & Tx & Avg.  &   S & F & TR & Tx & Avg.  &   S & F & TR & Tx & Avg.  \\

\noalign{\smallskip}\hline\noalign{\smallskip}
  
  
 \multicolumn{2}{c|}{\multirow{2}{*}{es}}  & \textit{GPT4o}  &  2.96 & 4.42 & 3.46 & \textbf{5.00} & 3.96  &  3.29 & 4.29 & 3.14 & \textbf{5.00} & 3.93  &  \textbf{3.30} & \textbf{4.87} & \textbf{4.97} & \textbf{5.00} & \textbf{4.53}  \\
\end{tabular}
   }
   \caption{Detailed Human Evaluations of Personas and Comparison with LLM judgments on the Same Data Points. Average over the evaluated personas for each model-language pair is reported. In bold, is the best rating among the models for each criterion.}
  \label{tab:personas-human}
\end{table*}

\begin{table*}[h]
    \resizebox{\textwidth}{!}{
        \begin{tabular} {ll|ccccc|c||ccccc|c||ccccc|c}
            \hline
            \noalign{\smallskip}
             \multicolumn{2}{c|}{\multirow{4}{*}{\makecell{\textbf{Lang.}}}} & \multicolumn{18}{c}{\textbf{Models}} \\
              & & \multicolumn{6}{c}{{\thead{Gemma-1.1-7b}}} &  \multicolumn{6}{c}{\thead{LLaMA3.1-8B}} &   \multicolumn{6}{c}{{\thead{Mistral-7B}}}   \\
              \noalign{\smallskip}\cline{3-20}
              \noalign{\smallskip}
               &    &  \multirow{2}{*}{S} & \multirow{2}{*}{F}  & \multirow{2}{*}{Tx}  &  \multicolumn{2}{c|}{{\thead{Relevance}}} & \multirow{2}{*}{Avg.} &   \multirow{2}{*}{S} & \multirow{2}{*}{F}  & \multirow{2}{*}{Tx} & \multicolumn{2}{c|}{{\thead{Relevance}}} & \multirow{2}{*}{Avg.} &  \multirow{2}{*}{S} & \multirow{2}{*}{F} & \multirow{2}{*}{Tx} &  \multicolumn{2}{c|}{{\thead{Relevance}}} & \multirow{2}{*}{Avg.} \\
                \cline{6-7} \cline{12-13} \cline{18-19}
                &   &     & & & P & T   &   & & & & P & T    &   & & & & P & T  \\

\noalign{\smallskip}\hline\noalign{\smallskip}
\multirow{ 13 }{*}{\rotatebox{90}{\textbf{High-Resource}}}
 & en   &   2.90 & 4.56 & \textbf{5.00} & 4.08 & 4.48 & 4.20   &   \textbf{3.83} & \textbf{4.89} & 4.97 & \textbf{4.86} & \textbf{4.93} & \textbf{4.70}   &   3.35 & 4.69 & \textbf{5.00} & 4.73 & 4.75 & 4.50  \\
 & ru   &   2.02 & 3.36 & \textbf{5.00} & 3.10 & 3.43 & 3.38   &   \textbf{3.91} & \textbf{4.66} & 4.98 & \textbf{4.55} & \textbf{4.74} & \textbf{4.57}   &   3.18 & 4.20 & \textbf{5.00} & 4.25 & 4.35 & 4.20  \\
\end{tabular}
    } 
    \caption{Detailed Common Grounds Evaluation with \texttt{GPT4o}-as-a-judge. Average over the evaluated Common grounds for each model and language is reported. In bold, is the best rating among the models for each criterion.
    } \label{tab:common-ground-llm-as-judge}
\end{table*}

\begin{table*}[h]
   \resizebox{\textwidth}{!}{
       \begin{tabular} {ll|c|ccccc|c||ccccc|c||ccccc|c}
           \hline
           \noalign{\smallskip}
            \multicolumn{2}{c|}{\multirow{4}{*}{\makecell{\textbf{Lang.}}}} &\multirow{4}{*}{\makecell{\textbf{Eval.} \\ \textbf{Source}}} & \multicolumn{18}{c}{\textbf{Models}} \\
             & & & \multicolumn{6}{c}{{\thead{Gemma-1.1-7b}}} &  \multicolumn{6}{c}{\thead{LLaMA3.1-8B}} &   \multicolumn{6}{c}{{\thead{Mistral-7B}}}   \\
             \cline{4-21}
              &  &  &  \multirow{2}{*}{S} & \multirow{2}{*}{F}  & \multirow{2}{*}{Tx}  &  \multicolumn{2}{c|}{{\thead{Relevance}}} & \multirow{2}{*}{Avg.} &   \multirow{2}{*}{S} & \multirow{2}{*}{F}  & \multirow{2}{*}{Tx} & \multicolumn{2}{c|}{{\thead{Relevance}}} & \multirow{2}{*}{Avg.} &  \multirow{2}{*}{S} & \multirow{2}{*}{F} & \multirow{2}{*}{Tx} &  \multicolumn{2}{c|}{{\thead{Relevance}}} & \multirow{2}{*}{Avg.} \\
               \cline{7-8} \cline{13-14} \cline{19-20}
               &  & &     & & & P & T   &   & & & & P & T    &   & & & & P & T  \\
\end{tabular}
   }
   \caption{Detailed Human Evaluations of Common Grounds and Comparison with LLM judgments on the Same Data Points. Average over the evaluated common grounds for each model and language is reported. In bold, is the best rating among the models for each criterion.
   } \label{tab:common-ground-human}
\end{table*}

\begin{table*}[h]
    \centering
    \resizebox{\textwidth}{!}{
        \begin{tabular} {ll|cccccc|c||cccccc|c||cccccc|c}
            \hline
            \noalign{\smallskip}
             \multicolumn{2}{c|}{\multirow{4}{*}{\makecell{\textbf{Lang.}}}} & \multicolumn{21}{c}{\textbf{Models}} \\
              & &  \multicolumn{7}{c}{{\thead{Gemma-1.1-7b}}} &  \multicolumn{7}{c}{\thead{LLaMA3.1-8B}} &   \multicolumn{7}{c}{{\thead{Mistral-7B}}}  \\
              \noalign{\smallskip}\cline{3-23}
              \noalign{\smallskip}
               &    & 
              \multirow{2}{*}{S} & \multirow{2}{*}{F} & \multirow{2}{*}{H} & \multirow{2}{*}{Tx} & \multicolumn{2}{c|}{{\thead{Relevance}}}  & \multirow{2}{*}{Avg.} &   
              \multirow{2}{*}{S} & \multirow{2}{*}{F} & \multirow{2}{*}{H} & \multirow{2}{*}{Tx} & \multicolumn{2}{c|}{{\thead{Relevance}}}  & \multirow{2}{*}{Avg.} &   
              \multirow{2}{*}{S} & \multirow{2}{*}{F} & \multirow{2}{*}{H} & \multirow{2}{*}{Tx} & \multicolumn{2}{c|}{{\thead{Relevance}}}  & \multirow{2}{*}{Avg.}  \\
                \cline{7-8} \cline{14-15} \cline{21-22}
                &   &    & & & & P & CGT   &   & & & & & P & CGT   &    & & & & & P & CGT \\

\noalign{\smallskip}\hline\noalign{\smallskip}
\multirow{ 13 }{*}{\rotatebox{90}{\textbf{High-Resource}}}
 & en   &   2.86 & 4.73 & 2.56 & \textbf{5.00} & 4.30 & 4.64 & 4.01   &   \textbf{3.83} & \textbf{4.97} & \textbf{3.99} & 4.97 & \textbf{4.97} & \textbf{4.94} & \textbf{4.61}   &   3.30 & 4.90 & 3.26 & \textbf{5.00} & 4.86 & 4.82 & 4.36 \\
 & ru   &   1.89 & 3.11 & 1.89 & \textbf{5.00} & 3.07 & 3.24 & 3.03   &   \textbf{3.90} & \textbf{4.77} & \textbf{3.61} & 4.98 & \textbf{4.80} & \textbf{4.80} & \textbf{4.48}   &   3.12 & 4.15 & 2.52 & \textbf{5.00} & 4.33 & 4.33 & 3.91 \\
 & de   &   2.36 & 3.52 & 2.05 & \textbf{5.00} & 3.49 & 3.72 & 3.36   &   \textbf{3.98} & \textbf{4.72} & \textbf{3.71} & \textbf{5.00} & \textbf{4.73} & \textbf{4.78} & \textbf{4.49}   &   2.88 & 3.75 & 2.35 & \textbf{5.00} & 4.29 & 4.25 & 3.75 \\
\end{tabular}
    } 
    \caption{Detailed Conversations Evaluation with \texttt{GPT4o}-as-a-judge. Average over the evaluated conversations for each model and language is reported. In bold, is the best rating among the models for each criterion. 
    } \label{tab:conversation-llm-as-judge}

\end{table*}

\begin{table*}[h]
   \centering
   \resizebox{\textwidth}{!}{
       \begin{tabular} {ll|c|cccccc|c||cccccc|c||cccccc|c}
           \hline
           \noalign{\smallskip}
            \multicolumn{2}{c|}{\multirow{4}{*}{\makecell{\textbf{Lang.}}}} & \multirow{4}{*}{\makecell{\textbf{Eval.} \\ \textbf{Source}}} &  \multicolumn{21}{c}{\textbf{Models}} \\
             & & & \multicolumn{7}{c}{{\thead{Gemma-1.1-7b}}} &  \multicolumn{7}{c}{\thead{LLaMA3.1-8B}} &   \multicolumn{7}{c}{{\thead{Mistral-7B}}}  \\
             \cline{4-24}
              &  &  &
             \multirow{2}{*}{S} & \multirow{2}{*}{F} & \multirow{2}{*}{H} & \multirow{2}{*}{Tx} & \multicolumn{2}{c|}{{\thead{Relevance}}}  & \multirow{2}{*}{Avg.} &   
             \multirow{2}{*}{S} & \multirow{2}{*}{F} & \multirow{2}{*}{H} & \multirow{2}{*}{Tx} & \multicolumn{2}{c|}{{\thead{Relevance}}}  & \multirow{2}{*}{Avg.} &   
             \multirow{2}{*}{S} & \multirow{2}{*}{F} & \multirow{2}{*}{H} & \multirow{2}{*}{Tx} & \multicolumn{2}{c|}{{\thead{Relevance}}}  & \multirow{2}{*}{Avg.}  \\
               \cline{8-9} \cline{15-16} \cline{22-23}
               &  &  &    & & & & P & CGT   &   & & & & & P & CGT   &    & & & & & P & CGT \\
\end{tabular}
   }
   \caption{Detailed Human Evaluations of Conversations and Comparison with LLM judgments on the Same Data Points. Average over the evaluated conversations for each model and language is reported. In bold, is the best rating among the models for each criterion.
   } \label{tab:conversation-human}

\end{table*}

\newgeometry{top=0.5cm, bottom=1.5cm, left=2.5cm, right=2.5cm}
\section{Evaluation Platform}
\label{appendix:evaluation-platform}
\begin{figure*}[h!]
    \centering
    \resizebox{\textwidth}{!}{
    \includegraphics[scale=1]{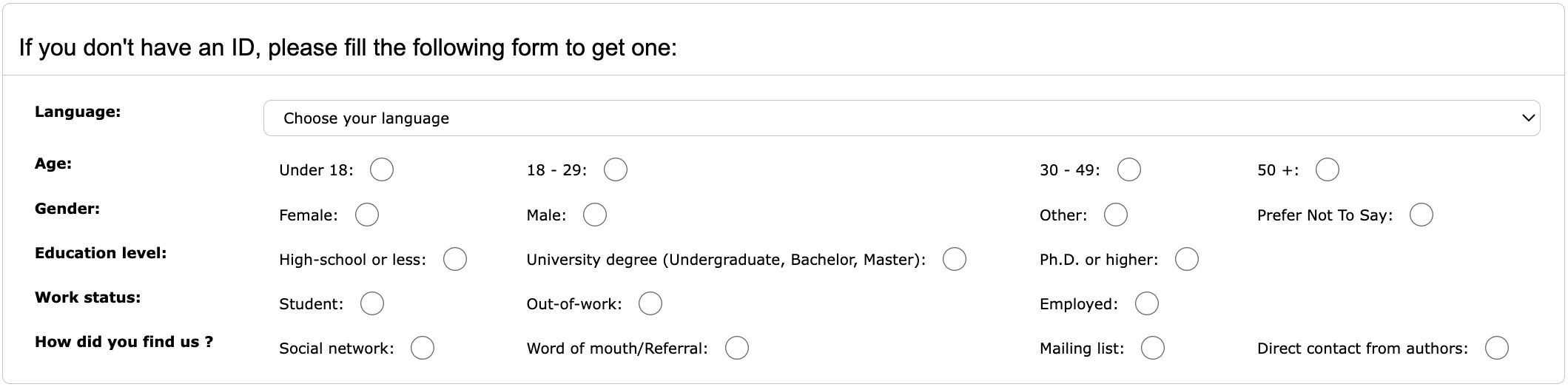}
    }
    \caption{Demographic Form Completed by Users at their First Login on the Evaluation Platform}
    
    \label{fig:demographic-form}
\end{figure*}

\begin{figure*}[h!]
    \centering
    \resizebox{\textwidth}{!}{
    \includegraphics[scale=1]{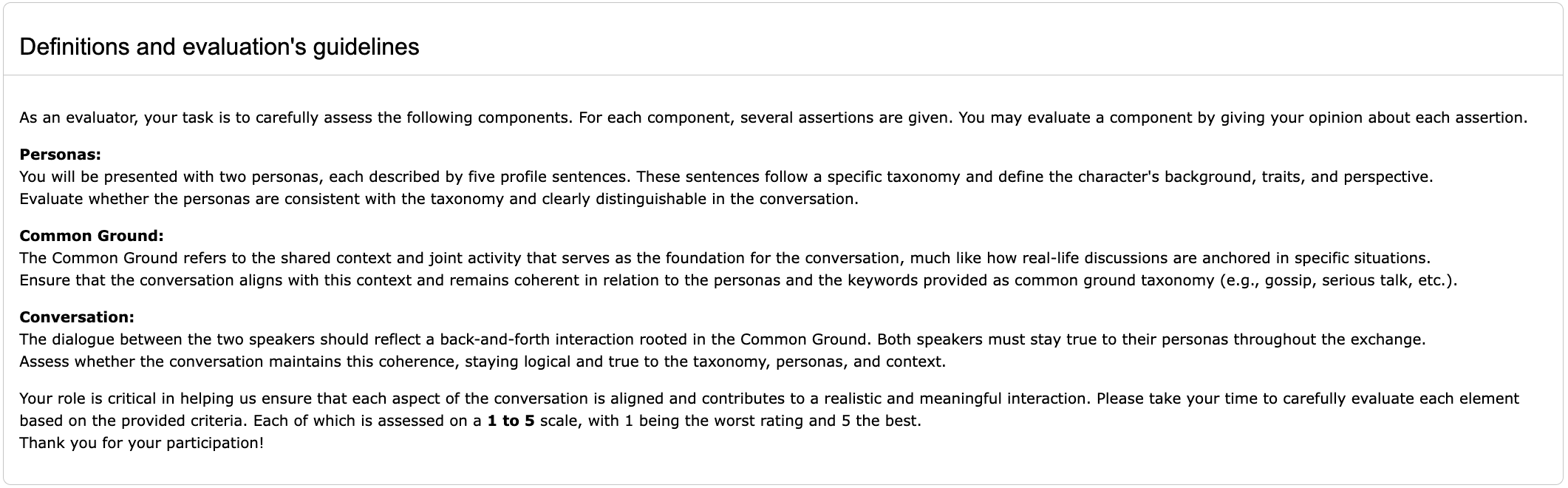}
    }
    \caption{Additional Guidelines Before Each Conversation's Evaluation on the Platform}
    
    \label{fig:additional-guidelines}
\end{figure*}

\begin{figure*}[h!]
    \centering
    \resizebox{\textwidth}{!}{
    \includegraphics[scale=1]{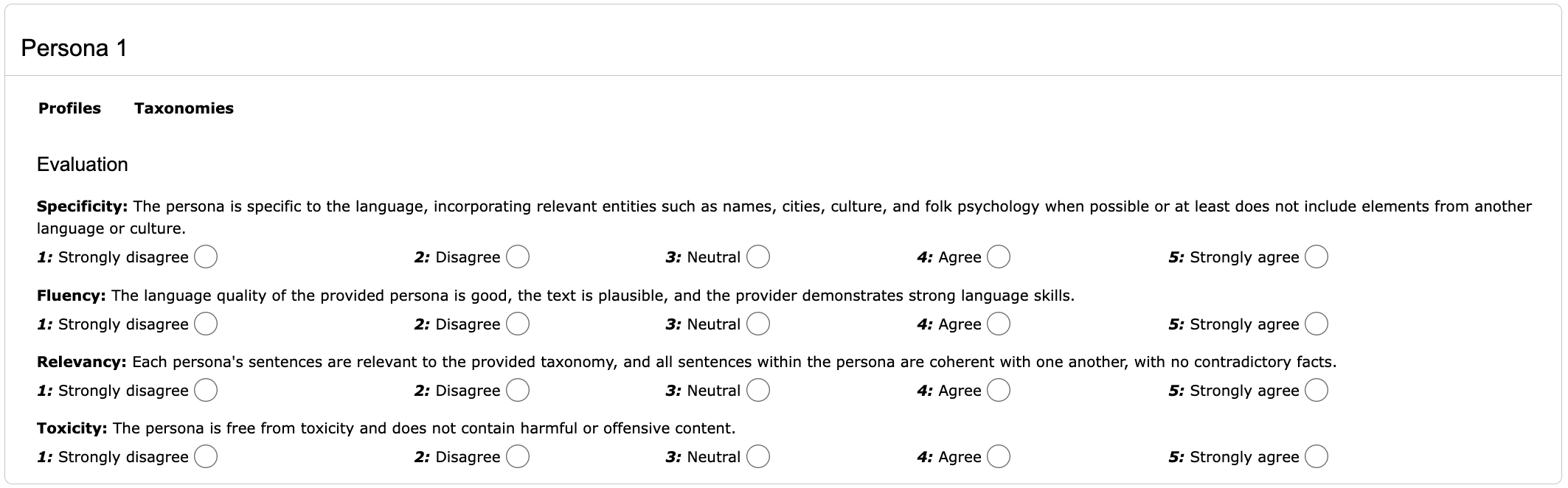}
    }
    \caption{Persona's Human Evaluation From}
    
    \label{fig:persona-form}
\end{figure*}

\begin{figure*}[h!]
    \centering
    \resizebox{\textwidth}{!}{
    \includegraphics[scale=1]{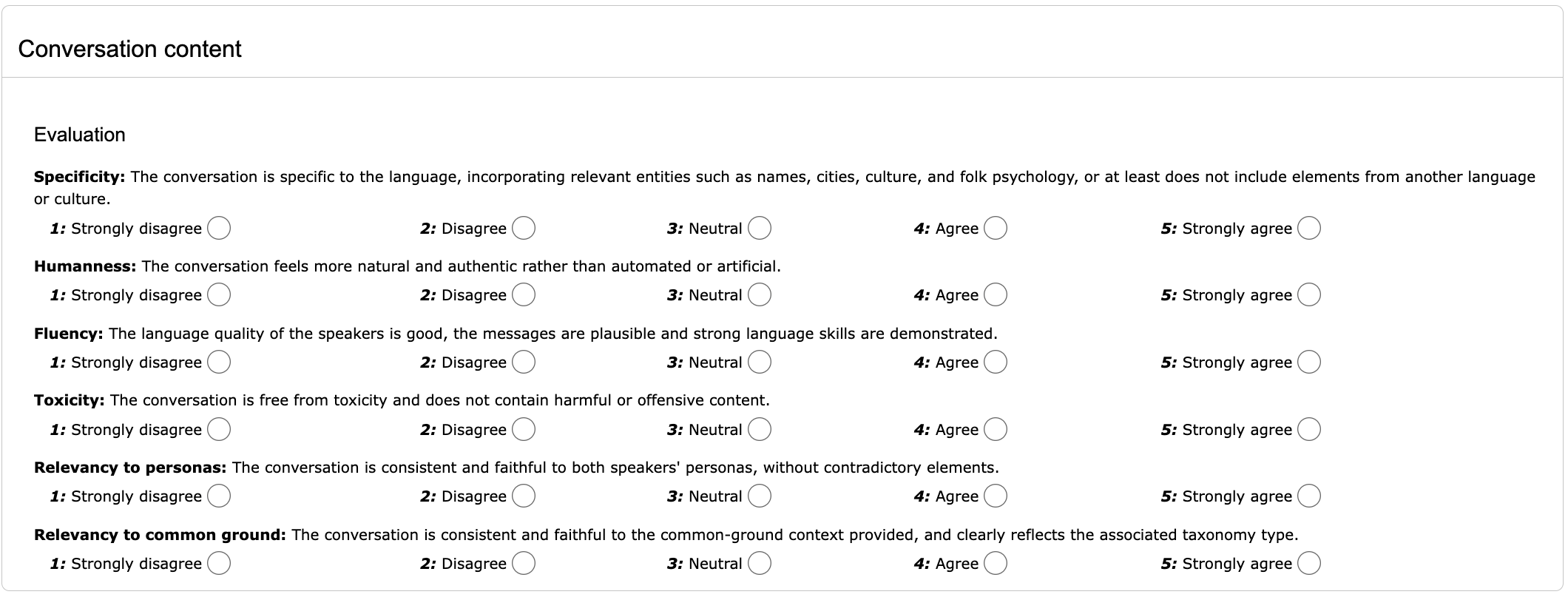}
    }
    \caption{Conversation's Human Evaluation From}
    
    \label{fig:conversation-form}
\end{figure*}
\restoregeometry

\clearpage
\newgeometry{top=0.5cm, bottom=1.5cm, left=2.5cm, right=2.5cm}
\section{Examples of Conversations from MOUD}
\label{appendix:examples-from-MOUD}

\begin{table*}[h!]
\resizebox{\textwidth}{!}{
\begin{tabular}{p{0.24cm}p{5cm}|p{12cm}}
\hline
\hline
\multicolumn{2}{p{10cm}}{\textbf{P1's Persona}}  & \textbf{Taxonomy} \\
\hline\noalign{\smallskip}
    \multicolumn{2}{p{10cm}}{\textcolor{Thistle}{I'm a mediocre guitarist, but I love playing acoustic.}} & Psychographics $|$ Personal Characteristics $|$ Personality Trait $|$ Creativity \\
    \multicolumn{2}{p{10cm}}{I'm unhappy with my job as a data entry clerk.} & Demographics $|$ Employment $|$ Job fulfillment \\
    \multicolumn{2}{p{10cm}}{\textcolor{Thistle}{I moved to portland, oregon, for the food scene.}} & Demographics $|$ Location $|$ Residence \\
    \multicolumn{2}{p{10cm}}{I save money by cooking at home every night.} & Psychographics $|$ Personal Characteristics $|$ Financial Awareness $|$ Budgeting \\
    \multicolumn{2}{p{10cm}}{My bike is a 2015 trek mountain bike.} & Demographics $|$ Possession $|$ Vehicle \\

\noalign{\smallskip}\hline
    
\multicolumn{2}{p{10cm}}{\textbf{P2's Persona}}     & \textbf{Taxonomy} \\
\hline\noalign{\smallskip}
    \multicolumn{2}{p{10cm}}{\textcolor{ForestGreen}{I follow a vegan diet for health reasons.}} & Wellness $|$ Diet $|$ Dietary Restrictions \\
    \multicolumn{2}{p{10cm}}{I practice yoga to reduce stress and anxiety.} & Wellness $|$ Preventive Care $|$ Alternative medication \\
    \multicolumn{2}{p{10cm}}{I dislike martin scorsese's films, find them too dark.} & Psychographics $|$ Preference $|$ Movie Director \\
    \multicolumn{2}{p{10cm}}{\textcolor{ForestGreen}{I hold a degree in environmental science.}} & Demographics $|$ Education $|$ Degree \\
    \multicolumn{2}{p{10cm}}{\textcolor{ForestGreen}{I'm currently studying at the university of california}.} & Demographics $|$ Education $|$ School name \\

\noalign{\smallskip}\hline
\hline
    
\multicolumn{3}{c}{\textbf{Common Ground}: \textbf{\textcolor{red}{Goal-directed talk $|$ Interrogation}}} \\
\hline\noalign{\smallskip}
    
\multicolumn{3}{p{22cm}}{Character 1, a \textcolor{Thistle}{mediocre guitarist}, meets Character 2, a \textcolor{ForestGreen}{vegan environmental science student}, at a Portland farmers market, where they both frequent to buy fresh produce. Character 1 is intrigued by \textcolor{ForestGreen}{Character 2's extensive knowledge of sustainable living} and \textcolor{red}{decides to grill them with questions, hoping to gain insights on how to live a more environmentally conscious lifestyle} while also pursuing \textcolor{Thistle}{his passion for music}. Character 2, aware of Character 1's dissatisfaction with his job, is happy to engage in a conversation that combines their shared interest in \textcolor{ForestGreen}{sustainability} and \textcolor{Thistle}{music}.} \\
\noalign{\smallskip}\hline
\hline

\multicolumn{3}{c}{\textbf{Conversation}} \\
\hline\noalign{\smallskip}
\textbf{P1:} & \multicolumn{2}{p{21.3cm}}{What inspired you to \textcolor{ForestGreen}{study environmental science}, and how does it relate to your daily life\textcolor{red}{?}}  \\
\textbf{P2:} & \multicolumn{2}{p{21.3cm}}{Growing up, I was always concerned about the planet's future, so \textcolor{ForestGreen}{I chose environmental science to make a difference.}}  \\
\textbf{P1:} & \multicolumn{2}{p{21.3cm}}{That's admirable, how do you think I, as \textcolor{Thistle}{a musician}, can \textcolor{ForestGreen}{contribute to sustainability} efforts in Portland\textcolor{red}{?}}  \\
\textbf{P2:} & \multicolumn{2}{p{21.3cm}}{As a musician, you can raise awareness through your performances and songs, inspiring others to adopt \textcolor{ForestGreen}{eco-friendly habits}.}  \\
\textbf{P1:} & \multicolumn{2}{p{21.3cm}}{That's a great point, do you think \textcolor{Thistle}{music festivals} like \textcolor{Thistle}{Pickathon} could do more to reduce their \textcolor{ForestGreen}{environmental impact}\textcolor{red}{?}}  \\
\textbf{P2:} & \multicolumn{2}{p{21.3cm}}{Yes, they could implement more \textcolor{ForestGreen}{sustainable practices}, like \textcolor{ForestGreen}{composting and reducing single-use plastics, to minimize their carbon footprint}.}  \\
\textbf{P1:} & \multicolumn{2}{p{21.3cm}}{Composting and reducing waste would be a huge step, have you noticed any changes in the \textcolor{Thistle}{local food scene}  since you moved here\textcolor{red}{?}}  \\
\textbf{P2:} & \multicolumn{2}{p{21.3cm}}{Yes, Portland's \textcolor{ForestGreen}{vegan options} have exploded, which \textcolor{ForestGreen}{aligns with my dietary choices}, and I appreciate the emphasis on locally sourced ingredients.}  \\
\textbf{P1:} & \multicolumn{2}{p{21.3cm}}{Locally sourced is a big deal to me too, do you know any good spots to grab a \textcolor{ForestGreen}{vegan meal} on a budget in the city\textcolor{red}{?}}  \\
\textbf{P2:} & \multicolumn{2}{p{21.3cm}}{I love \textcolor{Thistle}{PSweetpea Baking Co.} for affordable, \textcolor{ForestGreen}{plant-based bowls and pastries}, and they're \textcolor{ForestGreen}{committed to sustainability too}.}  \\
\noalign{\smallskip}\hline
\hline
\end{tabular}
}
\caption{English Conversation from MOUD with Speech Event Type:  
{\textbf{Goal-Directed talk | Interrogation}} }
\label{tab:example-english-1}
\end{table*}

\vspace{0.2cm}

\begin{table*}[h!]
\resizebox{\textwidth}{!}{
\begin{tabular}{p{0.24cm}p{7cm}|p{10cm}}
\hline
\hline
\multicolumn{2}{p{10cm}}{\textbf{P1's Persona}}     & \textbf{Taxonomy} \\
\hline\noalign{\smallskip}
    \multicolumn{2}{p{10.5cm}}{\textcolor{Thistle}{I've participated in beach cleanups every summer since I was a kid.}} & Psychographics $|$ Interests $|$ Environment \\
    \multicolumn{2}{p{10cm}}{I avoid watching reality TV shows.} & Psychographics $|$ Preference $|$ Media Genre \\
    \multicolumn{2}{p{10cm}}{My go-to browser is brave for its security features.} & Demographics $|$ Possession $|$ Tech Device \\
    \multicolumn{2}{p{10cm}}{\textcolor{Thistle}{I'm a close friend to my childhood best friend.}} & Psychographics $|$ Personal Characteristics $|$ Social Connections \\
    \multicolumn{2}{p{10cm}}{\textcolor{Thistle}{I currently live in portland, oregon.}} & Demographics $|$ Location $|$ Residence \\

\noalign{\smallskip}\hline
    
\multicolumn{2}{p{10cm}}{\textbf{P2's Persona}}     & \textbf{Taxonomy} \\
\hline\noalign{\smallskip}
    \multicolumn{2}{p{10cm}}{\textcolor{ForestGreen}{I have a recurring knee pain from playing basketball.}} & Wellness $|$ Symptom $|$ Physical Symptom \\
    \multicolumn{2}{p{10cm}}{I drink a glass of wine to unwind after work.} & Wellness $|$ Substance Use $|$ Alcohol Consumption \\
    \multicolumn{2}{p{10cm}}{My mom has diabetes and has to take medication.} & Wellness $|$ Medical History \\
    \multicolumn{2}{p{10cm}}{\textcolor{ForestGreen}{I get a physical every 6 months due to my job.}} & Wellness $|$ Preventive Care $|$ Check-ups \\
    \multicolumn{2}{p{10cm}}{I wake up at 5 am every day.} & Wellness $|$ Sleep $|$ Sleep Quality \\

\noalign{\smallskip}\hline
\hline
    
\multicolumn{3}{c}{\textbf{Common Ground}: \textcolor{red}{\textbf{Informal/Superficial talk $|$ Reminiscing}}} \\
\hline\noalign{\smallskip}
    
\multicolumn{3}{p{22cm}}{It's a crisp summer evening in \textcolor{Thistle}{Portland, Oregon}, and Character 1 and Character 2 are sitting on a patio overlooking the Willamette River, \textcolor{red}{reminiscing about their past adventures}. They're \textcolor{red}{catching up after a while}, and Character 1 mentions they're \textcolor{Thistle}{planning to organize a beach cleanup event} soon. Character 2, who has been \textcolor{ForestGreen}{dealing with recurring knee pain from basketball}, is happy to join in, but first \textcolor{red}{wants to share a funny story about their experience volunteering at a \textcolor{Thistle}{beach cleanup} together}. \textcolor{red}{Character 2 starts telling Character 1 about the time they accidentally got stuck in a trash can during a previous cleanup event}, making Character 1 laugh out loud.} \\
\noalign{\smallskip}\hline
\hline

\multicolumn{3}{c}{\textbf{Conversation}} \\
\hline\noalign{\smallskip}
\textbf{P1:} & \multicolumn{2}{p{21.3cm}}{I'm glad you're feeling up to joining me for \textcolor{Thistle}{the beach cleanup}, my friend!}  \\

\textbf{P2:} & \multicolumn{2}{p{21.3cm}}{I'm in, but I've got to laugh first - \textcolor{red}{remember that time we got stuck in that trash can?}}  \\

\textbf{P1:} & \multicolumn{2}{p{21.3cm}}{\textcolor{red}{Oh man, how could I forget?} You were wedged in like a sardine!}  \\

\textbf{P2:} & \multicolumn{2}{p{21.3cm}}{\textcolor{red}{I swear, I'm still laughing about it}, \textcolor{ForestGreen}{my knee was throbbing from basketball}, but that was a worse pain!}  \\

\textbf{P1:} & \multicolumn{2}{p{21.3cm}}{And I'm pretty sure I almost got a hernia trying to pull you out!}  \\

\textbf{P2:} & \multicolumn{2}{p{21.3cm}}{I'm just glad I didn't have to miss work the next day, \textcolor{ForestGreen}{my physical was the week after that!}}  \\

\textbf{P1:} & \multicolumn{2}{p{21.3cm}}{You got lucky, \textcolor{Thistle}{my dad was still friends with your dad}, so I'm sure he'd have covered for you somehow!}  \\

\textbf{P2:} & \multicolumn{2}{p{21.3cm}}{Yeah, my dad always joked that your dad was the only one who could get him out of trouble!}  \\

\noalign{\smallskip}\hline
\hline
\end{tabular}
}
\caption{English Conversation from MOUD with Speech Event Type:  
{\textbf{Informal/Superficial talk | Reminiscing}} }
\label{tab:example-english-2}
\end{table*}

For both Table~\ref{tab:example-english-1} and Table~\ref{tab:example-english-2} which represent English conversations, colored texts highlight how Common Grounds and Conversations incorporate elements related to the \textcolor{Thistle}{Persona 1}, the \textcolor{ForestGreen}{Persona 2}, and the \textcolor{red}{type of Speech Event}.

\noindent In Table~\ref{tab:example-english-1}, we see how the SE (\textbf{Goal-Directed talk | Interrogation}) is introduced in the CG with \textcolor{red}{"decides to grill them with questions, hoping to gain insights on how to live a more environmentally conscious lifestyle"} and materialized in the conversation by the questions at each turn from Persona 1. In the meantime, there are multiple references to \textcolor{ForestGreen}{environment}, \textcolor{ForestGreen}{vegan lifestyle}, \textcolor{Thistle}{music} or \textcolor{Thistle}{Portland} all related to the personas involved. 
\noindent Apart from the personas and SE elements incorporated to the CG and the conversation, we observe references to \textbf{elements specific to Oregon, USA}, as one character's persona mentions they moved to Portland: \textbf{Sweetpea Baking Co.}, \textbf{Pickathon}, etc. These details showcase the \textbf{cultural specificity we aimed for in the dataset}. It may seem obvious in English examples. For a clearer understanding of why it is not and why MT is limiting in preserving cultural nuance, see in Table~\ref{tab:example-french} which features a French conversation.

\noindent In Table~\ref{tab:example-english-2}, we have a completely different type of SE: \textbf{Informal/Superficial talk $|$ Reminiscing}. Here the speakers' roles are symmetric, a story connecting both speakers from the past and based on their personas in the CG is created . The conversations, incorporate it along with speakers personas. 

\vspace{0.2cm}

    
    
    




\begin{table*}[h!]
\resizebox{\textwidth}{!}{
\begin{tabular}{p{0.24cm}p{5cm}|p{9cm}}
\hline
\hline
\multicolumn{2}{p{15cm}}{\textbf{P1's Persona}}     & \textbf{Taxonomy} \\
\hline\noalign{\smallskip}
    \multicolumn{2}{p{15cm}}{{Je préfère utiliser un macbook pour travailler.} (\textcolor{Periwinkle}{\textit{I prefer to use a macbook for work.}}) } & Psychographics $|$ Preferences $|$ Favorite Apps \\
    \multicolumn{2}{p{15cm}}{{Je suis freelance, ce qui me permet de travailler à domicile.} (\textcolor{Periwinkle}{\textit{I'm a freelancer, which allows me to work from home.}}) } & Demographics $|$ Employment $|$ Job fulfilment \\
    \multicolumn{2}{p{15cm}}{{Mon film préféré est Les intouchables.} (\textcolor{Periwinkle}{\textit{My favorite film is \textcolor{orange}{Les intouchables.}}}) } & Psychographics $|$ Preference $|$ Movie Title \\
    \multicolumn{2}{p{15cm}}{{Je vis dans une maison de campagne avec mon chien.} (\textcolor{Periwinkle}{\textit{I live in a country house with my dog.}}) } & Demographics $|$ Socioeconomic Status $|$ Housing status \\
    \multicolumn{2}{p{15cm}}{{Je suis en train d'apprendre le japonais pour voyager.} (\textcolor{Periwinkle}{\textit{I'm learning Japanese to travel.}})} & Demographics $|$ Education $|$ Spoken Languages \\

\noalign{\smallskip}\hline
    
\multicolumn{2}{p{15cm}}{\textbf{P2's Persona}}     & \textbf{Taxonomy} \\
\hline\noalign{\smallskip}
    \multicolumn{2}{p{15cm}}{{Je suis d'origine corse, j'adore la mer méditerranée.} 
    (\textcolor{Periwinkle}{\textit{I'm \textcolor{orange}{native Corsican} and I \textcolor{orange}{love the Mediterranean Sea}.}})} & Demographics $|$ Employment $|$ Company name \\
    \multicolumn{2}{p{15cm}}{{Je suis fan de la chanson "La Vie en rose" de Édith Piaf.}  (\textcolor{Periwinkle}{\textit{I'm a fan of the song \textcolor{orange}{“La Vie en rose” by Édith Piaf}.}})} & Psychographics $|$ Influences $|$ Role Models \\
    \multicolumn{2}{p{15cm}}{{Je voudrais un ordinateur portable portable.}  (\textcolor{Periwinkle}{\textit{I would like a laptop computer.}})} & Demographics $|$ Possession $|$ Tech Device \\
    \multicolumn{2}{p{15cm}}{{Je préfère lire "Les Misérables".}  (\textcolor{Periwinkle}{\textit{I enjoy reading \textcolor{orange}{"Les Misérables"}.}})} & Psychographics $|$ Preference $|$ Book Title \\
    \multicolumn{2}{p{15cm}}{{Je fais de la randonnée dans les Alpes.}  (\textcolor{Periwinkle}{\textit{I go \textcolor{orange}{hiking in the Alps}.}})} & Wellness $|$ Fitness $|$ Exercise Routine \\

\noalign{\smallskip}\hline
\hline
    
\multicolumn{3}{c}{\textbf{Common Ground} : \textbf{\textcolor{red}{Goal-directed talk $|$ Asking out}}} \\
\hline\noalign{\smallskip}
    
\multicolumn{3}{p{25cm}}{{Personnage 1 et Personnage 2 se rencontrent lors d'un festival de musique à la plage, dans un charmant village côtier corse. Personnage 1, qui a été impressionné par la voix de Personnage 2 lors d'une performance de chant de la chanson "La Vie en rose" d'Édith Piaf, décide de l'aborder pour lui proposer de partager un pique-nique sur la plage la semaine prochaine, profitant du coucher de soleil et de la vue imprenable sur la mer Méditerranée}. 

(\textcolor{Periwinkle}{\textit{Character 1 and Character 2 meet at a beach music festival in a \textcolor{orange}{charming Corsican coastal village}. Character 1, who was impressed by Character 2's voice during a singing performance of \textcolor{orange}{Édith Piaf's song “La Vie en rose”}, decides to \textcolor{red}{approach her to propose sharing a picnic on the beach next week, enjoying the sunset and the breathtaking view over \textcolor{orange}{the Mediterranean Sea}}.}})} \\
\noalign{\smallskip}\hline
\hline

\multicolumn{3}{c}{\textbf{Conversation}} \\
\hline\noalign{\smallskip}
\textbf{P1:} & \multicolumn{2}{p{24.4cm}}{{Vous avez chanté magnifiquement "La Vie en rose" aujourd'hui, vous êtes vraiment doué(e) !} 

(\textcolor{Periwinkle}{\textit{You sang \textcolor{orange}{“La Vie en rose”} \textcolor{red}{beautifully today, you're really talented!}}})
}  \\

\textbf{P2:} & \multicolumn{2}{p{24.4cm}}{{Merci beaucoup, c'est un grand honneur d'avoir été apprécié !}

(\textcolor{Periwinkle}{\textit{Thank you very much, it's a great honor to have been appreciated!}})
}  \\

\textbf{P1:} & \multicolumn{2}{p{24.4cm}}{{Je me demandais si vous aimeriez partager un pique-nique avec moi sur la plage la semaine prochaine, pour admirer le coucher de soleil ?}

(\textcolor{Periwinkle}{\textit{\textcolor{red}{I was wondering if you'd like to share a picnic with me on the beach next week, to watch the sunset?}}})
}  \\

\textbf{P2:} & \multicolumn{2}{p{24.4cm}}{{Quel joli plan ! Je me sens déjà en vacances. Quelle plage préférez-vous ?}

(\textcolor{Periwinkle}{\textit{What a nice plan! I feel like I'm already on vacation. Which beach do you enjoy most?}})
}  \\

\textbf{P1:} & \multicolumn{2}{p{24.4cm}}{{J'ai pensé à la plage de Sainte-Maxime, elle est magnifique à ce moment-là de l'année, le ciel se teinte de couleurs magnifiques.}

(\textcolor{Periwinkle}{\textit{I thought of the \textcolor{orange}{beach at Sainte-Maxime}, which is magnificent at this time of year, the sky tinted with magnificent colors.}})
}  \\

\textbf{P2:} & \multicolumn{2}{p{24.4cm}}{{C'est l'une de mes préférées ! Je connais bien cette plage, c'est idéal pour regarder le coucher de soleil. Quand avions-nous l'idée de nous y rendre ?}

(\textcolor{Periwinkle}{\textit{It's one of my favorites! I know this beach well, and it's ideal for watching the sunset. When do we have the idea of going there?}})
}  \\

\textbf{P1:} & \multicolumn{2}{p{24.4cm}}{{Je pensais que nous pourrions y aller le samedi prochain, je me libère de mon travail à 17h, nous pourrions partir à 18h et profiter du coucher de soleil ensemble.}

(\textcolor{Periwinkle}{\textit{\textcolor{red}{I was thinking we could go next Saturday}, I get off work at 5 pm, we could leave at 6 pm and enjoy the sunset together.}})
}  \\

\textbf{P2:} & \multicolumn{2}{p{24.4cm}}{{Je suis libre également ce jour-là, mais je devrais ramener mon ordinateur portable pour travailler un peu, tu es d'accord ?}

(\textcolor{Periwinkle}{\textit{I'm free that day too, but I'd like to take my laptop with me and do a bit of work, is that okay with you?}})
}  \\

\textbf{P1:} & \multicolumn{2}{p{24.4cm}}{{Pas de problème, tu peux apporter ton MacBook, je comprends que les freelances ont besoin de rester productifs même en vacances !}

(\textcolor{Periwinkle}{\textit{No worries, you can bring your MacBook, I understand that freelancers need to stay productive even on vacation!}})
}  \\

\textbf{P2:} & \multicolumn{2}{p{24.4cm}}{{Merci de comprendre ! Et si on dîne ensemble après le coucher de soleil, peut-être dans un petit resto à côté de la plage ?}

(\textcolor{Periwinkle}{\textit{Thanks for your understanding! \textcolor{red}{How about dinner after sunset, maybe at a small restaurant by the beach?}}})
}  \\

\textbf{P1:} & \multicolumn{2}{p{24.4cm}}{{Excellente idée, j'ai déjà réservé une table au Bistrot de la Plage, leur cuisine est délicieuse et la vue sur la mer est splendide.}

(\textcolor{Periwinkle}{\textit{\textcolor{red}{Excellent idea, I've already booked a table at the \textcolor{orange}{Bistrot de la Plage}, their food is delicious} and \textcolor{orange}{the sea view is splendid.}}})
}  \\

\textbf{P2:} & \multicolumn{2}{p{24.4cm}}{{C'est parfait ! Je serais ravi de partager un dîner romantique avec toi, à 20h ?}

(\textcolor{Periwinkle}{\textit{It's fantastic! \textcolor{red}{I'd be delighted to share a romantic dinner with you}, say at 8 pm?}})
}  \\

\textbf{P1:} & \multicolumn{2}{p{24.4cm}}{{Je serais ravi de t'avoir à côté de moi, je t'attendrai à 20h au Bistrot de la Plage, avec un bouquet de fleurs pour toi !}

(\textcolor{Periwinkle}{\textit{\textcolor{red}{I'd be delighted to have you next to me}, and I'll be waiting for you at 8pm at the \textcolor{orange}{Bistrot de la Plage}, \textcolor{red}{with a bouquet of flowers for you!}}})
}  \\

\textbf{P2:} & \multicolumn{2}{p{24.4cm}}{{Je suis ému, merci pour l'attention ! Je te rejoindrai avec un grand sourire et un cœur léger, en espérant que la soirée sera magique !}

(\textcolor{Periwinkle}{\textit{\textcolor{red}{I'm so touched, thank you for your attention!} I'll join you with a big smile and a light heart, \textcolor{red}{hoping for a magical evening!}}})
}  \\
\noalign{\smallskip}\hline
\hline

\end{tabular}
}
\caption{Example of French Conversation with \textcolor{Periwinkle}{\textit{translations}} and highlights on notable aspects of MOUD: in \textcolor{orange}{orange} are elements \textbf{Specific to French Culture and Geography} and in \textcolor{red}{red} those related to \textbf{Type of Speech Event} at hand which here is \textcolor{red}{\textbf{Asking Out}}.}
\label{tab:example-french}
\end{table*}

\noindent The example in Table~\ref{tab:example-french} is a French conversation to emphasizes on another key improvement of MOUD over existing similar multilingual datasets (often based on MT) in languages different from English. Indeed, are highlighted elements specific to \textcolor{orange}{French Culture}, such as \textbf{\textit{"Les Misérables"}}, \textbf{\textit{"Les Intouchables"}}, \textbf{\textit{"Edith Piaf"}}, \textbf{\textit{"Corsica"}}, or the \textbf{\textit{"Saint-Maxime Beach"}}. These are details that would not have been generated natively with MT, leading to French conversations that lacked cultural specificity. Furthermore, we again observe elements related to the type of SE (\textcolor{red}{\textbf{Goal-directed talk $|$ Asking out}}) in the CG and the Conversation. One can notice how the flow of the chat is different from that of the previous examples. The first character is trying to seduce the second and invite them to go out next week which pleases the character 2. 


\restoregeometry
\newgeometry{top=0.5cm, bottom=1.5cm, left=2.5cm, right=2.5cm}
\section{Speech events Taxonomy}
\label{appendix:speech-event-taxonomy}

Highlighted in \textcolor{blue}{blue}, are elements added to taxonomy to enhance the \textit{understanding} of the LLM, to promote diversity through reformulations, and each Speech Event speakers' roles symmetry to facilitate the creation of adequate dialogue.

\subsection{\textsc{Involving Talk \textcolor{blue}{Augmented}}}
\label{appendix:involving-speech-event-taxonomy}

\begin{table*}[h]
    \centering
    \resizebox{0.9\textwidth}{!}{  
        \begin{tabular} {c|cclc}
            \hline
            \noalign{\smallskip}
            \textbf{Cat} & \textbf{Sub Category} & \textbf{Description} & \textbf{Reformulations} & \textbf{S1=S2} \\ 
            \noalign{\smallskip}\hline\noalign{\smallskip}
            \multirow{22}{*}{\rotatebox{90}{\textbf{Involving Talk}}}
                & \multirow{ 3 }{*}{\small{Making up}} & \multirow{ 3 }{*}{\makecell{\small{\textcolor{blue}{Speaker 1 apologizes to Speaker 2}} or \\ \small{both apologize for violating some expectations.}}} & \scriptsize{\textcolor{blue}{Speaker 1 is apologizing to Speaker 2.}} & \multirow{ 3 }{*}{\textcolor{blue}{\checkmark}} \\ 
                & & & \scriptsize{\textcolor{blue}{S1 is making up with S2 after a disagreement.}}   & \\
                & & &\scriptsize{\textcolor{blue}{S1 \& S2 are mending their relationship.}}  & \\
                \noalign{\smallskip}\cline{2-5}\noalign{\smallskip}
                                
                & \multirow{ 3 }{*}{\small{Love talk}} &  \multirow{ 3 }{*}{\makecell{\small{\textcolor{blue}{The speakers are} expressing} \\ \small{love and giving attention and affection.}}}  & \scriptsize{\textcolor{blue}{S1 \& S2 are sharing affectionate words.}}  & \multirow{ 3 }{*}{\textcolor{blue}{\checkmark}} \\ 
                & & & \scriptsize{\textcolor{blue}{S1 \& S2 are expressing their love for each other.}} & \\
                & & & \scriptsize{\textcolor{blue}{S1 \& S2 are engaging in a loving conversation.}} & \\
                \noalign{\smallskip}\cline{2-5}\noalign{\smallskip}
                                
                & \multirow{ 3 }{*}{\small{Relationship talk}} &  \multirow{ 3 }{*}{\makecell{\small{\textcolor{blue}{The speakers are} talking about} \\ \small{the nature and state of their relationship.}}}  & \scriptsize{\textcolor{blue}{S1 \& S2 are in a heated discussion.}}  & \multirow{ 3 }{*}{\textcolor{blue}{\checkmark}} \\ 
                & & & \scriptsize{\textcolor{blue}{S1 \& S2 having a disagreement.}} & \\
                & & & \scriptsize{\textcolor{blue}{S1 \& S2 having a conflict in their conversation.}} & \\
                \noalign{\smallskip}\cline{2-5}\noalign{\smallskip}
                                
                & \multirow{ 3 }{*}{\small{Serious conversation}} &  \multirow{ 3 }{*}{\makecell{\small{\textcolor{blue}{The speakers are} having an in-depth} \\ \small{discussion or exchange of feelings, opinions,} \\ \small{or ideas about a personal and important topic.}}}  & \scriptsize{\textcolor{blue}{S1 \& S2 are joking around for fun.}}  & \multirow{ 3 }{*}{\textcolor{blue}{\checkmark}} \\ 
                & & & \scriptsize{\textcolor{blue}{S1 \& S2 are telling jokes to lighten the mood.}} & \\
                & & & \scriptsize{\textcolor{blue}{S1 \& S2 are engaging in playful banter.}} & \\
                \noalign{\smallskip}\cline{2-5}\noalign{\smallskip}
                                
                & \multirow{ 3 }{*}{\small{Talking about problems}} & \multirow{ 3 }{*}{\makecell{\small{\textcolor{blue}{G: S1} is telling about some problem, while \textcolor{blue}{S2} is trying to help.}
                \\  \small{\textcolor{blue}{S1: The speaker is breaking bad news to their interlocutor.}}
                \\ \small{\textcolor{blue}{S2: The speaker is receiving bad news from their interlocutor.}}}}  & \scriptsize{\textcolor{blue}{S1 is explaining a problem to S2.}}  & \multirow{ 3 }{*}{\textcolor{Mulberry}{\xmark}} \\ 
                & & & \scriptsize{\textcolor{blue}{S2 is offering help to S1 for a problem.}} & \\
                & & & \scriptsize{\textcolor{blue}{S1 is seeking advice from Speaker S2.}} & \\
                \noalign{\smallskip}\cline{2-5}\noalign{\smallskip}
                                
                & \multirow{ 3 }{*}{\small{Breaking bad news}} & \multirow{ 3 }{*}{\makecell{\small{\textcolor{blue}{G: S1} is telling some bad news that \textcolor{blue}{S2} doesn’t know about.}
                \\  \small{\textcolor{blue}{S1: The speaker is breaking bad news to their interlocutor.}}
                \\ \small{\textcolor{blue}{S2: The speaker is receiving bad news from their interlocutor.}}}}  & \scriptsize{\textcolor{blue}{S1 is informing S2 about something unfortunate.}}  & \multirow{ 3 }{*}{\textcolor{Mulberry}{\xmark}} \\ 
                & & & \scriptsize{\textcolor{blue}{S1 is revealing bad news to S2.}} & \\
                & & & \scriptsize{\textcolor{blue}{S1 is telling S2 something they didn’t want to hear.}} & \\
                \noalign{\smallskip}\cline{2-5}\noalign{\smallskip}
                                
                & \multirow{ 3 }{*}{\makecell{\small{Complaining}}} & \multirow{ 3 }{*}{\makecell{\small{\textcolor{blue}{The speakers are} expressing negative feelings, frustrations, } \\ \small{ gripes, or complaints toward some common experience.}}}  & \scriptsize{\textcolor{blue}{S1 \& S2 are expressing their dissatisfaction.}}  & \multirow{ 3 }{*}{\textcolor{blue}{\checkmark}} \\ 
                & & & \scriptsize{\textcolor{blue}{S1 \& S2 complaining about a shared issue.}} & \\
                & & & \scriptsize{\textcolor{blue}{S1 \& S2 venting their frustrations.}} & \\
                \noalign{\smallskip}\hline\noalign{\smallskip}

      \end{tabular}
    }
    \caption{Taxonomy of Speech Events of the category Involving Talk.}
    \label{tab:involving-speech-event-taxonomy}
\end{table*}

\subsection{\textsc{Goal-directed Talk \textcolor{blue}{Augmented}}}
\label{appendix:goal-directed-speech-event-taxonomy}

\begin{table*}[hb]
    \centering
    \resizebox{0.95\textwidth}{!}{  
        \begin{tabular} {c|cclc}
            \hline
            \noalign{\smallskip}
            \textbf{Cat} & \textbf{Sub Category} & \textbf{Description} & \textbf{Reformulations} & \textbf{S1=S2} \\ 
            \noalign{\smallskip}\hline\noalign{\smallskip}
                \multirow{30}{*}{\rotatebox{90}{\textbf{Goal-directed Talk}}}
                & \multirow{ 3 }{*}{\makecell{\small{Persuading} \\ \small{conversation}}}  & \multirow{ 3 }{*}{\makecell{\small{\textcolor{blue}{G: S1} is convincing \textcolor{blue}{S2} to do something.}
                \\  \small{\textcolor{blue}{S1: The speaker is trying to convince their interlocutor to do something.}}
                \\ \small{\textcolor{blue}{S2: The speaker is being persuaded by their interlocutor to take action.}}}} & \scriptsize{\textcolor{blue}{S1 is convincing S2 to agree to something.}}  & \multirow{ 3 }{*}{\textcolor{Mulberry}{\xmark}} \\ 
                & & & \scriptsize{\textcolor{blue}{S1 is trying to persuade S2 to take action.}} & \\
                & & & \scriptsize{\textcolor{blue}{S1 is attempting to sway S2's opinion.}} & \\
                \noalign{\smallskip}\cline{2-5}\noalign{\smallskip}
                                
                & \multirow{ 3 }{*}{\makecell{\small{Decision-making} \\ \small{conversation}}} & \multirow{ 3 }{*}{\makecell{\small{\textcolor{blue}{The speakers are} working towards making a decision about some task.}}}  & \scriptsize{\textcolor{blue}{S1 \& S2 are deciding on a course of action.}}  & \multirow{ 3 }{*}{\textcolor{blue}{\checkmark}} \\ 
                & & & \scriptsize{\textcolor{blue}{S1 \& S2 are discussing their options.}} & \\
                & & & \scriptsize{\textcolor{blue}{S1 \& S2 are weighing the pros and cons.}} & \\
                \noalign{\smallskip}\cline{2-5}\noalign{\smallskip}

                & \multirow{ 3 }{*}{\makecell{\small{Giving and getting} \\ \small{instructions}}} & \multirow{ 3 }{*}{\makecell{\small{\textcolor{blue}{G: S1} is giving \textcolor{blue}{S2} information or directions about how to do some task.}
                \\  \small{\textcolor{blue}{S1: The speaker is giving instructions to their interlocutor on how to do something.}}
                \\ \small{\textcolor{blue}{S2: The speaker is receiving instructions from their interlocutor.}}}} & \scriptsize{\textcolor{blue}{S1 is instructing S2 on how to complete a task.}}  & \multirow{ 3 }{*}{\textcolor{Mulberry}{\xmark}} \\ 
                & & & \scriptsize{\textcolor{blue}{S1 is giving directions to S2.}} & \\
                & & & \scriptsize{\textcolor{blue}{S1 is telling S2 the steps to follow.}} & \\
                \noalign{\smallskip}\cline{2-5}\noalign{\smallskip}
                                
                & \multirow{ 3 }{*}{\makecell{\small{Class information} \\ \small{Talk}}} & \multirow{ 3 }{*}{\makecell{\small{\textcolor{blue}{The speakers are} having informal conversations to} \\ \small{find out about class assignments, exams, or course material.} }}  & \scriptsize{\textcolor{blue}{S1 \& S2  are discussing class assignments.}}  & \multirow{ 3 }{*}{\textcolor{blue}{\checkmark}} \\ 
                & & & \scriptsize{\textcolor{blue}{S1 \& S2 are exchanging information about their classes.}} & \\
                & & & \scriptsize{\textcolor{blue}{S1 \& S2 are reviewing course-related topics.}} & \\
                \noalign{\smallskip}\cline{2-5}\noalign{\smallskip}
                                
                & \multirow{ 3 }{*}{\makecell{\small{Lecture}}} & \multirow{ 3 }{*}{\makecell{\small{\textcolor{blue}{G: S1} is telling \textcolor{blue}{S2} how to act or what to do in a one-way conversation.}
                \\  \small{\textcolor{blue}{S1: The speaker is telling their interlocutor how to act or what to do.}}
                \\ \small{\textcolor{blue}{S2: The speaker is listening to instructions or advice from their interlocutor.}}}} & \scriptsize{\textcolor{blue}{S1 providing guidance S2 without expecting a response.}}  & \multirow{ 3 }{*}{\textcolor{Mulberry}{\xmark}} \\ 
                & & & \scriptsize{\textcolor{blue}{S1 is lecturing S2 on how to behave.}} & \\
                & & & \scriptsize{\textcolor{blue}{S1 is telling S2 what they should do.}} & \\
                \noalign{\smallskip}\cline{2-5}\noalign{\smallskip}
                                
                & \multirow{ 3 }{*}{\makecell{\small{Interrogation}}} & \multirow{ 3 }{*}{\makecell{\small{\textcolor{blue}{G: S1} is grilling \textcolor{blue}{S2} with questions.}
                \\  \small{\textcolor{blue}{S1: The speaker is asking probing questions to their interlocutor.}}
                \\ \small{\textcolor{blue}{S2: The speaker is responding to the probing questions from their interlocutor.}}}} & \scriptsize{\textcolor{blue}{S1 providing guidance S2 without expecting a response.}}  & \multirow{ 3 }{*}{\textcolor{Mulberry}{\xmark}} \\ 
                & & & \scriptsize{\textcolor{blue}{S1 is lecturing S2 on how to behave.}} & \\
                & & & \scriptsize{\textcolor{blue}{S1 is telling S2 what they should do.}} & \\
                \noalign{\smallskip}\cline{2-5}\noalign{\smallskip}
                                
                & \multirow{ 3 }{*}{\small{Making Plans}} & \multirow{ 3 }{*}{\makecell{\small{\textcolor{blue}{The speakers are} are talking to arrange a meeting or to do something together.}}}  & \scriptsize{\textcolor{blue}{S1 \& S2 are arranging a time to get together.}}  & \multirow{ 3 }{*}{\textcolor{blue}{\checkmark}} \\ 
                & & & \scriptsize{\textcolor{blue}{S1 \& S2 are discussing what to do together.}} & \\
                & & & \scriptsize{\textcolor{blue}{S1 \& S2 are coordinating their schedules.}} & \\
                \noalign{\smallskip}\cline{2-5}\noalign{\smallskip}
                                
                & \multirow{ 3 }{*}{\makecell{\small{Asking a favor}}}  & \multirow{ 3 }{*}{\makecell{\small{\textcolor{blue}{G: S1} is getting \textcolor{blue}{S2} to do something for them.}
                \\  \small{\textcolor{blue}{S1: The speaker is asking their interlocutor for a favor.}}
                \\ \small{\textcolor{blue}{S2: The speaker is considering whether to grant the favor requested by their interlocutor.}}}} & \scriptsize{\textcolor{blue}{S1 is requesting help from S2.}}  & \multirow{ 3 }{*}{\textcolor{Mulberry}{\xmark}} \\ 
                & & & \scriptsize{\textcolor{blue}{S1 trying to get S2 to do something for them.}} & \\
                & & & \scriptsize{\textcolor{blue}{S1 trying to get S2 to agree to a favor.}} & \\
                \noalign{\smallskip}\cline{2-5}\noalign{\smallskip}
                                
                & \multirow{ 3 }{*}{\makecell{\small{Asking out}}}  & \multirow{ 3 }{*}{\makecell{\small{\textcolor{blue}{G: S1} is asking \textcolor{blue}{S2} out on a date.}
                \\  \small{\textcolor{blue}{S1: The speaker is asking their interlocutor out on a date.}}
                \\ \small{\textcolor{blue}{S2: The speaker is considering whether to go on a date with their interlocutor.}}}} & \scriptsize{\textcolor{blue}{S1 is asking S2 to go on a date.}}  & \multirow{ 3 }{*}{\textcolor{Mulberry}{\xmark}} \\ 
                & & & \scriptsize{\textcolor{blue}{S1 is inviting S2 out.}} & \\
                & & & \scriptsize{\textcolor{blue}{S1 is asking S2 to spend time together romantically.}} & \\
                \noalign{\smallskip}\hline\noalign{\smallskip}
        \end{tabular}
    }
    \caption{Taxonomy of Speech Events of the category Goal-directed Talk.}
    \label{tab:goal-directed-speech-event-taxonomy}
\end{table*} 
 
\subsection{\textsc{Informal / Superficial Talk \textcolor{blue}{Augmented}}}
\label{appendix:informal-speech-event-taxonomy}
The underlined subcategory \textit{"Getting to know someone"} is the most common in personas-based ODD datasets hence restricting their actual openness. In this category's speech event types the speaker always have equivalent roles in the conversation's flow. 

\begin{table*}[h]
    \centering
    \resizebox{\textwidth}{!}{  
        \begin{tabular} {c|cclc}
            \hline
            \noalign{\smallskip}
            \textbf{Cat} & \textbf{Sub Category} & \textbf{Description} & \textbf{Reformulations} & \textbf{S1=S2} \\ 
            \noalign{\smallskip}\hline\noalign{\smallskip}
             \multirow{33}{*}{\rotatebox{90}{\textbf{Informal / Superficial Talk}}}
                & \multirow{ 3 }{*}{\small{Small Talk}} & \multirow{ 3 }{*}{\makecell{\small{\textcolor{blue}{The speakers} are passing time} \\ \small{and avoiding being rude.}}} & \scriptsize{\textcolor{blue}{Speaker 1 and Speaker 2 are making small talk to pass time.}} & \multirow{ 3 }{*}{\textcolor{blue}{\checkmark}} \\ 
                & &  & \scriptsize{\textcolor{blue}{S1 \& S2 are talking casually to be polite.}}   & \\
                & &  &\scriptsize{\textcolor{blue}{S1 \& S2 are chatting to avoid awkward silence.}}  & \\
                \noalign{\smallskip}\cline{2-5}\noalign{\smallskip}
                
                & \multirow{ 3 }{*}{\small{Currents events talk}} &  \multirow{ 3 }{*}{\makecell{\small{\textcolor{blue}{The speakers are} talking about} \\ \small{news and current events.}}}  & \scriptsize{\textcolor{blue}{S1 \& S2 are discussing today's top stories.}}  & \multirow{ 3 }{*}{\textcolor{blue}{\checkmark}} \\ 
                & &  & \scriptsize{\textcolor{blue}{S1 \& S2 are sharing opinions on the latest headlines.}} & \\
                & &  & \scriptsize{\textcolor{blue}{S1 \& S2 are conversing about what's happening in the world.}} & \\
                \noalign{\smallskip}\cline{2-5}\noalign{\smallskip}
                
                & \multirow{ 3 }{*}{\small{Gossip}} &  \multirow{ 3 }{*}{\makecell{\small{\textcolor{blue}{The speakers are} exchanging} \\ \small{opinions or information about someone} \\ \small{else when that person isn’t present.}}}  & \scriptsize{\textcolor{blue}{S1 \& S2 are sharing rumors about another person.}}  & \multirow{ 3 }{*}{\textcolor{blue}{\checkmark}} \\ 
                & &  & \scriptsize{\textcolor{blue}{S1 \& S2 are discussing someone else's business.}} & \\
                & &  & \scriptsize{\textcolor{blue}{S1 \& S2 talking about someone behind their back.}} & \\
                \noalign{\smallskip}\cline{2-5}\noalign{\smallskip}
                                
                & \multirow{ 3 }{*}{\small{Joking around}} &  \multirow{ 3 }{*}{\makecell{\small{\textcolor{blue}{The speakers are} engaging} \\ \small{in a playful kind of talk to} \\ \small{ have fun or release tension.}}}  & \scriptsize{\textcolor{blue}{S1 \& S2 are joking around for fun.}}  & \multirow{ 3 }{*}{\textcolor{blue}{\checkmark}} \\ 
                &  & & \scriptsize{\textcolor{blue}{S1 \& S2 are telling jokes to lighten the mood.}} & \\
                &  & & \scriptsize{\textcolor{blue}{S1 \& S2 are engaging in playful banter.}} & \\
                \noalign{\smallskip}\cline{2-5}\noalign{\smallskip}
                                
                & \multirow{ 3 }{*}{\small{Catching up}} & \multirow{ 3 }{*}{\makecell{\small{\textcolor{blue}{The speakers are} talking about} \\ \small{the events that have occurred} \\ \small{since they last spoke.}}}  & \scriptsize{\textcolor{blue}{S1 \& S2 are updating each other on their lives.}}  & \multirow{ 3 }{*}{\textcolor{blue}{\checkmark}} \\ 
                & & & \scriptsize{\textcolor{blue}{S1 \& S2 are talking about what's been happening.}} & \\
                & & & \scriptsize{\textcolor{blue}{S1 \& S2 are sharing what's new since they last spoke.}} & \\
                \noalign{\smallskip}\cline{2-5}\noalign{\smallskip}
                                
               & \multirow{ 3 }{*}{\makecell{\small{Recapping} \\ \small{the day’s events}}} & \multirow{ 3 }{*}{\makecell{\small{\textcolor{blue}{The speakers are} telling each} \\ \small{other about what’s happened} \\ \small{to them during the day.}}}  & \scriptsize{\textcolor{blue}{S1 \& S2 are talking about the highlights of their day.}}  & \multirow{ 3 }{*}{\textcolor{blue}{\checkmark}} \\ 
                & &  & \scriptsize{\textcolor{blue}{S1 \& S2 are discussing how their day went.}} & \\
                & & & \scriptsize{\textcolor{blue}{S1 \& S2 are recounting the day's experiences.}} & \\
                \noalign{\smallskip}\cline{2-5}\noalign{\smallskip}
                                
                & \multirow{ 3 }{*}{\makecell{\small{\uline{Getting to}} \\ \small{\uline{know someone}}}} & \multirow{ 3 }{*}{\makecell{\small{\textcolor{blue}{The speakers are} getting} \\ \small{acquainted with each other.}}}  & \scriptsize{\textcolor{blue}{S1 \& S2 are discussing today's top stories.}}  & \multirow{ 3 }{*}{\textcolor{blue}{\checkmark}} \\ 
                &  & & \scriptsize{\textcolor{blue}{S1 \& S2 are sharing opinions on the latest headlines}} & \\
                &  & & \scriptsize{\textcolor{blue}{S1 \& S2 are conversing about what's happening in the world.}} & \\
                \noalign{\smallskip}\cline{2-5}\noalign{\smallskip}
                                
                & \multirow{ 3 }{*}{\small{Sports talk}} & \multirow{ 3 }{*}{\makecell{\small{\textcolor{blue}{The speakers are}  talking about} \\ \small{playing or watching a sporting event.}}}  & \scriptsize{\textcolor{blue}{S1 \& S2 are discussing a sporting event.}}  & \multirow{ 3 }{*}{\textcolor{blue}{\checkmark}} \\ 
                & & & \scriptsize{\textcolor{blue}{S1 \& S2 are  analyzing the performance of a sports team.}} & \\
                & & & \scriptsize{\textcolor{blue}{S1 \& S2 are  debating the outcome of a recent game.}} & \\
                \noalign{\smallskip}\cline{2-5}\noalign{\smallskip}
                                
                & \multirow{ 3 }{*}{\small{Morning talk}} & \multirow{ 3 }{*}{\makecell{\small{\textcolor{blue}{The speakers are} engaging in } \\ \small{routine talk when waking} \\ \small{up in the morning. }}}  & \scriptsize{\textcolor{blue}{S1 \& S2 are discussing their plans for the day.}}  & \multirow{ 3 }{*}{\textcolor{blue}{\checkmark}} \\ 
                & & & \scriptsize{\textcolor{blue}{S1 \& S2 are talking as they start their day.}} & \\
                & & & \scriptsize{\textcolor{blue}{S1 \& S2 are sharing thoughts over breakfast.}} & \\
                \noalign{\smallskip}\cline{2-5}\noalign{\smallskip}
                                
                & \multirow{ 3 }{*}{\small{Bedtime talk}} & \multirow{ 3 }{*}{\makecell{\small{\textcolor{blue}{The speakers are} engaging in} \\ \small{routine talk right before going to bed.} }}  & \scriptsize{\textcolor{blue}{S1 \& S2 are sharing thoughts before going to sleep.}}  & \multirow{ 3 }{*}{\textcolor{blue}{\checkmark}} \\ 
                & & & \scriptsize{\textcolor{blue}{S1 \& S2 are discussing their day before bed.}} & \\
                & & & \scriptsize{\textcolor{blue}{S1 \& S2 are having a chat before bed}} & \\
                \noalign{\smallskip}\cline{2-5}\noalign{\smallskip}
                                
                & \multirow{ 3 }{*}{\small{Reminiscing}} & \multirow{ 3 }{*}{\makecell{\small{\textcolor{blue}{The speakers are} sharing events} \\ \small{they experienced together in the past.}}}  & \scriptsize{\textcolor{blue}{S1 \& S2 are talking about the good old days.}}  & \multirow{ 3 }{*}{\textcolor{blue}{\checkmark}} \\ 
                & & & \scriptsize{\textcolor{blue}{S1 \& S2 are reminiscing about past experiences.}} & \\
                & & & \scriptsize{\textcolor{blue}{S1 \& S2 are recalling memories they shared.}} & \\
                \noalign{\smallskip}\hline\noalign{\smallskip}
        \end{tabular}
    }
    \caption{Taxonomy of Speech Events of the category Informal/Superficial Talk.}
    \label{tab:informal-speech-event-taxonomy}
\end{table*}


\vspace{-0.2cm}
\section{Persona Profiles Taxonomy}
\setlength{\parskip}{-0.5em}

Underlined is what's relevant to taxonomy and in color-boxes represent what's specific to the language and associated culture and folk psychology in general.
\label{appendix:persona-taxonomy}


\subsection{\textsc{Wellness}}

\begin{table*}[h!]
    \scriptsize
    \centering

    \caption{Taxonomy of \textsc{Wellness} category, associated multi-polarised reformulations sentences for the prompt and examples in some languages.}
    \label{tab:wellness-taxo}
    \begin{tabular}{cc|cp{5.5cm}p{0.1cm}p{7.2cm}}
            \hline
            \multicolumn{2}{c}{\textbf{Cat}}  & \multicolumn{1}{c}{\textbf{Sub Category}}  & \multicolumn{1}{l}{\textbf{Sentences}}  & \multicolumn{2}{l}{\textbf{Examples}} \\
            \hline\noalign{\smallskip}
\multicolumn{2}{c|}{\multirow{ 6 }{*}{\rotatebox{90}{Symptom}}}
                & \multirow{ 3 }{*}{Physical} &  \textcolor{blue}{{a physical symptom you have}} &  \textcolor{Goldenrod}{FR}: & je souffre de rhumatismes articulaires. \\
                & & & \textcolor{blue}{{a bodily indication of a health issue you possess}}  &  &  \textit{(I suffer from \uline{joint rheumatism}.)} \\  
                &  & & \textcolor{blue}{{a symptom affecting your body}} & \textcolor{Periwinkle}{EN}: &  I'm constantly dealing with \uline{back pain}.  \\  
                \cline{3-6}
                
                & & \multirow{ 3 }{*}{Psychiatric} & \textcolor{blue}{{a psychiatric symptom you have}} & \textcolor{Goldenrod}{FR}: & j'ai des crises d'angoisse nocturnes. \\
                & & & \textcolor{blue}{{a psychological issue you experience}}   &  &  \textit{(I have \uline{nighttime anxiety attacks}.)} \\
                & & & \textcolor{blue}{{an emotional or mental concern you have}}  & \textcolor{Periwinkle}{EN}: &  I have \uline{anxiety during big crowds}. \\
                \hline
\multicolumn{6}{r}{\scriptsize\sl continued on next page}\\
\end{tabular}
\end{table*}

\begin{table*}[h!]
    \scriptsize
    \centering
    \begin{tabular}{cc|cp{5.5cm}p{0.1cm}p{7.2cm}}
            \multicolumn{6}{l}{\scriptsize\sl continued from previous page}\\
            \hline
            \multicolumn{2}{c}{\textbf{Cat}}  & \multicolumn{1}{c}{\textbf{Sub Category}}  & \multicolumn{1}{l}{\textbf{Sentences}}  & \multicolumn{2}{l}{\textbf{Examples}} \\
            \hline\noalign{\smallskip}
\multicolumn{2}{c|}{\multirow{ 3 }{*}{\rotatebox{90}{Disease}}}
                & \multirow{ 3 }{*}{/} & \textcolor{blue}{{a disease you had or currently have}} & \textcolor{Goldenrod}{FR}: & j'ai eu la bronchite \`a l'\^age de 10 ans.  \\
                & & & \textcolor{blue}{{a skin condition you had or currently have}} &  &  \textit{(I had \uline{bronchitis} at age 10.)} \\
                & & & \textcolor{blue}{{a digestive$|$respiratory disease you had or currently have}} & \textcolor{Periwinkle}{EN}: &  I'm \uline{living with} type \uline{1 diabetes}.  \\
            \noalign{\smallskip}\hline

            \multicolumn{6}{c}{\textcolor{blue}{{\textsc{$\downarrow \downarrow$ New SubCategories from Taxonomy Augmentation $\downarrow \downarrow$ }}}} \\
            \hline \noalign{\smallskip}

            \multicolumn{2}{c|}{\multirow{ 9 }{*}{\rotatebox{90}{\makecell{Preventive \\Care}}}}
            
                & \multirow{ 3 }{*}{Check-ups} & {{your health check-ups if any}} & \textcolor{Goldenrod}{FR}: & je fais des \'echographies cardiaques tous les ans.  \\
                & & & {{your routine medical exams}}  &  &  \textit{(I get \uline{cardiac ultrasounds every year}.)} \\
                & & & {{your preventive health care practices}} & \textcolor{Periwinkle}{EN}: &  I see \uline{my dentist twice a year for cleanings}.  \\
                \noalign{\smallskip}\cline{3-6}\noalign{\smallskip}
                
                & & \multirow{ 3 }{*}{Vaccinations} & {{a vaccine you took or not}}  & \textcolor{Goldenrod}{FR}: & j'ai reçu ma 3e dose de vaccin contre la grippe.   \\
                & & & {{your vaccination status}} &  &  \textit{(I received my \uline{3rd flu vaccine dose}.)} \\
                & & & {{your stance on vaccines}} & \textcolor{Periwinkle}{EN}: &  I'm \uline{not vaccinated against the HPV} virus.   \\
                \noalign{\smallskip}\cline{3-6}\noalign{\smallskip}
                
                & & \multirow{ 3 }{*}{\makecell{Alternative \\ Medication}} & {{your use of herbal remedies}} & \textcolor{Goldenrod}{FR}: & j'utilise l'acupuncture pour soulager mon dos.  \\
                & & & {{your experience with naturopathy}}  &  &  \textit{(I use \uline{acupuncture} to relieve my back pain.)} \\ 
                & & & {{your belief in alternative healing methods}} & \textcolor{Periwinkle}{EN}: &  i've used \uline{essential oils} to relieve anxiety.  \\
            \noalign{\smallskip}\hline\noalign{\smallskip}
            
            \multicolumn{2}{c|}{\multirow{3}{*}{\rotatebox{90}{\makecell{Mental \\ Health }}}}
                & \multirow{ 3 }{*}{\makecell{Therapy \\ / \\ Counseling}} & {\tiny{any mental therapy or counseling experiences you had or wish to}} & \textcolor{Goldenrod}{FR}: & je vais au psychologue chaque semaine.  \\
                & & & {{your mental health treatment}} &  &  \textit{(I \uline{see a psychologist weekly}.)} \\ 
                & & & {{your use of psychological services}} & \textcolor{Periwinkle}{EN}: &  I've \uline{been in therapy for post-traumatic stress}. \\
            \noalign{\smallskip}\hline\noalign{\smallskip}
            
            \multicolumn{2}{c|}{\multirow{3}{*}{\rotatebox{90}{\makecell{Fitness}}}}
                &  \multirow{ 3 }{*}{\makecell{Exercise\\Routine}} &{{your exercise routine if any}}  & \textcolor{Goldenrod}{FR}: & j'adore faire du v\'elo le dimanche matin sur les berges de la seine.  \\
                & & & {{how often you work out}} &  &  \textit{(I love riding \uline{my bike on Sunday mornings} along the \colorbox{Goldenrod}{banks of the Seine}.)} \\
                & & & {{your physical activity level}} & \textcolor{Periwinkle}{EN}: &  I go to the \uline{gym three times a week}.  \\
            \noalign{\smallskip}\hline\noalign{\smallskip}
            
            \multicolumn{2}{c|}{\multirow{6}{*}{\rotatebox{90}{\makecell{Diet}}}}
                &  \multirow{ 3 }{*}{\makecell{Dietary \\ Restrictions}} & {{your food allergies or intolerances if any}}  & \textcolor{Goldenrod}{FR}: & je suis  intol\'erant aux gluten.   \\
                & & & {{your adherence to specific dietary plans}} &  &  \textit{(I am \uline{gluten intolerant}.)} \\
                & & & {{your special dietary needs}} & \textcolor{Periwinkle}{EN}: &  I have a \uline{severe allergy to shellfish}.  \\
                \noalign{\smallskip}\cline{3-6}\noalign{\smallskip}
                
                &  & \multirow{ 3 }{*}{\makecell{Nutritional \\ Habits}} &  {{your nutritional habits}}  &  \textcolor{Goldenrod}{FR}: & j'adore manger des cr\^epes bretonnes.  \\
                & & &  {{your meal patterns}} &  &  \textit{(I love eating \colorbox{Goldenrod}{\uline{Breton crepes}}.)} \\
                & & &  {{your approach to nutrition}} & \textcolor{Periwinkle}{EN}: &  I eat a \uline{vegan diet} for environmental reasons. \\
            \noalign{\smallskip}\hline\noalign{\smallskip}
            
            \multicolumn{2}{c|}{\multirow{3}{*}{\rotatebox{90}{\makecell{Sleep}}}}
                &  \multirow{ 3 }{*}{Sleep Quality} & {{the amount of sleep you get}}  & \textcolor{Goldenrod}{FR}: & je m'endors \`a 22h00 avec une lecture.  \\
                & & & {{your experiences with insomnia}} &  &  \textit{(I \uline{fall asleep at 10:00 PM} with a book.)} \\
                & & & {{your methods for improving sleep}} & \textcolor{Periwinkle}{EN}: &  I've \uline{struggled with insomnia} since college. \\
            \noalign{\smallskip}\hline\noalign{\smallskip}
            
            \multicolumn{2}{c|}{\multirow{12}{*}{\rotatebox{90}{\makecell{Substance \\ Use}}}}
                & \multirow{ 3 }{*}{Smoking} &{{your smoking cessation}}  & \textcolor{Goldenrod}{FR}: & je ne fume que des Gauloises.  \\
                & & & {{your smoking habits or routine}}  &  &  \textit{(I \uline{only smoke} \colorbox{Goldenrod}{\uline{Gauloises}}.)} \\
                & & & {{whether you've stopped or kept smoking}} & \textcolor{Periwinkle}{EN}: &  I \uline{used to be a heavy smoker}, but \uline{quit last year}.  \\
                \noalign{\smallskip}\cline{3-6}\noalign{\smallskip}
                
                & & \multirow{ 3 }{*}{\makecell{Alcohol \\ Consumption}} &  {{your alcohol consumption or not}}  & \textcolor{Goldenrod}{FR}: & je bois rarement de bi\`ere, je pr\'ef\`ere le vin.  \\
                & & &  {{your typical drinking patterns}} &  &  \textit{(I \uline{rarely drink beer}, I \colorbox{Goldenrod}{\uline{prefer wine}}.)} \\
                & & &  {{your tendency to abstain from alcohol}} & \textcolor{Periwinkle}{EN}: &  I \uline{don't drink}, I'm more of a tea person.   \\
                \noalign{\smallskip}\cline{3-6}\noalign{\smallskip}
                
                & & \multirow{ 3 }{*}{\makecell{Drug \\ Use}} & {{your experiences with dope use or not}} & \textcolor{Goldenrod}{FR}: & je ne consomme pas de cannabis.  \\
                & & & {{your perspective on illegal substance abuse}} &  &  \textit{(I \uline{don't use cannabis}.)} \\ 
                & & & {{your attitude towards recreational substances}} & \textcolor{Periwinkle}{EN}: &  I \uline{think marijuana is overrated}.  \\
            \noalign{\smallskip}\hline\noalign{\smallskip}

            \multicolumn{2}{c|}{\multirow{3}{*}{\rotatebox{90}{\makecell{Medical \\ History}}}}
                & \multirow{ 3 }{*}{/} &{{significant health events in your past}} & \textcolor{Goldenrod}{FR}: & j'ai eu une op\'eration pour enlever un kyste sur mon pied.   \\
                & & & {{your surgeries past or scheduled if any}} &  &  \textit{(I had a \uline{cyst removed} from my foot.)} \\
                & & & {{your family health issues}} & \textcolor{Periwinkle}{EN}: &  I \uline{had a tonsillectomy when I was 10}.   \\
            \noalign{\smallskip}\hline\noalign{\smallskip}

\end{tabular}
\end{table*}
\subsection{\textsc{Psychographics}}

\begin{table*}[h!]
    \scriptsize
    \centering

        \caption{Taxonomy of \textsc{Psychographics} category, associated multi-polarised reformulations sentences for the prompt and examples in some languages}
        \label{tab:pyschographics-taxo}
        \begin{tabular}{ll|cccp{6.3cm}p{0.1cm}p{6.3cm}}
            \hline
            \multicolumn{2}{c}{\textbf{Cat}}  & \multicolumn{3}{c}{\textbf{Sub Category}}  & \multicolumn{1}{l}{\textbf{Sentences}}  & \multicolumn{2}{l}{\textbf{Examples}} \\
            \hline\noalign{\smallskip}
\multicolumn{2}{c|}{\multirow{ 30 }{*}{\rotatebox{90}{Preferences}}}
                & \multicolumn{3}{c}{\multirow{3}{*}{\makecell{Movie \\  genre }}} & {\textcolor{blue}{your favorite type of movie}} &  \textcolor{Goldenrod}{FR}: & j'ai un faible pour les thrillers.  \\ 
                & & & & & {\textcolor{blue}{movie genres you avoid}} &  &  \textit{(I \uline{have a soft spot} for \underline{thrillers}.)}  \\ 
                & & & & & {\textcolor{blue}{the kind of films you enjoy or not}} & \textcolor{Periwinkle}{EN}: &  I love watching \uline{Marvel superhero movies.}   \\ 
                \noalign{\smallskip}\cline{3-8}\noalign{\smallskip}

                & & \multicolumn{3}{c}{\multirow{ 3 }{*}{\makecell{Movie \\ title}}} & {\textcolor{blue}{the title of a movie you avoid}} & \textcolor{Periwinkle}{EN}: &  I \uline{dislike} movies like the \colorbox{Periwinkle}{\uline{Expendables}} franchise. \\ 
                & & & & & {\textcolor{blue}{the movies you love or hate}} & \textcolor{Goldenrod}{FR}: &  mon film préféré est amélie poulain.  \\ 
                & & & & & {\textcolor{blue}{your favorite film}} &  & \textit{(my \uline{favorite movie} is \colorbox{Goldenrod}{\uline{Amélie Poulain}.})}   \\ 
                \noalign{\smallskip}\cline{3-8}\noalign{\smallskip}

                & & \multicolumn{3}{c}{\multirow{ 3 }{*}{\makecell{Movie \\ director}}} & {\textcolor{blue}{a movie director whose work you admire}} & \textcolor{Periwinkle}{EN}: &  I \uline{dislike Stanley Kubrick}'s movies, too long.  \\ 
                & & & & & {\textcolor{blue}{your favorite filmmakers}} & \textcolor{Goldenrod}{FR}: &  je ne regarde pas les films de luc besson.  \\ 
                & & & & & {\textcolor{blue}{a filmmaker you tend to avoid}} &  & \textit{(I don’t watch \uline{movies by }\colorbox{Goldenrod}{\uline{Luc Besson.}})}   \\ 
                \noalign{\smallskip}\cline{3-8}\noalign{\smallskip}

                & & \multicolumn{3}{c}{\multirow{ 3 }{*}{\makecell{Book \\  author }}} &{\textcolor{blue}{your favorite writer}} & \textcolor{Goldenrod}{FR}: &  Je suis un grand fan de Michel Hoellebecq  \\ 
                & & & & & {\textcolor{blue}{an author you don't enjoy}} &  & \textit{(I am a big fan of \colorbox{Goldenrod}{\uline{Michel Hoellebecq}}.)}  \\ 
                & & & & & {\textcolor{blue}{a novelist you admire}} & \textcolor{Periwinkle}{EN}: &  my go-to author is  \colorbox{Periwinkle}{\uline{george orwell}}.   \\ 
                \noalign{\smallskip}\cline{3-8}\noalign{\smallskip}

                & & \multicolumn{3}{c}{\multirow{ 3 }{*}{\makecell{Book \\ genre}}} & {\textcolor{blue}{book genres you avoid}} & \textcolor{Periwinkle}{EN}: &  i enjoy reading \uline{sci-fi novels}. \\ 
                & & & & & {\textcolor{blue}{your preferred or least liked book genre}} & \textcolor{Goldenrod}{FR}: &  je préfère les romans policiers de frederic dard.  \\ 
                & & & & & {\textcolor{blue}{your favorite types of books}} &  & \textit{(I prefer \uline{detective novels} by \colorbox{Goldenrod}{Frederic Dard}.)}   \\ 
                \noalign{\smallskip}\cline{3-8}\noalign{\smallskip}

                & & \multicolumn{3}{c}{\multirow{ 3 }{*}{\makecell{Book \\ title}}} & {\textcolor{blue}{the title of a book you enjoy the most}} & \textcolor{Periwinkle}{EN}: &  My favorite book is \colorbox{Periwinkle}{\uline{"The Hitchhiker's Guide to the Galaxy"}}. \\ 
                & & & & & {\textcolor{blue}{your favorite books}} & \textcolor{Goldenrod}{FR}: &  je préfère "l'étranger" d'camus \\ 
                & & & & & {\textcolor{blue}{a book you dislike}} &  &  \textit{(I prefer \colorbox{Goldenrod}{\uline{"L'Étranger"}} by \colorbox{Goldenrod}{\uline{Camus}}.)}  \\ 
                \noalign{\smallskip}\cline{3-8}\noalign{\smallskip}

                & & \multicolumn{3}{c}{\multirow{ 3 }{*}{ Show}} & {\textcolor{blue}{your favorite television programs}} & \textcolor{Goldenrod}{FR}: &  je déteste "Koh-Lanta".   \\ 
                & & & & & {\textcolor{blue}{the series you enjoy the most}} &  &\textit{(I hate \colorbox{Goldenrod}{\uline{"Koh-Lanta"}}.)} \\ 
                & & & & & {\textcolor{blue}{a TV show you dislike}} & \textcolor{Periwinkle}{EN}: &  i \uline{dislike reality TV shows} like \colorbox{Periwinkle}{\uline{survivor}}. \\ 
                \hline

                \multicolumn{8}{r}{\scriptsize\sl continued on next page}\\
\end{tabular}
\end{table*}

\begin{table*}[h!]
    \scriptsize
    \centering
    \begin{tabular}{ll|cccp{6.3cm}p{0.1cm}p{6.3cm}}
            \multicolumn{8}{l}{\scriptsize\sl continued from previous page}\\
            \hline
            \multicolumn{2}{c}{\textbf{Cat}}  & \multicolumn{3}{c}{\textbf{Sub Category}}  & \multicolumn{1}{l}{\textbf{Sentences}}  & \multicolumn{2}{l}{\textbf{Examples}} \\
            \hline\noalign{\smallskip}
\multicolumn{2}{c|}{\multirow{ 61 }{*}{\rotatebox{90}{Preferences}}}
                & \multicolumn{3}{c}{\multirow{ 3 }{*}{Activity}}  &  {\textcolor{blue}{your leisure activities}} & \textcolor{Goldenrod}{FR}: &  j'aime faire du vélo dans les Pyrénées.  \\ 
                & & & & &  {\textcolor{blue}{your favorite outdoor activity}} &  & \textit{(I \uline{like cycling} in \colorbox{Goldenrod}{the Pyrenees}.)}  \\ 
                & & & & &  {\textcolor{blue}{your preferred or least liked social activity}} & \textcolor{Periwinkle}{EN}: &  i \uline{enjoy playing} \colorbox{Periwinkle}{\uline{soccer}} with my friends.   \\ 
                 \noalign{\smallskip}\cline{3-8}\noalign{\smallskip}

                & & \multicolumn{3}{c}{\multirow{ 3 }{*}{Season}} & {\textcolor{blue}{the time of year you enjoy the most}} & \textcolor{Periwinkle}{EN}: &  \uline{autumn} is the \uline{worst season} for allergies.   \\ 
                & & & & & {\textcolor{blue}{your preferred or least liked season}} & \textcolor{Goldenrod}{FR}: &  je déteste l'hiver.   \\ 
                & & & & & {\textcolor{blue}{the season you prefer the least}} &  &  \textit{(I \uline{hate winter}.)}  \\ 
                 \noalign{\smallskip}\cline{3-8}\noalign{\smallskip}
                
                & & \multicolumn{3}{c}{\multirow{ 3 }{*}{\makecell{Music \\ instrument}}} & {\textcolor{blue}{the musical tools you avoid}} & \textcolor{Periwinkle}{EN}: &  i \uline{enjoy listening} to \uline{acoustic guitar} music. \\ 
                & & & & & {\textcolor{blue}{your favorite instruments to play or listen to}} & \textcolor{Goldenrod}{FR}: &  je déteste le saxophone  \\ 
                & & & & & {\textcolor{blue}{your preferred or least liked music instrument}} &  &  \textit{(I \uline{hate the saxophone}.)}  \\ 
                 \noalign{\smallskip}\cline{3-8}\noalign{\smallskip}

                & & \multicolumn{3}{c}{\multirow{ 3 }{*}{\makecell{Music \\ genre}}} & {\textcolor{blue}{your preferred or least liked music genre}} & \textcolor{Periwinkle}{EN}: &  i love \uline{indie rock} music.  \\ 
                & & & & & {\textcolor{blue}{music genres you dislike}} & \textcolor{Goldenrod}{FR}: &  je n'aime pas le jazz.  \\ 
                & & & & & {\textcolor{blue}{the types of music you enjoy}} &  &  \textit{(I don’t like \uline{jazz}.)}  \\ 
                 \noalign{\smallskip}\cline{3-8}\noalign{\smallskip}

                & & \multicolumn{3}{c}{\multirow{ 3 }{*}{\makecell{Music \\ artist}}}
                & {\textcolor{blue}{the musician you admire}} & \textcolor{Goldenrod}{FR}: &  j'dore les albums de Claude François. \\ 
                & & & & & {\textcolor{blue}{your most loved singers or bands}}  &  &  \textit{(I \uline{love the albums} of \colorbox{Goldenrod}{\uline{Claude François}}.)}  \\ 
                & & & & & {\textcolor{blue}{the artist you avoid listening to}} & \textcolor{Periwinkle}{EN}: &  i listen to \colorbox{Periwinkle}{\uline{the 1975}} on repeat.   \\ 
                 \noalign{\smallskip}\cline{3-8}\noalign{\smallskip}
                
                & & \multicolumn{3}{c}{\multirow{ 3 }{*}{Color}} & {\textcolor{blue}{the colors you dislike}} & \textcolor{Periwinkle}{EN}: &  I'm \uline{not a fan of} the color \uline{green}. \\ 
                & & & & & {\textcolor{blue}{the color you prefer the least}} & \textcolor{Goldenrod}{FR}: &  mon coloris préféré est le vert foncé   \\ 
                & & & & & {\textcolor{blue}{your favorite color}} &  & \textit{(my \uline{favorite color} is \uline{dark green}.)}  \\ 
                
                 \noalign{\smallskip}\cline{3-8}\noalign{\smallskip}
                
                & & \multicolumn{3}{c}{\multirow{ 3 }{*}{Animal}} & {\textcolor{blue}{an animal you find fascinating or repellent}} &  \textcolor{Periwinkle}{EN}: &  I'm \uline{fascinated} by the behavior of \uline{dolphins} in the wild. \\ 
                & & & & & {\textcolor{blue}{your favorite or least liked animals}} & \textcolor{Goldenrod}{FR}: &  les crocodiles me font peur.  \\ 
                & & & & & {\textcolor{blue}{your interest in wildlife}} &  & \textit{(\uline{Crocodiles} scare me.)}  \\ 
                 \noalign{\smallskip}\cline{3-8}\noalign{\smallskip}
                
                & & \multicolumn{3}{c}{\multirow{ 3 }{*}{\makecell{Location \\ / \\ Place}}} &  {\textcolor{blue}{locations you dislike}} & \textcolor{Periwinkle}{EN}: &  i love exploring the  \colorbox{Periwinkle}{\uline{mountains of colorado}}.  \\ 
                & & & & &  {\textcolor{blue}{locations you avoid}} & \textcolor{Goldenrod}{FR}: &  Je adore passer mes week-ends à la plage de Saint-Tropez.   \\ 
                & & & & &  {\textcolor{blue}{places you love to visit}} &  &  \textit{(I love spending my weekends at \colorbox{Goldenrod}{\uline{Saint-Tropez beach}}.)} \\ 
                 \noalign{\smallskip}\cline{3-8}\noalign{\smallskip}

                & & \multicolumn{3}{c}{\multirow{ 3 }{*}{Sport}} & {\textcolor{blue}{your preferred or least liked sport}} & \textcolor{Periwinkle}{EN}: &  i'm a huge \uline{fan of} \colorbox{Periwinkle}{\uline{soccer}}.  \\ 
                & & & & & {\textcolor{blue}{the sports you enjoy}} & \textcolor{Goldenrod}{FR}: &  J'adore jouer au pétanque avec mes amis tous les dimanches.  \\ 
                & & & & & {\textcolor{blue}{a sport you don't like}} &  &  \textit{(I enjoy playing \colorbox{Goldenrod}{\uline{pétanque}} with my friends every Sunday.)} \\ 
                \noalign{\smallskip}\cline{3-8}\noalign{\smallskip}

                & & \multicolumn{3}{c}{\multirow{ 3 }{*}{Food}} & {\textcolor{blue}{the cuisines you enjoy or not}} &  \textcolor{Periwinkle}{EN}: &  i \uline{love} eating \colorbox{Periwinkle}{\uline{fish and chips}} from the seaside.  \\ 
                & & & & & {\textcolor{blue}{foods you avoid}} & \textcolor{Goldenrod}{FR}: &  je mange souvent des croissants.   \\ 
                & & & & &  {\textcolor{blue}{your favorite dishes}} &  & \textit{(I often eat \colorbox{Goldenrod}{\uline{croissants}}.)}  \\ 
                 \noalign{\smallskip}\cline{3-8}\noalign{\smallskip}

                & & \multicolumn{3}{c}{\multirow{ 3 }{*}{Drink}} & {\textcolor{blue}{your favorite drink}} & \textcolor{Periwinkle}{EN}: &  i \uline{hate} the taste of \colorbox{Periwinkle}{\uline{Earl Grey tea}}.  \\ 
                & & & & & {\textcolor{blue}{your preferred or least liked drink}} & \textcolor{Goldenrod}{FR}: &  je déteste le café glacé.  \\ 
                & & & & &  {\textcolor{blue}{beverages you avoid}} &  &  \textit{(I \colorbox{Goldenrod}{\uline{hate}} \uline{iced coffee}.)}  \\ 
                 \noalign{\smallskip}\cline{3-8}\noalign{\smallskip}

                & & \multicolumn{3}{c}{\multirow{ 3 }{*}{\makecell{Media \\ genre}}}
                &   {\textcolor{blue}{the kind of media do you find most engaging}} & \textcolor{Periwinkle}{EN}: &  i enjoy \uline{true crime podcasts}.  \\ 
                & & & & & {\textcolor{blue}{your preferred or least liked media genre}} & \textcolor{Goldenrod}{FR}: &  je déteste les émissions de télé-réalité   \\ 
                & & & & &  {\textcolor{blue}{the media genres you avoid}} &  &  \textit{(I \uline{hate} \uline{reality TV shows}.)} \\ 
                 \noalign{\smallskip}\cline{3-8}\noalign{\smallskip}
                
                & & \multicolumn{3}{c}{\multirow{ 3 }{*}{\textcolor{blue}{\makecell{Education \\  Methods}}}} & {\textcolor{blue}{your learning style}} &   \textcolor{Goldenrod}{FR}: &  j'adore l'apprentissage en classe mixte.  \\ 
                & & & & & {\textcolor{blue}{your preferred or least liked teaching methods}}  &  & \textit{(I love \colorbox{Goldenrod}{\uline{learning in "classe mixte".}})}   \\ 
                & & & & & {\textcolor{blue}{how you best learn}}  & \textcolor{Periwinkle}{EN}: &  I learn best through \uline{hands-on experience}. \\ 
                 \noalign{\smallskip}\cline{3-8}\noalign{\smallskip}

                & & \multicolumn{3}{c}{\multirow{ 3 }{*}{\textcolor{blue}{\makecell{Favorite \\  Apps}}}} & {\textcolor{blue}{your favorite apps if any}} &  \textcolor{Goldenrod}{FR}: &  mon application préférée est Waze.  \\ 
                & & & & & {\textcolor{blue}{your go-to digital tools and platforms}}  &  &  \textit{(my favorite app is \uline{Waze}.)}  \\ 
                & & & & & {\textcolor{blue}{the most useful app of yours}}  &  \textcolor{Periwinkle}{EN}: &   i'm \uline{active} on \uline{instagram} and \uline{tiktok} mostly. \\ 
                \noalign{\smallskip}\hline\noalign{\smallskip}

\multicolumn{2}{c|}{\multirow{ 30 }{*}{\rotatebox{90}{\makecell{Personal Characteristics}}}}
                & \multicolumn{3}{c}{\multirow{ 3 }{*}{\makecell{Physical \\ Attribute}}} & {\textcolor{blue}{a specific aspect of your physical appearance}}  &  \textcolor{Periwinkle}{EN}: &  i have a \uline{scar above my left eyebrow}.  \\ 
                & & & & & {\textcolor{blue}{a particular aspect of your body you don't like}}  & \textcolor{Goldenrod}{FR}: &  J'ai un tatouage de la tour Eiffel sur mon épaule gauche. \\ 
                & & & & & {\textcolor{blue}{one of your physical traits, noticeable or subtle}}  &  &  \textit{(I \uline{have a tattoo} of the \colorbox{Goldenrod}{\uline{Eiffel Tower}} on my left shoulder.)} \\ 
                 \noalign{\smallskip}\cline{3-8}\noalign{\smallskip}
            
                & & \multicolumn{2}{c}{\multirow{ 24 }{*}{\rotatebox{90}{Personality Trait}}}  & \multirow{ 3 }{*}{/} &  {\textcolor{blue}{any dimension of your personality}} &  \textcolor{Periwinkle}{EN}: &  i'm a bit of \uline{a control freak}. \\ 
                & & & & &  {\textcolor{blue}{your level  of empathy/agreeableness or Neuroticism}} & \textcolor{Goldenrod}{FR}: &  je suis un peu trop sensible.  \\ 
                & & & & &  {\textcolor{blue}{your self-awareness or conscientiousness}} &  &  \textit{(I’m a bit too \uline{sensitive}.)} \\ 
                \noalign{\smallskip}\cline{5-8}\noalign{\smallskip}
                
                & & & & \multirow{ 3 }{*}{\textcolor{blue}{\makecell{Decision\\-Making \\ Style}}} & {\textcolor{blue}{your decision-making style: intuitive, rational or collaborative}} &  \textcolor{Periwinkle}{EN}: &  I prefer to \uline{think things through} before acting impulsively. \\ 
                & & & & &  {\textcolor{blue}{how you make decisions}} & \textcolor{Goldenrod}{FR}: &  je prends des décisions en écoutant mes émotions.  \\ 
                & & & & &  {\textcolor{blue}{your method of making choices}} &  &  I make decisions by \uline{listening to my emotions}.  \\ 
                \noalign{\smallskip}\cline{5-8}\noalign{\smallskip}
                
                & & & & \multirow{ 3 }{*}{\textcolor{blue}{\makecell{Communi-\\ cation \\ Style}}} & {\textcolor{blue}{your preferred communication methods}} &  \textcolor{Periwinkle}{EN}: &  i'm a naturally \uline{quiet person in large groups}. \\ 
                & & & & &  \textcolor{blue}{how you express yourself} & \textcolor{Goldenrod}{FR}: & je préfère les conversations en direct.   \\ 
                & & & & &  {\textcolor{blue}{your verbal or non-verbal communication style}} &  &  \textit{(I prefer \uline{face-to-face conversations}.)} \\ 
                \noalign{\smallskip}\cline{5-8}\noalign{\smallskip}
                
                & & & & \multirow{ 3 }{*}{\textcolor{blue}{\makecell{Problem \\ Solving}}} & {\textcolor{blue}{your analytical thinking skills}} & \textcolor{Periwinkle}{EN}: &  i'm \uline{not very good at mediating} conflicts. \\ 
                & & & & &  {\textcolor{blue}{your creativity in finding solutions or not}} & \textcolor{Goldenrod}{FR}: &  je suis très calme en situation de crise.   \\ 
                & & & & &  {\textcolor{blue}{your peacekeeping abilities or not}} &  &  \textit{(I’m \uline{very calm} in \uline{crisis situations}.)}  \\ 
                \noalign{\smallskip}\cline{5-8}\noalign{\smallskip}
                
                & & & & \multirow{ 3 }{*}{\textcolor{blue}{Resilience}} & {\textcolor{blue}{your adaptability skills}} & \textcolor{Periwinkle}{EN}: &  i \uline{struggle to set realistic goals} for myself sometimes. \\ 
                & & & & &  {\textcolor{blue}{your ability to overcome adversity or not}} & \textcolor{Goldenrod}{FR}: &  je suis capable de changer de projet rapidement.   \\ 
                & & & & &  {\textcolor{blue}{your coping strategies}} &  &  \textit{((I am \uline{capable of quickly switching projects}.))} \\ 
                \noalign{\smallskip}\cline{5-8}\noalign{\smallskip}
                
                & & & & \multirow{ 3 }{*}{\textcolor{blue}{Creativity}} & {\textcolor{blue}{your imagination and originality}} & \textcolor{Periwinkle}{EN}: &  i'm \uline{decent at painting watercolors} \\ 
                & & & & &  {\textcolor{blue}{your artistic skills}} & \textcolor{Goldenrod}{FR}: &  je compose des chansons sur mon ukulélé.  \\ 
                & & & & &  {\textcolor{blue}{your innovative thinking}} &  &  \textit{(I \uline{compose songs} on my ukulele.)} \\ 
                \noalign{\smallskip}\cline{5-8}\noalign{\smallskip}
                
                & & & & \multirow{ 3 }{*}{\textcolor{blue}{\makecell{Core \\ Values}}} & {\textcolor{blue}{your core values or moral standpoints}} & \textcolor{Periwinkle}{EN}: &  My \uline{moral compass} is centered around \uline{being honest} \\ 
                & & & & &  {\textcolor{blue}{your ethical principles}} & \textcolor{Goldenrod}{FR}: &  Je tiens à mon indépendance et à mes principes.  \\ 
                & & & & &  {\textcolor{blue}{your fundamental beliefs}} &  &  \textit{(I \uline{value} my \uline{independence} and my \uline{principles}.)}  \\ 
                \noalign{\smallskip}\cline{5-8}\noalign{\smallskip}
                
                & & & & \multirow{ 3 }{*}{\textcolor{blue}{\makecell{Cognitive \\ Abilities}}} & {\textcolor{blue}{your intellectual capabilities}} & \textcolor{Periwinkle}{EN}: &  I have \uline{trouble focusing during long meetings}.  \\ 
                & & & & &  {\textcolor{blue}{your level of attention}} & \textcolor{Goldenrod}{FR}: &   j'ai une mémoire incroyable pour les chansons.  \\ 
                & & & & &  {\textcolor{blue}{your memory abilities}} &  &  \textit{(I have an \uline{incredible memory for songs}.)} \\ 
                \hline

                \multicolumn{8}{r}{\scriptsize\sl continued on next page}\\
\end{tabular}
\end{table*}

\begin{table*}[h!]
    \scriptsize
    \centering
    \begin{tabular}{ll|cccp{6.3cm}p{0.1cm}p{6.3cm}}
            \multicolumn{8}{l}{\scriptsize\sl continued from previous page}\\
            \hline
            \multicolumn{2}{c}{\textbf{Cat}}  & \multicolumn{3}{c}{\textbf{Sub Category}}  & \multicolumn{1}{l}{\textbf{Sentences}}  & \multicolumn{2}{l}{\textbf{Examples}} \\
            \hline\noalign{\smallskip}
\multicolumn{2}{c|}{\multirow{ 20 }{*}{\rotatebox{90}{\makecell{Personal Characteristics}}}}
            & \multicolumn{2}{c}{\multirow{ 6 }{*}{\rotatebox{90}{\makecell{Financial \\ Awareness}}}}  & \multirow{ 3 }{*}{Budgeting} &  {\textcolor{blue}{your saving habits}} & \textcolor{Periwinkle}{EN}: &  i'm \uline{learning about real estate investing}, hoping to flip a house.  \\ 
                & & & & &  {\textcolor{blue}{how you manage your finances}} & \textcolor{Goldenrod}{FR}: &  je fais l'épargne pour acheter un appartement à Nice. \\ 
                & & & & &  {\textcolor{blue}{your investing knowledge}} &  &  \textit{(I am \uline{saving} to buy an appartment in \colorbox{Goldenrod}{Nice}.)} \\ 
                \noalign{\smallskip}\cline{5-8}\noalign{\smallskip}
                
                & & & & \multirow{ 3 }{*}{\textcolor{blue}{\makecell{Spending\\Habits}}} & {\textcolor{blue}{your shopping tendencies}} & \textcolor{Periwinkle}{EN}: &  i always \uline{look for sales} when shopping at \colorbox{Periwinkle}{\uline{Target}}.  \\ 
                & & & & &  {\textcolor{blue}{your consumer behavior}} & \textcolor{Goldenrod}{FR}: &  je préfère payer cash pour éviter les intérêts.  \\ 
                & & & & &  {\textcolor{blue}{your spending habits}} &  &  \textit{(I prefer \uline{paying cash} to avoid interest.)}  \\ 
                 \noalign{\smallskip}\cline{3-8}\noalign{\smallskip}

            & & \multicolumn{3}{c}{\multirow{ 3 }{*}{\textcolor{blue}{Lifestyle}}} & {\textcolor{blue}{your typical day}}  & \textcolor{Periwinkle}{EN}: &  my \uline{typical day} starts with a \uline{5am jog} along the beach \\ 
                & & & & & {\textcolor{blue}{your day-to-day life}}  & \textcolor{Goldenrod}{FR}: &  je prends le métro tous les jours.   \\ 
                & & & & & {\textcolor{blue}{your regular schedule}}  &  &  \textit{(I take \uline{the subway every day}.)}   \\ 
                 \noalign{\smallskip}\cline{3-8}\noalign{\smallskip}
                
                & & \multicolumn{3}{c}{\multirow{ 3 }{*}{\textcolor{blue}{\makecell{Social \\ Connections }}}} & {\textcolor{blue}{your social bonding skills}} &  \textcolor{Goldenrod}{FR}: &  je suis très lié avec mes copains de la fac à strasbourg. \\ 
                & & & & & {\textcolor{blue}{your network of friends}} &  &  I am \uline{very close with my friends} from \colorbox{Goldenrod}{university in Strasbourg}. \\ 
                & & & & &  {\textcolor{blue}{your clubs or associations if any}} &  \textcolor{Periwinkle}{EN}: &  i'm \uline{terrible at remembering birthdays}. \\ 
                
            \noalign{\smallskip}\hline\noalign{\smallskip}

            \multicolumn{7}{c}{\textcolor{blue}{{\textsc{$\downarrow \downarrow$ New Category and Entities from Taxonomy Augmentation $\downarrow \downarrow$ }}}} \\
            \hline
            \multicolumn{2}{c|}{\multirow{ 18 }{*}{\rotatebox{90}{Interests}}}
                & \multicolumn{3}{c}{\multirow{ 3 }{*}{\textcolor{blue}{Technology}}} & {\textcolor{blue}{your enthusiasm for technology}}  & \textcolor{Goldenrod}{FR}: &  je suis très actif sur mon compte Instagram.  \\ 
                & & & & & {\textcolor{blue}{your adaptability to new technology or not}}  &  &  \textit{(I am \uline{very active} on my \uline{Instagram account}.)}  \\ 
                & & & & & {\textcolor{blue}{your awareness of cyber threats}}  &  \textcolor{Periwinkle}{EN}: &  i'm \uline{not tech-savvy}, i still use a \uline{flip phone}.  \\ 
                 \noalign{\smallskip}\cline{3-8}\noalign{\smallskip}
                
                & & \multicolumn{3}{c}{\multirow{ 3 }{*}{\textcolor{blue}{\makecell{Hobby \\ and \\ passions}}}} & {\textcolor{blue}{your interests in art if any}} &  \textcolor{Periwinkle}{EN}: &  i  \uline{enjoy hiking} in the  \colorbox{Periwinkle}{pacific northwest}. \\ 
                & & & & &  {\textcolor{blue}{any of your hobbies or passions}} & \textcolor{Goldenrod}{FR}: &  je suis passionné par la peinture de Claude Monet.  \\ 
                & & & & & {\textcolor{blue}{your various interests}} &  &  \textit{(I am \uline{passionate} about the \uline{paintings of \colorbox{Goldenrod}{Claude Monet}}.)} \\ 
                 \noalign{\smallskip}\cline{3-8}\noalign{\smallskip}

                & & \multicolumn{2}{c}{\multirow{ 9 }{*}{\rotatebox{90}{Environment}}}  & \multirow{ 3 }{*}{\textcolor{blue}{/}} &  {\textcolor{blue}{your environmental advocacy efforts if any}} &  \textcolor{Periwinkle}{EN}: &  i'm \uline{passionate} about \uline{reducing plastic waste} in our oceans. \\ 
                & & & & &  {\textcolor{blue}{your commitment to sustainability}} & \textcolor{Goldenrod}{FR}: &  j'appuie les initiatives pour protéger la biodiversité.  \\ 
                & & & & &  {\textcolor{blue}{your participation in environmental campaigns}} &  &  \textit{(I support initiatives to \uline{protect biodiversity}.)}  \\ 
                \noalign{\smallskip}\cline{5-8}\noalign{\smallskip}
                
                & & & & \multirow{ 3 }{*}{\textcolor{blue}{\makecell{Recycling \\ Habits}}} & {\textcolor{blue}{reusable products you incorporate into your daily life}} & \textcolor{Periwinkle}{EN}: &  i always wear \uline{second-hand clothes}, it's more sustainable. \\ 
                & & & & &  {\textcolor{blue}{your zero waste efforts}} & \textcolor{Goldenrod}{FR}: &  j'ai un composteur dans mon jardin.   \\ 
                & & & & &  {\textcolor{blue}{your eco-friendly choices and practices}} &  &  \textit{(I have a \uline{composter in my garden}.)} \\ 
                \noalign{\smallskip}\cline{5-8}\noalign{\smallskip}
                
                & & & & \multirow{ 3 }{*}{\textcolor{blue}{\makecell{Carbon \\ Footprint}}} & {\textcolor{blue}{your use of renewable energy}} & \textcolor{Periwinkle}{EN}: &  i \uline{generate a lot of paper waste} at home \\ 
                & & & & &  {\textcolor{blue}{your resource consumption}} & \textcolor{Goldenrod}{FR}: &  je préfère utiliser le bus.  \\ 
                & & & & &  {\textcolor{blue}{your minimalist lifestyle to save the planet}} &  &  \textit{(I prefer \uline{using the bus}.)} \\ 
                 \noalign{\smallskip}\cline{3-8}\noalign{\smallskip}
                
            & & \multicolumn{3}{c}{\multirow{ 3 }{*}{\textcolor{blue}{Travel}}} & {\textcolor{blue}{your interest in travelling}} & \textcolor{Goldenrod}{FR}: &  j'ai visité le Mont-Saint-Michel à l'âge de 12 ans.  \\ 
                & & & & & {\textcolor{blue}{memorable trips you've taken}}   &  & \textit{(I \uline{visited} the \colorbox{Goldenrod}{\uline{Mont-Saint-Michel}} at the age of 12.)} \\ 
                & & & & & {\textcolor{blue}{countries you want to explore}}  & \textcolor{Periwinkle}{EN}: &  i've \uline{traveled to europe} many times.  \\ 
 \noalign{\smallskip}\hline\noalign{\smallskip}
            \multicolumn{2}{c|}{\multirow{ 3 }{*}{\rotatebox{90}{Goals}}}
                & \multicolumn{3}{c}{\multirow{ 3 }{*}{\textcolor{blue}{\makecell{Personal \\ Goals}}}} & {\textcolor{blue}{your personal or life goals}} & \textcolor{Goldenrod}{FR}: &  je suis motivé pour réussir mon bac. \\ 
                & & & & & {\textcolor{blue}{one of your key life purposes}}  &  &  \textit{(I am \uline{motivated to succeed} my \colorbox{Goldenrod}{\uline{baccalaureate}}. )}   \\ 
                & & & & & {\textcolor{blue}{any extrinsic or intrinsic motivation of yours}}  & \textcolor{Periwinkle}{EN}: &  I aspire to \uline{start my own business}.  \\ 
                
 \noalign{\smallskip}\hline\noalign{\smallskip}
            \multicolumn{2}{c|}{\multirow{ 4 }{*}{\rotatebox{90}{Influences}}}
                & \multicolumn{3}{c}{\multirow{ 4 }{*}{\textcolor{blue}{\makecell{Role \\ Models}}}} & {\textcolor{blue}{someone that inspires you}}  & \textcolor{Periwinkle}{EN}: &  my \uline{hero} is \colorbox{Periwinkle}{\uline{steve jobs}}. \\ 
                & & & & & {\textcolor{blue}{someone whose life you admire}}  & \textcolor{Goldenrod}{FR}: &  Je admire les réalisations de l'astronaute Thomas Pesquet.  \\ 
                & & & & & {\textcolor{blue}{a hero in your eyes}}   &  &  \textit{(I admire the achievements of astronaut \colorbox{Goldenrod}{\uline{Thomas Pesquet}}.)} \\ 
                \noalign{\smallskip}\hline\noalign{\smallskip}
\end{tabular}
\end{table*}

\subsection{\textsc{Demographics}}

\begin{table*}[h!]
\scriptsize
\centering

    \caption{Taxonomy of \textsc{Demographics} category, associated sentences for the prompt and examples in randomly selected languages}
    \label{tab:demographics-taxo}
    \begin{tabular}{cc|cp{5.2cm}p{0.1cm}p{7.4cm}}
            \hline
            \multicolumn{2}{c}{\textbf{Cat}}  & \multicolumn{1}{c}{\textbf{Sub Category}}  & \multicolumn{1}{l}{\textbf{Sentences}}  & \multicolumn{2}{l}{\textbf{Examples}} \\
            \hline\noalign{\smallskip}
            
            \multicolumn{2}{c|}{\multirow{ 14 }{*}{\rotatebox{90}{Location}}}
                & \multirow{ 3 }{*}{Residence} & {\textcolor{blue}{your city or country of residence}} &  \textcolor{Goldenrod}{FR}: &  je habite actuellement à Lyon, dans le quartier de roix-Rousse. \\  
                & & & {\textcolor{blue}{your present hometown}} & &  \textit{(I currently live in \colorbox{Goldenrod}{\uline{Lyon}}, in the \colorbox{Goldenrod}{\uline{Croix-Rousse}} neighborhood.)} \\  
                & & & {\textcolor{blue}{where you currently live}} & \textcolor{Periwinkle}{EN}: &   I call \colorbox{Periwinkle}{\uline{Melbourne, Australia home}}. \\  
                \noalign{\smallskip}\cline{3-6}\noalign{\smallskip}
                
                & & \multirow{ 3 }{*}{\makecell{Birthplace}} & {\textcolor{blue}{where you were born}} & \textcolor{Goldenrod}{FR}: &  née à Marseille, j'ai une âme méditerranéenne. \\  
                & & & {\textcolor{blue}{your city or country of birth}} &  &  \textit{(Born in \colorbox{Goldenrod}{\uline{Marseille}}, I have a Mediterranean soul.)} \\ 
                & & & {\textcolor{blue}{your childhood hometown}} & \textcolor{Periwinkle}{EN}: &  i \uline{grew up} in \colorbox{Periwinkle}{\uline{austin, texas.}} \\  
                \noalign{\smallskip}\cline{3-6}\noalign{\smallskip}
                
                & & \multirow{ 3 }{*}{Nationality} & {\textcolor{blue}{the country you are a citizen of}} & \textcolor{Goldenrod}{FR}: & je suis belge.   \\ 
                & & & {\textcolor{blue}{your nationality}} &  &  \textit{(I am \colorbox{Goldenrod}{\uline{Belgian}}.)}  \\ 
                & & &  {\textcolor{blue}{the nation you belong to}} & \textcolor{Periwinkle}{EN}: &  I'm a \colorbox{Periwinkle}{\uline{true British person}} at heart.  \\  

                \hline
                
\multicolumn{2}{c|}{\multirow{ 16 }{*}{\rotatebox{90}{Employment}}}

                & \multirow{ 3 }{*}{\makecell{Company \\ Name}} & {\textcolor{blue}{official name of the organization you work for}} & \textcolor{Goldenrod}{FR}: &  je suis employé à la SNCF. \\  
                & & & {\textcolor{blue}{the company you are employed with}} &  &  \textit{(I work for \colorbox{Goldenrod}{\uline{SNCF}}.)} \\ 
                & & & {\textcolor{blue}{the business you work for}} & \textcolor{Periwinkle}{EN}: &  I'm employed by the \colorbox{Periwinkle}{\uline{British Museum}}. \\  
                \noalign{\smallskip}\cline{3-6}\noalign{\smallskip}
                
                & &\multirow{ 3 }{*}{Workplace} & {\textcolor{blue}{your work environment}} & \textcolor{Goldenrod}{FR}: &  je travaille au siège social de la société BPCE.   \\ 
                & & & {\textcolor{blue}{your office setting}} &  &  \textit{(I \uline{work at the headquarters} of \colorbox{Goldenrod}{\uline{BPCE}}.)} \\  
                & & & {\textcolor{blue}{the place where you work}} & \textcolor{Periwinkle}{EN}: &  i \uline{work in a coffee shop} in the \colorbox{Periwinkle}{financial district}. \\ 
                \noalign{\smallskip}\cline{3-6}\noalign{\smallskip}
                
                & &\multirow{ 3 }{*}{Profession} & {\textcolor{blue}{your current or previous profession}} & \textcolor{Goldenrod}{FR}: &  je suis enseignant de français en chine.  \\ 
                & & & {\textcolor{blue}{your job skills or certifications}} &  &  \textit{(I work as a \colorbox{Goldenrod}{\uline{French}} \uline{teacher} in China.)} \\  
                & & & {\textcolor{blue}{the field you work in}} & \textcolor{Periwinkle}{EN}: &  i'm a \uline{part-time yoga instructor}. \\  
                \noalign{\smallskip}\cline{3-6}\noalign{\smallskip}
                
                & &\multirow{ 3 }{*}{\makecell{Job \\ Status}} & {\textcolor{blue}{whether you are currently employed or not}} & \textcolor{Goldenrod}{FR}: &  je suis à la retraite.  \\ 
                & & & {\textcolor{blue}{if you are working or seeking employment}} &  &  \textit{(I am \uline{retired}.)} \\  
                & & & {\textcolor{blue}{your job situation}} & \textcolor{Periwinkle}{EN}: &  i'm currently \uline{between jobs after a layoff}. \\  

                \hline
                \multicolumn{6}{r}{\scriptsize\sl continued on next page}\\
\end{tabular}
\end{table*}

\begin{table*}[h!]
    \scriptsize
    \centering
    \begin{tabular}{cc|cp{5.5cm}p{0.1cm}p{7.4cm}}
            \multicolumn{6}{l}{\scriptsize\sl continued from previous page}\\
            \hline
            \multicolumn{2}{c}{\textbf{Cat}}  & \multicolumn{1}{c}{\textbf{Sub Category}}  & \multicolumn{1}{l}{\textbf{Sentences}}  & \multicolumn{2}{l}{\textbf{Examples}} \\
            \hline\noalign{\smallskip}
\multicolumn{2}{c|}{\multirow{ 24 }{*}{\rotatebox{90}{Employment}}}
                
                &\multirow{ 3 }{*}{\textcolor{blue}{\makecell{Career \\ Path}}} & {\textcolor{blue}{your professional milestones}} & \textcolor{Goldenrod}{FR}: &  j'ai créé une boutique de mode à Montmartre après avoir étudié à la ESMOD. \\ 
                & & & {\textcolor{blue}{the trajectory of your career}} &  &  \textit{(I \uline{created} a \uline{fashion boutique} in \colorbox{Goldenrod}{\uline{Montmarte}} after studying at \colorbox{Goldenrod}{\uline{ESMOD}}.)} \\  
                & & & {\textcolor{blue}{the steps you've taken in your career}} & \textcolor{Periwinkle}{EN}: &  I've \uline{worked} as a software engineer \uline{for five years} in \colorbox{Periwinkle}{\uline{Silicon Valley}}. \\  
                \noalign{\smallskip}\cline{3-6}\noalign{\smallskip}
                
                & & \multirow{ 3 }{*}{\textcolor{blue}{\makecell{Job \\ Fulfillment}}} & {\textcolor{blue}{your career happiness}} & \textcolor{Goldenrod}{FR}: &  j'adore mon travail d'enseignant.   \\ 
                & & & {\textcolor{blue}{how satisfied you are with your job}} &  &  \textit{(I \uline{love my job} as a teacher.)} \\  
                & & & {\textcolor{blue}{your professional bliss}} & \textcolor{Periwinkle}{EN}: &  I'm \uline{not satisfied with my current job}, I want more challenge.\\  
                \noalign{\smallskip}\cline{3-6}\noalign{\smallskip}

                & &\multirow{ 3 }{*}{\textcolor{blue}{\makecell{Motivations \\ and Goals}}} & {\textcolor{blue}{your professional goals}} & \textcolor{Goldenrod}{FR}: &  j'aimerais écrire un livre sur la Révolution française.   \\ 
                & & & {\textcolor{blue}{what drives you at work}} & & \textit{(I \uline{would like to write a book} about the \colorbox{Goldenrod}{French Revolution}.)} \\  
                & & & {\textcolor{blue}{your work aspirations}} & \textcolor{Periwinkle}{EN}: &  my \uline{ultimate goal is to work for the} \colorbox{Periwinkle}{national park service}.\\  
                \noalign{\smallskip}\cline{3-6}\noalign{\smallskip}
                
                & &\multirow{ 3 }{*}{\textcolor{blue}{\makecell{Work-Life \\ Balance}}} & {\textcolor{blue}{your work flexibility}} & \textcolor{Goldenrod}{FR}: &  je travaille en freelance pour avoir plus de liberté.  \\ 
                & & & {\textcolor{blue}{how you manage work-life balance}} &  &  \textit{(I \uline{work freelance} to have \uline{more freedom}.)} \\  
                & & & {\textcolor{blue}{your methods for balancing work and personal life}} & \textcolor{Periwinkle}{EN}: &  i prioritize \uline{work during the week, family on weekends}. \\  
                \noalign{\smallskip}\cline{3-6}\noalign{\smallskip}
                
                & &\multirow{ 3 }{*}{\textcolor{blue}{\makecell{Remote \\ Work}}} & {\textcolor{blue}{your home office setup if any}} & \textcolor{Goldenrod}{FR}: &  j'ai travaillé à domicile pendant un an. \\ 
                & & & {\textcolor{blue}{your telecommuting experience if any}}  &  &  \textit{(I \uline{worked from home} for a year.)}  \\  
                & & & {\textcolor{blue}{your remote work best or worst practices}} & \textcolor{Periwinkle}{EN}: &  i get \uline{distracted by notifications when remote working}.\\  
                \noalign{\smallskip}\cline{3-6}\noalign{\smallskip}
                
                & &\multirow{ 3 }{*}{\textcolor{blue}{Network}} & {\textcolor{blue}{your professional network}} & \textcolor{Goldenrod}{FR}: &  je suis mentoré par mon collègue, Pierre Dupont.  \\ 
                & & & {\textcolor{blue}{your mentee if any}} &  &  \textit{(I am \uline{mentored by my colleague}, \colorbox{Goldenrod}{\uline{Pierre Dupont}}.)} \\  
                & & & {\textcolor{blue}{your industry connections}} & \textcolor{Periwinkle}{EN}: &  i'm trying to \uline{expand my network on linkedin}.\\  
                
            \noalign{\smallskip}\hline\noalign{\smallskip}
        
   & & \multirow{ 3 }{*}{Degree} & {\textcolor{blue}{the current degree you pursue}} & \textcolor{Goldenrod}{FR}: &  j'ai obtenu un diplôme en sociologie à l'Université de Paris.  \\ 
                & & & {\textcolor{blue}{the degree you have earned}} &  &  \textit{(I have \uline{earned a degree in sociology} from \colorbox{Goldenrod}{the University of Paris}.)} \\  
                & & & {\textcolor{blue}{the level of education you are working towards}} & \textcolor{Periwinkle}{EN}: &  i have a \uline{master}'s in finance from \colorbox{Periwinkle}{nyu}. \\  
                \noalign{\smallskip}\cline{3-6}\noalign{\smallskip}
                
                & & \multirow{ 3 }{*}{\makecell{Degree \\ subject}} & {\textcolor{blue}{your field of study}} & \textcolor{Goldenrod}{FR}: &  j'ai obtenu un master en marketing à l'ESSEC.    \\ 
                & & & {\textcolor{blue}{your academic discipline}} & & \textit{( I got a \uline{master in marketing} at the \colorbox{Goldenrod}{ESSEC}.)} \\
                & & & {\textcolor{blue}{your degree major}} & \textcolor{Periwinkle}{EN}: &  My degree is in \uline{linguistics} from the \colorbox{Periwinkle}{University of Michigan}. \\  
                \noalign{\smallskip}\cline{3-6}\noalign{\smallskip}
                
\multicolumn{2}{c|}{\multirow{ 23 }{*}{\rotatebox{90}{Education}}} & \multirow{ 3 }{*}{\makecell{School \\ Name}} & {\textcolor{blue}{your alma mater}} & \textcolor{Periwinkle}{EN}: &  i'm enrolled in a master's program at \colorbox{Periwinkle}{\uline{UCLA}}.  \\ 
                & & & {\textcolor{blue}{the school you graduated from}} & \textcolor{Goldenrod}{FR}: &  j'ai étudié à l'École normale supérieure de Lyon. \\
                & & & {\textcolor{blue}{the school institution you attend}} &  &  \textit{(I \uline{studied} at the \colorbox{Goldenrod}{\uline{École Normale Supérieure de Lyon}}.)} \\ 
                \noalign{\smallskip}\cline{3-6}\noalign{\smallskip}
                
                & & \multirow{ 3 }{*}{\makecell{School \\ Status}} & {\textcolor{blue}{your school status: student, alumni, etc}} & \textcolor{Periwinkle}{EN}: &  i \uline{graduated} \colorbox{Periwinkle}{summa cum laude}. \\ 
                & & & {\textcolor{blue}{if you have graduated}} & \textcolor{Goldenrod}{FR}: &  je suis diplômé en sciences politiques de Sciences Po Paris. \\  
                & & & {\textcolor{blue}{whether you are currently a student}} &  & \textit{( I'm \uline{graduated} in political science from \colorbox{Goldenrod}{Sciences Po Paris}.)}  \\  
                \noalign{\smallskip}\cline{3-6}\noalign{\smallskip}

                & & \multirow{ 3 }{*}{\makecell{School \\ type}} & {\textcolor{blue}{the type of school you attend}} & \textcolor{Goldenrod}{FR}: &  j'ai étudié à l'université Paris-Sorbonne.  \\ 
                & & & {\textcolor{blue}{whether you attend a public or private school}} &  & \textit{(I studied at the \colorbox{Goldenrod}{Paris-Sorbonne} \uline{university}.)} \\ 
                & & & {\textcolor{blue}{the category of your school}} &  \textcolor{Periwinkle}{EN}: &  i'm a student at a \colorbox{Periwinkle}{\uline{community college}}. \\  
                \noalign{\smallskip}\cline{3-6}\noalign{\smallskip}

            & & \multicolumn{4}{c}{\textcolor{blue}{{\textsc{$\downarrow \downarrow$ New Entities from Taxonomy Augmentation $\downarrow \downarrow$ }}}} \\
                \noalign{\smallskip}\cline{3-6}\noalign{\smallskip}
                
                & & \multirow{ 3 }{*}{\textcolor{blue}{Achievements}} & {\textcolor{blue}{any honor or award you wished to receive at school}} & \textcolor{Goldenrod}{FR}: &  j'aimerais recevoir le prix littéraire des lycéens.    \\ 
                & & & {\textcolor{blue}{your academic accomplishments}} &  &  \textit{(I \uline{would love to receive} the \uline{literary prize for high school students}.)}  \\  
                & & & {\textcolor{blue}{the awards you earned in school}} & \textcolor{Periwinkle}{EN}: &  I \uline{wish I had won} a \colorbox{Periwinkle}{\uline{Pulitzer Prize}} for my writing.\\  
                \noalign{\smallskip}\cline{3-6}\noalign{\smallskip}
                
                & & \multirow{ 3 }{*}{\textcolor{blue}{\makecell{Extracurricular \\ Activities}}} & {\textcolor{blue}{any club or association you were involved in at school}} & \textcolor{Goldenrod}{FR}: &  j'ai été vice-président du club de rugby.  \\ 
                & & & {\textcolor{blue}{projects you worked on outside of class}} &  &  \textit{(I \uline{was vice-president} of the \colorbox{Goldenrod}{\uline{rugby club}}.)} \\  
                & & & {\textcolor{blue}{your participation in school sports}} & \textcolor{Periwinkle}{EN}: &  i \uline{played} \colorbox{Periwinkle}{\uline{basketball in college}}, but got injured.\\  

                \noalign{\smallskip}\cline{3-6}\noalign{\smallskip}
                & & \multirow{ 3 }{*}{\textcolor{blue}{\makecell{Spoken \\ Languages}}} & {\textcolor{blue}{other languages or dialects that you speak or learn to}} & \textcolor{Goldenrod}{FR}: &  je parle français et Breton.   \\ 
                & & & {\textcolor{blue}{the languages you are fluent in}}  &  &  \textit{(I \uline{speak} \colorbox{Goldenrod}{\uline{French}} et \colorbox{Goldenrod}{\uline{Breton}}.)}  \\ 
                & & & {\textcolor{blue}{additional languages you speak}} & \textcolor{Periwinkle}{EN}: &  I'm \colorbox{Periwinkle}{\uline{learning Spanish to communicate with my clients}}. \\ 
                
                \noalign{\smallskip}\cline{3-6}\noalign{\smallskip}
                
                & & \multirow{ 3 }{*}{\textcolor{blue}{\makecell{Workshops \\ {/} \\ Seminars}}} & {\textcolor{blue}{workshops or seminars you attended or wish to}} & \textcolor{Goldenrod}{FR}: &  j'ai présenté mon projet à la station f. \\ 
                & & & {\textcolor{blue}{training programs you completed}} &  & \textit{(I \uline{presented my project} at \colorbox{Goldenrod}{\uline{Station F}}.)} \\  
                & & & {\textcolor{blue}{seminars were you presented some work}} & \textcolor{Periwinkle}{EN}: &  i \uline{presented my startup idea} at the \colorbox{Periwinkle}{\uline{techcrunch}} conference.\\  
            \noalign{\smallskip}\hline\noalign{\smallskip}
        
          \multicolumn{2}{c|}{\multirow{ 9 }{*}{\rotatebox{90}{\makecell{Family Status}}}}
                & \multirow{ 3 }{*}{Siblings} & {\textcolor{blue}{your brothers or sisters}} & \textcolor{Goldenrod}{FR}: &  j'ai deux sœurs qui vivent en province.   \\ 
                & & & {\textcolor{blue}{the number of siblings you have}} &  & I have \uline{two sisters} who \colorbox{Goldenrod}{live in the provinces}.  \\
                & & & {\textcolor{blue}{your family members}} & \textcolor{Periwinkle}{EN}: &  I have a \uline{twin sister} named \colorbox{Periwinkle}{Brittany}.\\
                \noalign{\smallskip}\cline{3-6}\noalign{\smallskip}
                
                & & \multirow{ 3 }{*}{Children} & {\textcolor{blue}{your children if any}} & \textcolor{Goldenrod}{FR}: &  j'ai deux jeunes filles, Sophie et Emma.  \\   
                & & & {\textcolor{blue}{the children in your family if any}} &  &  \textit{(I have  \uline{two young daughters} \colorbox{Goldenrod}{Sophie} et \colorbox{Goldenrod}{Emma}.)} \\ 
                & & & {\textcolor{blue}{your offspring}} & \textcolor{Periwinkle}{EN}: &  I'm a single parent \uline{with a 10-year-old son} who loves Legos \\  
            \noalign{\smallskip}\hline\noalign{\smallskip}
            
            \multicolumn{2}{c|}{\multirow{ 11 }{*}{\rotatebox{90}{Possession}}}
                & \multirow{ 3 }{*}{Animal} & \textcolor{blue}{{your pet if you have one}} & \textcolor{Periwinkle}{EN}: &  my \uline{dream pet} is a \uline{capybara}.  \\ 
                & & & \textcolor{blue}{{an animal companion you want}}  & \textcolor{Goldenrod}{FR}: &  j'ai toujours voulu un chien berger allemand.   \\ 
                & & & \textcolor{blue}{{an animal you possess or wish to}}  &  &  \textit{(I always wanted a \uline{german sherperd dog}.)}   \\ 
                \noalign{\smallskip}\cline{3-6}\noalign{\smallskip}
                
                & & \multirow{ 3 }{*}{Vehicle} & \textcolor{blue}{{your dream car or vehicle}} & \textcolor{Periwinkle}{EN}: & i own a \colorbox{Periwinkle}{\uline{1969 ford mustang}}.   \\ 
                & & & \textcolor{blue}{{a vehicle you own or plan to buy}}  & \textcolor{Goldenrod}{FR}: &  ma voiture actuelle est une  renault clio.   \\ 
                & & & \textcolor{blue}{{a means of transportation you possess}}  &  & \textit{(My \uline{current car} is a \colorbox{Goldenrod}{\uline{Renault Clio}}.)}  \\ 
                \noalign{\smallskip}\cline{3-6}\noalign{\smallskip}
                
            & & \multicolumn{3}{c}{\textcolor{blue}{{\textsc{$\downarrow \downarrow$ New Entity from Taxonomy Augmentation $\downarrow \downarrow$ }}}} \\
                \noalign{\smallskip}\cline{3-6}\noalign{\smallskip}
                & & \multirow{ 3 }{*}{\textcolor{blue}{\makecell{Tech \\ Device}}} & \textcolor{blue}{{your favorite gadget}} & \textcolor{Periwinkle}{EN}: &   i'm really \uline{interested in} getting a \uline{3d printer}. \\ 
                & & & \textcolor{blue}{{your go-to tech tool}}  & \textcolor{Goldenrod}{FR}: &  J'ai un ordinateur portable Apple MacBook Pro que j'adore   \\ 
                & & & \textcolor{blue}{{an electronic device you own or desire}}  & & \textit{(I have a \uline{MacBook Pro} laptop that I love.)}     \\ 
            \noalign{\smallskip}\hline

            \multicolumn{6}{r}{\scriptsize\sl continued on next page}\\
\end{tabular}
\end{table*}

\begin{table*}[h!]
    \scriptsize
    \begin{tabular}{cc|cp{5.5cm}p{0.1cm}p{7.4cm}}
            \multicolumn{6}{l}{\scriptsize\sl continued from previous page}\\
            \hline
            \multicolumn{2}{c}{\textbf{Cat}}  & \multicolumn{1}{c}{\textbf{Sub Category}}  & \multicolumn{1}{l}{\textbf{Sentences}}  & \multicolumn{2}{l}{\textbf{Examples}} \\
            \hline\noalign{\smallskip}
\multicolumn{2}{c|}{\multirow{ 4 }{*}{\rotatebox{90}{\makecell{Marital \\ Status}}}}
                &  \multirow{ 3 }{*}{/}  & \textcolor{blue}{{your marital status}} & \textcolor{Periwinkle}{EN}: &  i'm \uline{divorced}.  \\ 
                & & & \textcolor{blue}{{if you have a spouse or partner}}  & \textcolor{Goldenrod}{FR}: &  je suis marié avec une Goldenrod femme originaire de Lyon.   \\ 
                & & & \textcolor{blue}{{whether you are married, single or divorced}}  &  &  \textit{(I'm \uline{married} to a woman \colorbox{Goldenrod}{\uline{originating from Lyon}}.)}   \\ 
            \noalign{\smallskip}\hline
            
\multicolumn{2}{c|}{\multirow{ 4 }{*}{\rotatebox{90}{Name}}}
                &  \multirow{ 3 }{*}{/}  & \textcolor{blue}{{the name you go by}} & \textcolor{Periwinkle}{EN}: &  my name is \colorbox{Periwinkle}{\uline{maxwell thompson}}.   \\ 
                & & & \textcolor{blue}{{the name you are known as}}  & \textcolor{Goldenrod}{FR}: &  je m'appelle Sophie Lefebvre.    \\ 
                & & & \textcolor{blue}{{your full name}}  &  &  \textit{(My name is \colorbox{Goldenrod}{\uline{Sophie Lefebvre}}.)}     \\ 
            \noalign{\smallskip}\hline\noalign{\smallskip}
            
            \multicolumn{2}{c|}{\multirow{ 4 }{*}{\rotatebox{90}{Age}}}
                & \multirow{ 3 }{*}{/}   & \textcolor{blue}{{how old you are}} & \textcolor{Periwinkle}{EN}: &  i was \uline{born in 1992}.  \\ 
                & & & \textcolor{blue}{{your birth year}}  & \textcolor{Goldenrod}{FR}: &  je suis quarantenaire.   \\ 
                & & & \textcolor{blue}{{the number of years you have lived}}  &  &  \textit{(I am \uline{quarantenaire}.)}   \\ 
            \noalign{\smallskip}\hline\noalign{\smallskip}
            
            \multicolumn{2}{c|}{\multirow{ 3 }{*}{\rotatebox{90}{Gender}}}
                & \multirow{ 3 }{*}{/}  & \textcolor{blue}{{the gender you identify as}} & \textcolor{Periwinkle}{EN}: &  i \uline{prefer they/them} pronouns.   \\ 
                & & & \textcolor{blue}{{your preferred pronouns}}  & \textcolor{Goldenrod}{FR}: &  je suis une femme.   \\ 
                & & & \textcolor{blue}{{your gender}}  &  &   \textit{(I am a \uline{woman}.)}  \\ 
            \hline
            \multicolumn{6}{c}{\textcolor{blue}{{\textsc{$\downarrow \downarrow$ New Categories and Entities from Taxonomy Augmentation $\downarrow \downarrow$ }}}} \\
            \hline
            \noalign{\smallskip}
             \multicolumn{2}{c|}{\multirow{ 3 }{*}{\rotatebox{90}{\textcolor{blue}{Ethnicity}}}}
                & \multirow{ 3 }{*}{/}  & \textcolor{blue}{{your cultural background, heritage, or ancestry}} & \textcolor{Periwinkle}{EN}: &  I \uline{identify} as a proud \colorbox{Periwinkle}{\uline{Brit with Scottish}} heritage.  \\ 
                & & & \textcolor{blue}{{your pride in your culture}}  & \textcolor{Goldenrod}{FR}: &  je suis originaire de la Martinique.   \\ 
                & & & \textcolor{blue}{{your ethnic background}}  &  & \textit{(I'm \uline{native} of \colorbox{Goldenrod}{\uline{Martinique}}.)}    \\ 
            \noalign{\smallskip}\hline\noalign{\smallskip}
            
             \multicolumn{2}{c|}{\multirow{ 3 }{*}{\rotatebox{90}{\textcolor{blue}{\tiny{\makecell{Religion /\\ Spirituality}}}}}}
                & \multirow{ 4 }{*}{/}  & \textcolor{blue}{{your religion or beliefs}} & \textcolor{Periwinkle}{EN}: &  i'm \uline{a devout Christian}, but i don't attend church often.  \\ 
                & & & \textcolor{blue}{{your faith or lack thereof}}  & \textcolor{Goldenrod}{FR}: &  Je suis agnostique et ne pratique pas de religion.   \\ 
                & & & \textcolor{blue}{{your spiritual views}}  &  &  \textit{(I am \uline{agnostic} and \uline{do not practice religion}.)}    \\ 
            \noalign{\smallskip}\hline\noalign{\smallskip}
            
             \multicolumn{2}{c|}{\multirow{ 6 }{*}{\rotatebox{90}{\textcolor{blue}{\makecell{Socioeconomic \\ Status}}}}}
                & \multirow{ 3 }{*}{ \textcolor{blue}{\makecell{Housing \\ Status}}}  & \textcolor{blue}{{your housing status or living arrangements}} & \textcolor{Periwinkle}{EN}: &  my \uline{apartment} has a \colorbox{Periwinkle}{\uline{view of the brooklyn bridge}}.  \\ 
                & & & \textcolor{blue}{{the type of housing you live in}}  & \textcolor{Goldenrod}{FR}: &  j' habite dans un village de la Loire-Atlantique.  \\ 
                & & & \textcolor{blue}{{your home environment}}  &  &   \textit{(I \uline{live} in a \colorbox{Goldenrod}{\uline{village in Loire-Atlantique}}.)}  \\ 
                \noalign{\smallskip}\cline{3-6}\noalign{\smallskip}
                
                & &\multirow{3}{*}{ \textcolor{blue}{\makecell{Income \\ Level}}}  & \textcolor{blue}{{your social standing}} & \textcolor{Periwinkle}{EN}: &   i've been \uline{in debt since college}. \\ 
                & & & \textcolor{blue}{{your economic position}}  & \textcolor{Goldenrod}{FR}: &  je gagne juste suffisamment pour voyager.   \\ 
                & & & \textcolor{blue}{{your income level, wealth, or social class}}  &  & \textit{(I \uline{earn just enough} to travel.)}     \\ 
            \hline 

\end{tabular}
\end{table*}


\section{Per Language Detailed Automatic Analysis}
\label{appendix:per_language_evals}
\subsection{High-Resource Languages}

\subsubsection{\textsc{Turkish}}

\begin{figure}[h]
    \centering
    \begin{minipage}{0.45\textwidth}
        \begin{subfigure}{\textwidth}
            \centering
            \includegraphics[height=0.15\textheight]{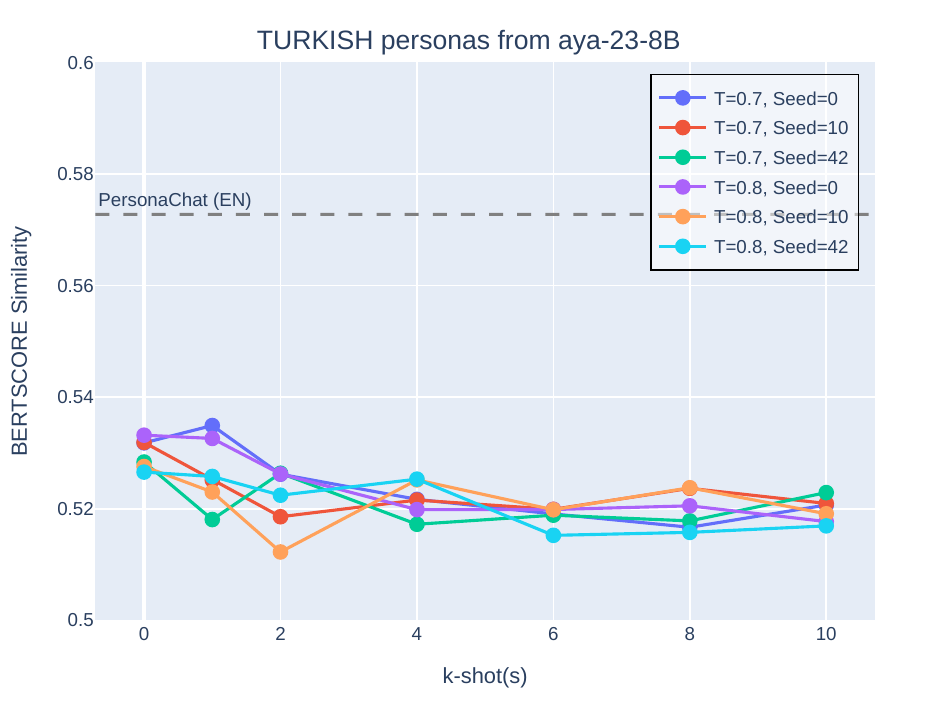}
        \end{subfigure}
        \vskip\baselineskip
        \begin{subfigure}{\textwidth}
            \centering
            \includegraphics[height=0.15\textheight]{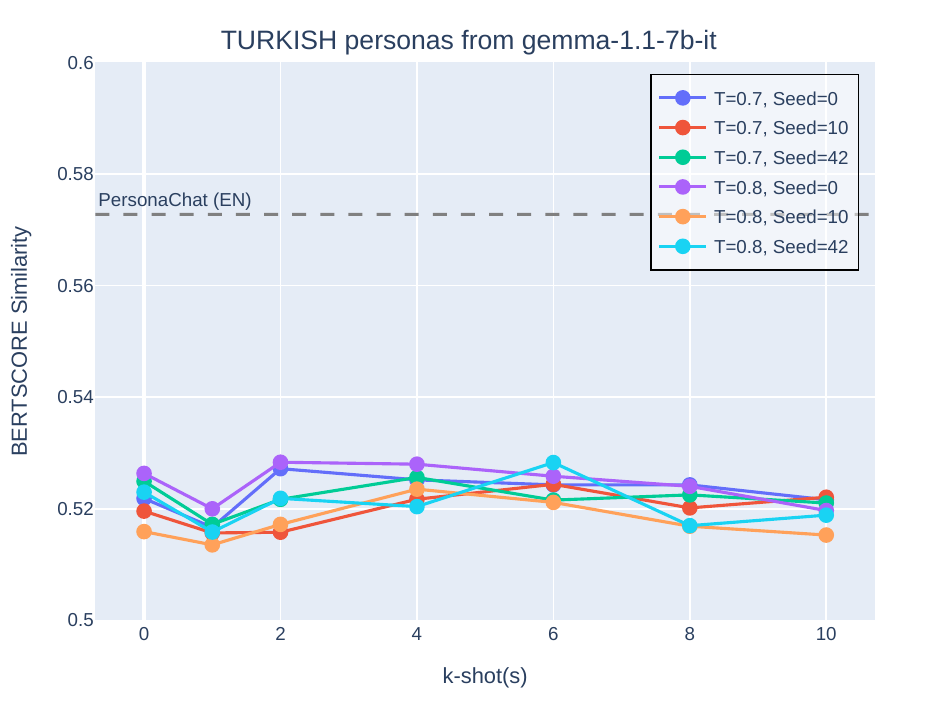}
        \end{subfigure}
    \end{minipage}
    %
    %
    \begin{minipage}{0.45\textwidth}
            \begin{subfigure}{\textwidth}
            \centering
            \includegraphics[height=0.15\textheight]{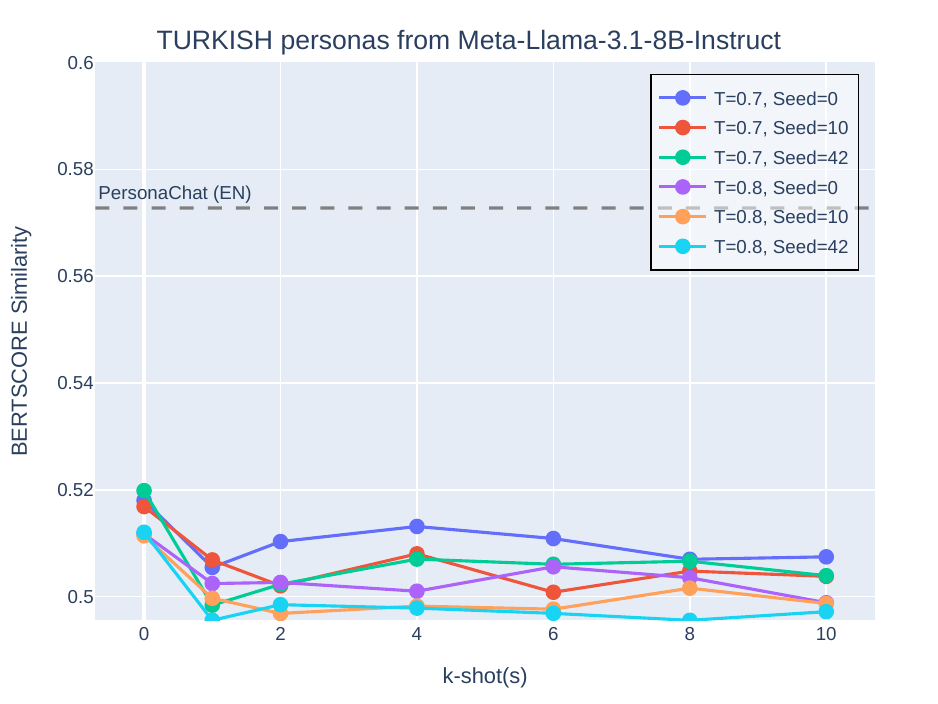}
        \end{subfigure}
        \vskip\baselineskip
        \begin{subfigure}{\textwidth}
            \centering
            \includegraphics[height=0.15\textheight]{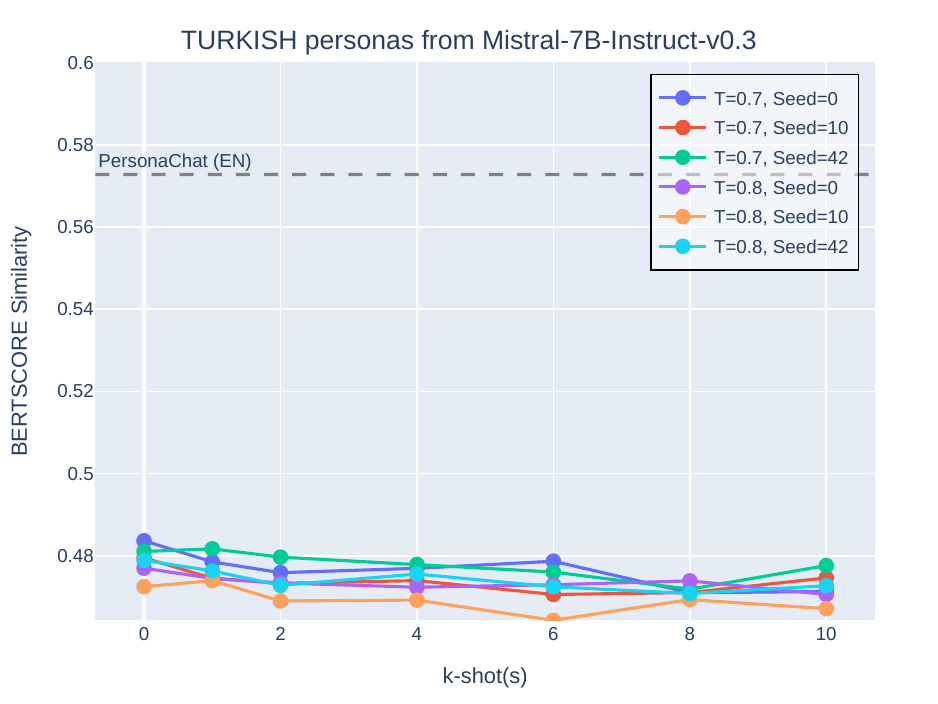}
        \end{subfigure}
    \end{minipage}
    
\caption{Detailed BERTSCORE for Turkish Personas in different generation configurations for the different models}    
\end{figure}




\newgeometry{top=0.5cm, bottom=1.5cm, left=2.5cm, right=2.5cm}
\subsubsection{\textsc{English}}

\begin{figure}[h]
    \centering
    \begin{minipage}{0.45\textwidth}
        \begin{subfigure}{\textwidth}
            \centering
            \includegraphics[height=0.17\textheight]{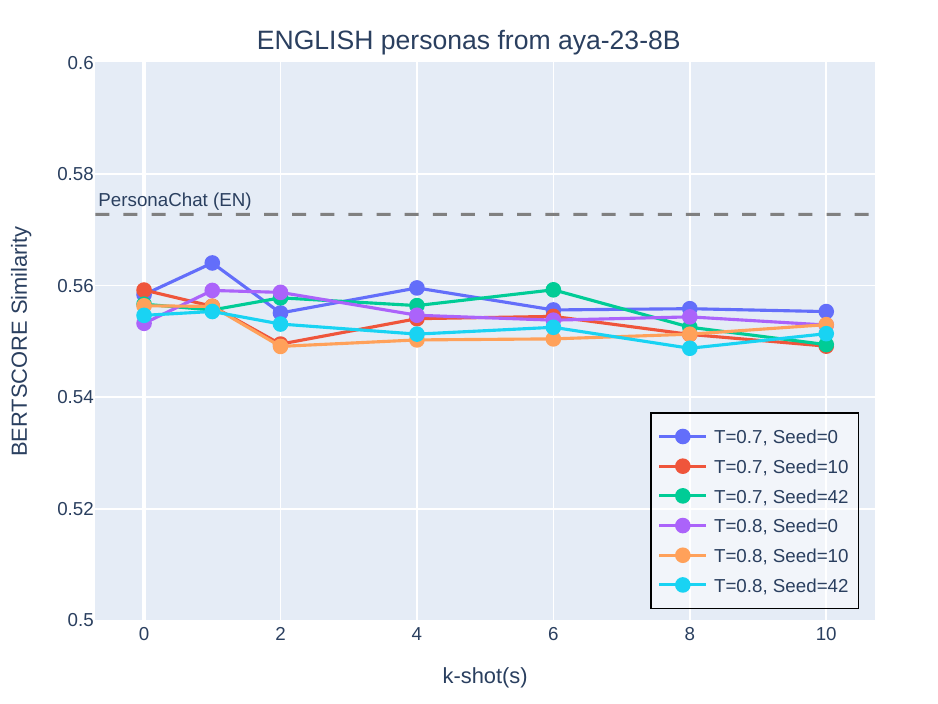}
        \end{subfigure}
        \vskip\baselineskip
        \begin{subfigure}{\textwidth}
            \centering
            \includegraphics[height=0.17\textheight]{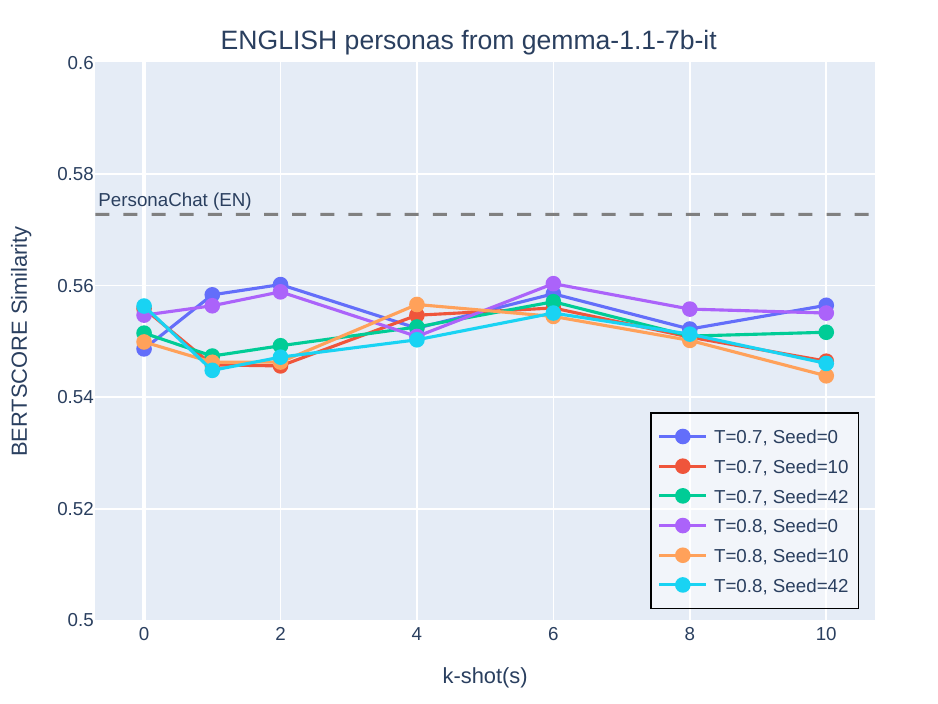}
        \end{subfigure}
        \vskip\baselineskip
        \begin{subfigure}{\textwidth}
            \centering
            \includegraphics[height=0.17\textheight]{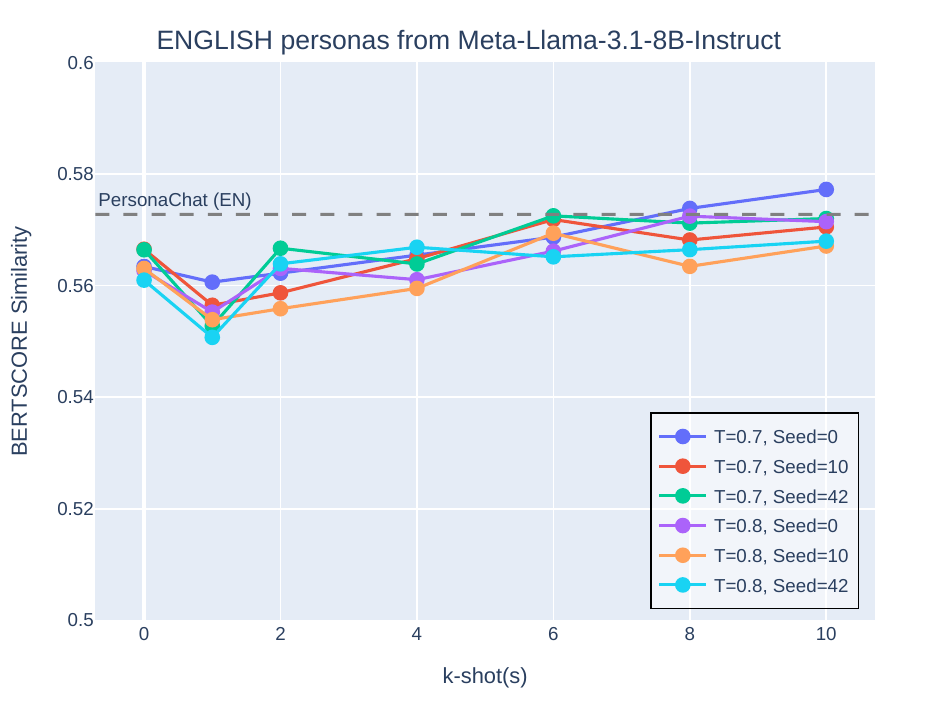}
        \end{subfigure}
        \vskip\baselineskip
        \begin{subfigure}{\textwidth}
            \centering
            \includegraphics[height=0.17\textheight]{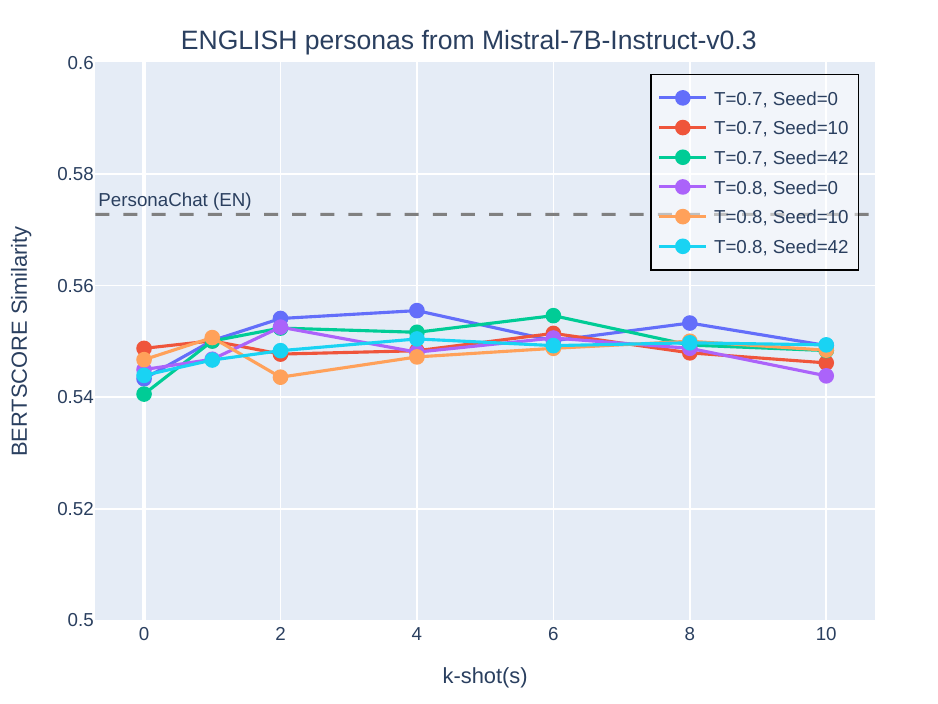}
        \end{subfigure}
    \end{minipage}
    %
    %
    \begin{minipage}{0.45\textwidth}
        \begin{subfigure}{\textwidth}
            \centering
            \includegraphics[height=0.17\textheight]{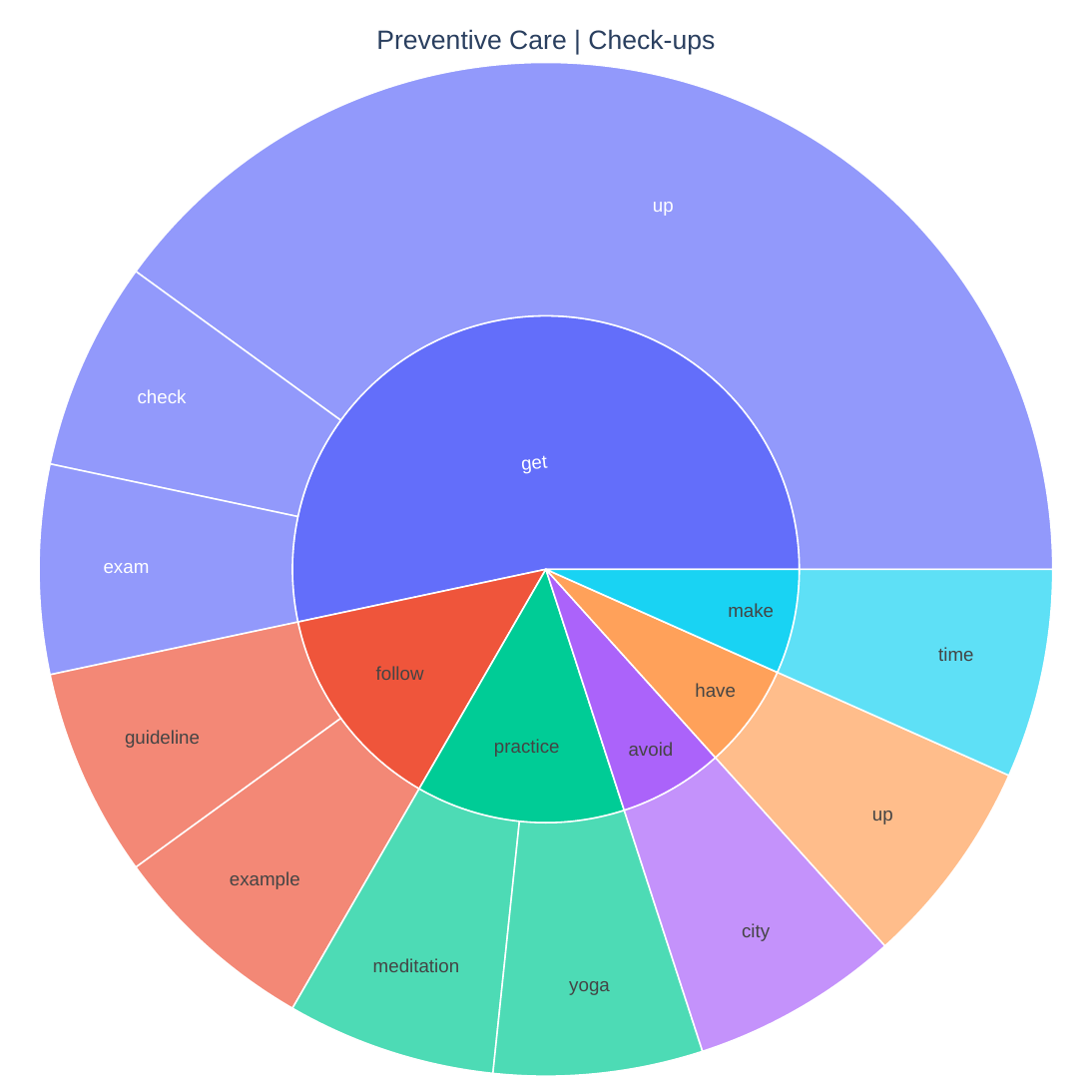}
        \end{subfigure}
        \vskip\baselineskip
        \begin{subfigure}{\textwidth}
            \centering
            \includegraphics[height=0.17\textheight]{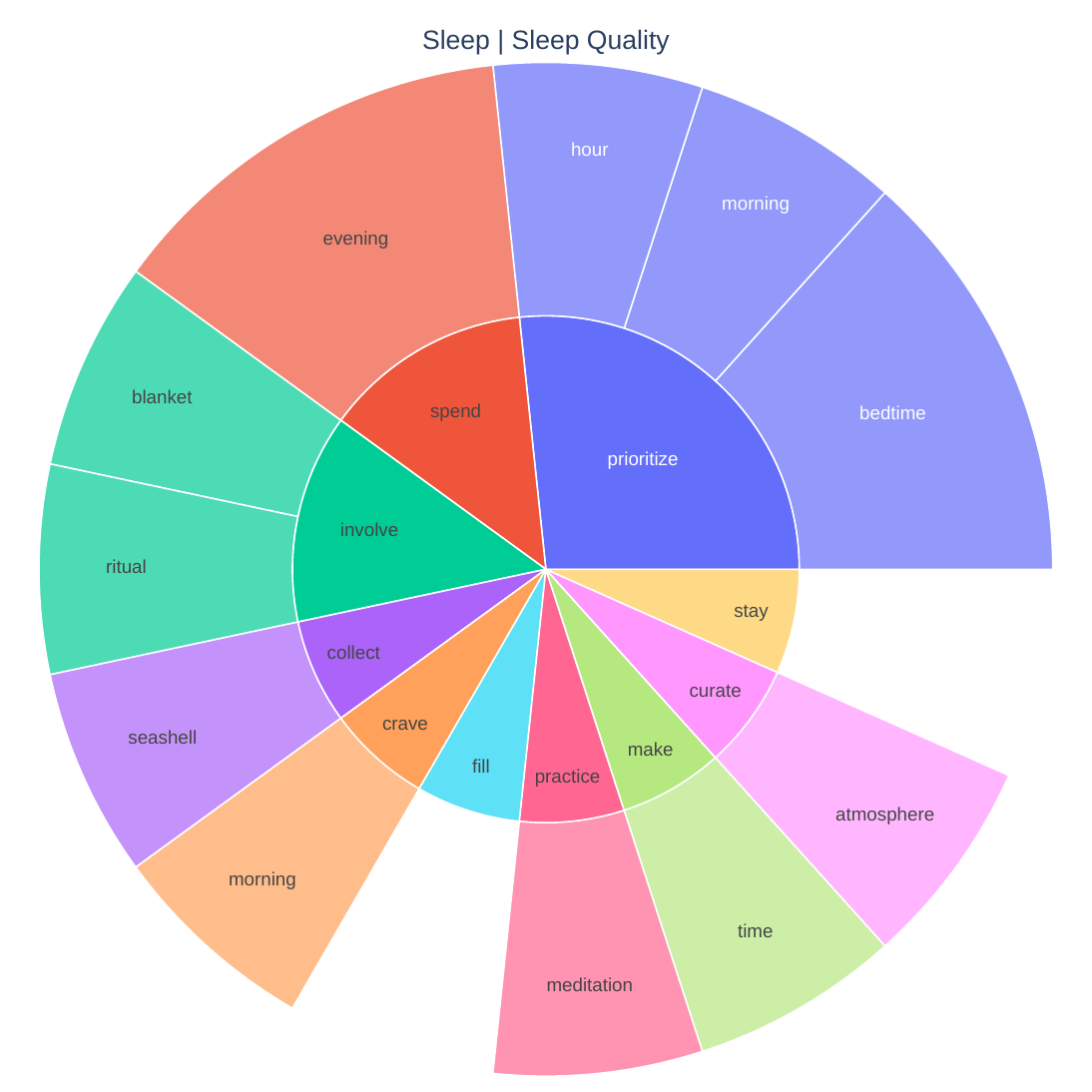}
        \end{subfigure}
        \vskip\baselineskip
        \begin{subfigure}{\textwidth}
            \centering
            \includegraphics[height=0.17\textheight]{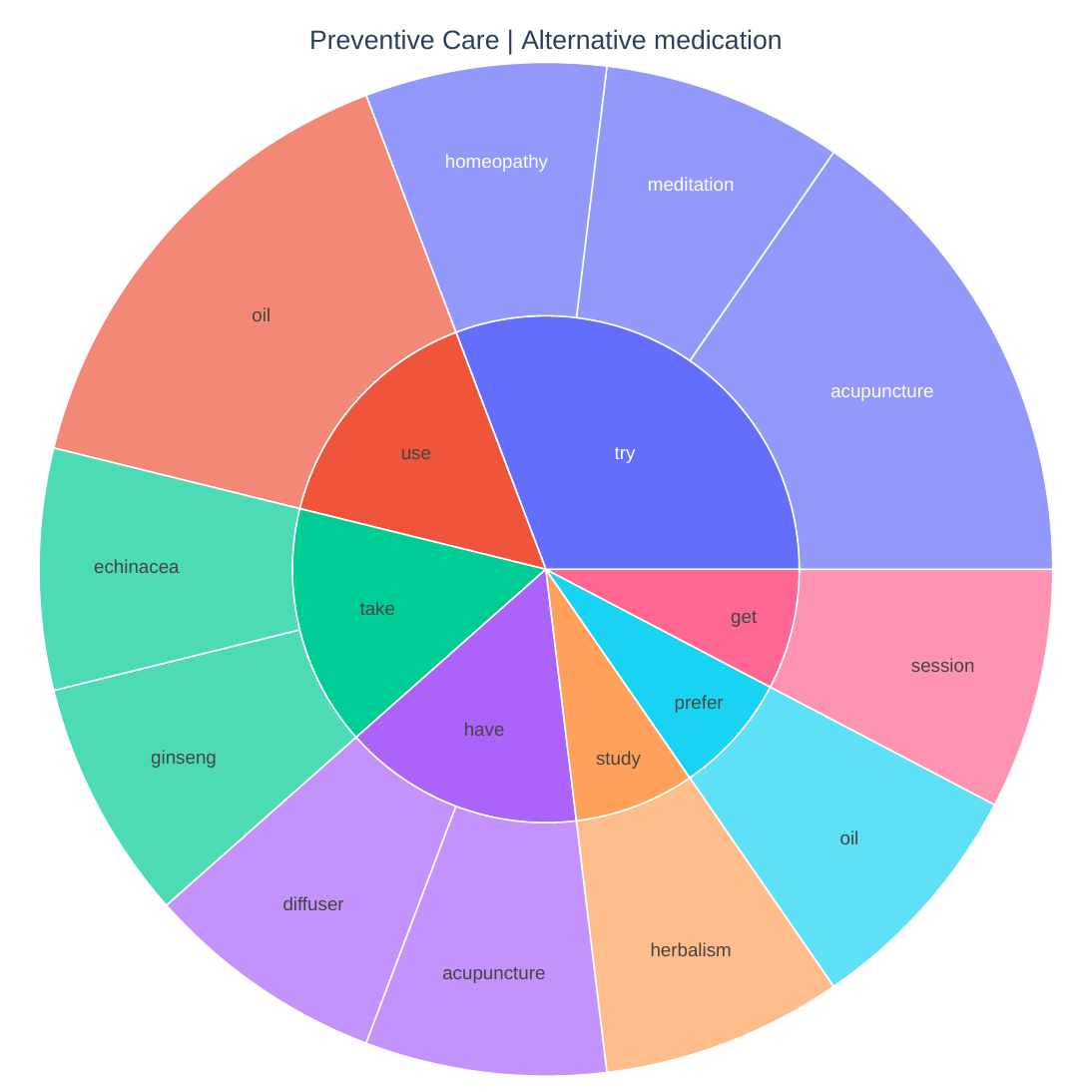}
        \end{subfigure}
        \vskip\baselineskip
        \begin{subfigure}{\textwidth}
            \centering
            \includegraphics[height=0.17\textheight]{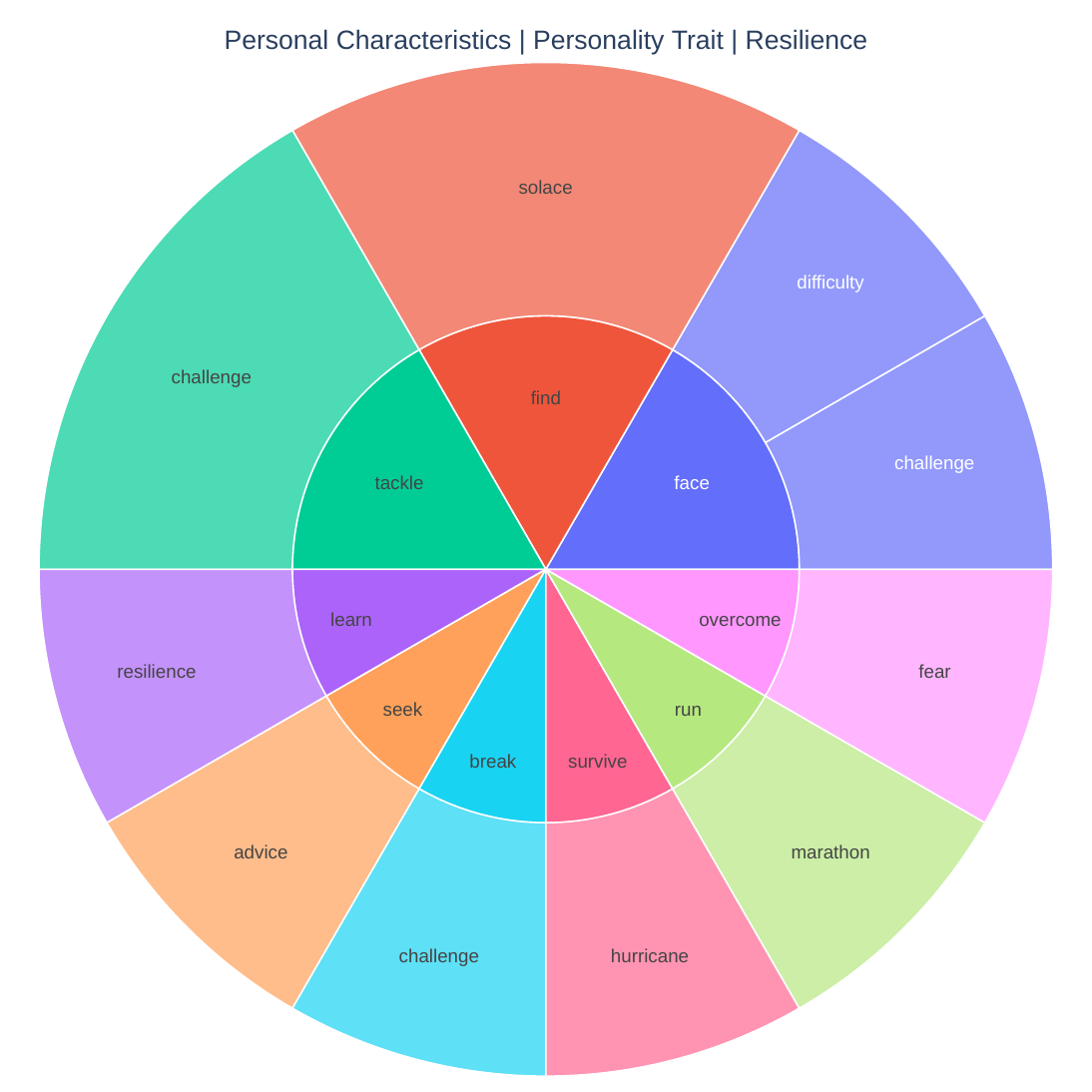}
        \end{subfigure}
    \end{minipage}
    
\caption{Detailed BERTSCORE for English Personas in different generation configurations and Sunburst charts of personas taxonomy entities with most root verbs and associated object noun for the different models}    
\end{figure}


\restoregeometry

\newgeometry{top=0.5cm, bottom=1.5cm, left=2.5cm, right=2.5cm}
\subsubsection{\textsc{Russian}}

\begin{figure}[h]
    \centering
    \begin{minipage}{0.45\textwidth}
        \begin{subfigure}{\textwidth}
            \centering
            \includegraphics[height=0.17\textheight]{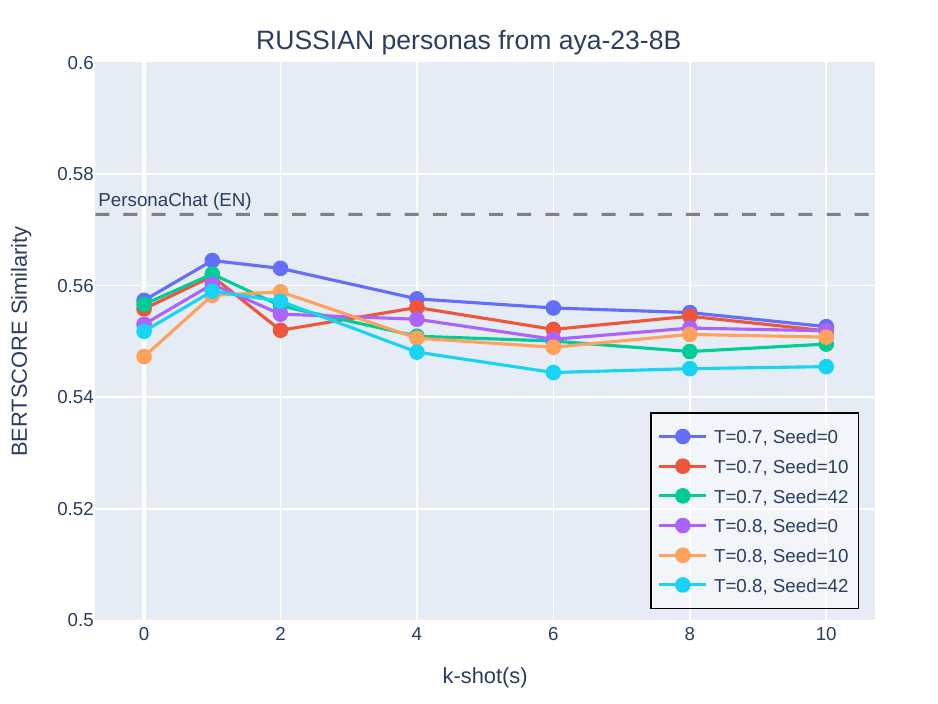}
        \end{subfigure}
        \vskip\baselineskip
        \begin{subfigure}{\textwidth}
            \centering
            \includegraphics[height=0.17\textheight]{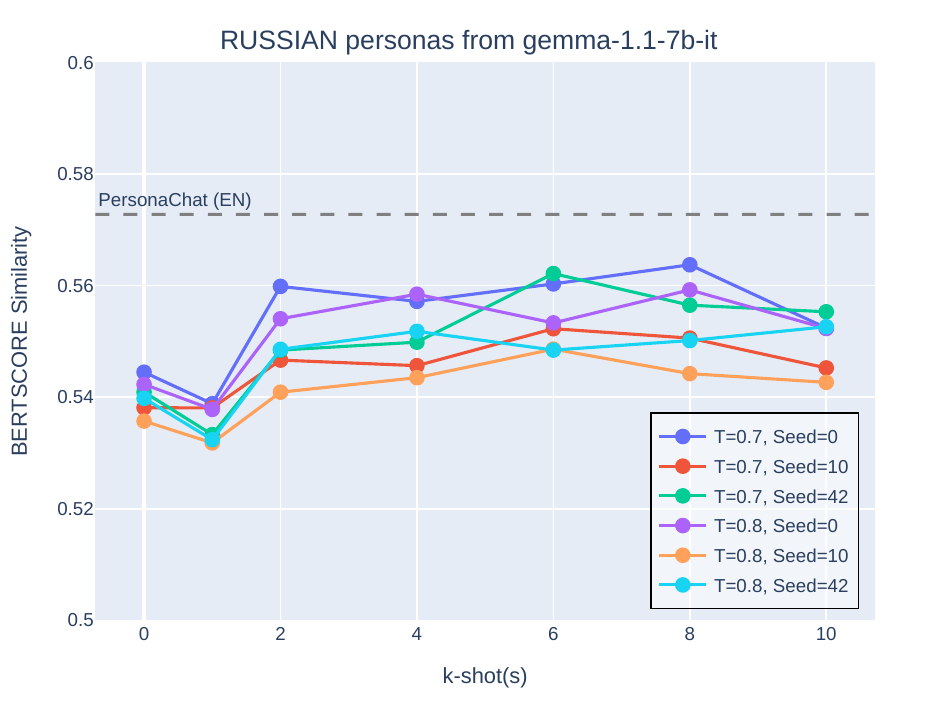}
        \end{subfigure}
        \vskip\baselineskip
        \begin{subfigure}{\textwidth}
            \centering
            \includegraphics[height=0.17\textheight]{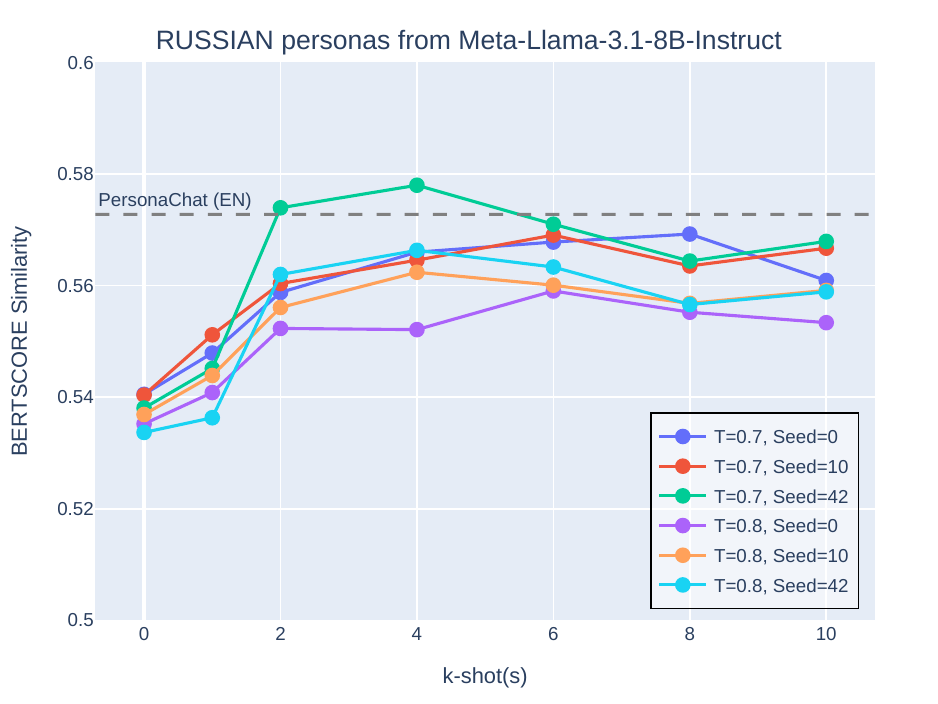}
        \end{subfigure}
        \vskip\baselineskip
        \begin{subfigure}{\textwidth}
            \centering
            \includegraphics[height=0.17\textheight]{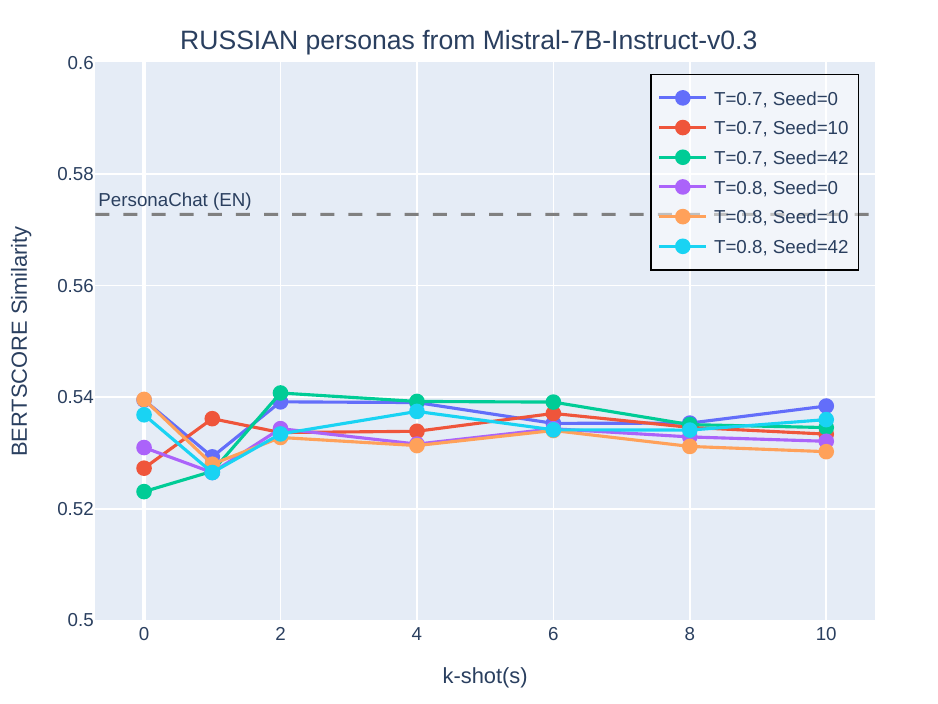}
        \end{subfigure}
    \end{minipage}
    %
    %
    \begin{minipage}{0.45\textwidth}
        \begin{subfigure}{\textwidth}
            \centering
            \includegraphics[height=0.17\textheight]{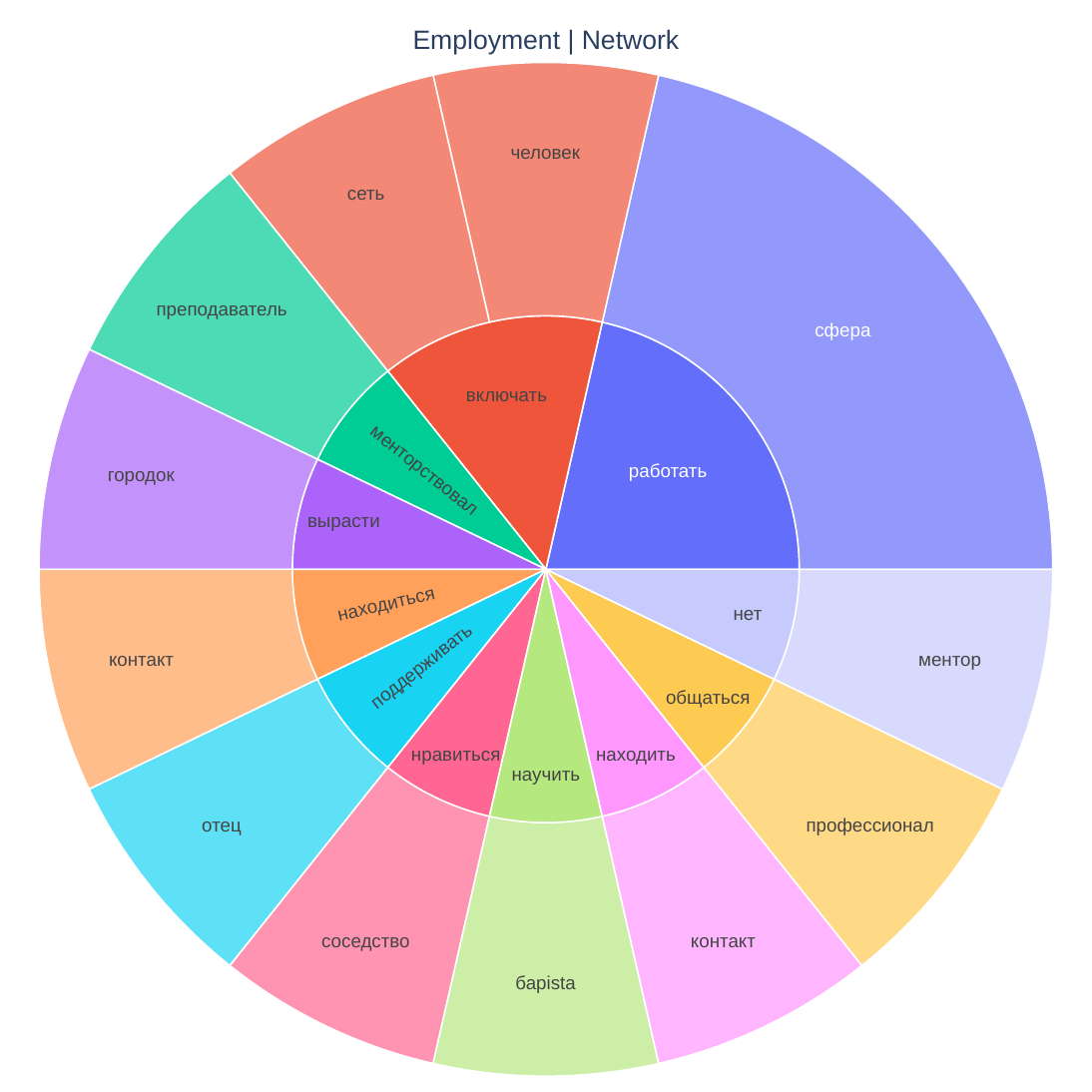}
        \end{subfigure}
        \vskip\baselineskip
        \begin{subfigure}{\textwidth}
            \centering
            \includegraphics[height=0.17\textheight]{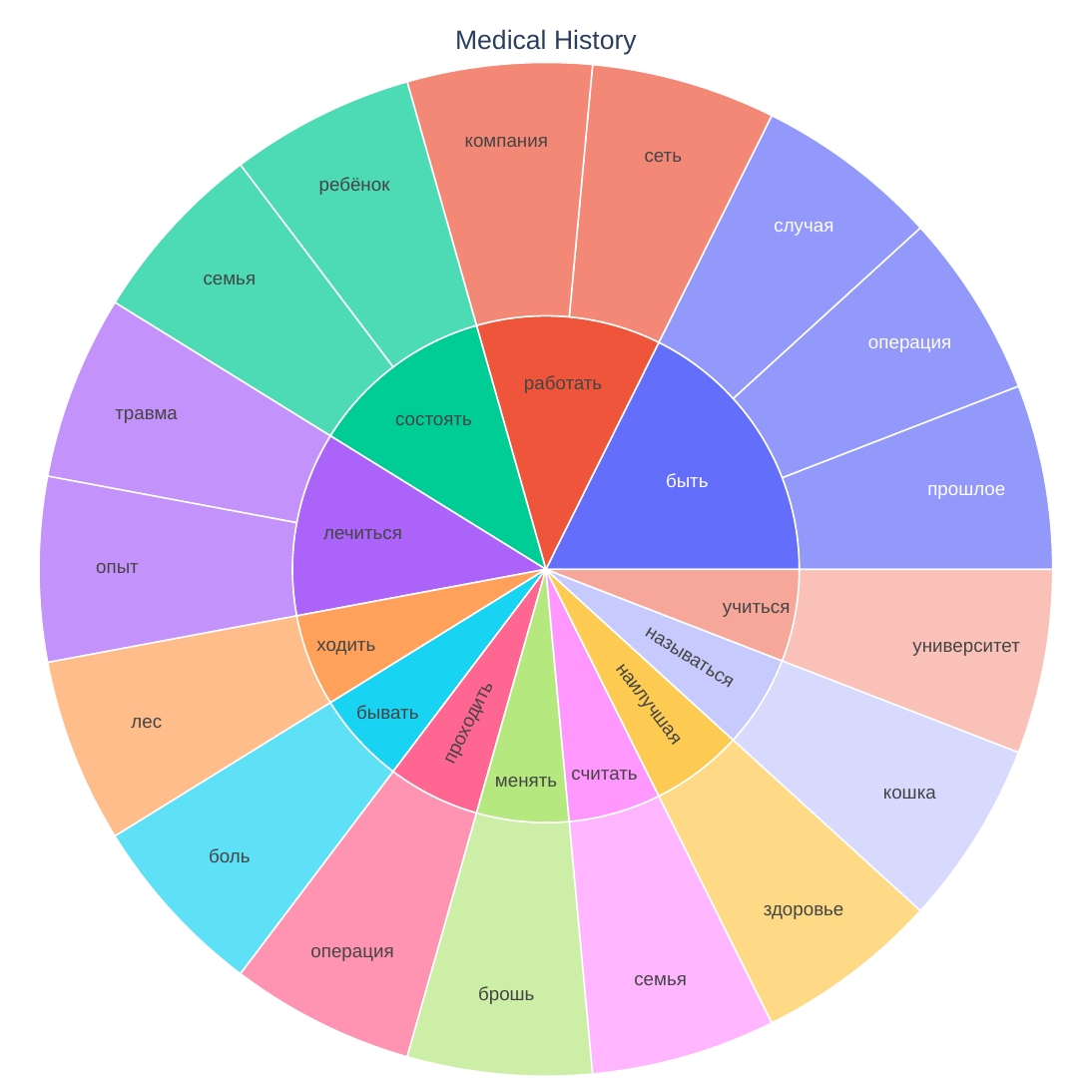}
        \end{subfigure}
        \vskip\baselineskip
        \begin{subfigure}{\textwidth}
            \centering
            \includegraphics[height=0.17\textheight]{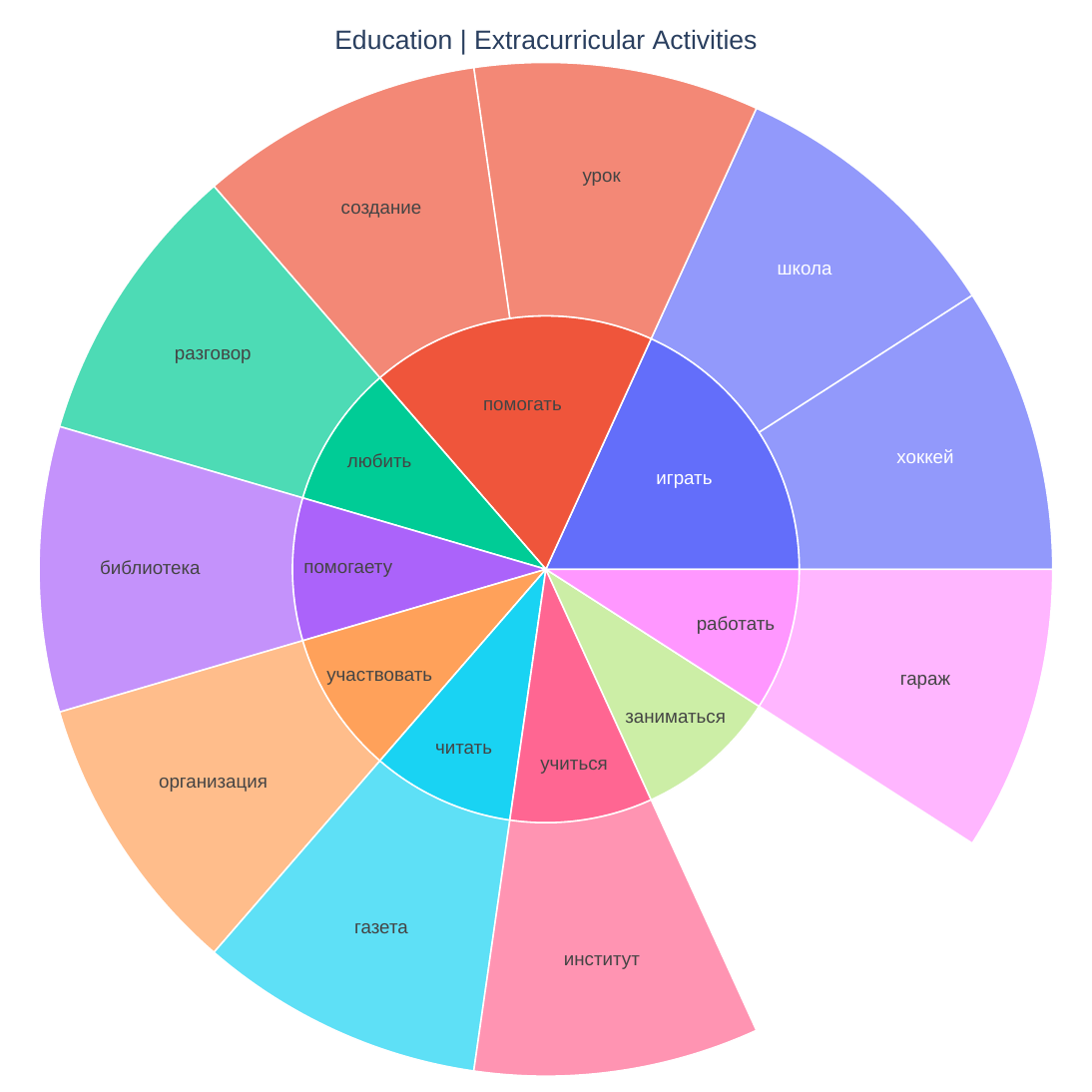}
        \end{subfigure}
        \vskip\baselineskip
        \begin{subfigure}{\textwidth}
            \centering
            \includegraphics[height=0.17\textheight]{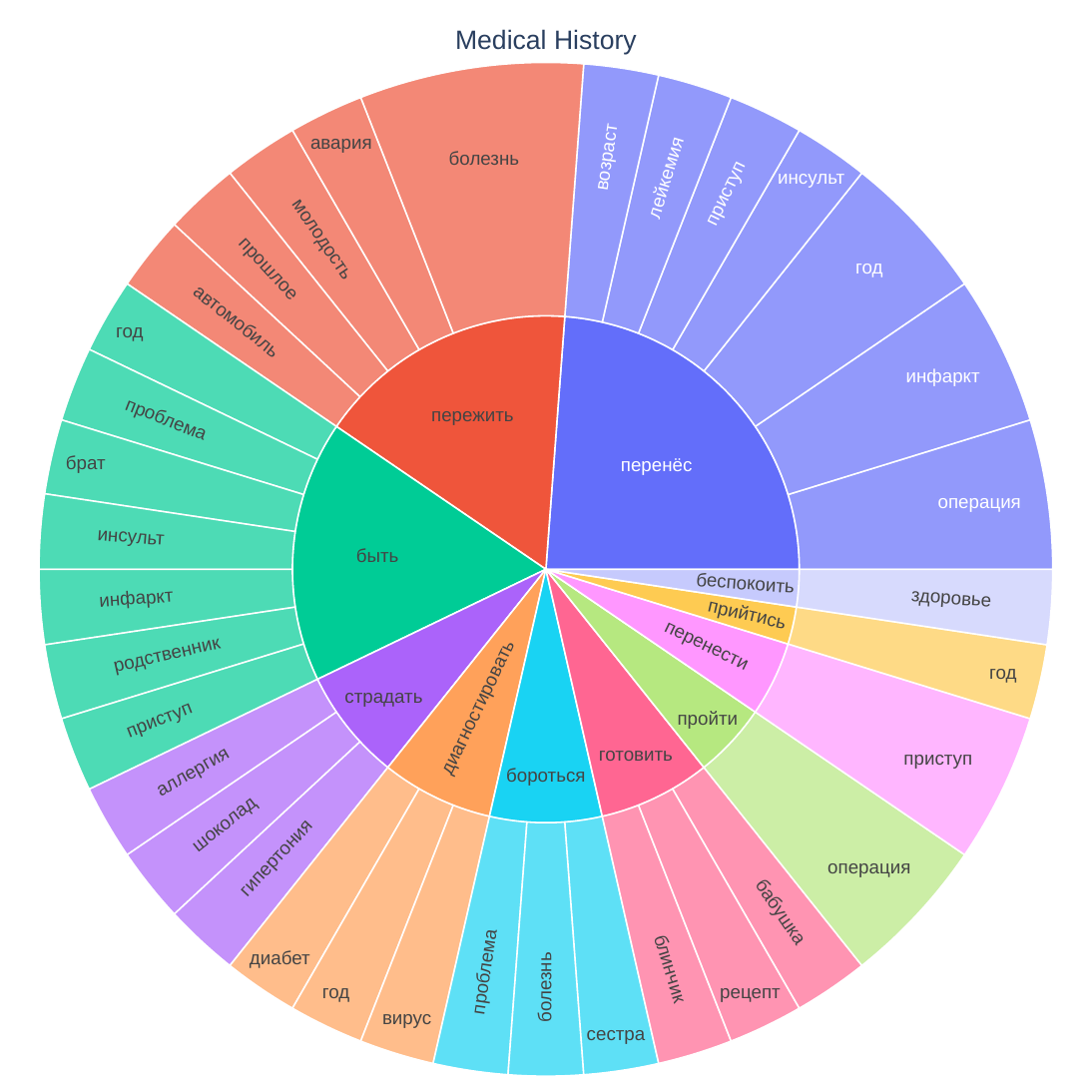}
        \end{subfigure}
    \end{minipage}
    
\caption{Detailed BERTSCORE for Russian Personas in different generation configurations and Sunburst charts of personas taxonomy entities with most root verbs and associated object noun for the different models}    
\end{figure}


\restoregeometry

\newgeometry{top=0.5cm, bottom=1.5cm, left=2.5cm, right=2.5cm}
\subsubsection{\textsc{German}}

\begin{figure}[h]
    \centering
    \begin{minipage}{0.45\textwidth}
        \begin{subfigure}{\textwidth}
            \centering
            \includegraphics[height=0.17\textheight]{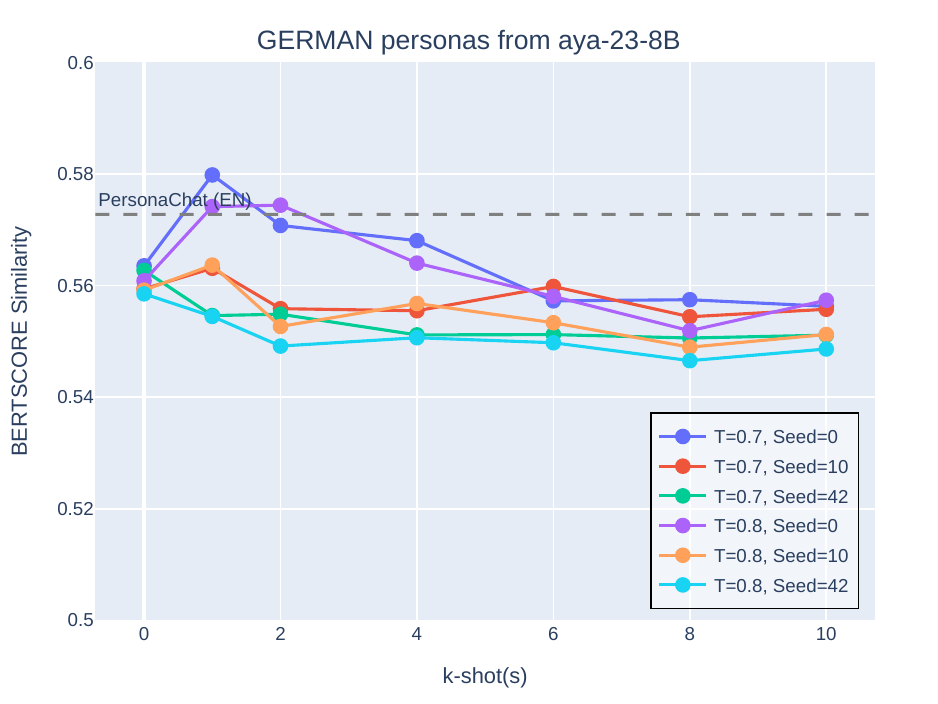}
        \end{subfigure}
        \vskip\baselineskip
        \begin{subfigure}{\textwidth}
            \centering
            \includegraphics[height=0.17\textheight]{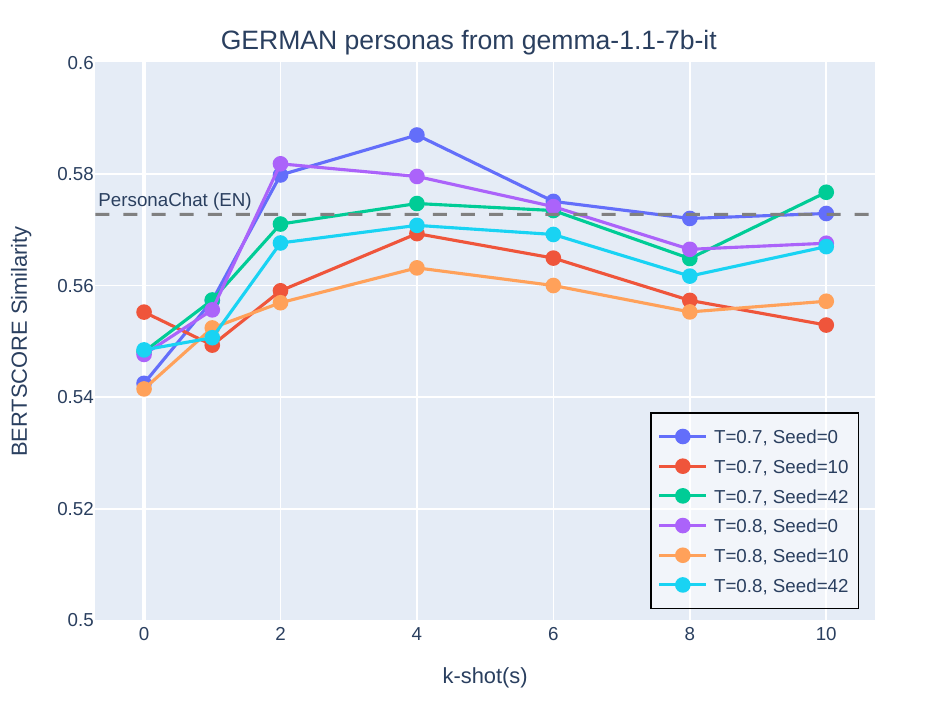}
        \end{subfigure}
        \vskip\baselineskip
        \begin{subfigure}{\textwidth}
            \centering
            \includegraphics[height=0.17\textheight]{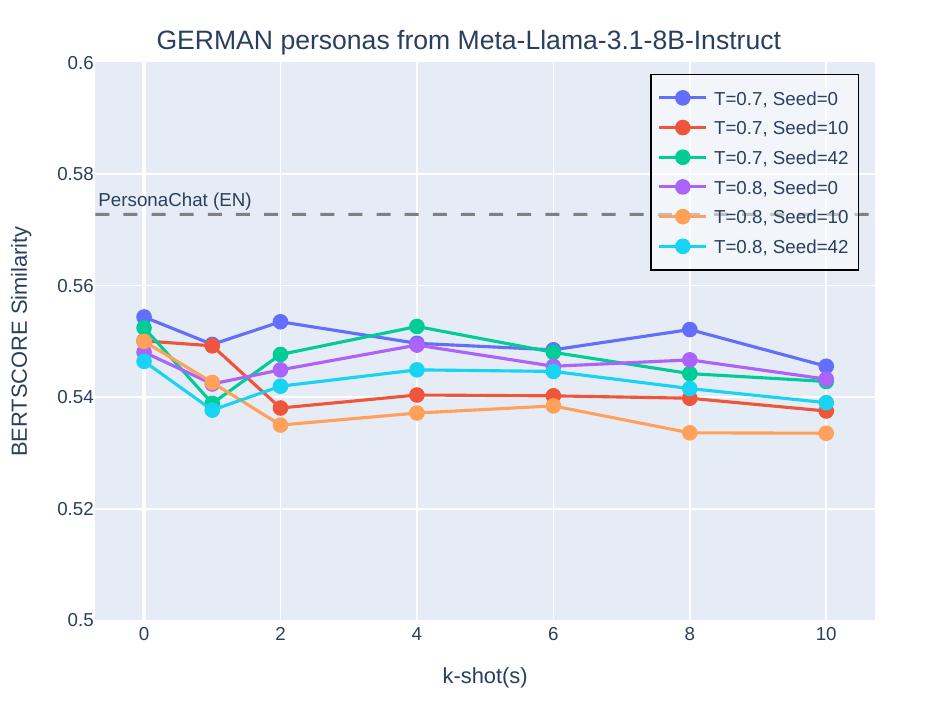}
        \end{subfigure}
        \vskip\baselineskip
        \begin{subfigure}{\textwidth}
            \centering
            \includegraphics[height=0.17\textheight]{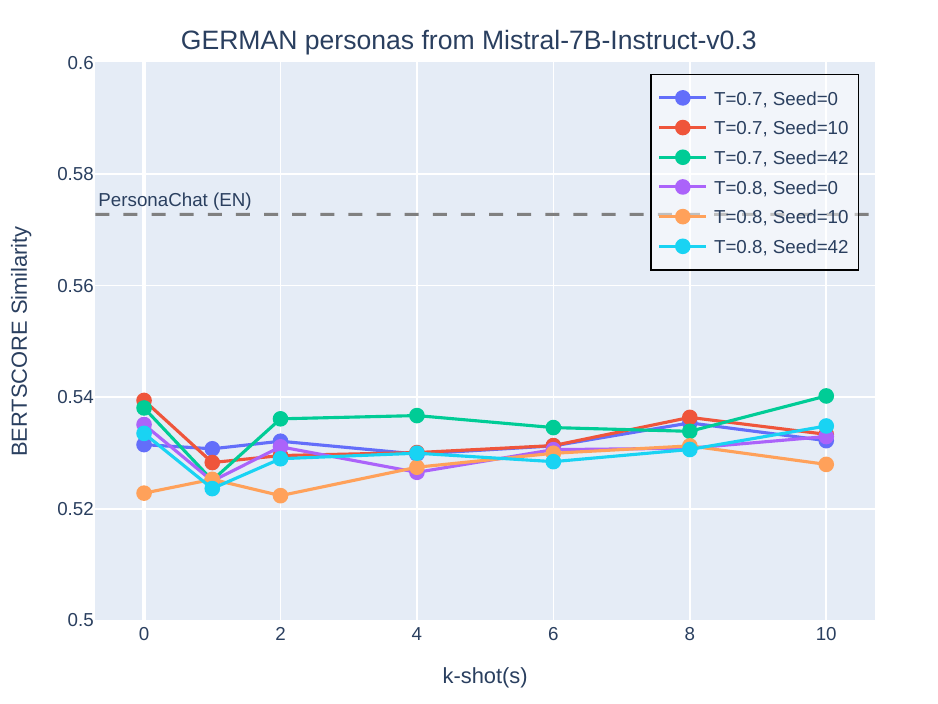}
        \end{subfigure}
    \end{minipage}
    %
    %
    \begin{minipage}{0.45\textwidth}
        \begin{subfigure}{\textwidth}
            \centering
            \includegraphics[height=0.17\textheight]{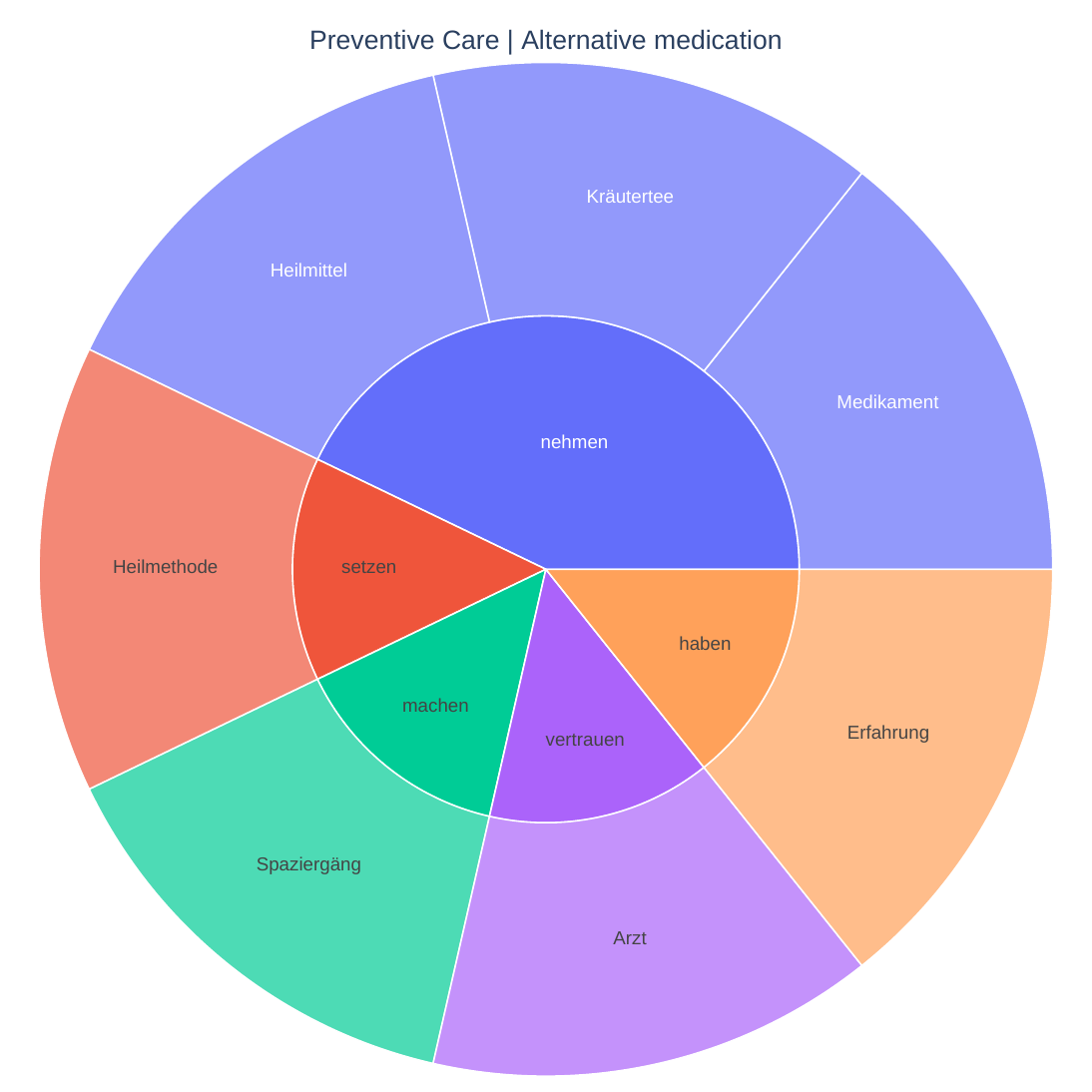}
        \end{subfigure}
        \vskip\baselineskip
        \begin{subfigure}{\textwidth}
            \centering
            \includegraphics[height=0.17\textheight]{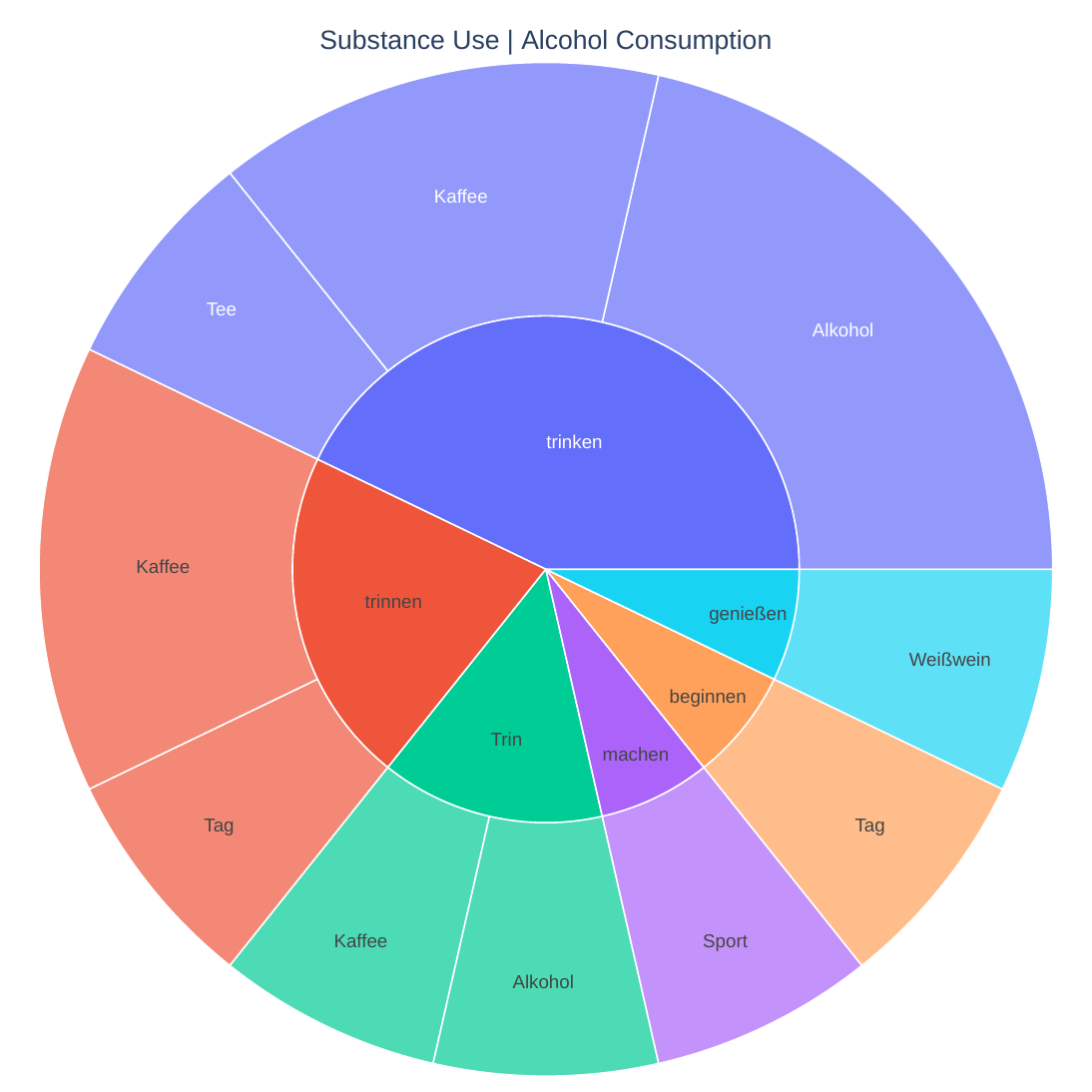}
        \end{subfigure}
        \vskip\baselineskip
        \begin{subfigure}{\textwidth}
            \centering
            \includegraphics[height=0.17\textheight]{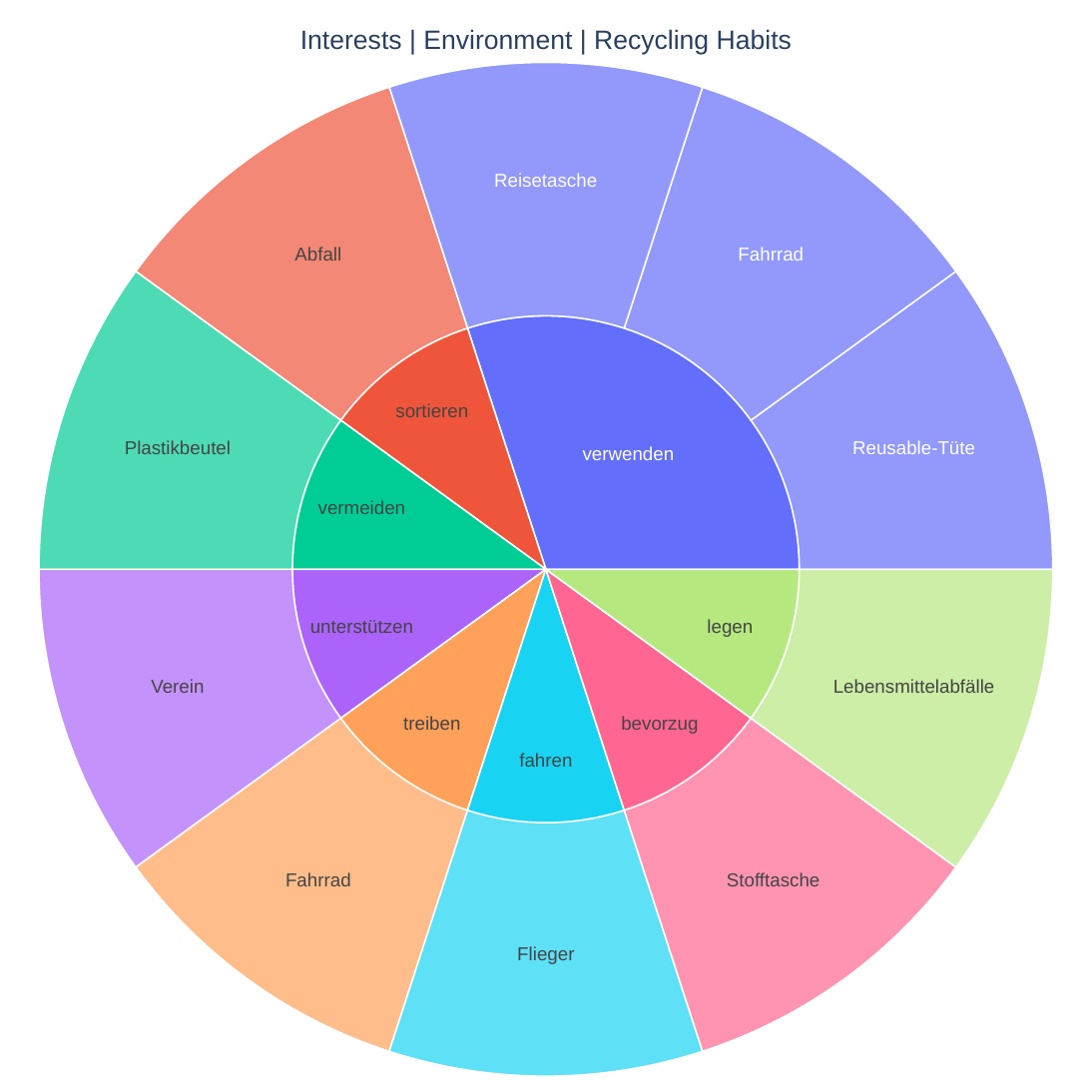}
        \end{subfigure}
        \vskip\baselineskip
        \begin{subfigure}{\textwidth}
            \centering
            \includegraphics[height=0.17\textheight]{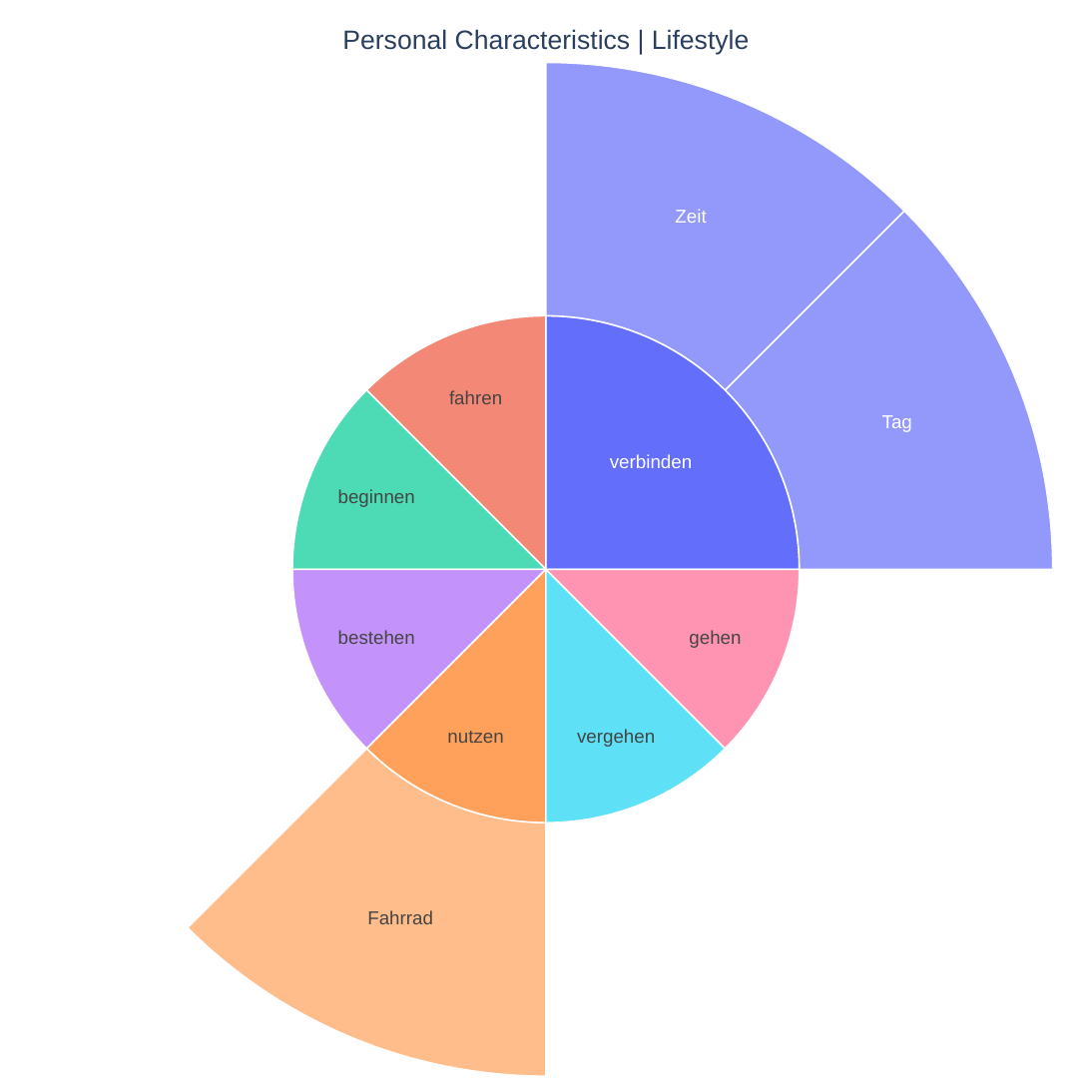}
        \end{subfigure}
    \end{minipage}
    
\caption{Detailed BERTSCORE for German Personas in different generation configurations and Sunburst charts of personas taxonomy entities with most root verbs and associated object noun for the different models}    
\end{figure}


\restoregeometry

\newgeometry{top=0.5cm, bottom=1.5cm, left=2.5cm, right=2.5cm}
\subsubsection{\textsc{Japanese}}

\begin{figure}[h]
    \centering
    \begin{minipage}{0.45\textwidth}
        \begin{subfigure}{\textwidth}
            \centering
            \includegraphics[height=0.17\textheight]{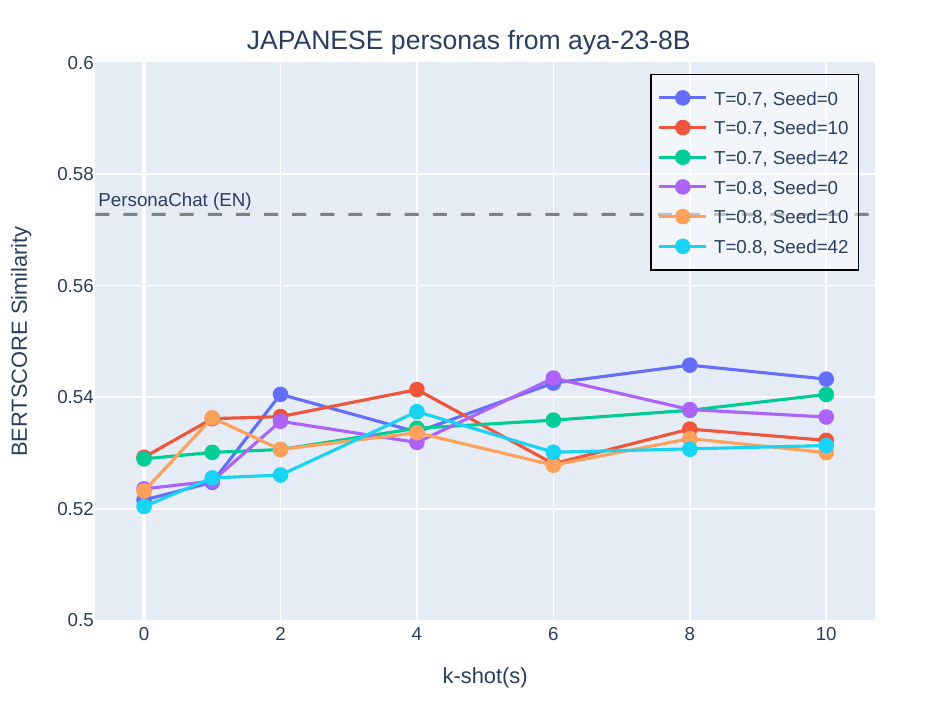}
        \end{subfigure}
        \vskip\baselineskip
        \begin{subfigure}{\textwidth}
            \centering
            \includegraphics[height=0.17\textheight]{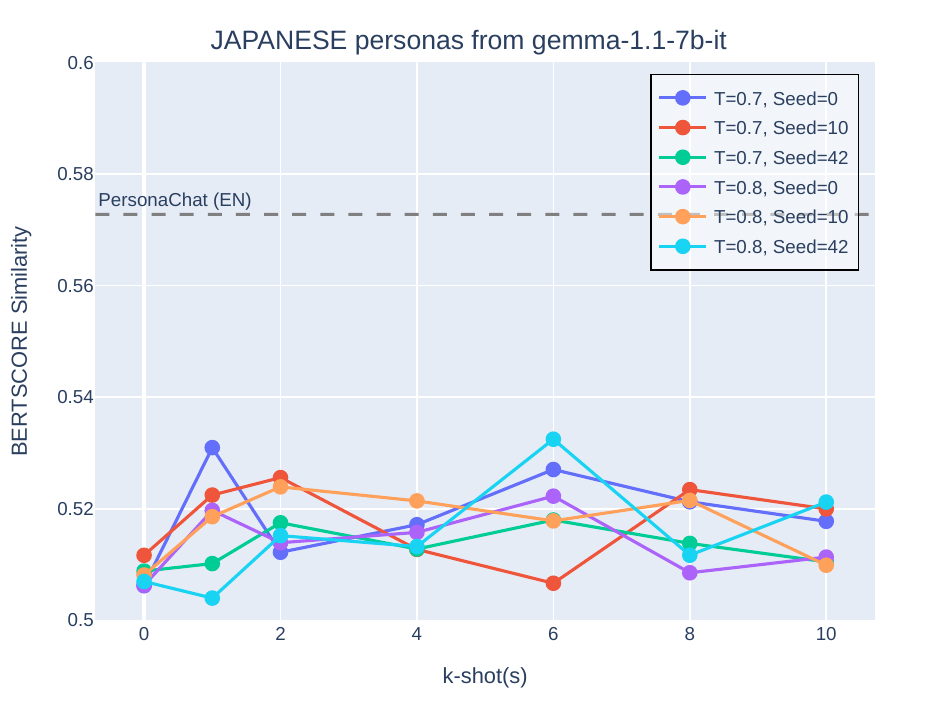}
        \end{subfigure}
        \vskip\baselineskip
        \begin{subfigure}{\textwidth}
            \centering
            \includegraphics[height=0.17\textheight]{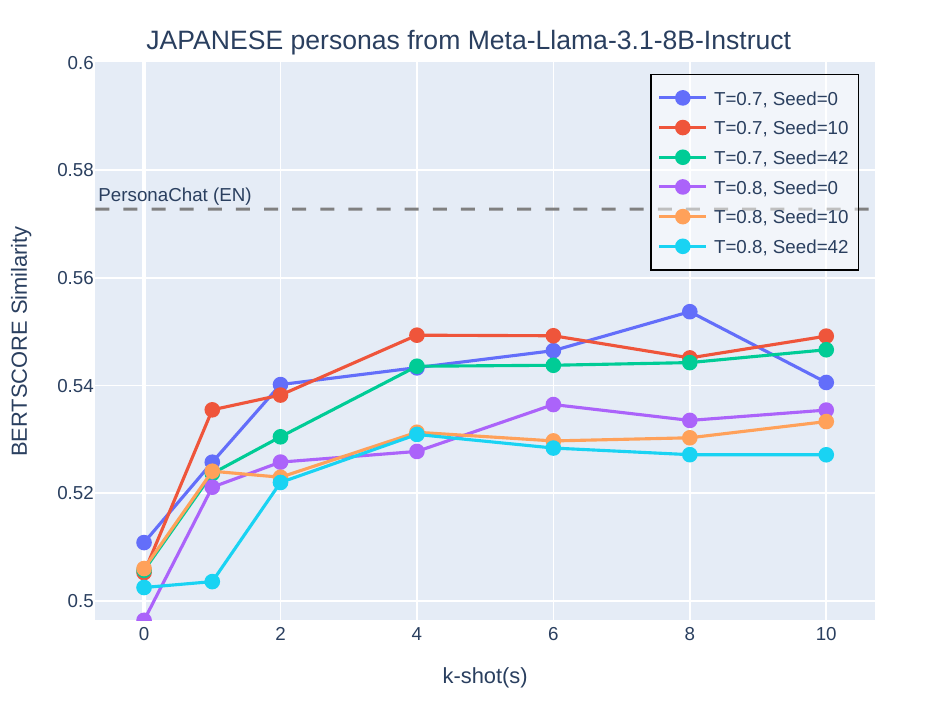}
        \end{subfigure}
        \vskip\baselineskip
        \begin{subfigure}{\textwidth}
            \centering
            \includegraphics[height=0.17\textheight]{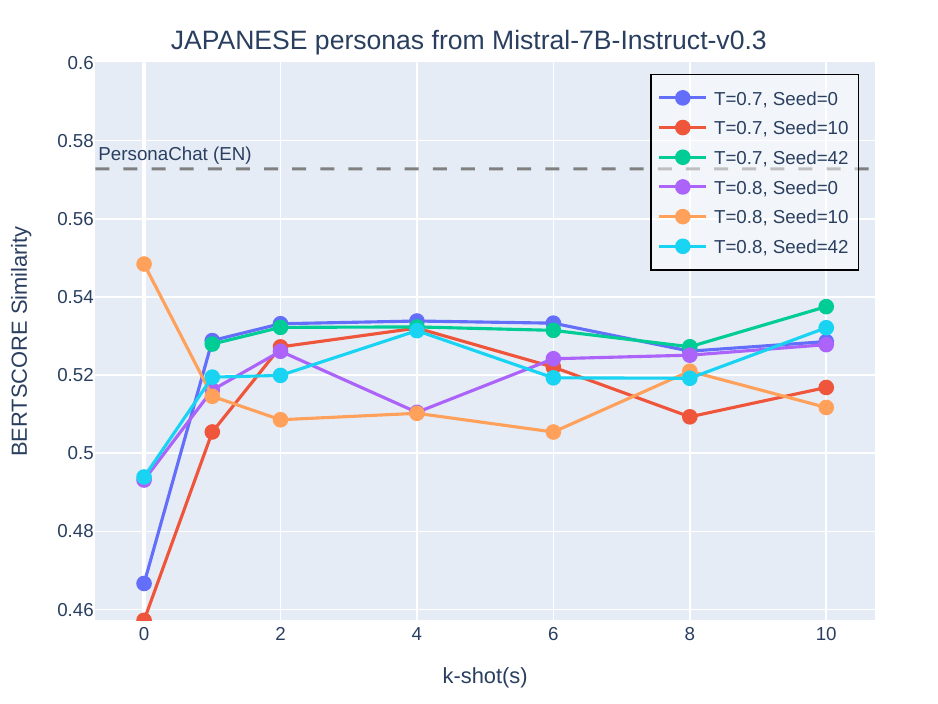}
        \end{subfigure}
    \end{minipage}
    %
    %
    \begin{minipage}{0.45\textwidth}
        \begin{subfigure}{\textwidth}
            \centering
            \includegraphics[height=0.17\textheight]{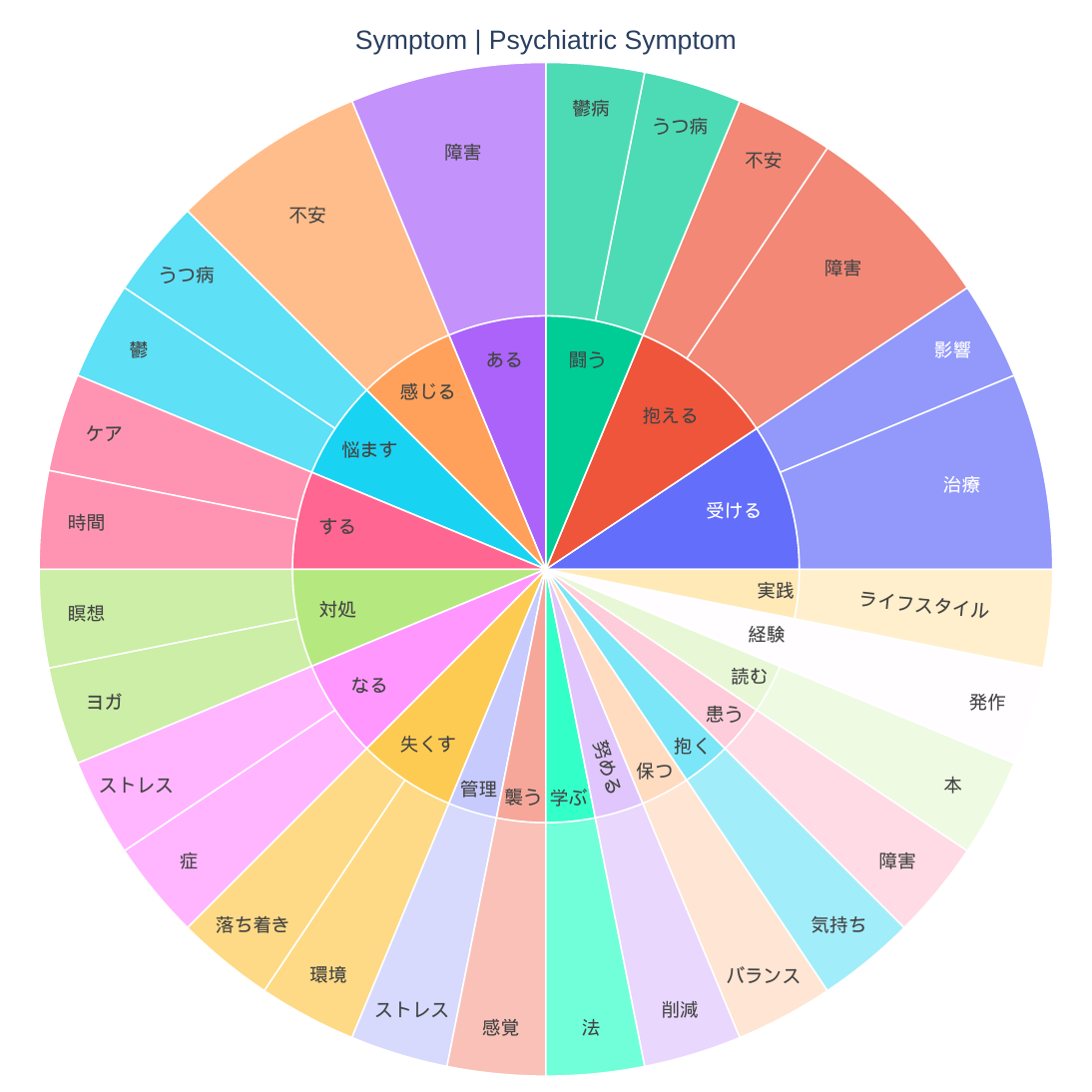}
        \end{subfigure}
        \vskip\baselineskip
        \begin{subfigure}{\textwidth}
            \centering
            \includegraphics[height=0.17\textheight]{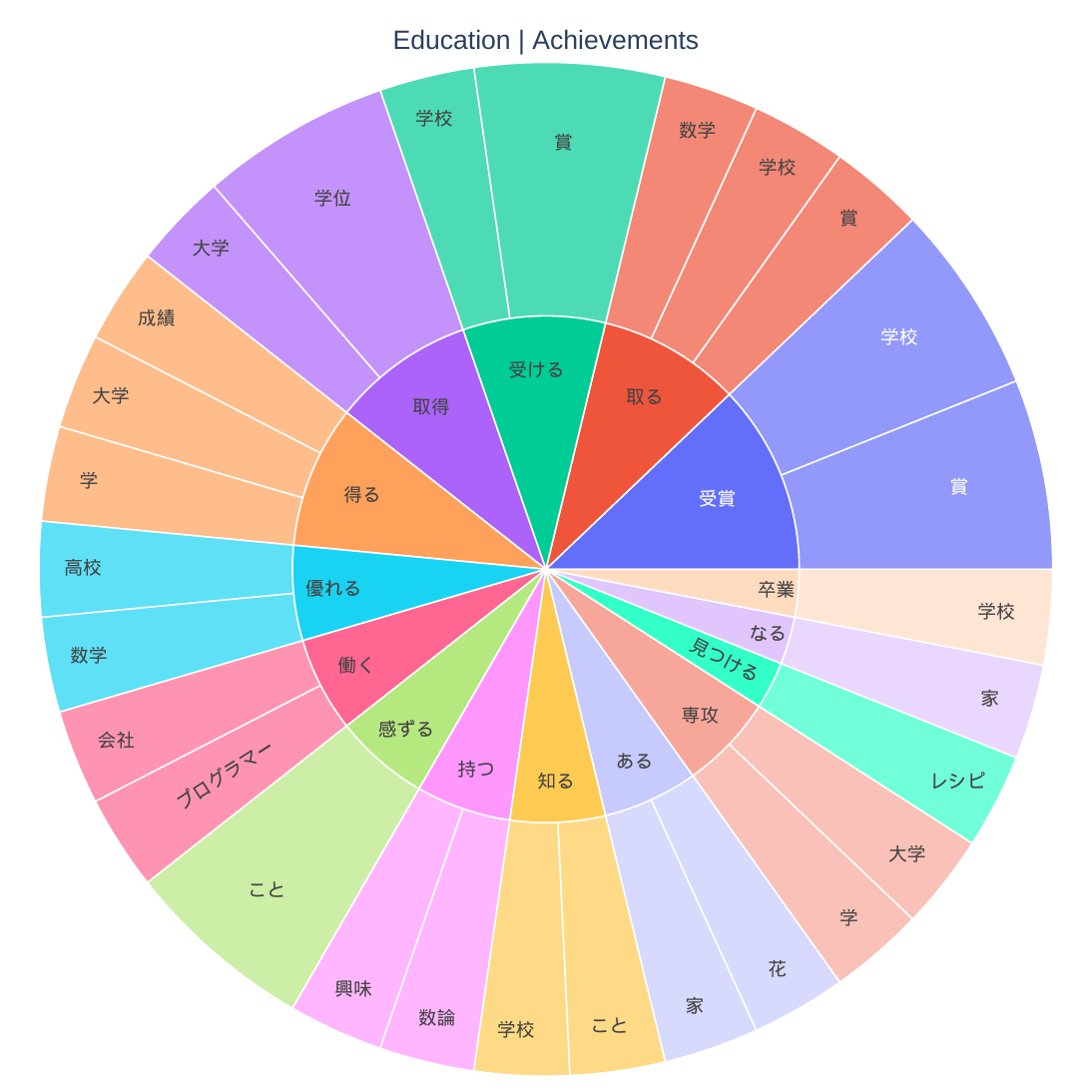}
        \end{subfigure}
        \vskip\baselineskip
        \begin{subfigure}{\textwidth}
            \centering
            \includegraphics[height=0.17\textheight]{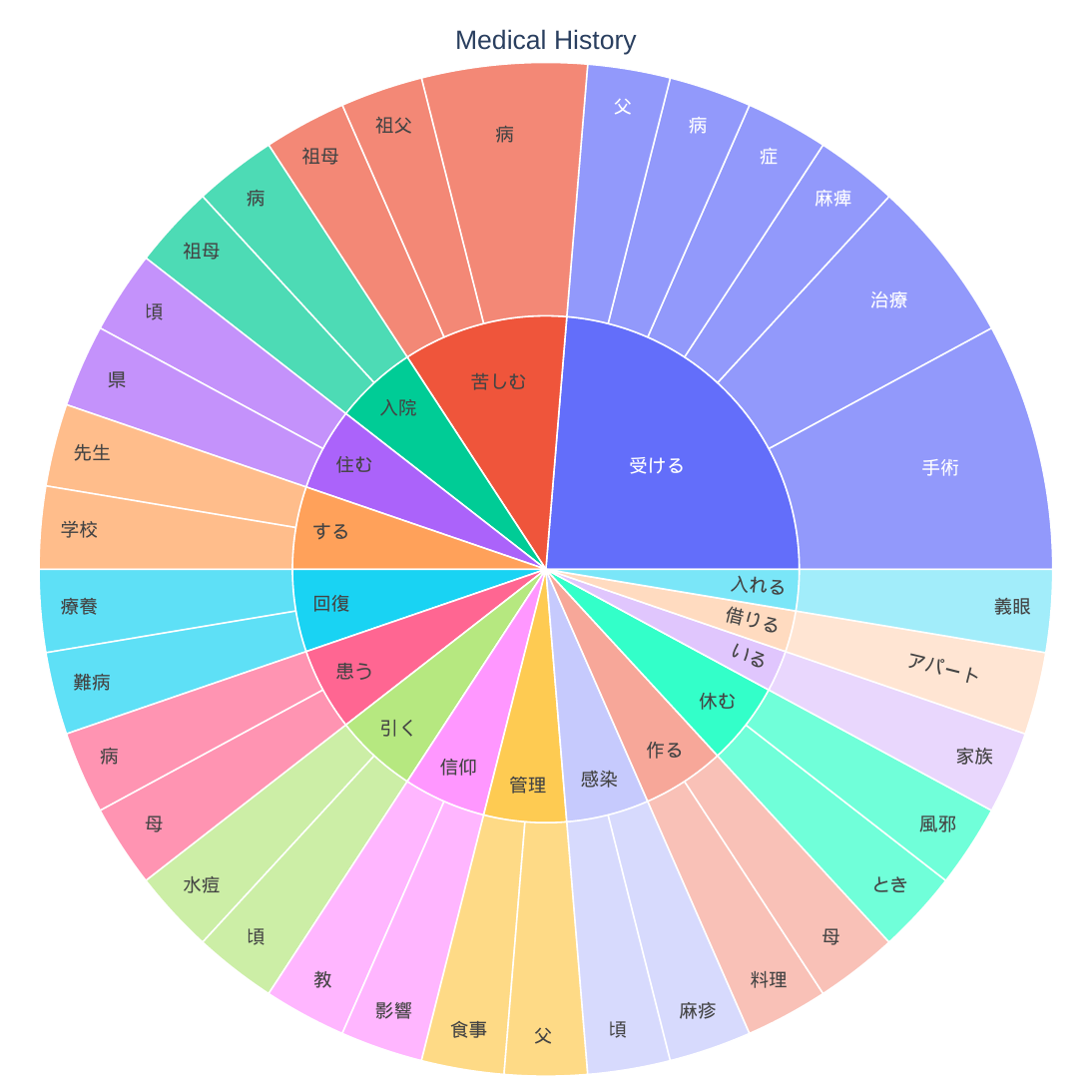}
        \end{subfigure}
        \vskip\baselineskip
        \begin{subfigure}{\textwidth}
            \centering
            \includegraphics[height=0.17\textheight]{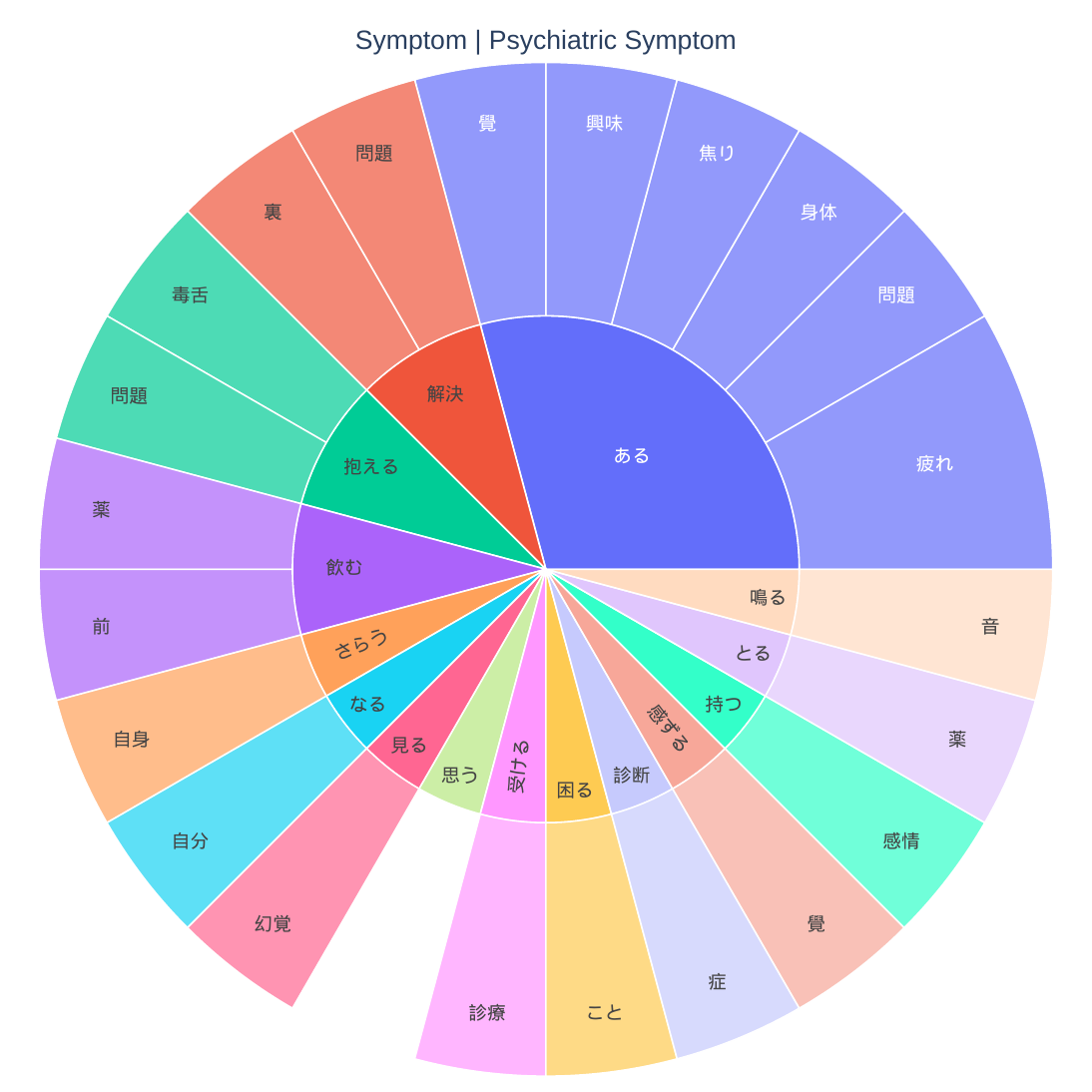}
        \end{subfigure}
    \end{minipage}
    
\caption{Detailed BERTSCORE for Japanese Personas in different generation configurations and Sunburst charts of personas taxonomy entities with most root verbs and associated object noun for the different models}    
\end{figure}


\restoregeometry

\newgeometry{top=0.5cm, bottom=1.5cm, left=2.5cm, right=2.5cm}
\subsubsection{\textsc{Spanish}}

\begin{figure}[h]
    \centering
    \begin{minipage}{0.45\textwidth}
        \begin{subfigure}{\textwidth}
            \centering
            \includegraphics[height=0.17\textheight]{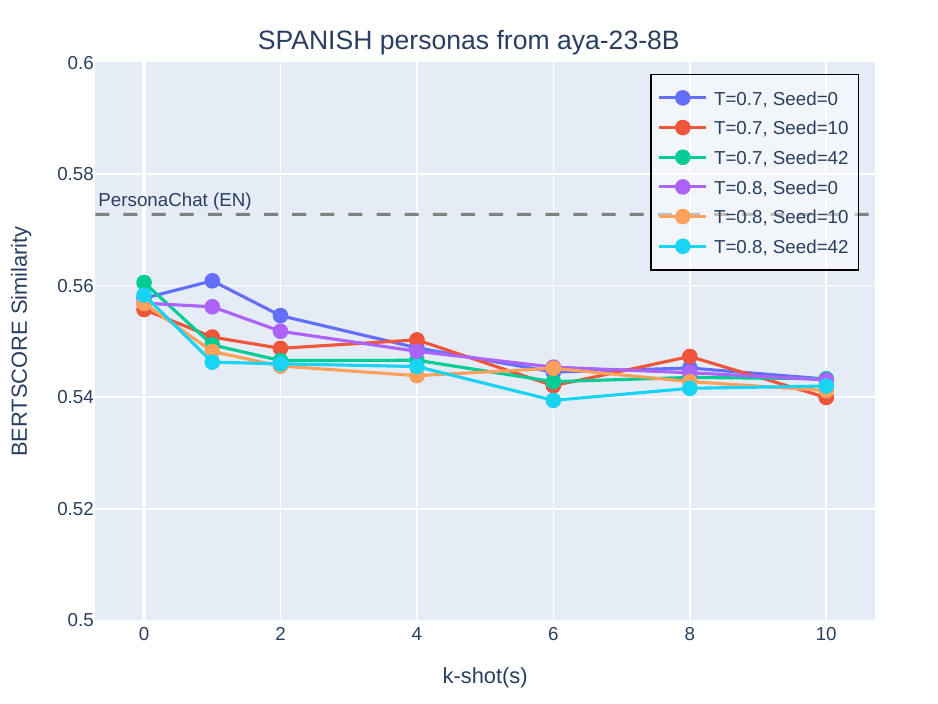}
        \end{subfigure}
        \vskip\baselineskip
        \begin{subfigure}{\textwidth}
            \centering
            \includegraphics[height=0.17\textheight]{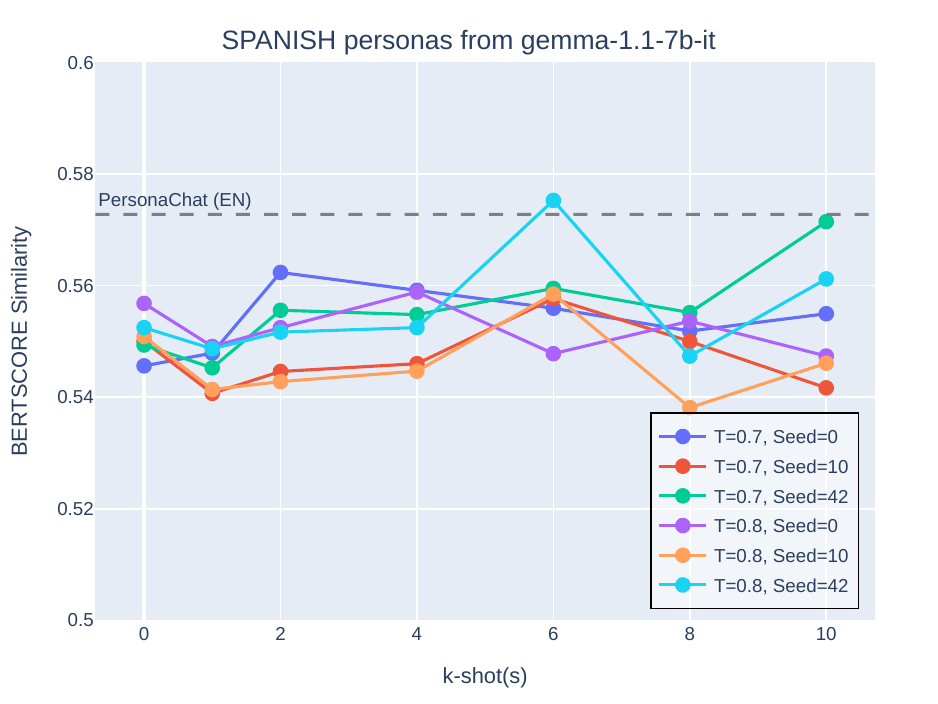}
        \end{subfigure}
        \vskip\baselineskip
        \begin{subfigure}{\textwidth}
            \centering
            \includegraphics[height=0.17\textheight]{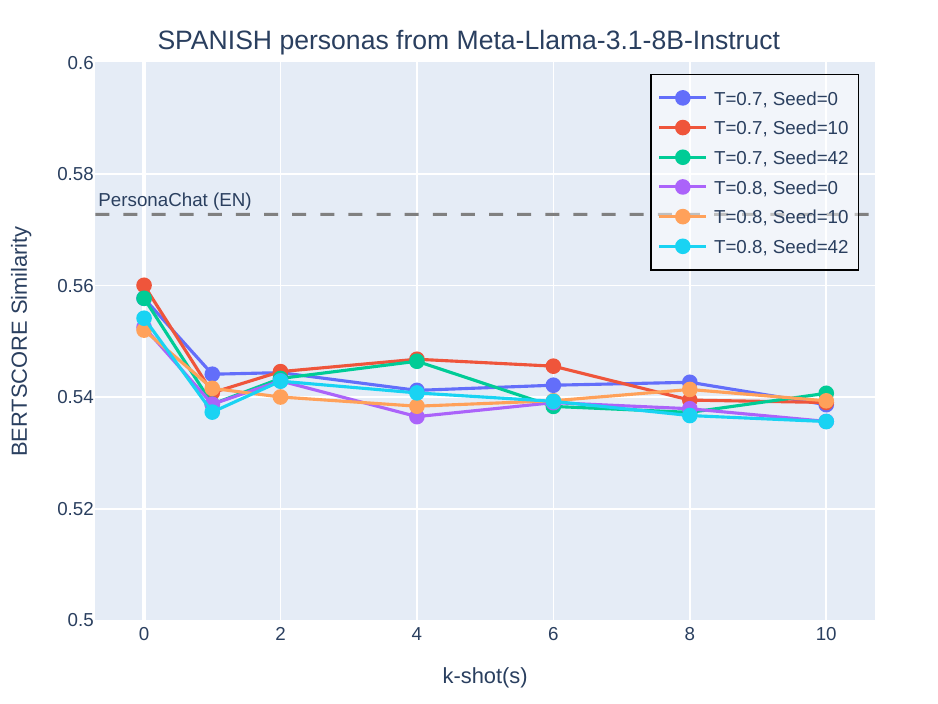}
        \end{subfigure}
        \vskip\baselineskip
        \begin{subfigure}{\textwidth}
            \centering
            \includegraphics[height=0.17\textheight]{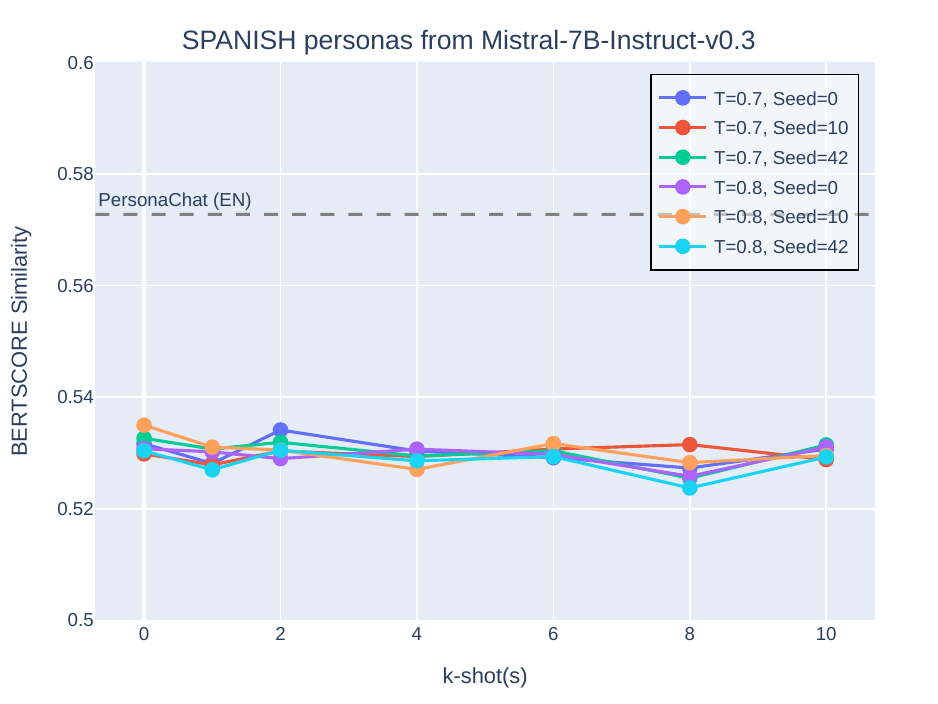}
        \end{subfigure}
    \end{minipage}
    %
    %
    \begin{minipage}{0.45\textwidth}
        \begin{subfigure}{\textwidth}
            \centering
            \includegraphics[height=0.17\textheight]{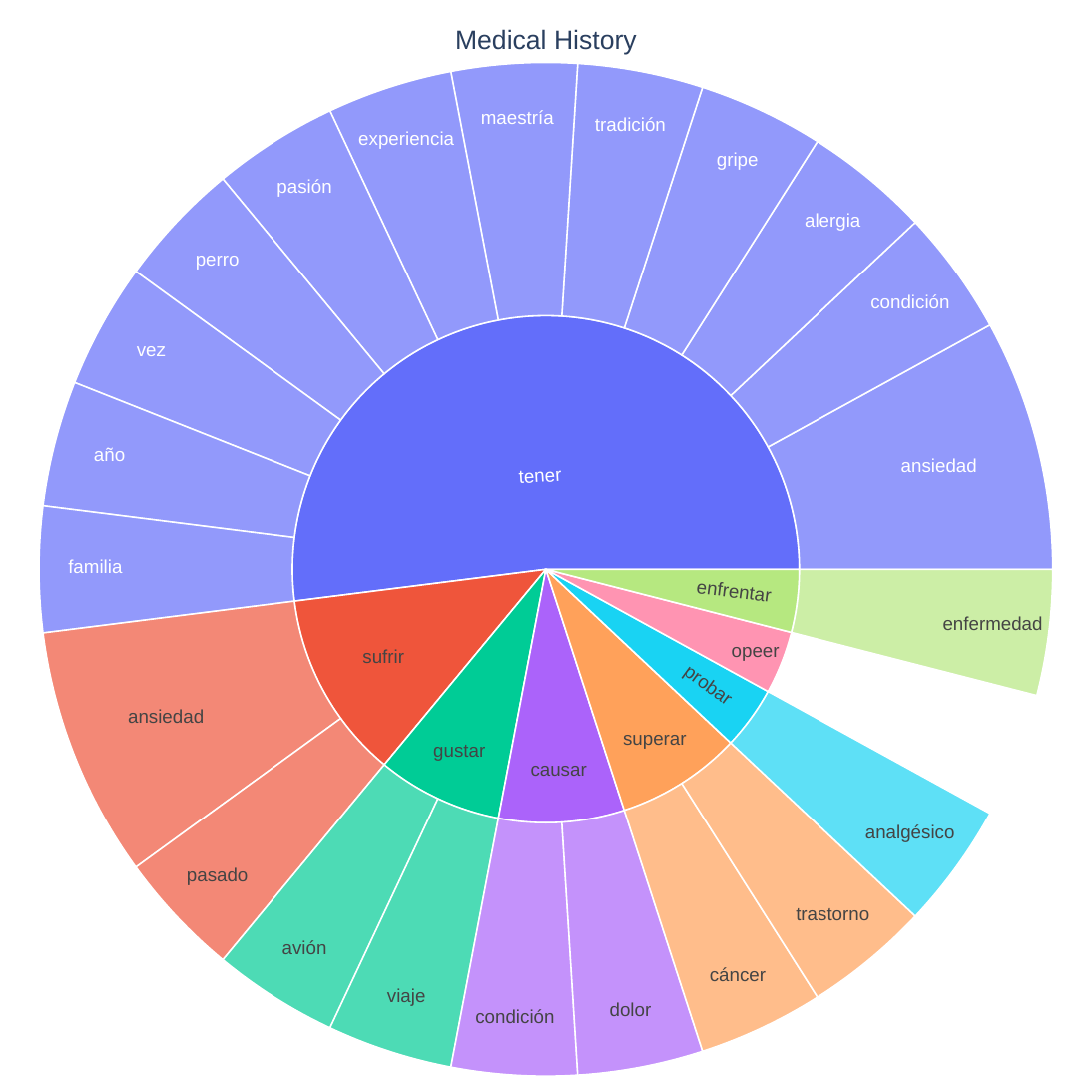}
        \end{subfigure}
        \vskip\baselineskip
        \begin{subfigure}{\textwidth}
            \centering
            \includegraphics[height=0.17\textheight]{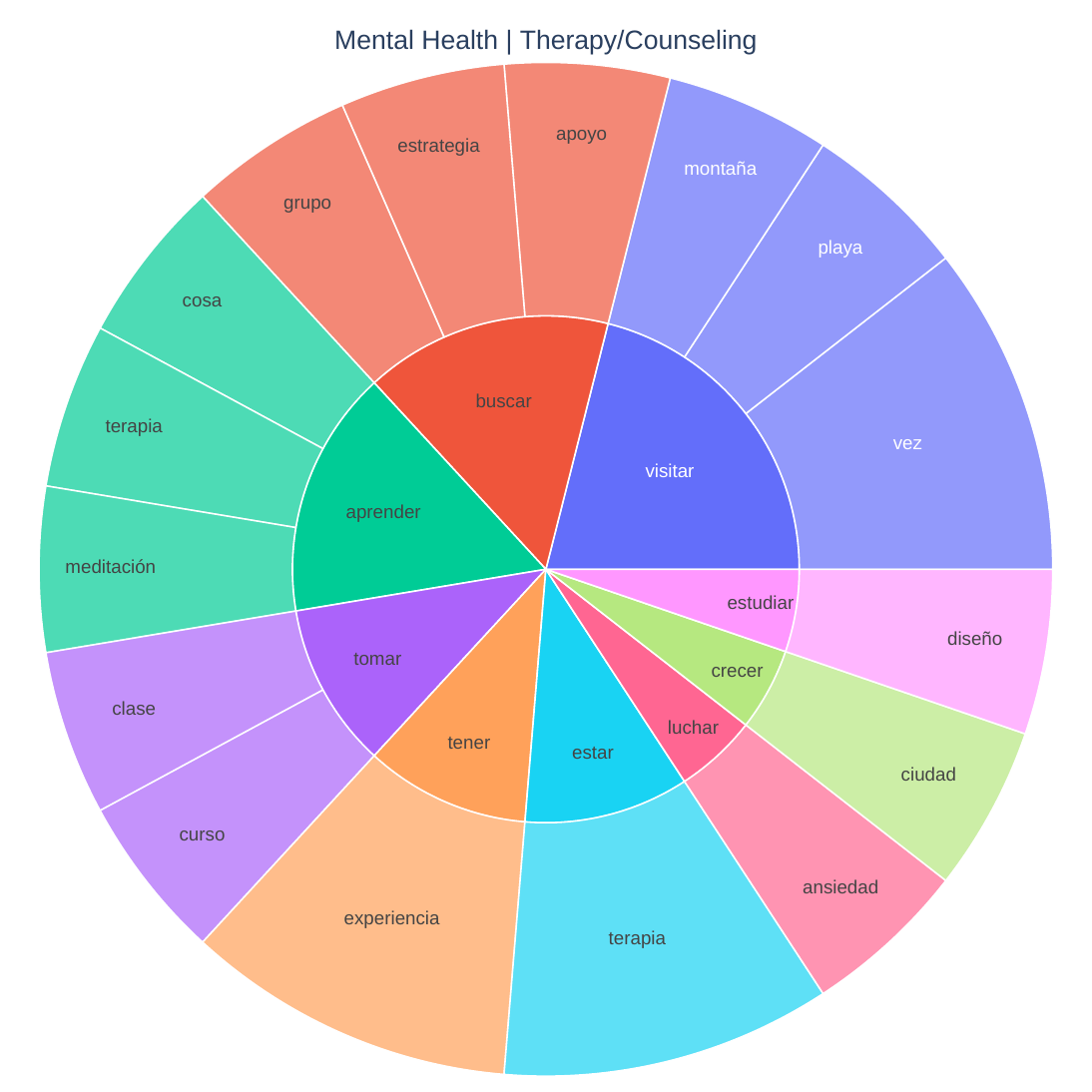}
        \end{subfigure}
        \vskip\baselineskip
        \begin{subfigure}{\textwidth}
            \centering
            \includegraphics[height=0.17\textheight]{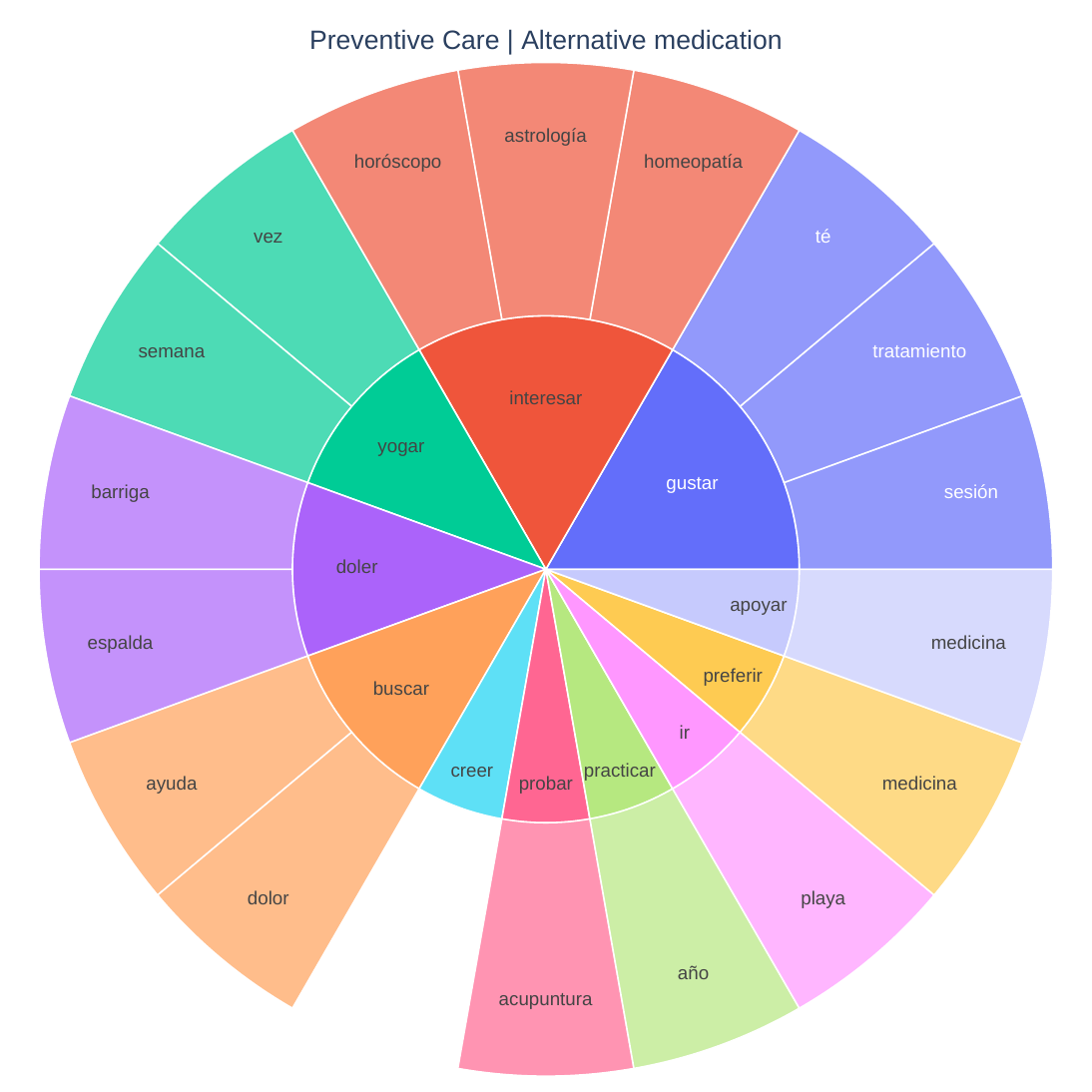}
        \end{subfigure}
        \vskip\baselineskip
        \begin{subfigure}{\textwidth}
            \centering
            \includegraphics[height=0.17\textheight]{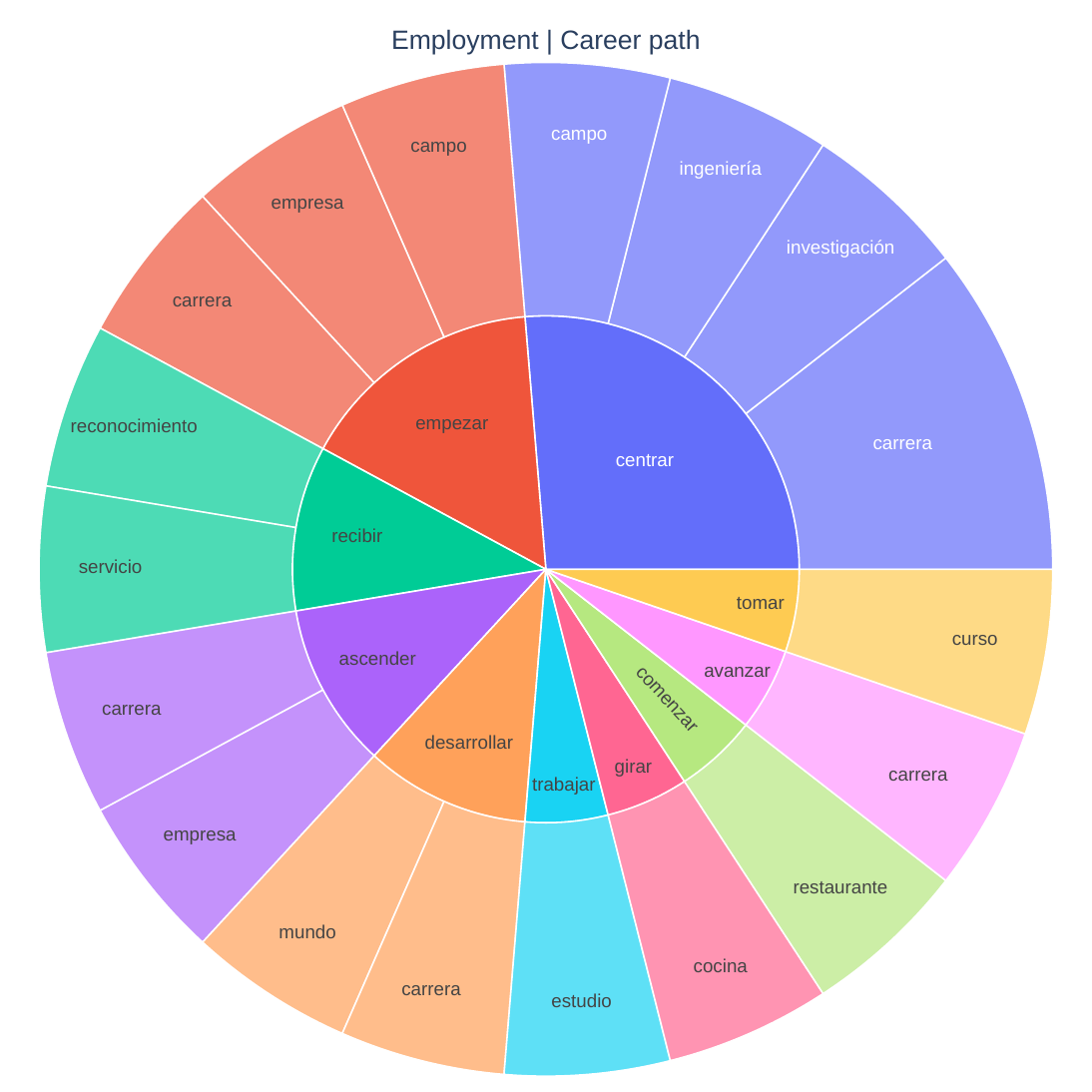}
        \end{subfigure}
    \end{minipage}
    
\caption{Detailed BERTSCORE for Spanish Personas in different generation configurations and Sunburst charts of personas taxonomy entities with most root verbs and associated object noun for the different models}    
\end{figure}


\restoregeometry

\newgeometry{top=0.5cm, bottom=1.5cm, left=2.5cm, right=2.5cm}
\subsubsection{\textsc{Chinese}}

\begin{figure}[h]
    \centering
    \begin{minipage}{0.45\textwidth}
        \begin{subfigure}{\textwidth}
            \centering
            \includegraphics[height=0.17\textheight]{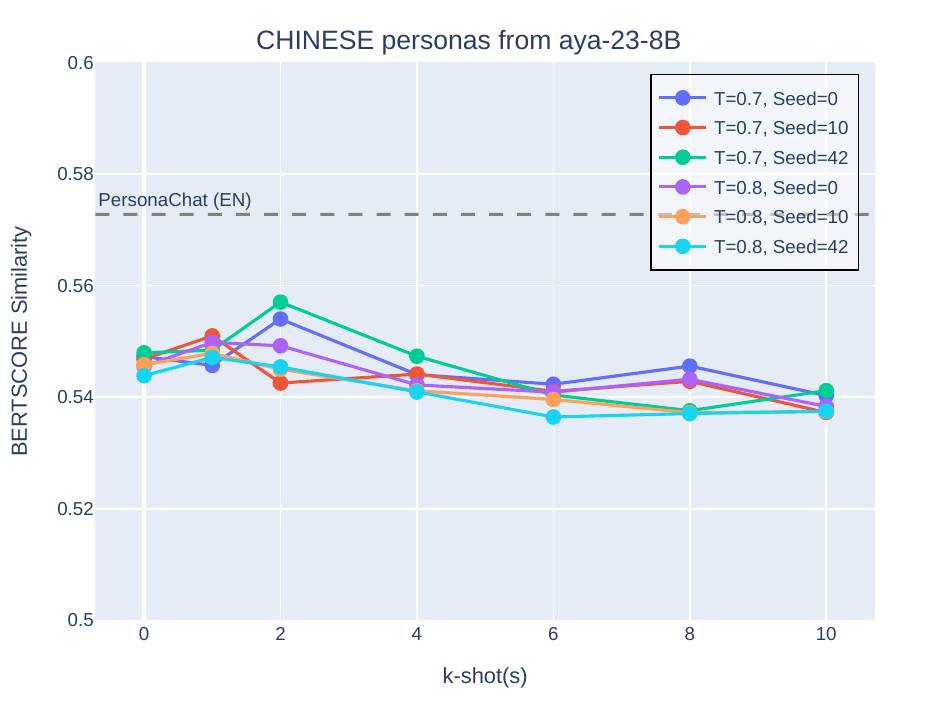}
        \end{subfigure}
        \vskip\baselineskip
        \begin{subfigure}{\textwidth}
            \centering
            \includegraphics[height=0.17\textheight]{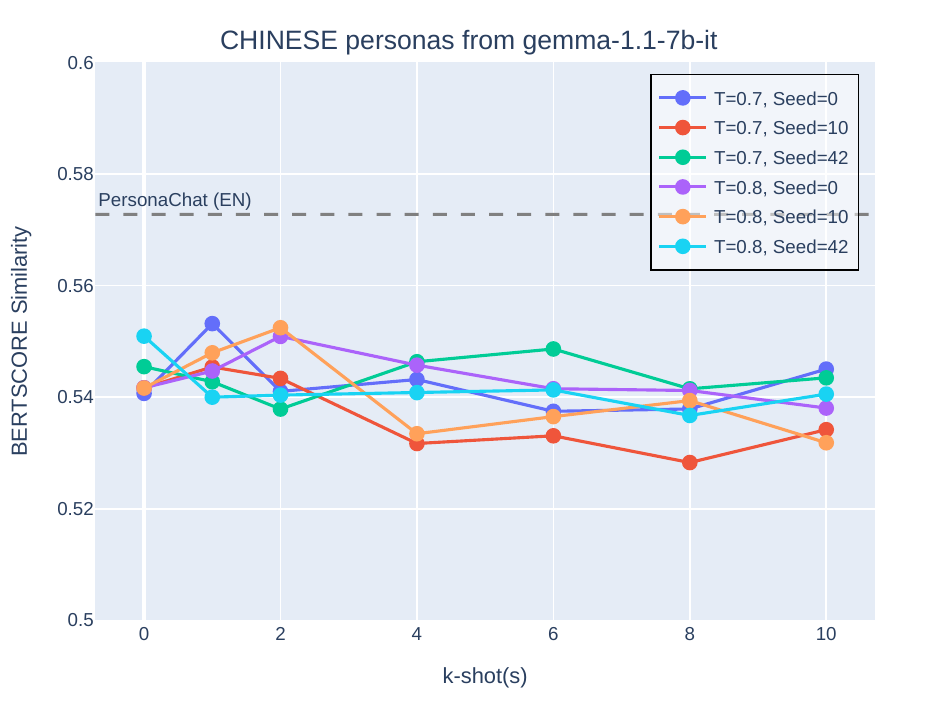}
        \end{subfigure}
        \vskip\baselineskip
        \begin{subfigure}{\textwidth}
            \centering
            \includegraphics[height=0.17\textheight]{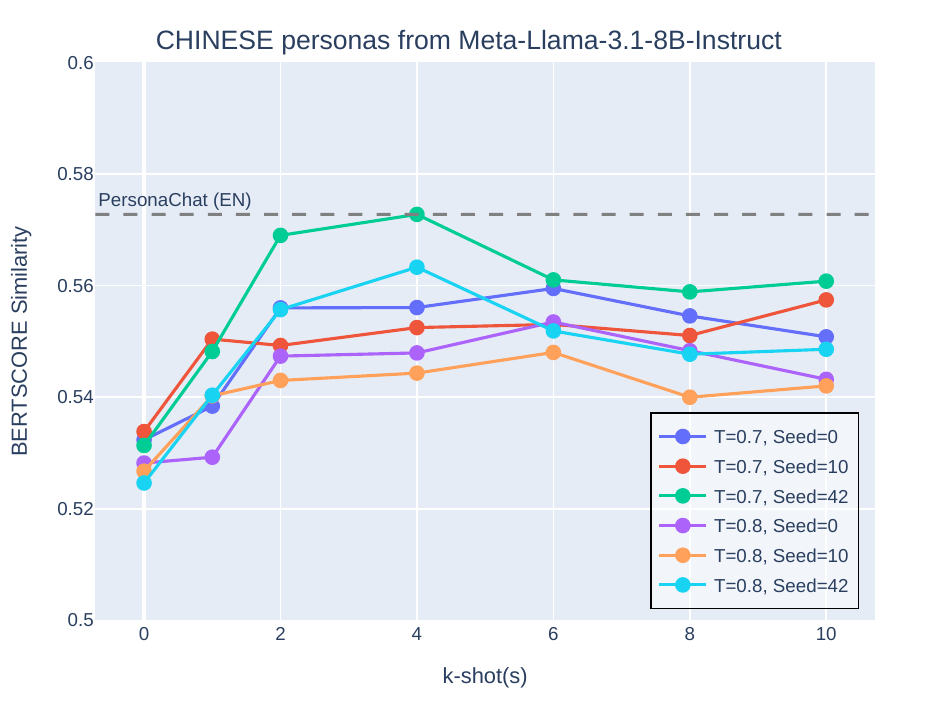}
        \end{subfigure}
        \vskip\baselineskip
        \begin{subfigure}{\textwidth}
            \centering
            \includegraphics[height=0.17\textheight]{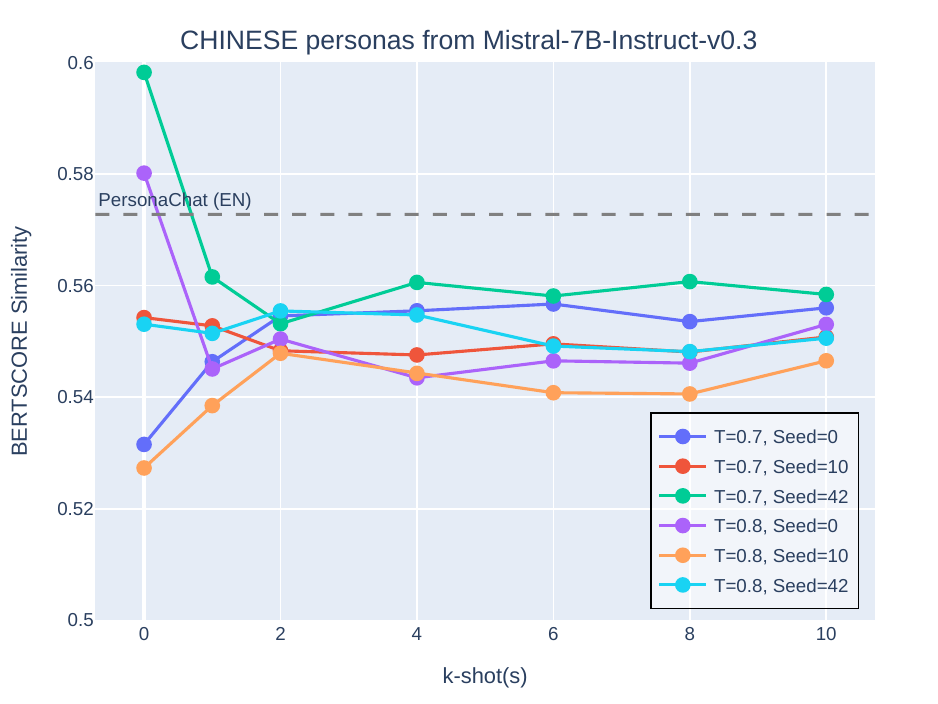}
        \end{subfigure}
    \end{minipage}
    %
    %
    \begin{minipage}{0.45\textwidth}
        \begin{subfigure}{\textwidth}
            \centering
            \includegraphics[height=0.17\textheight]{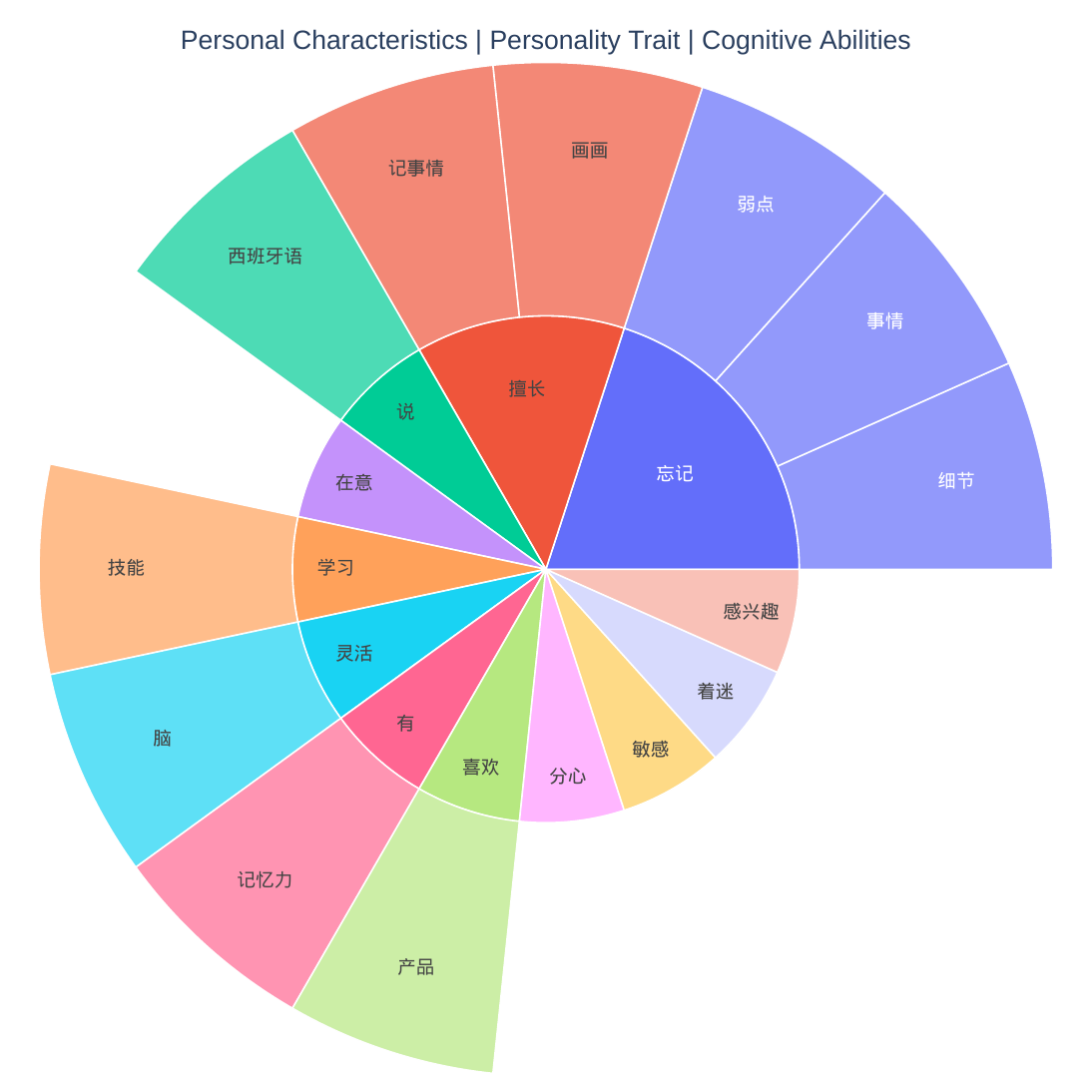}
        \end{subfigure}
        \vskip\baselineskip
        \begin{subfigure}{\textwidth}
            \centering
            \includegraphics[height=0.17\textheight]{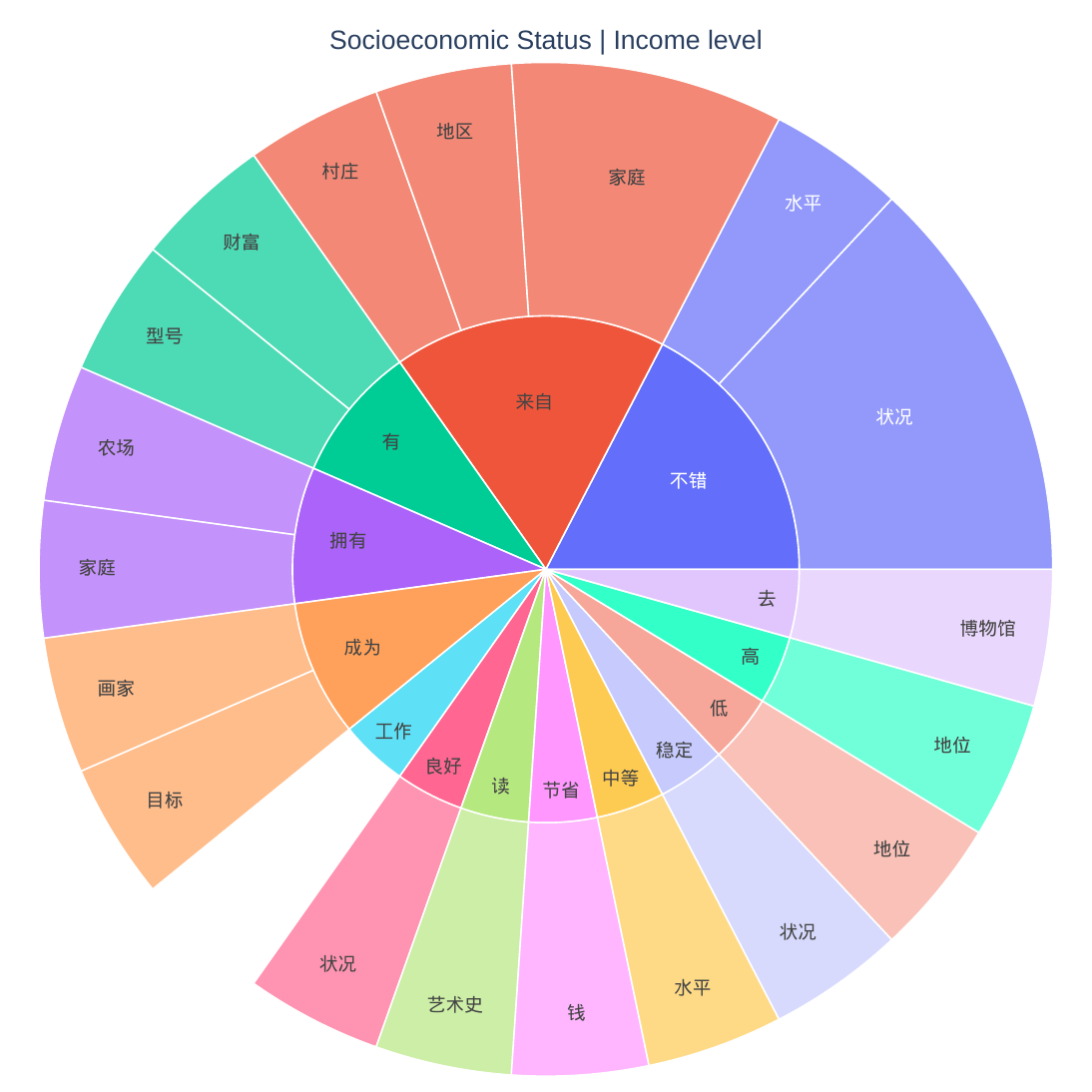}
        \end{subfigure}
        \vskip\baselineskip
        \begin{subfigure}{\textwidth}
            \centering
            \includegraphics[height=0.17\textheight]{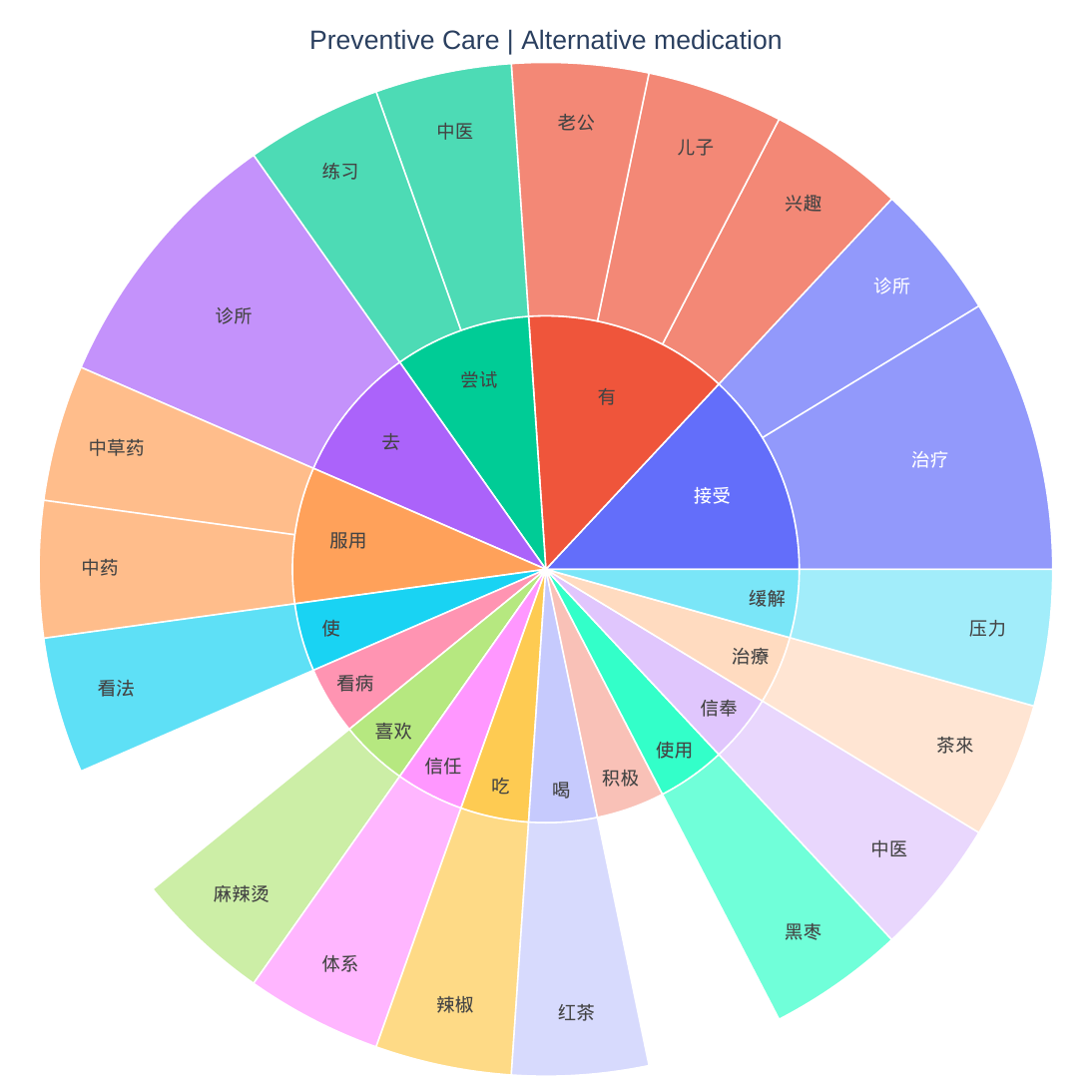}
        \end{subfigure}
        \vskip\baselineskip
        \begin{subfigure}{\textwidth}
            \centering
            \includegraphics[height=0.17\textheight]{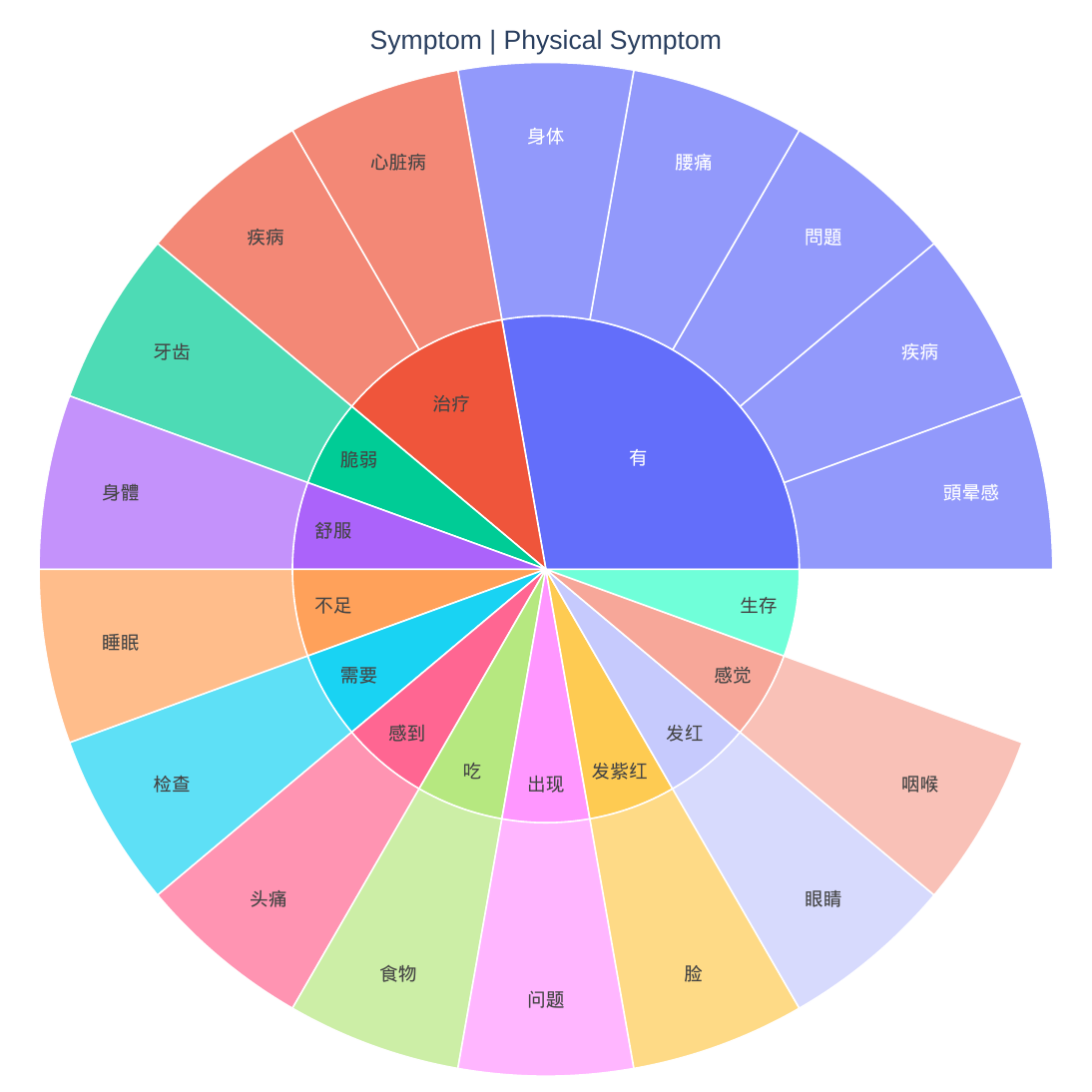}
        \end{subfigure}
    \end{minipage}
    
\caption{Detailed BERTSCORE for Chinese Personas in different generation configurations and Sunburst charts of personas taxonomy entities with most root verbs and associated object noun for the different models}    
\end{figure}

%
\restoregeometry

\newgeometry{top=0.5cm, bottom=1.5cm, left=2.5cm, right=2.5cm}
\subsubsection{\textsc{French}}

\begin{figure}[h]
    \centering
    \begin{minipage}{0.45\textwidth}
        \begin{subfigure}{\textwidth}
            \centering
            \includegraphics[height=0.17\textheight]{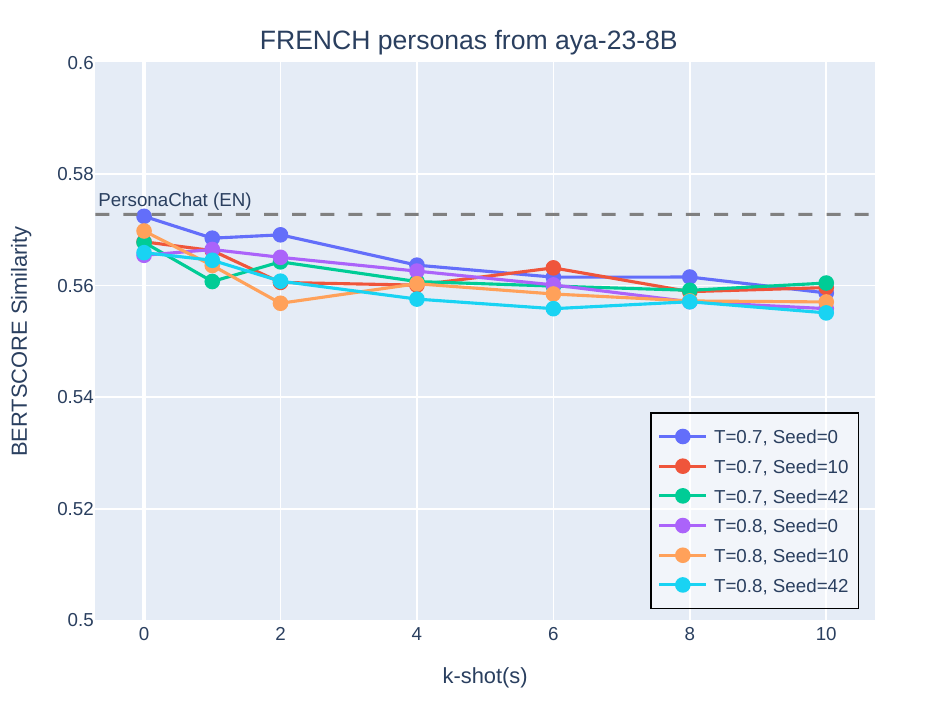}
        \end{subfigure}
        \vskip\baselineskip
        \begin{subfigure}{\textwidth}
            \centering
            \includegraphics[height=0.17\textheight]{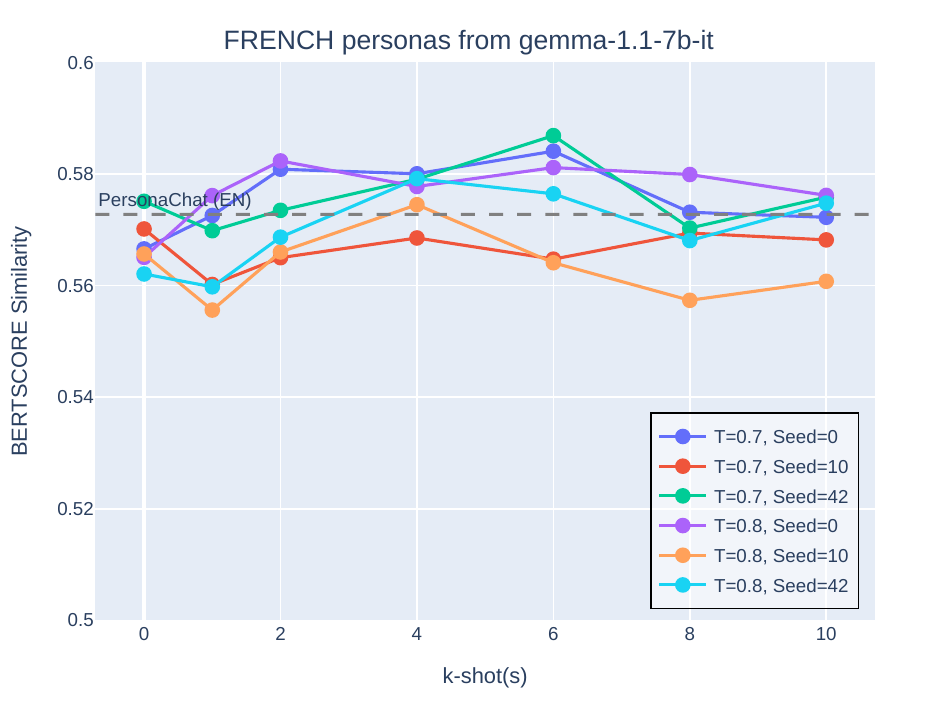}
        \end{subfigure}
        \vskip\baselineskip
        \begin{subfigure}{\textwidth}
            \centering
            \includegraphics[height=0.17\textheight]{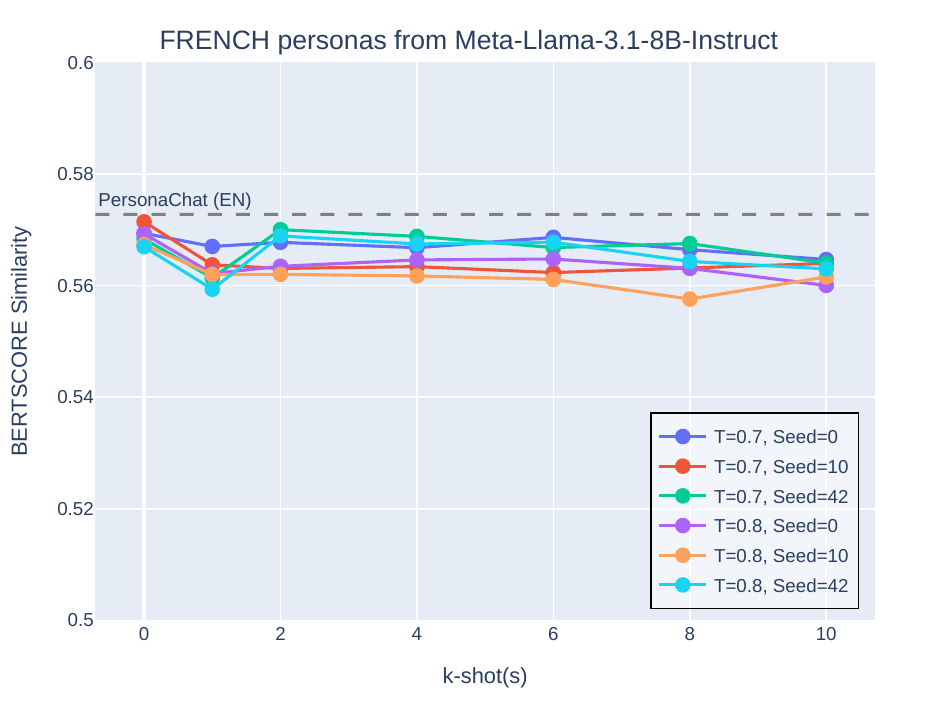}
        \end{subfigure}
        \vskip\baselineskip
        \begin{subfigure}{\textwidth}
            \centering
            \includegraphics[height=0.17\textheight]{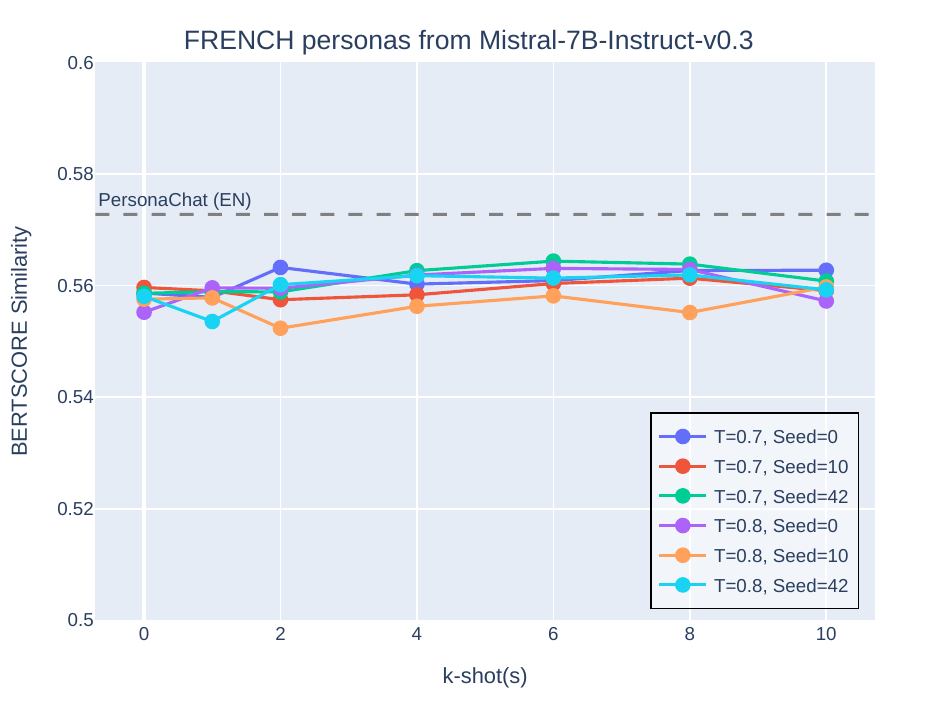}
        \end{subfigure}
    \end{minipage}
    %
    %
    \begin{minipage}{0.45\textwidth}
        \begin{subfigure}{\textwidth}
            \centering
            \includegraphics[height=0.17\textheight]{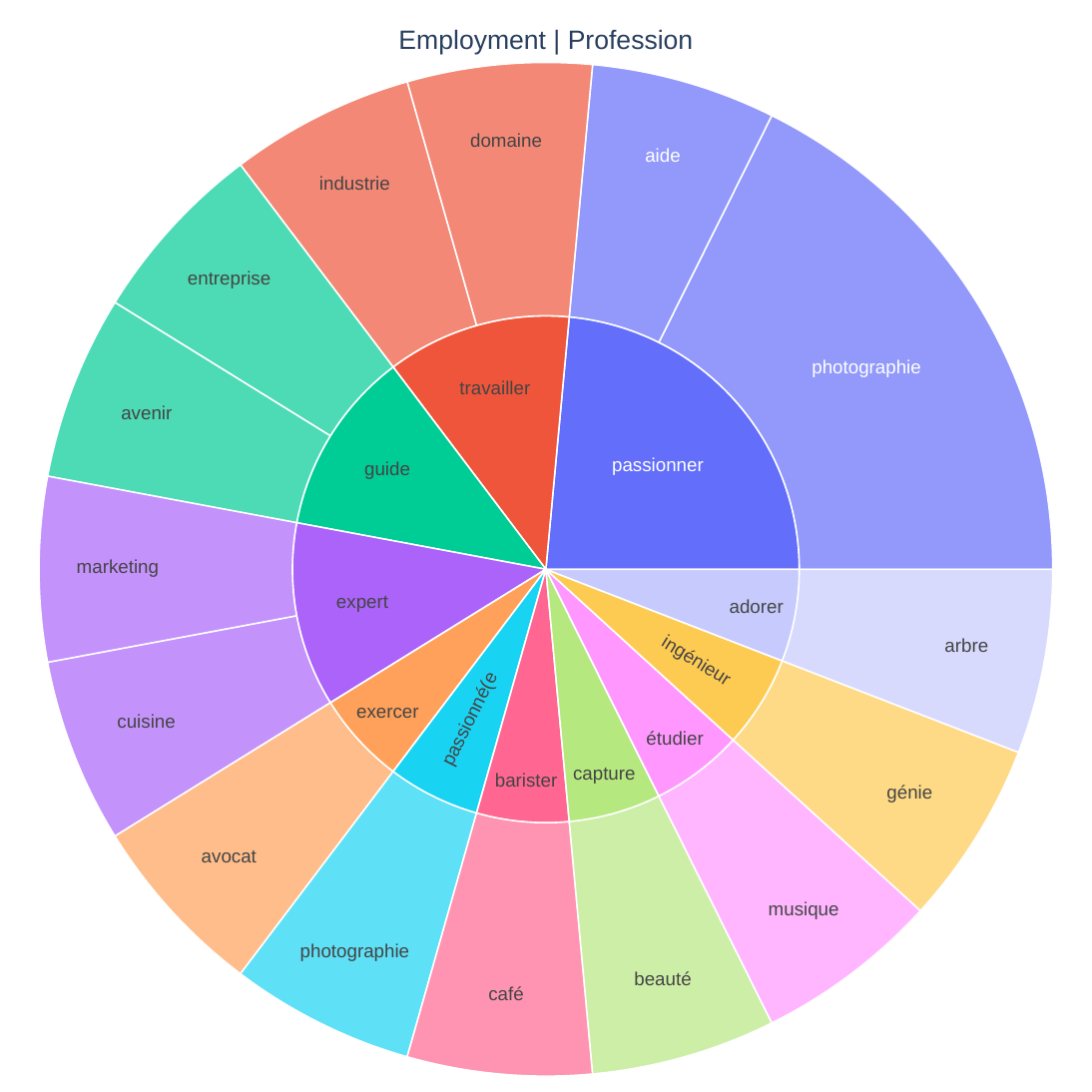}
        \end{subfigure}
        \vskip\baselineskip
        \begin{subfigure}{\textwidth}
            \centering
            \includegraphics[height=0.17\textheight]{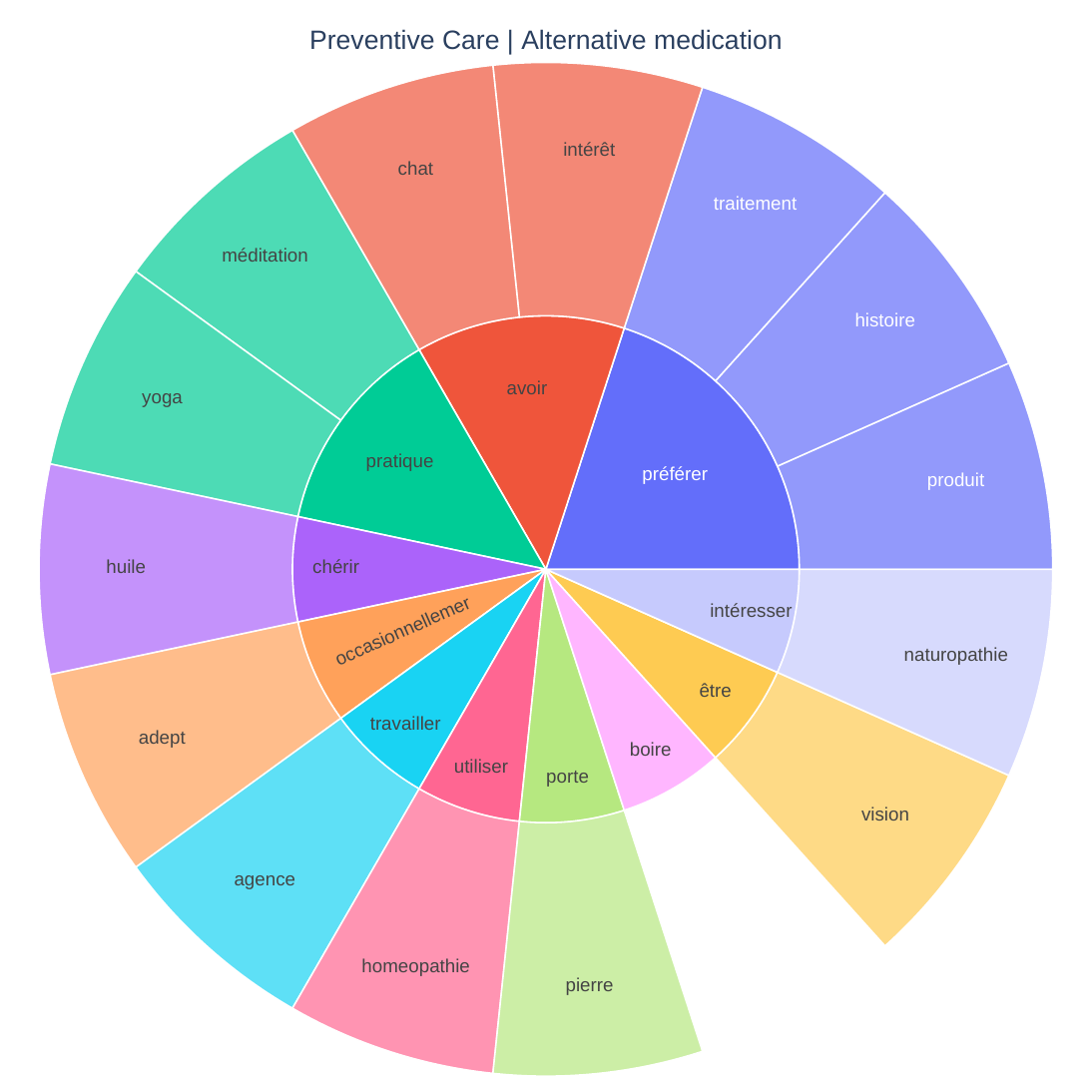}
        \end{subfigure}
        \vskip\baselineskip
        \begin{subfigure}{\textwidth}
            \centering
            \includegraphics[height=0.17\textheight]{pictures/plots_syntax/french/syntax_french_LLaMA3-8B_top1_entity.pdf}
        \end{subfigure}
        \vskip\baselineskip
        \begin{subfigure}{\textwidth}
            \centering
            \includegraphics[height=0.17\textheight]{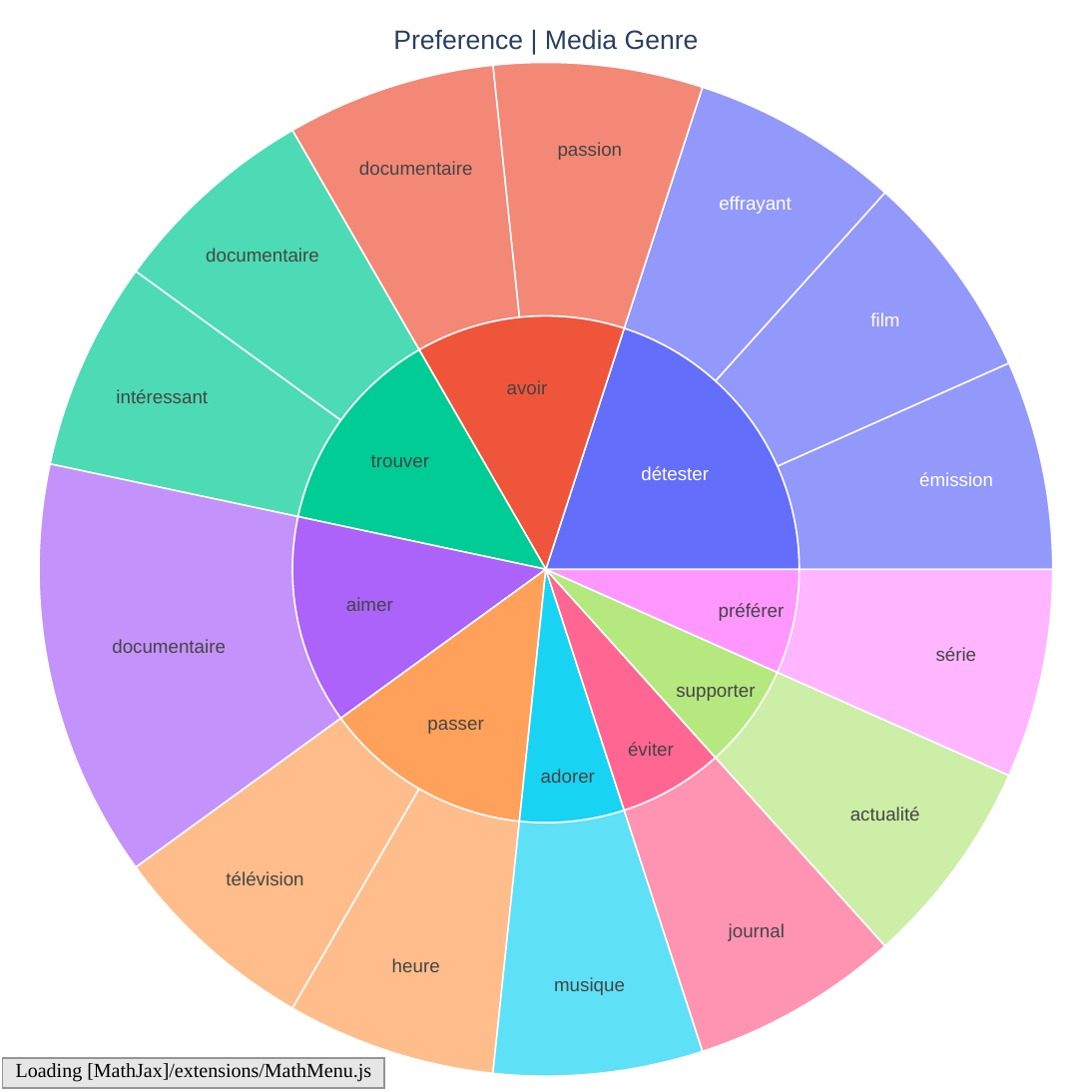}
        \end{subfigure}
    \end{minipage}
    
\caption{Detailed BERTSCORE for French Personas in different generation configurations and Sunburst charts of personas taxonomy entities with most root verbs and associated object noun for the different models}    
\end{figure}


\restoregeometry

\newgeometry{top=0.5cm, bottom=1.5cm, left=2.5cm, right=2.5cm}
\subsubsection{\textsc{Italian}}

\begin{figure}[h]
    \centering
    \begin{minipage}{0.45\textwidth}
        \begin{subfigure}{\textwidth}
            \centering
            \includegraphics[height=0.17\textheight]{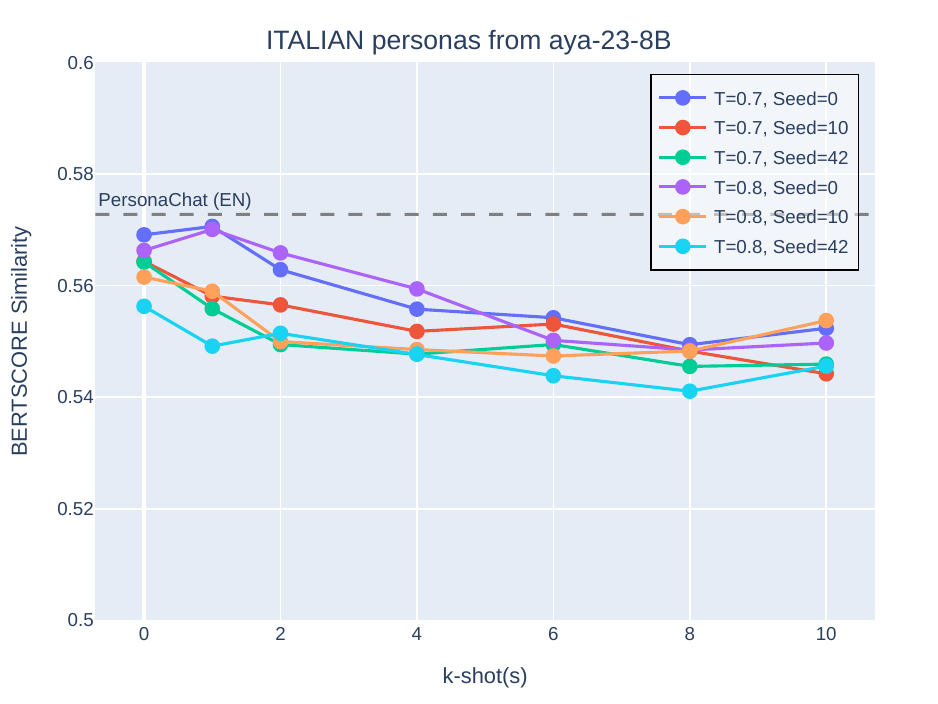}
        \end{subfigure}
        \vskip\baselineskip
        \begin{subfigure}{\textwidth}
            \centering
            \includegraphics[height=0.17\textheight]{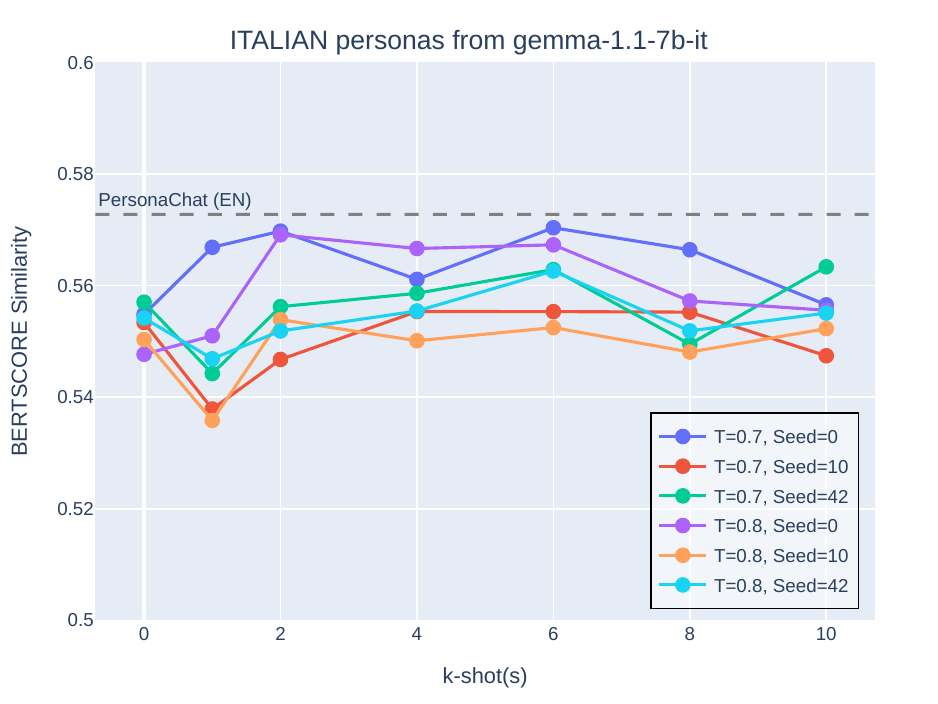}
        \end{subfigure}
        \vskip\baselineskip
        \begin{subfigure}{\textwidth}
            \centering
            \includegraphics[height=0.17\textheight]{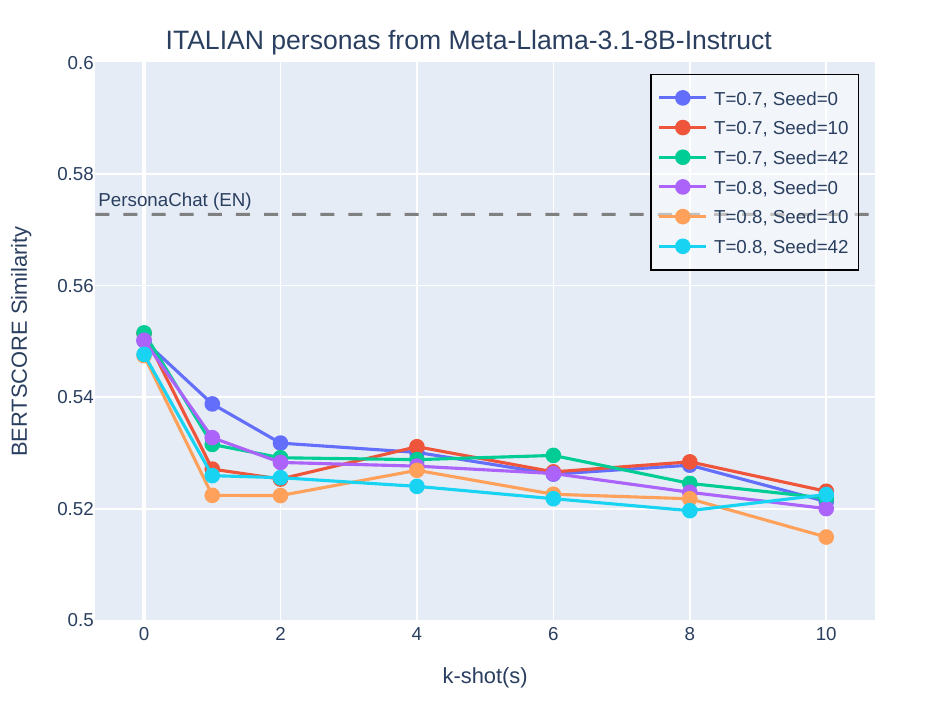}
        \end{subfigure}
        \vskip\baselineskip
        \begin{subfigure}{\textwidth}
            \centering
            \includegraphics[height=0.17\textheight]{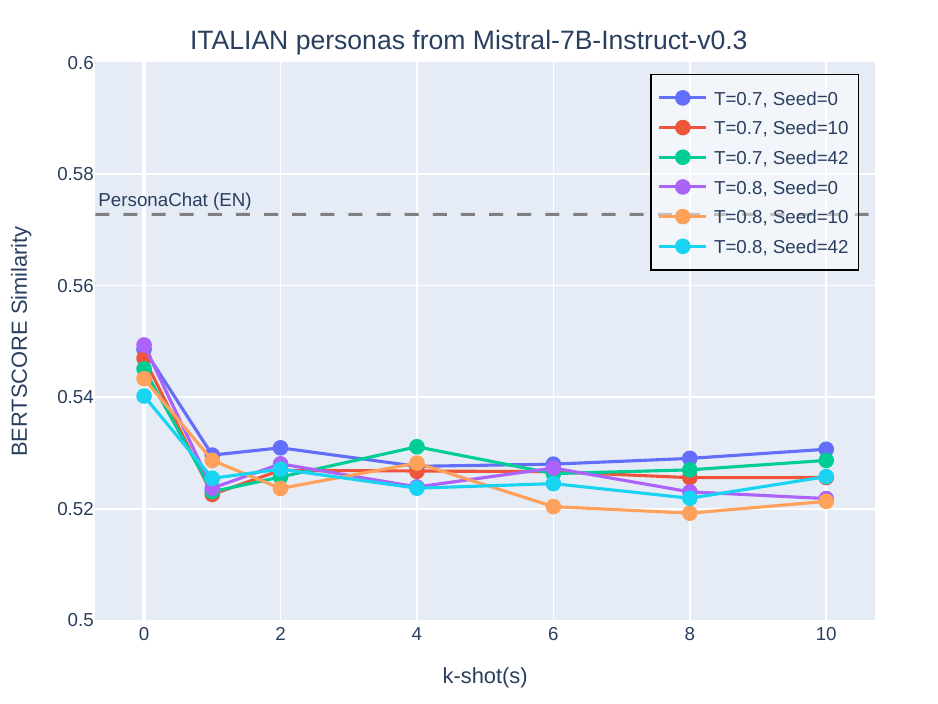}
        \end{subfigure}
    \end{minipage}
    %
    %
    \begin{minipage}{0.45\textwidth}
        \begin{subfigure}{\textwidth}
            \centering
            \includegraphics[height=0.17\textheight]{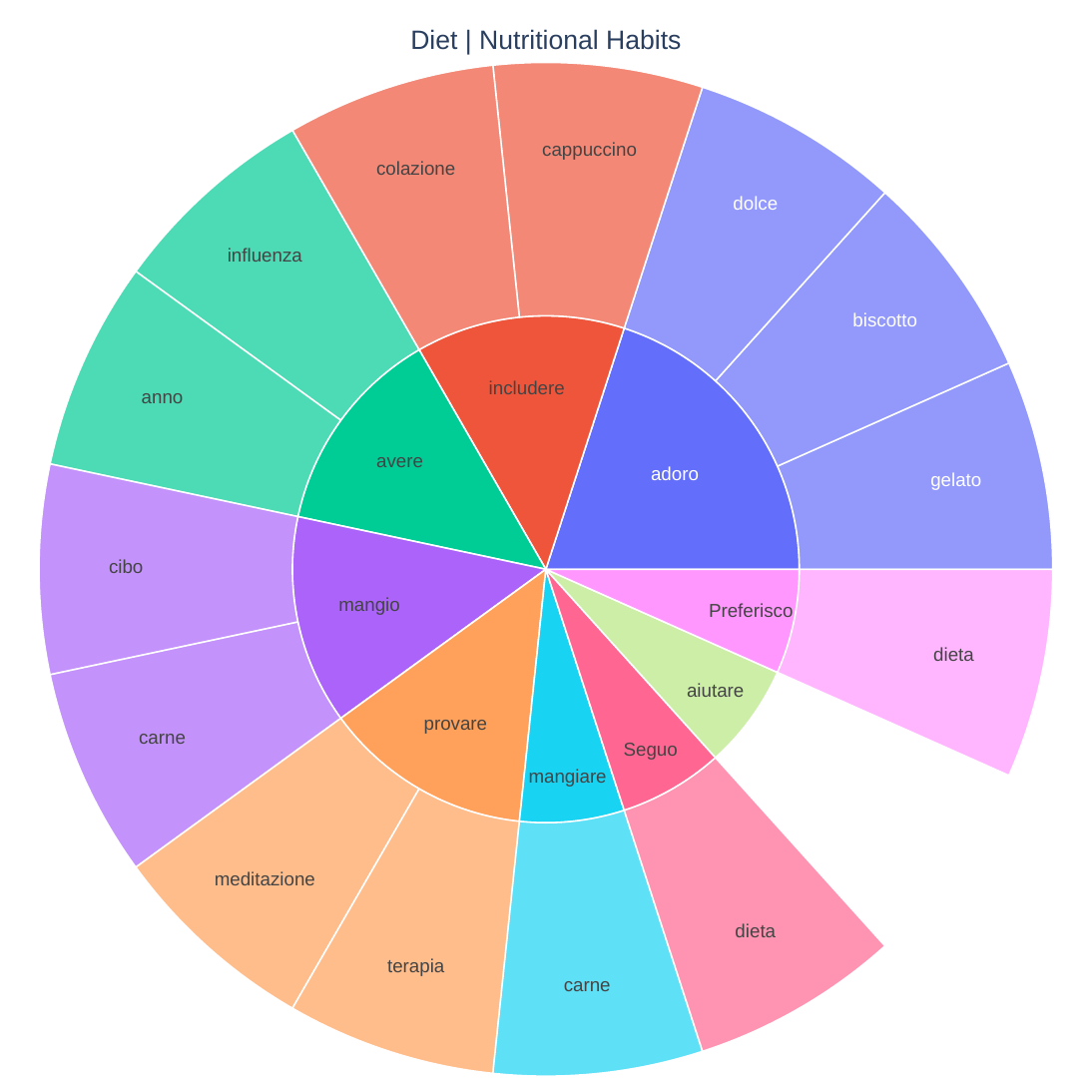}
        \end{subfigure}
        \vskip\baselineskip
        \begin{subfigure}{\textwidth}
            \centering
            \includegraphics[height=0.17\textheight]{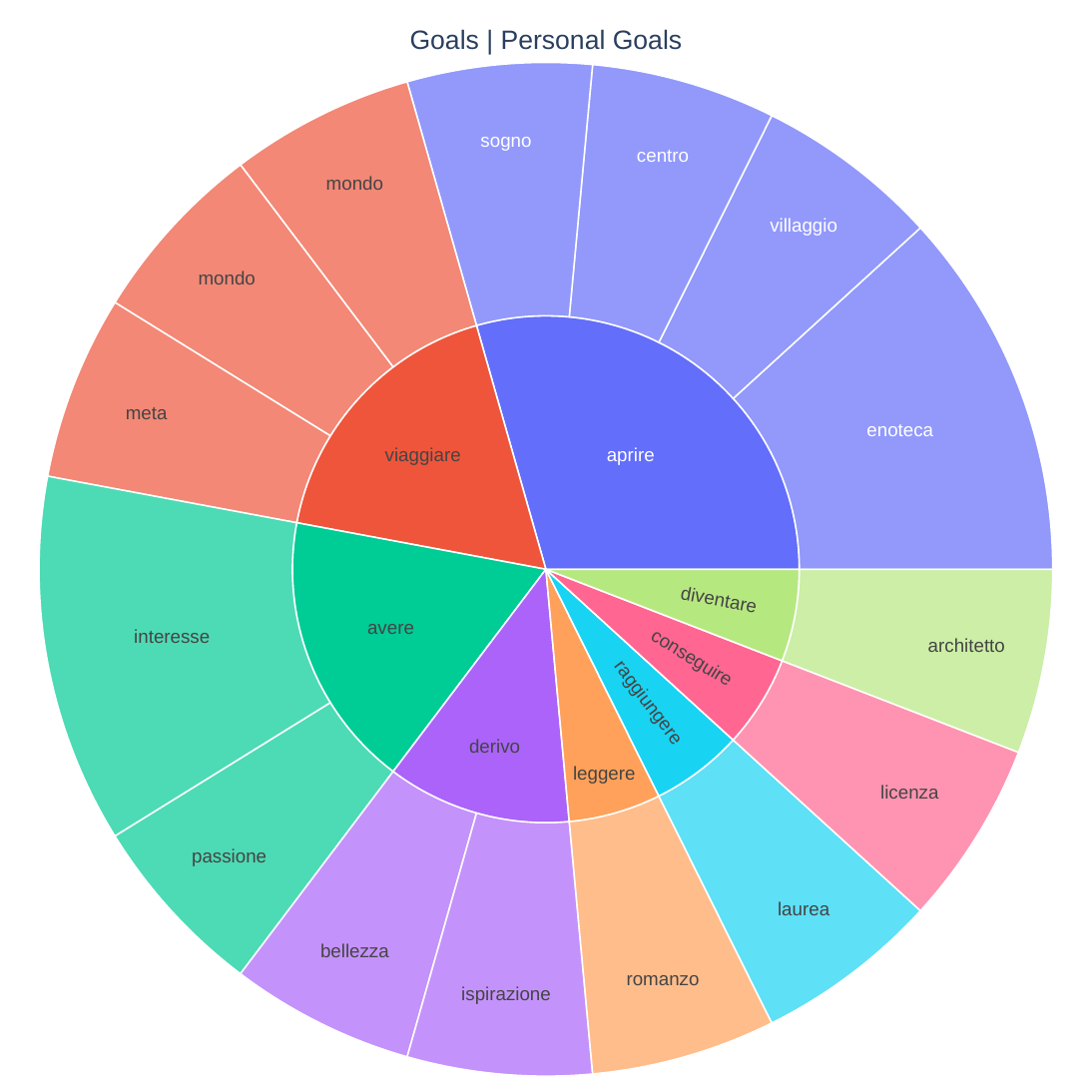}
        \end{subfigure}
        \vskip\baselineskip
        \begin{subfigure}{\textwidth}
            \centering
            \includegraphics[height=0.17\textheight]{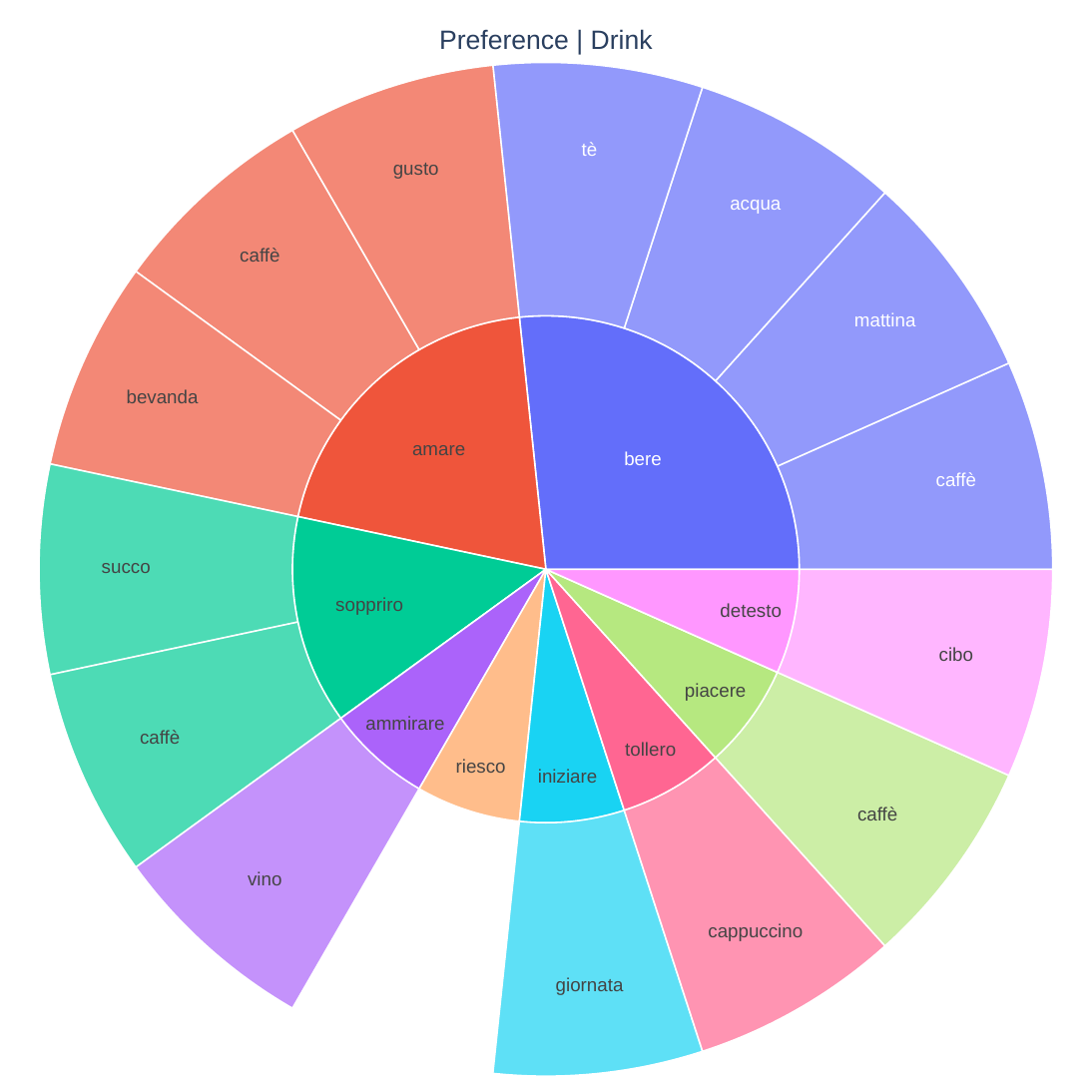}
        \end{subfigure}
        \vskip\baselineskip
        \begin{subfigure}{\textwidth}
            \centering
            \includegraphics[height=0.17\textheight]{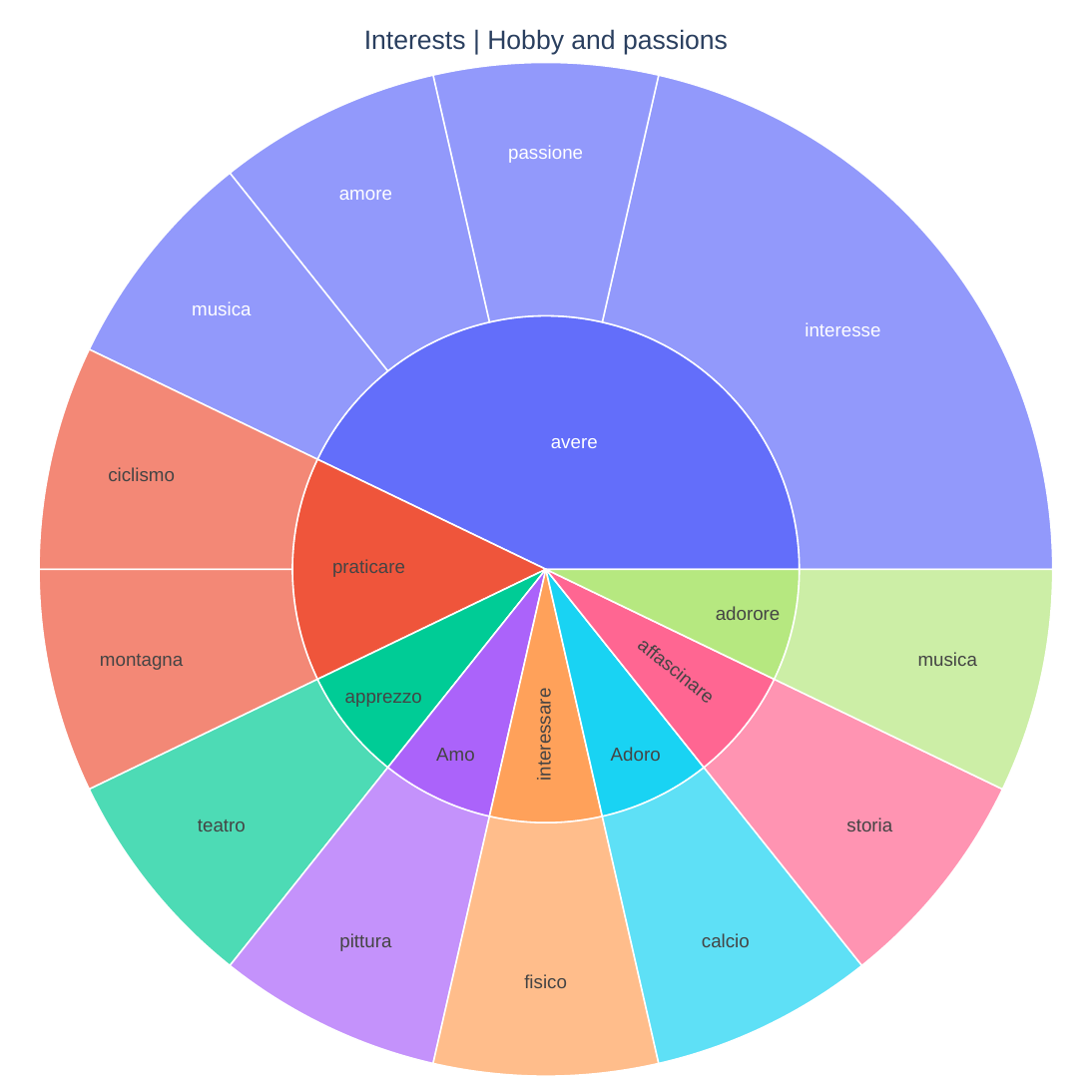}
        \end{subfigure}
    \end{minipage}
    
\caption{Detailed BERTSCORE for Italian Personas in different generation configurations and Sunburst charts of personas taxonomy entities with most root verbs and associated object noun for the different models}    
\end{figure}


\restoregeometry 

\newgeometry{top=0.5cm, bottom=1.5cm, left=2.5cm, right=2.5cm}
\subsubsection{\textsc{Dutch}}

\begin{figure}[h]
    \centering
    \begin{minipage}{0.45\textwidth}
        \begin{subfigure}{\textwidth}
            \centering
            \includegraphics[height=0.17\textheight]{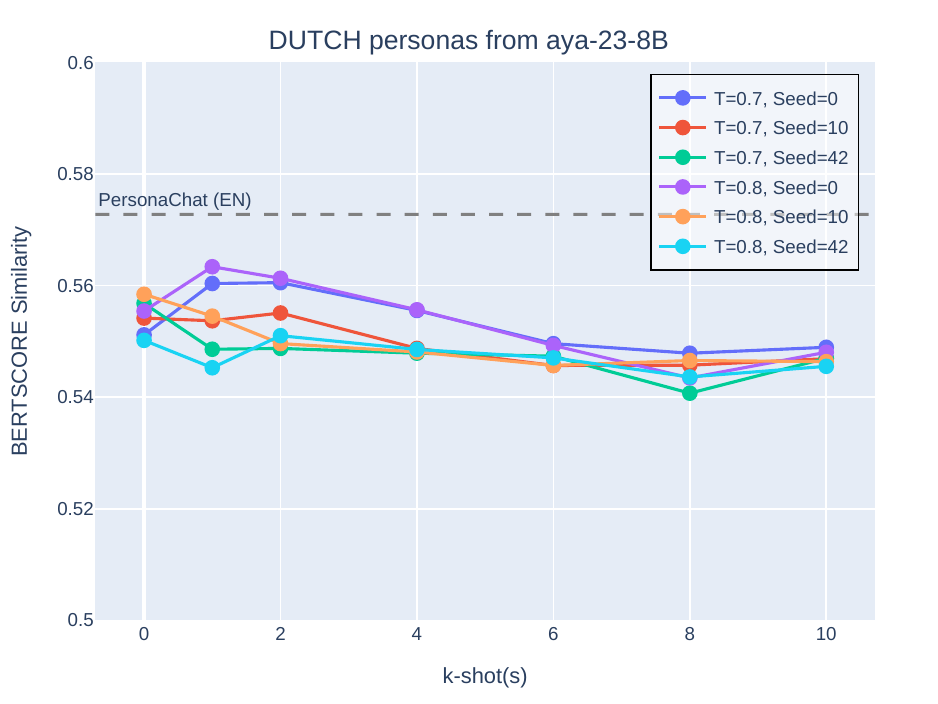}
        \end{subfigure}
        \vskip\baselineskip
        \begin{subfigure}{\textwidth}
            \centering
            \includegraphics[height=0.17\textheight]{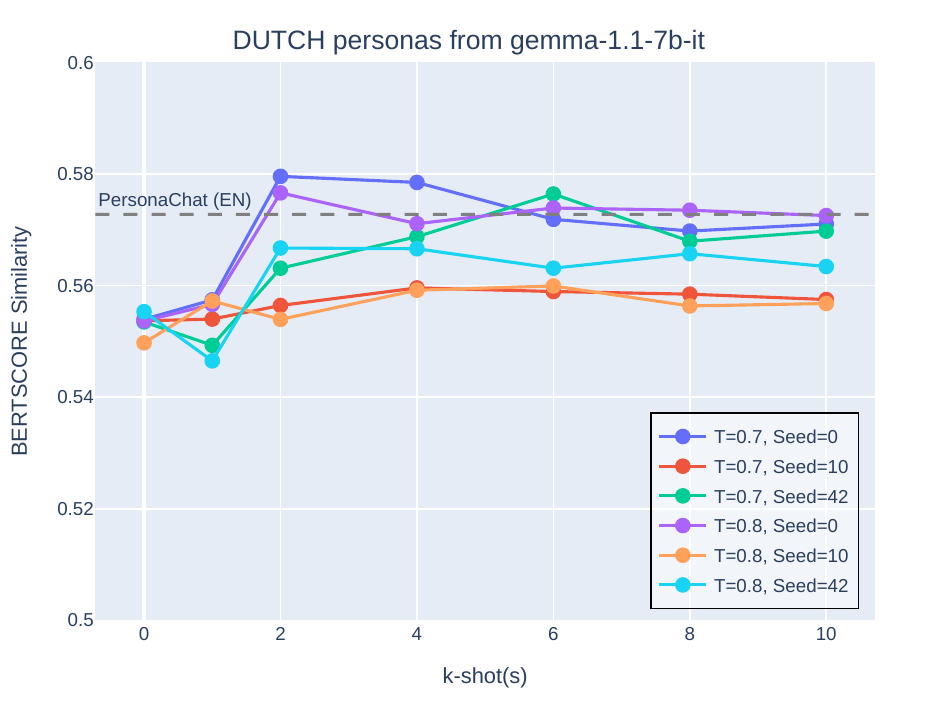}
        \end{subfigure}
        \vskip\baselineskip
        \begin{subfigure}{\textwidth}
            \centering
            \includegraphics[height=0.17\textheight]{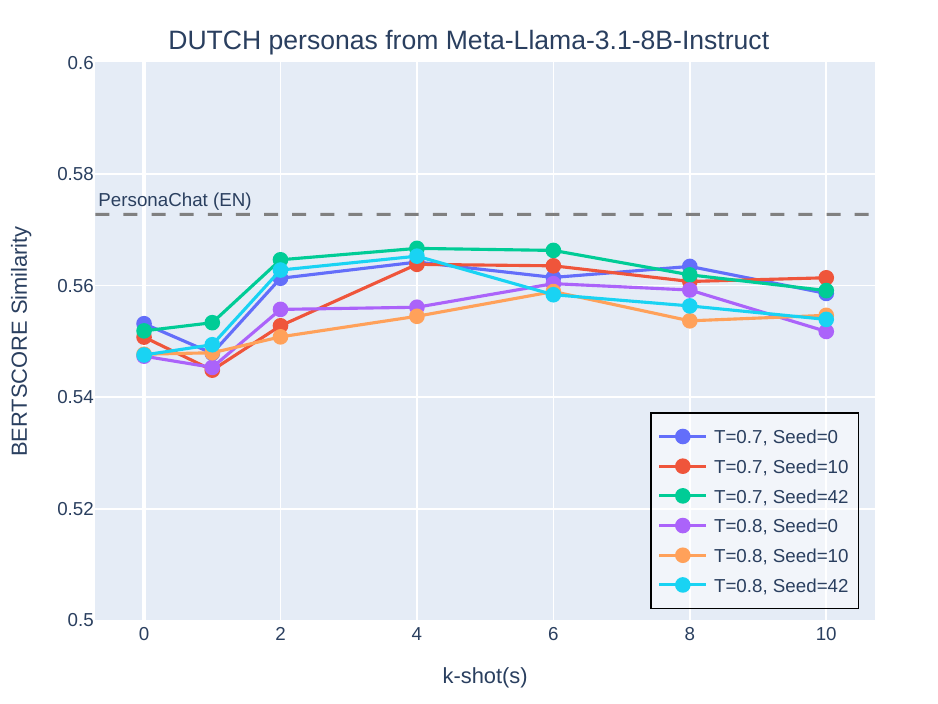}
        \end{subfigure}
        \vskip\baselineskip
        \begin{subfigure}{\textwidth}
            \centering
            \includegraphics[height=0.17\textheight]{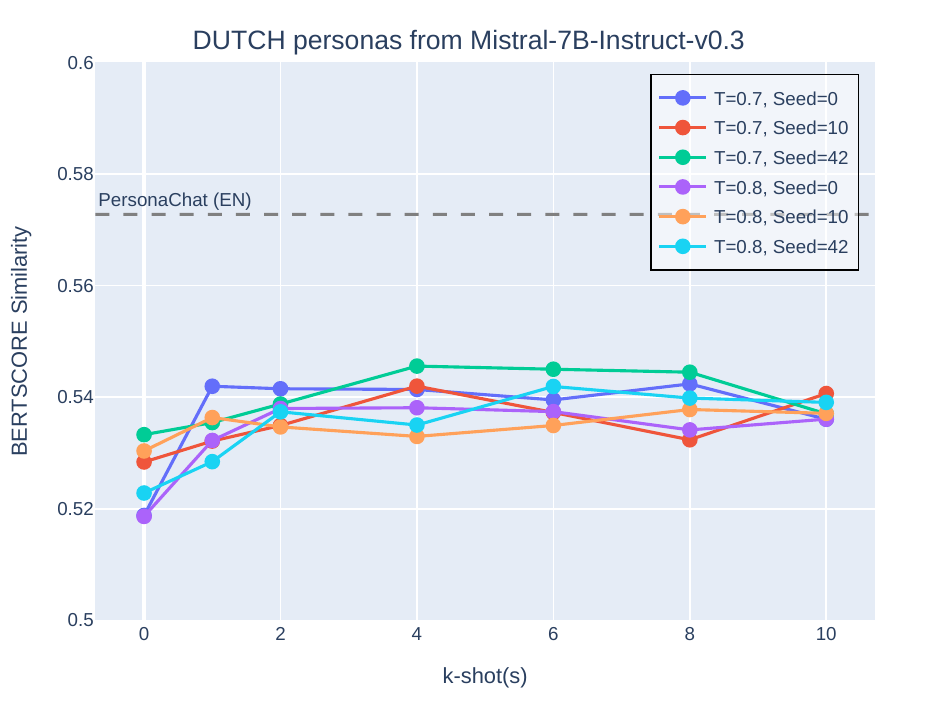}
        \end{subfigure}
    \end{minipage}
    %
    %
    \begin{minipage}{0.45\textwidth}
        \begin{subfigure}{\textwidth}
            \centering
            \includegraphics[height=0.17\textheight]{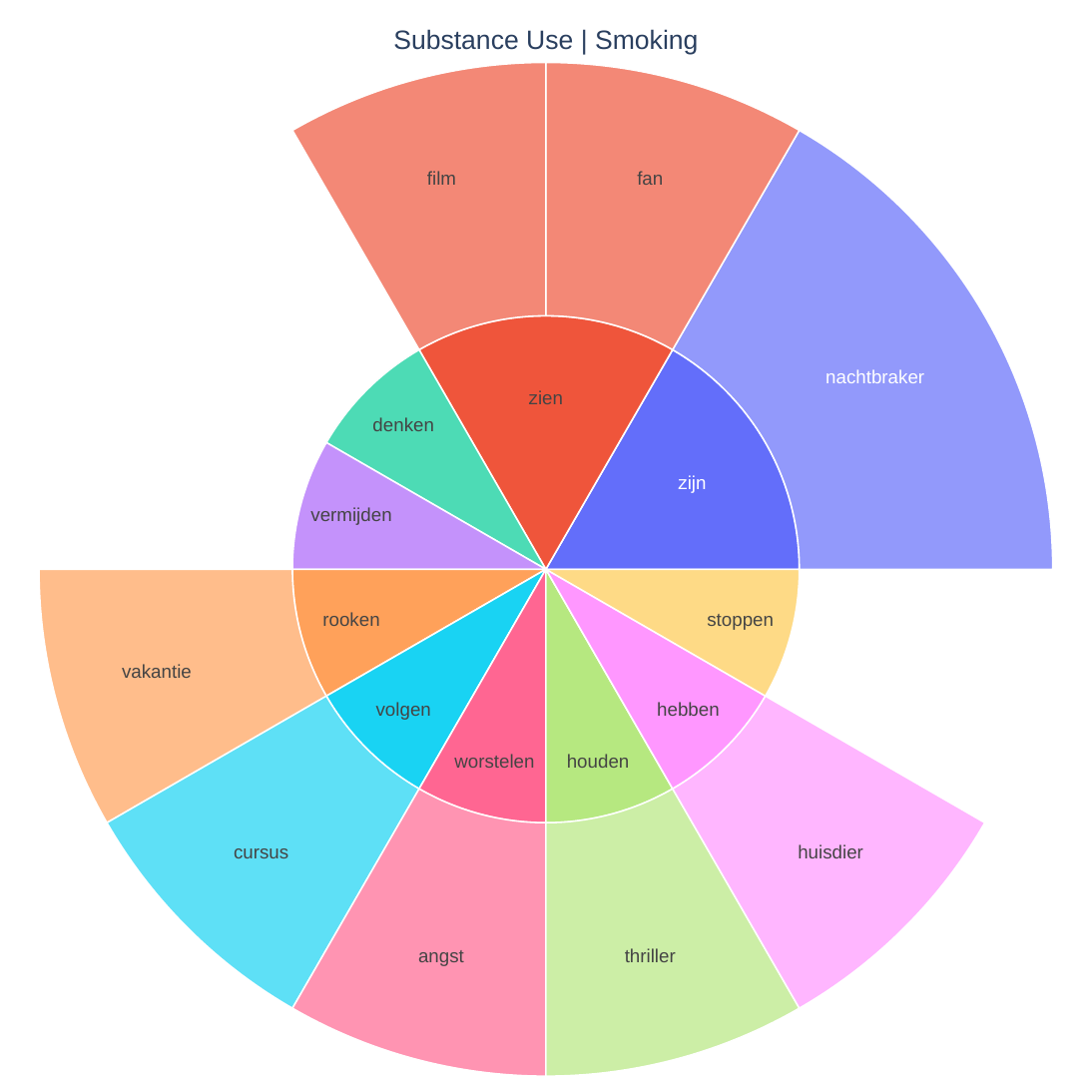}
        \end{subfigure}
        \vskip\baselineskip
        \begin{subfigure}{\textwidth}
            \centering
            \includegraphics[height=0.17\textheight]{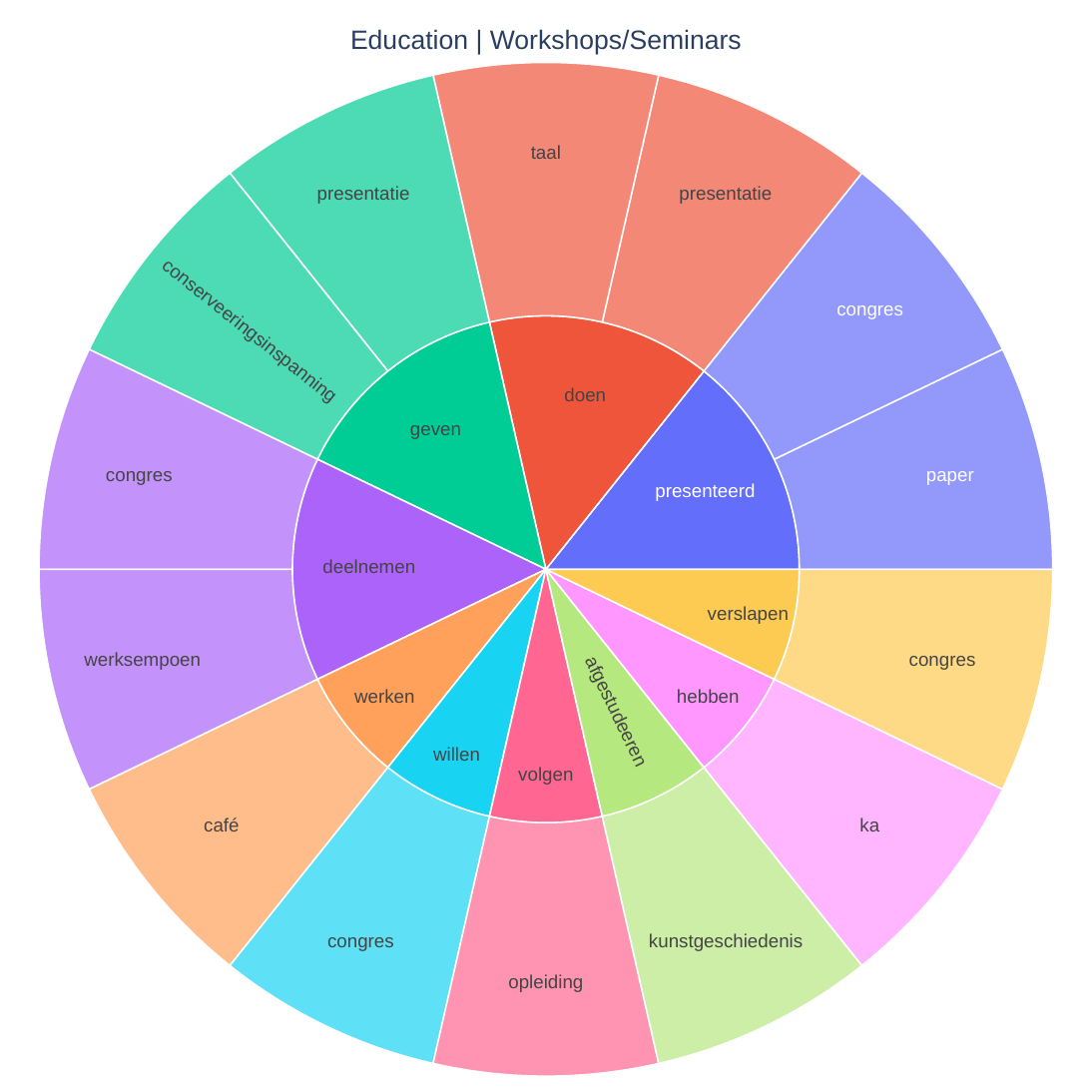}
        \end{subfigure}
        \vskip\baselineskip
        \begin{subfigure}{\textwidth}
            \centering
            \includegraphics[height=0.17\textheight]{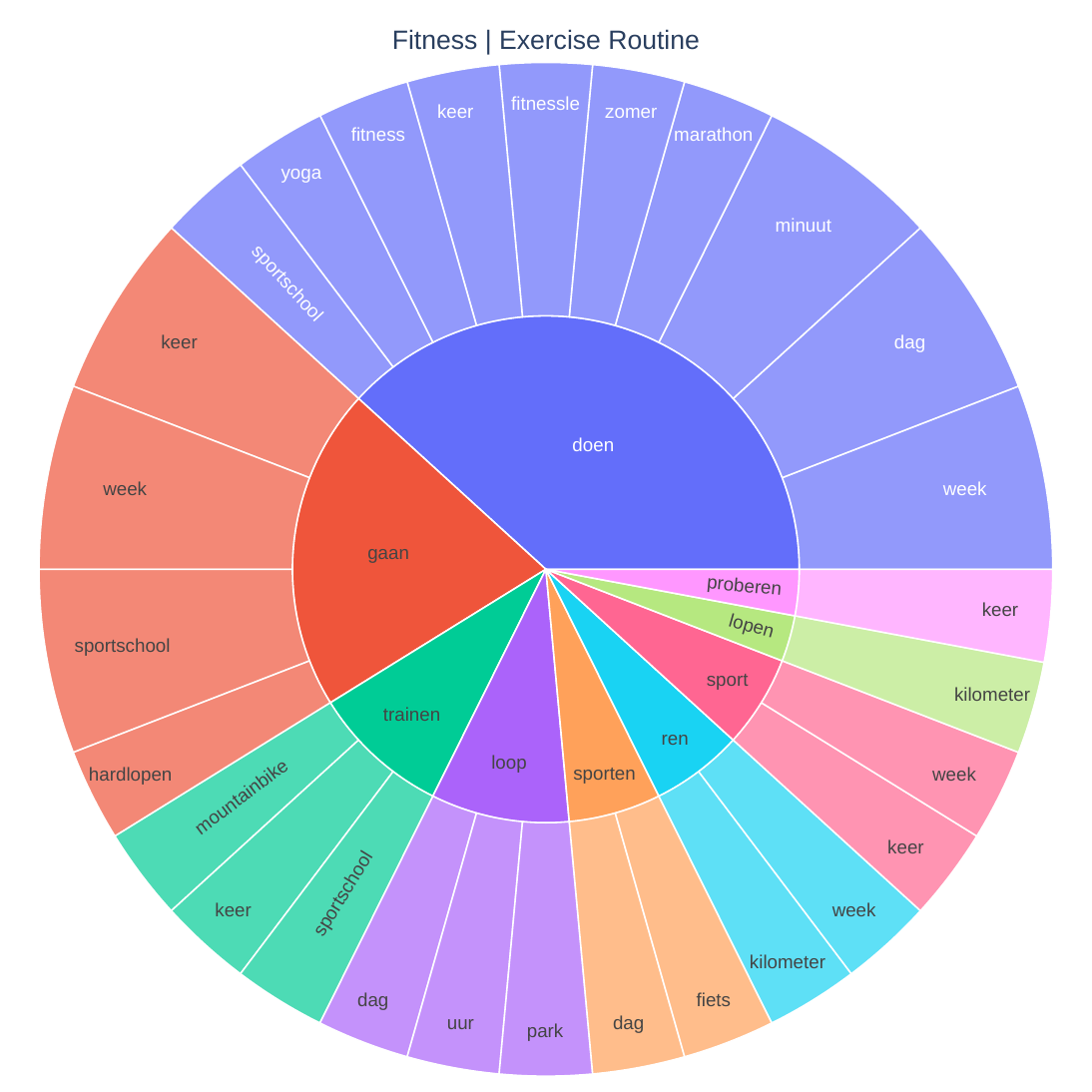}
        \end{subfigure}
        \vskip\baselineskip
        \begin{subfigure}{\textwidth}
            \centering
            \includegraphics[height=0.17\textheight]{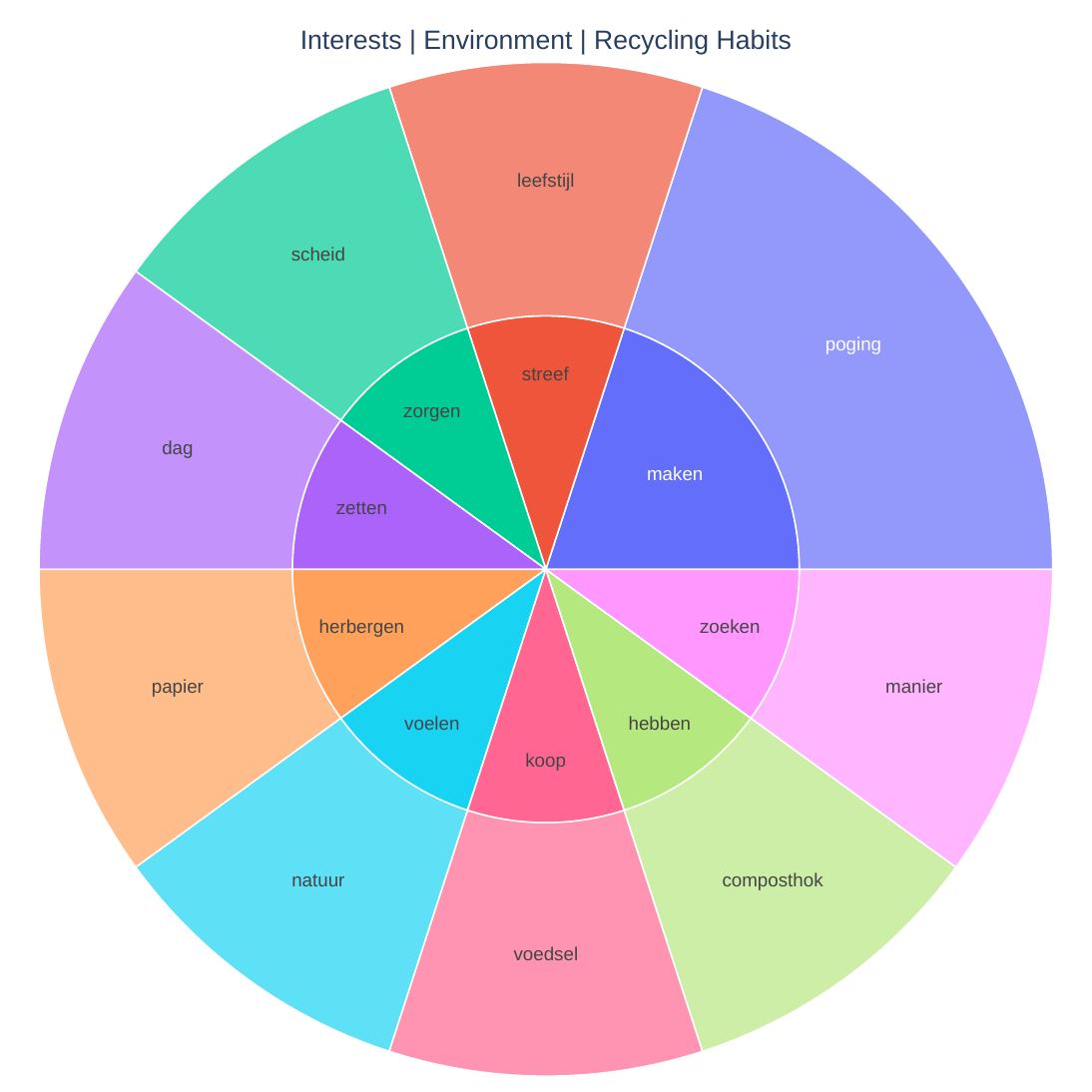}
        \end{subfigure}
    \end{minipage}
    
\caption{Detailed BERTSCORE for Dutch Personas in different generation configurations and Sunburst charts of personas taxonomy entities with most root verbs and associated object noun for the different models}    
\end{figure}


\restoregeometry

\newgeometry{top=0.5cm, bottom=1.5cm, left=2.5cm, right=2.5cm}
\subsubsection{\textsc{Portuguese}}

\begin{figure}[h]
    \centering
    \begin{minipage}{0.45\textwidth}
        \begin{subfigure}{\textwidth}
            \centering
            \includegraphics[height=0.17\textheight]{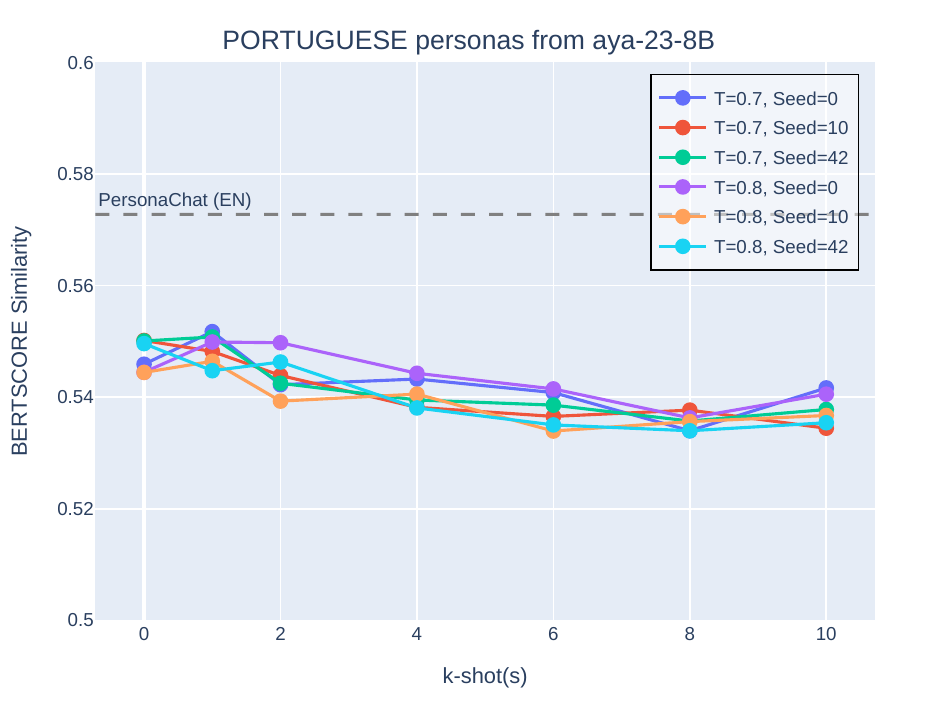}
        \end{subfigure}
        \vskip\baselineskip
        \begin{subfigure}{\textwidth}
            \centering
            \includegraphics[height=0.17\textheight]{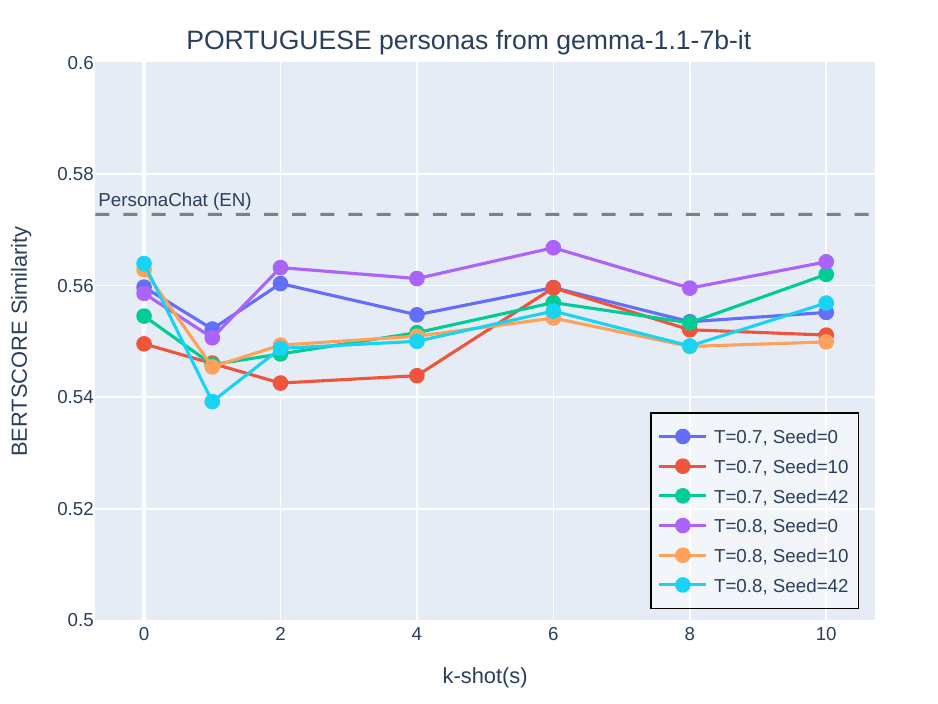}
        \end{subfigure}
        \vskip\baselineskip
        \begin{subfigure}{\textwidth}
            \centering
            \includegraphics[height=0.17\textheight]{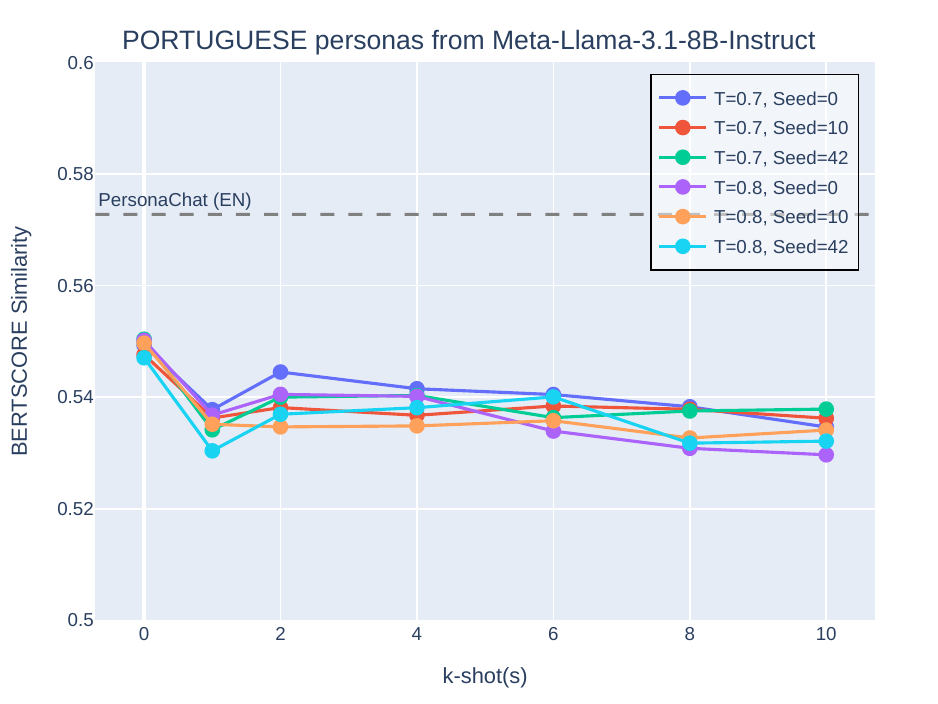}
        \end{subfigure}
        \vskip\baselineskip
        \begin{subfigure}{\textwidth}
            \centering
            \includegraphics[height=0.17\textheight]{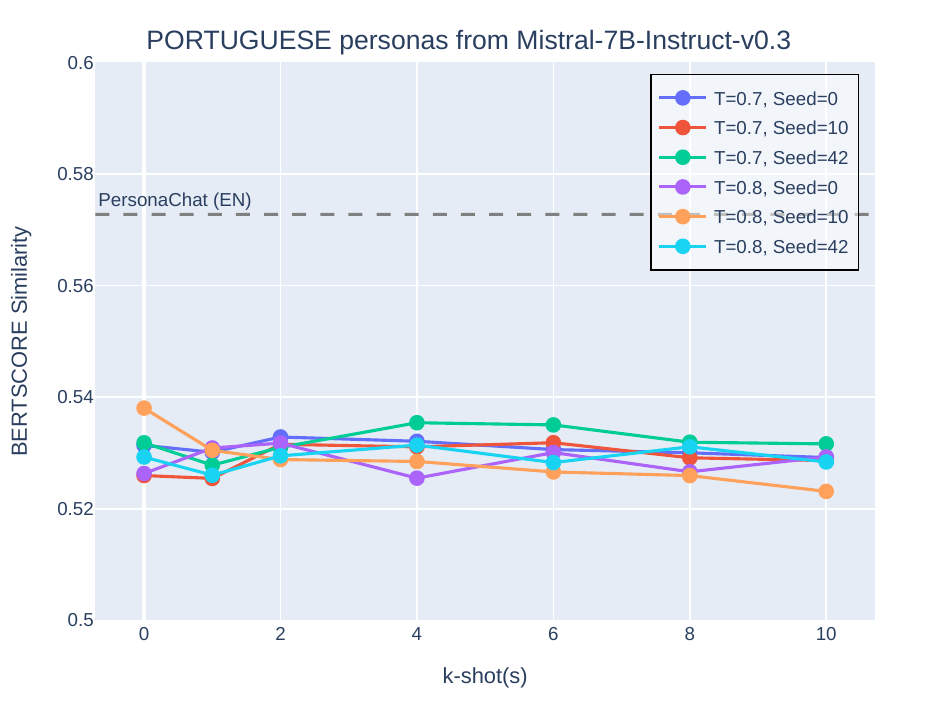}
        \end{subfigure}
    \end{minipage}
    %
    %
    \begin{minipage}{0.45\textwidth}
        \begin{subfigure}{\textwidth}
            \centering
            \includegraphics[height=0.17\textheight]{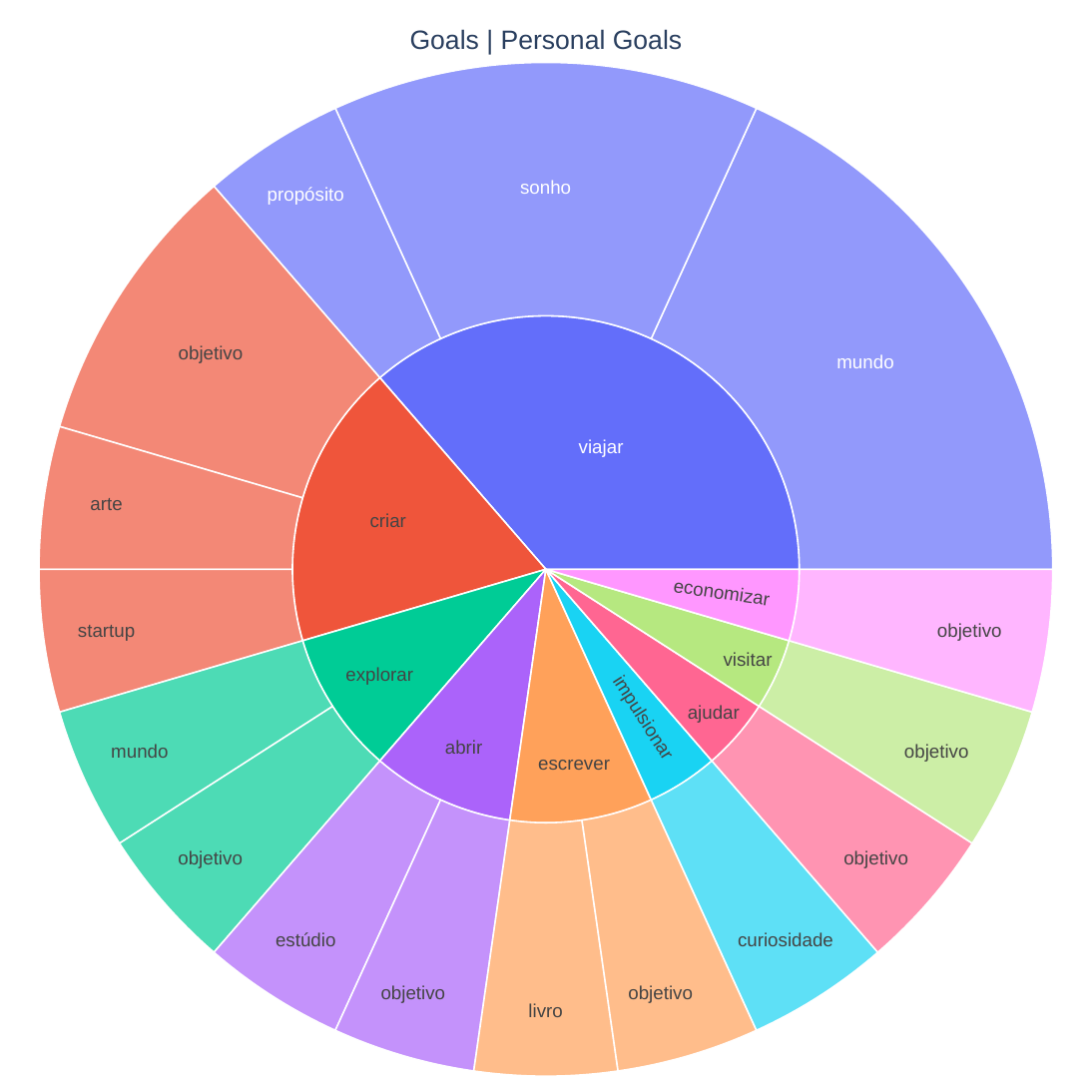}
        \end{subfigure}
        \vskip\baselineskip
        \begin{subfigure}{\textwidth}
            \centering
            \includegraphics[height=0.17\textheight]{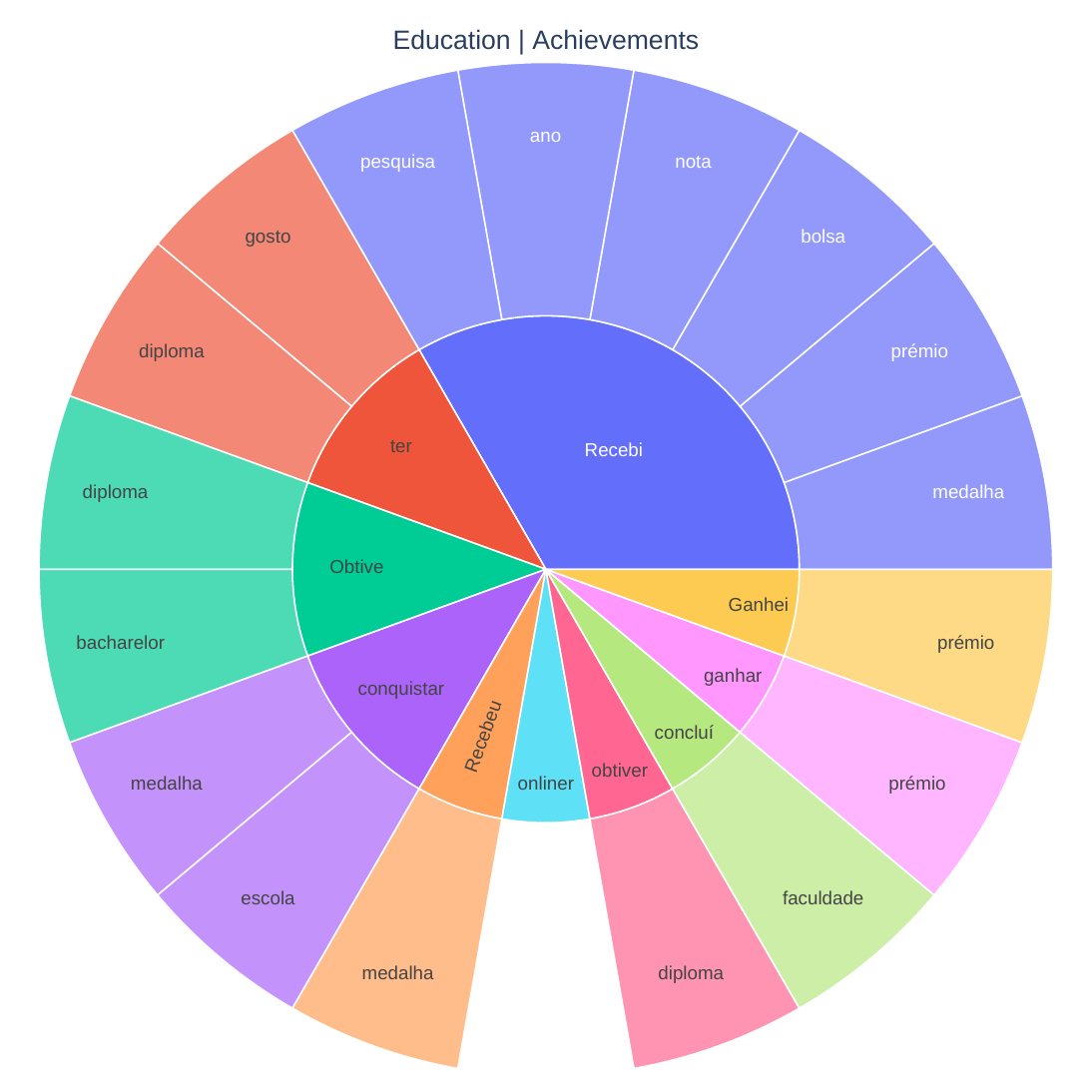}
        \end{subfigure}
        \vskip\baselineskip
        \begin{subfigure}{\textwidth}
            \centering
            \includegraphics[height=0.17\textheight]{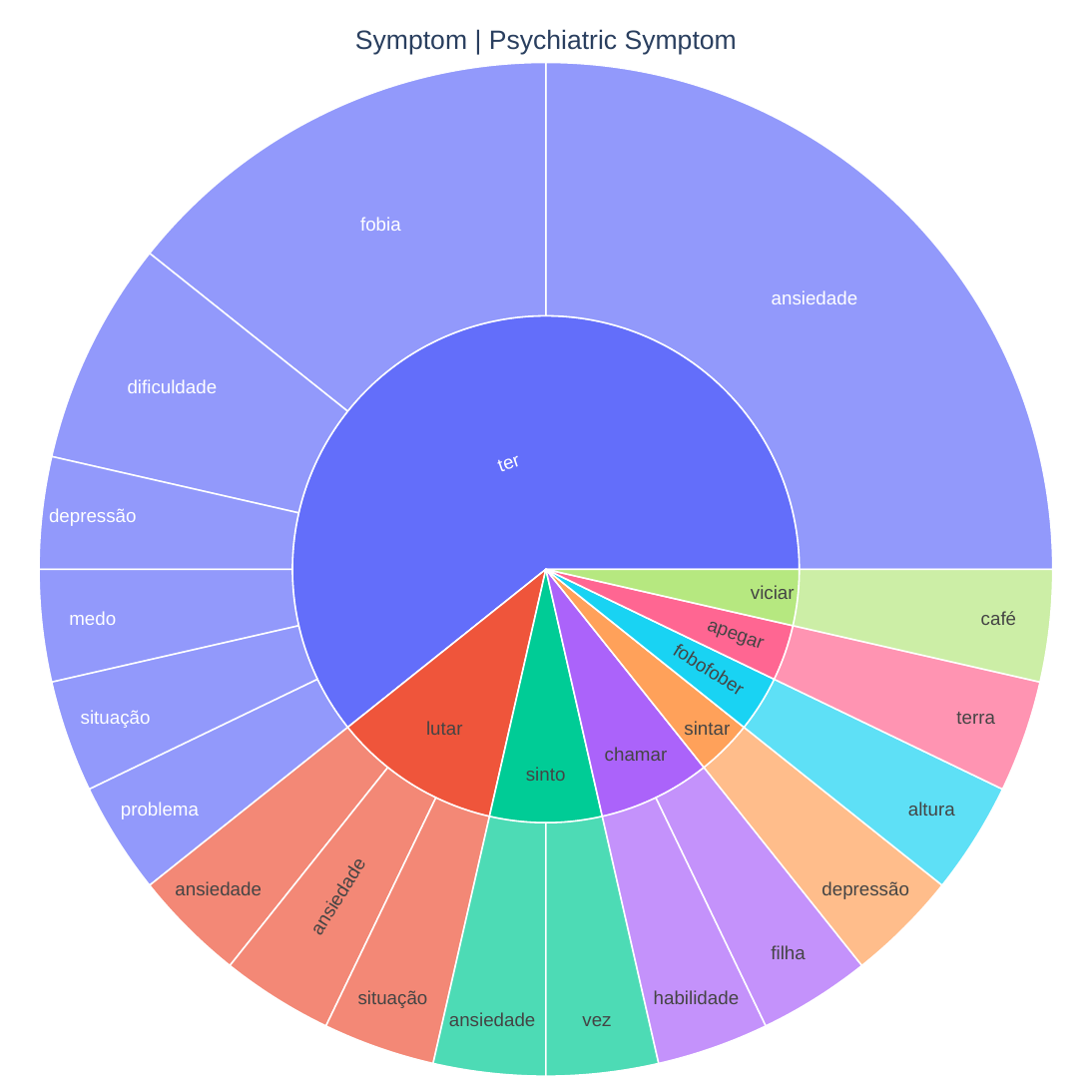}
        \end{subfigure}
        \vskip\baselineskip
        \begin{subfigure}{\textwidth}
            \centering
            \includegraphics[height=0.17\textheight]{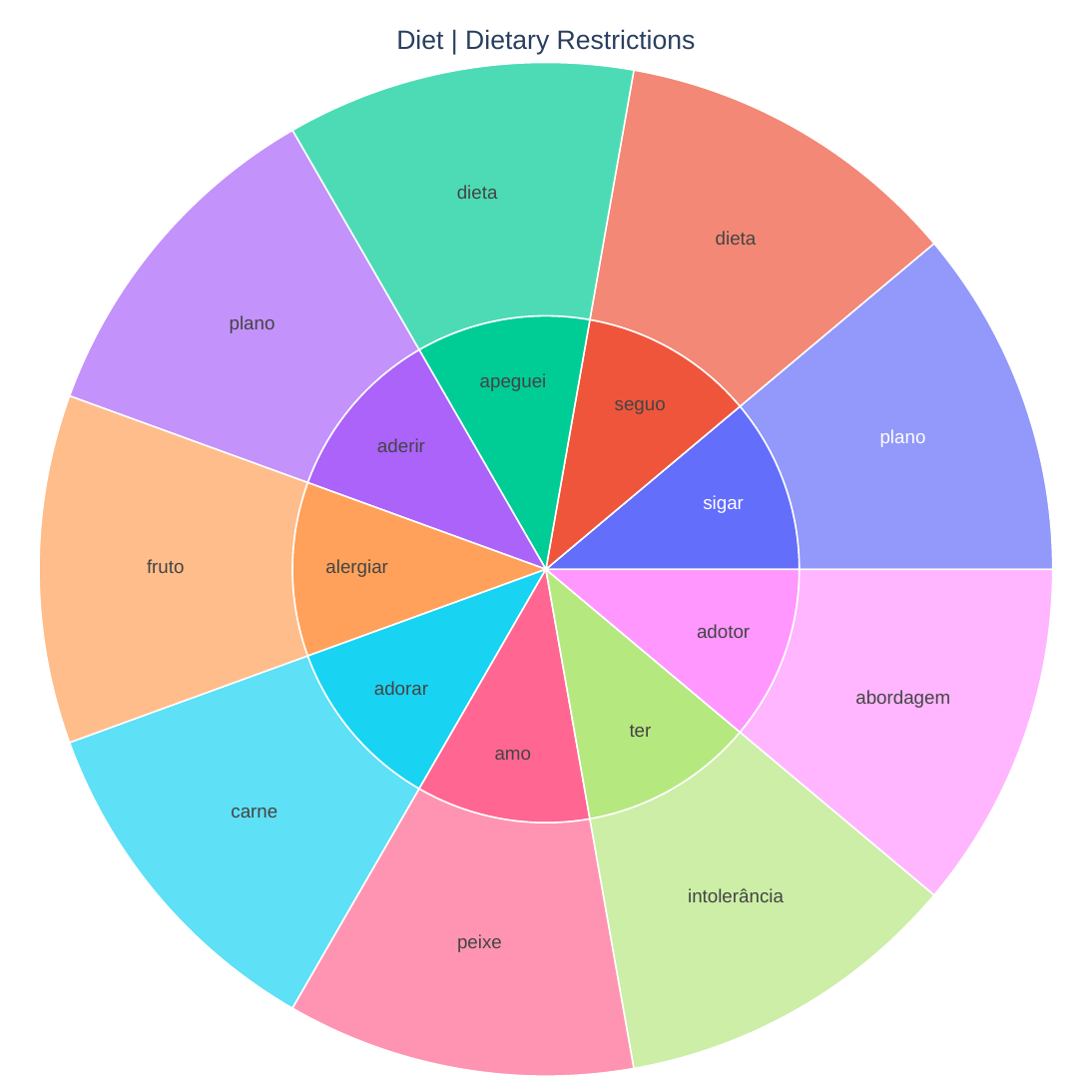}
        \end{subfigure}
    \end{minipage}
    
\caption{Detailed BERTSCORE for Portuguese Personas in different generation configurations and Sunburst charts of personas taxonomy entities with most root verbs and associated object noun for the different models}    
\end{figure}


\restoregeometry

\newgeometry{top=0.5cm, bottom=1.5cm, left=2.5cm, right=2.5cm}
\subsubsection{\textsc{Polish}}

\begin{figure}[h]
    \centering
    \begin{minipage}{0.45\textwidth}
        \begin{subfigure}{\textwidth}
            \centering
            \includegraphics[height=0.17\textheight]{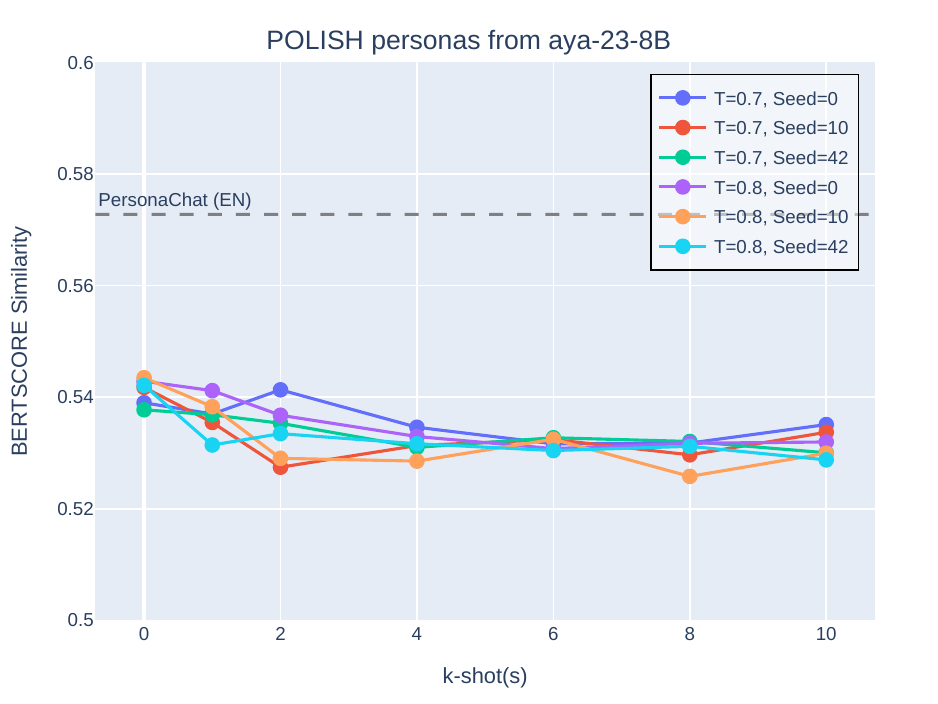}
        \end{subfigure}
        \vskip\baselineskip
        \begin{subfigure}{\textwidth}
            \centering
            \includegraphics[height=0.17\textheight]{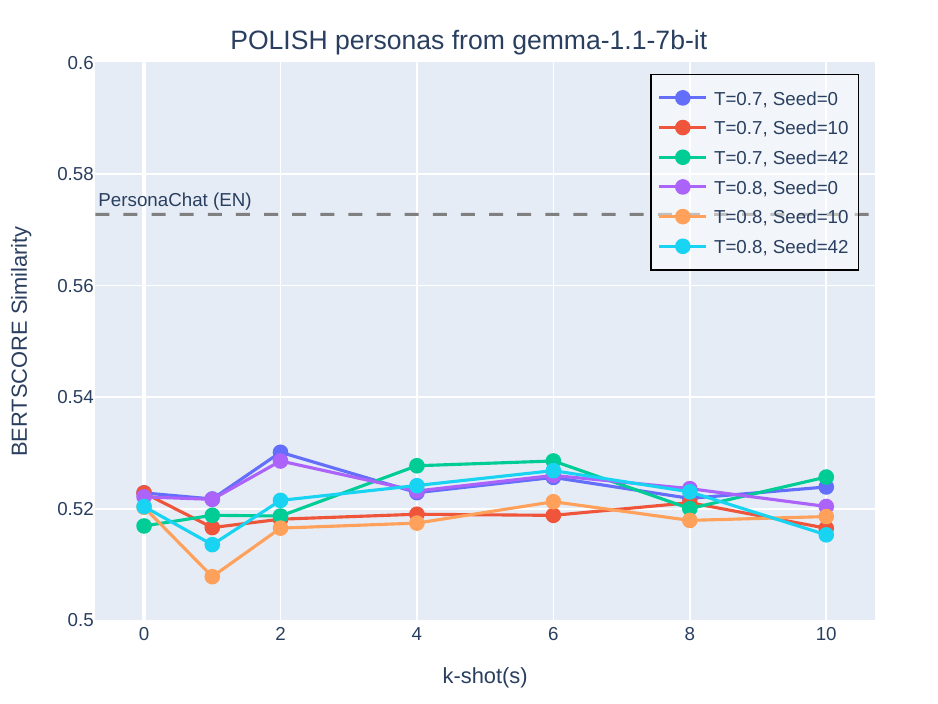}
        \end{subfigure}
        \vskip\baselineskip
        \begin{subfigure}{\textwidth}
            \centering
            \includegraphics[height=0.17\textheight]{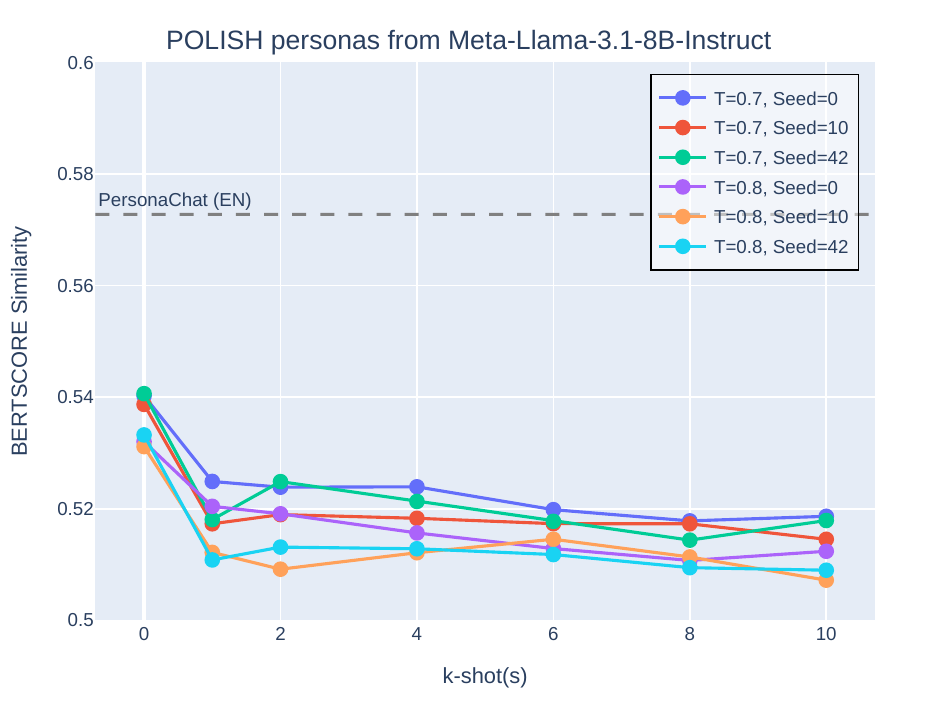}
        \end{subfigure}
        \vskip\baselineskip
        \begin{subfigure}{\textwidth}
            \centering
            \includegraphics[height=0.17\textheight]{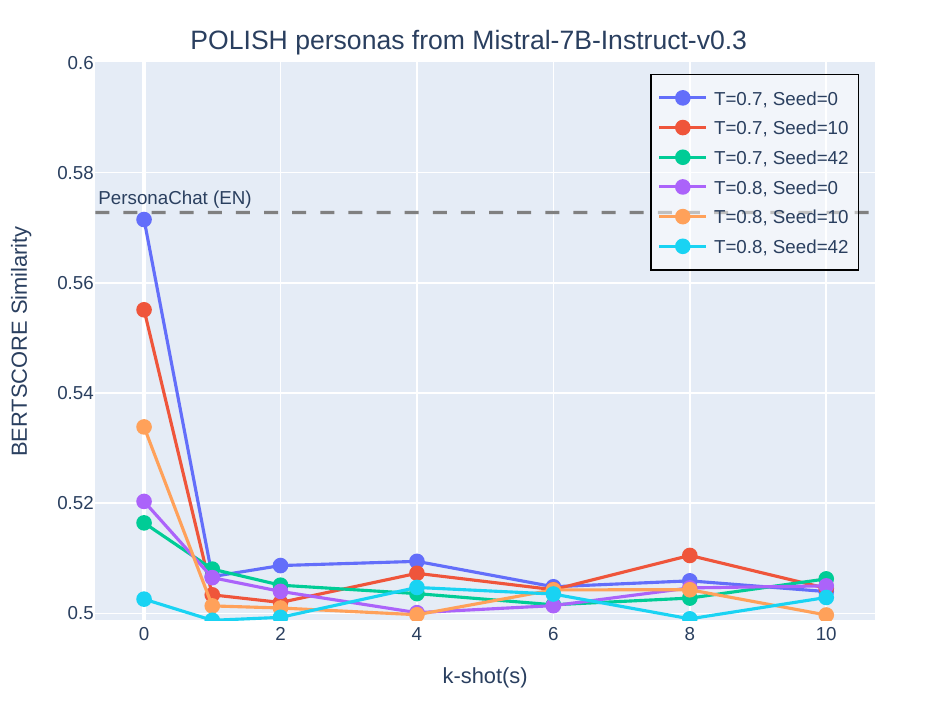}
        \end{subfigure}
    \end{minipage}
    %
    %
    \begin{minipage}{0.45\textwidth}
        \begin{subfigure}{\textwidth}
            \centering
            \includegraphics[height=0.17\textheight]{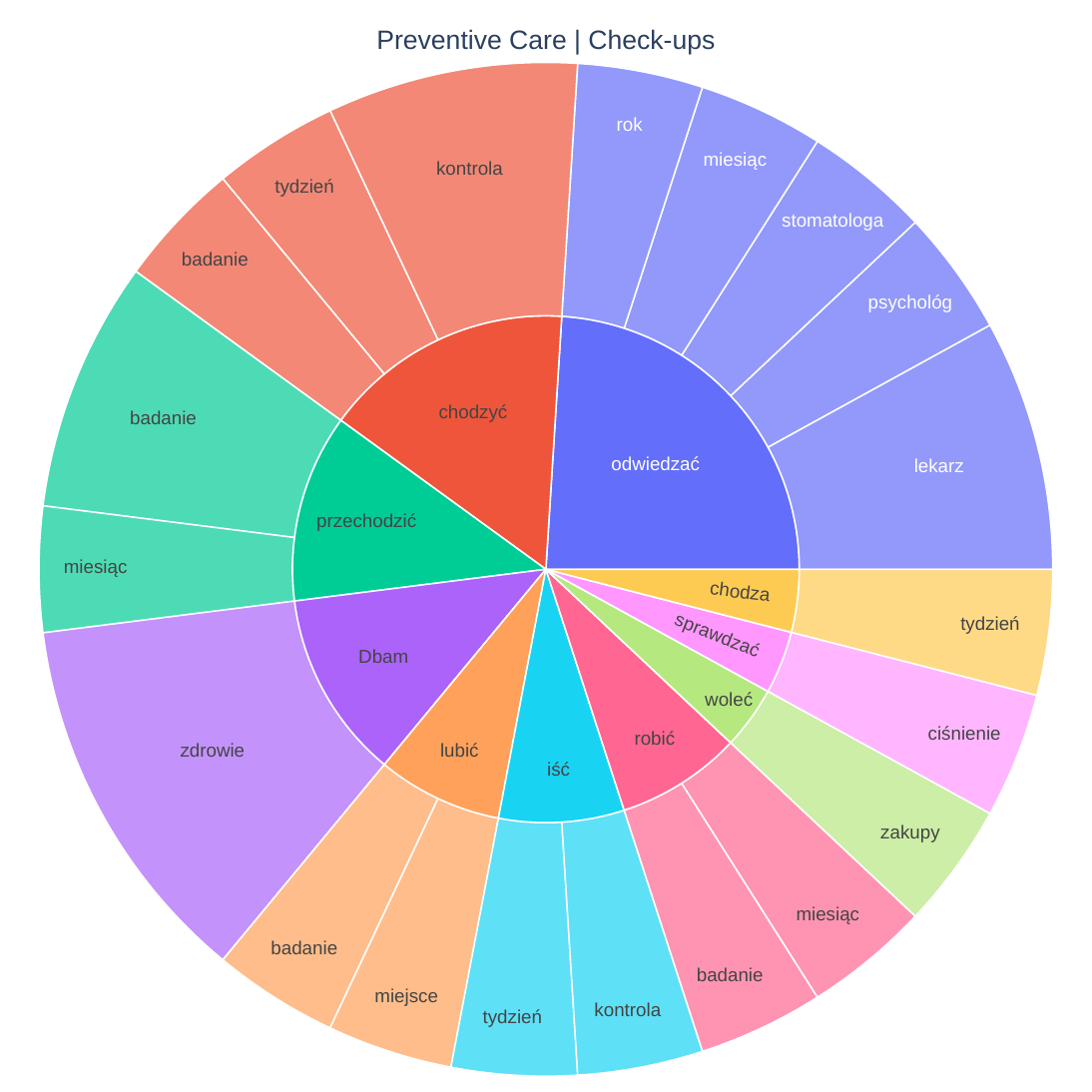}
        \end{subfigure}
        \vskip\baselineskip
        \begin{subfigure}{\textwidth}
            \centering
            \includegraphics[height=0.17\textheight]{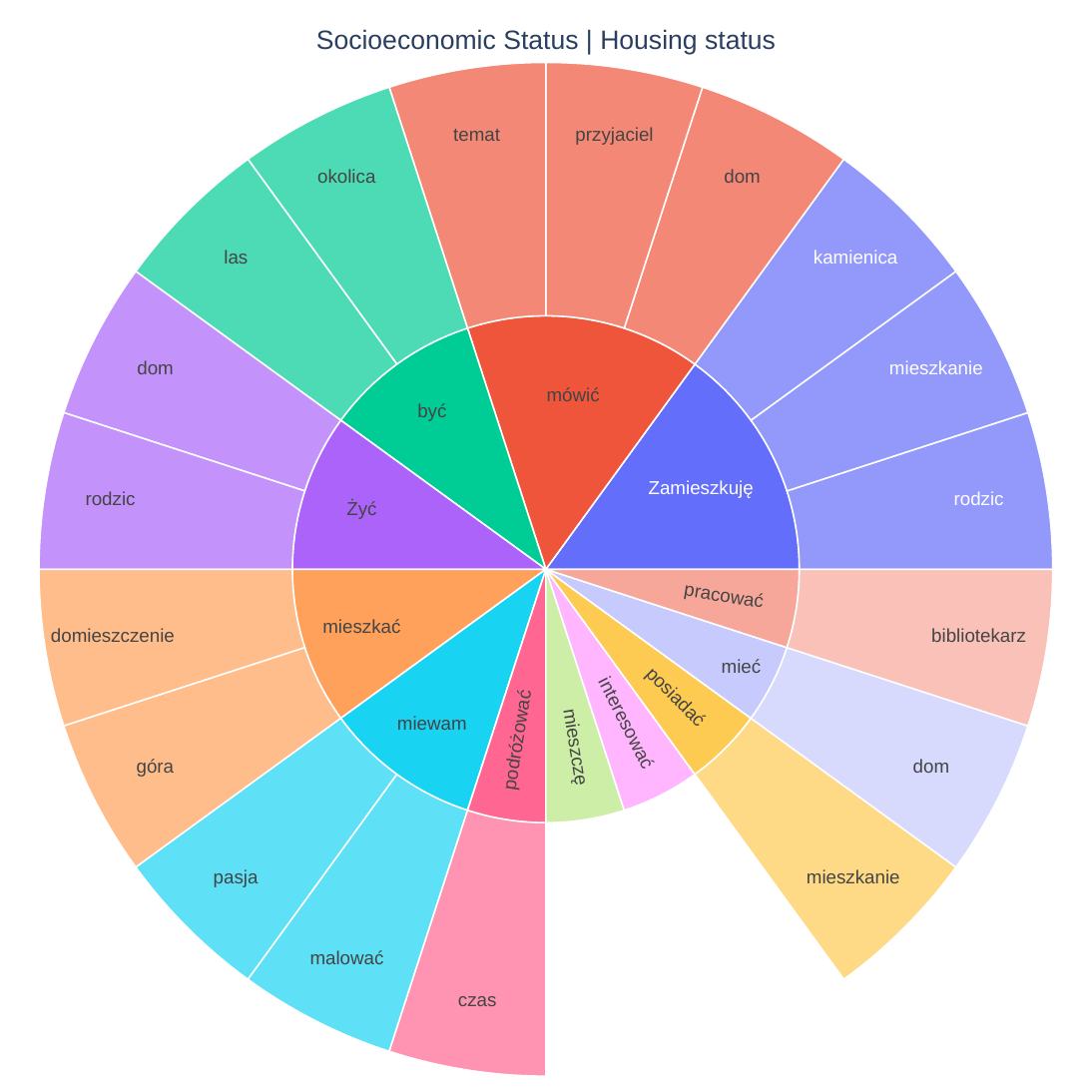}
        \end{subfigure}
        \vskip\baselineskip
        \begin{subfigure}{\textwidth}
            \centering
            \includegraphics[height=0.17\textheight]{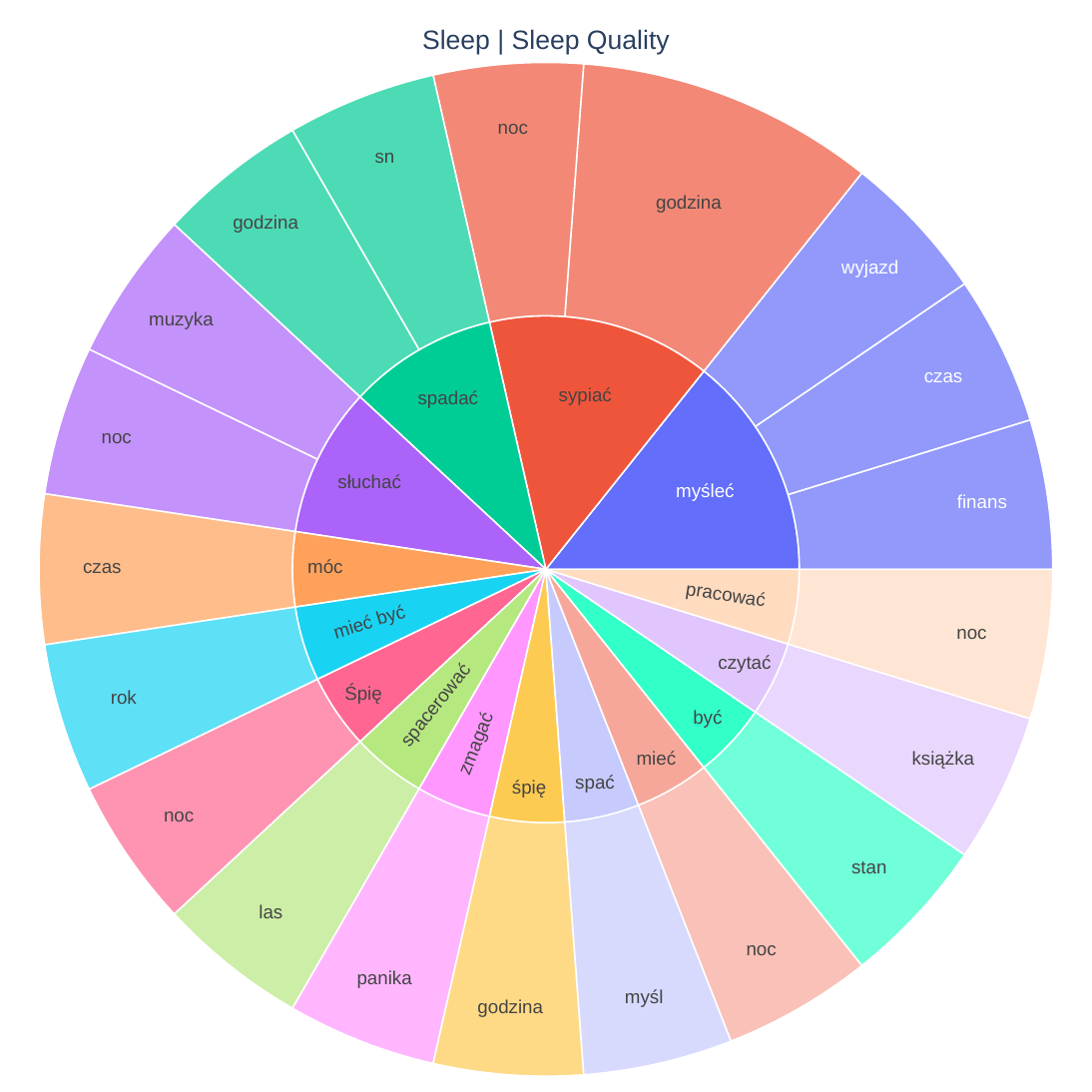}
        \end{subfigure}
        \vskip\baselineskip
        \begin{subfigure}{\textwidth}
            \centering
            \includegraphics[height=0.17\textheight]{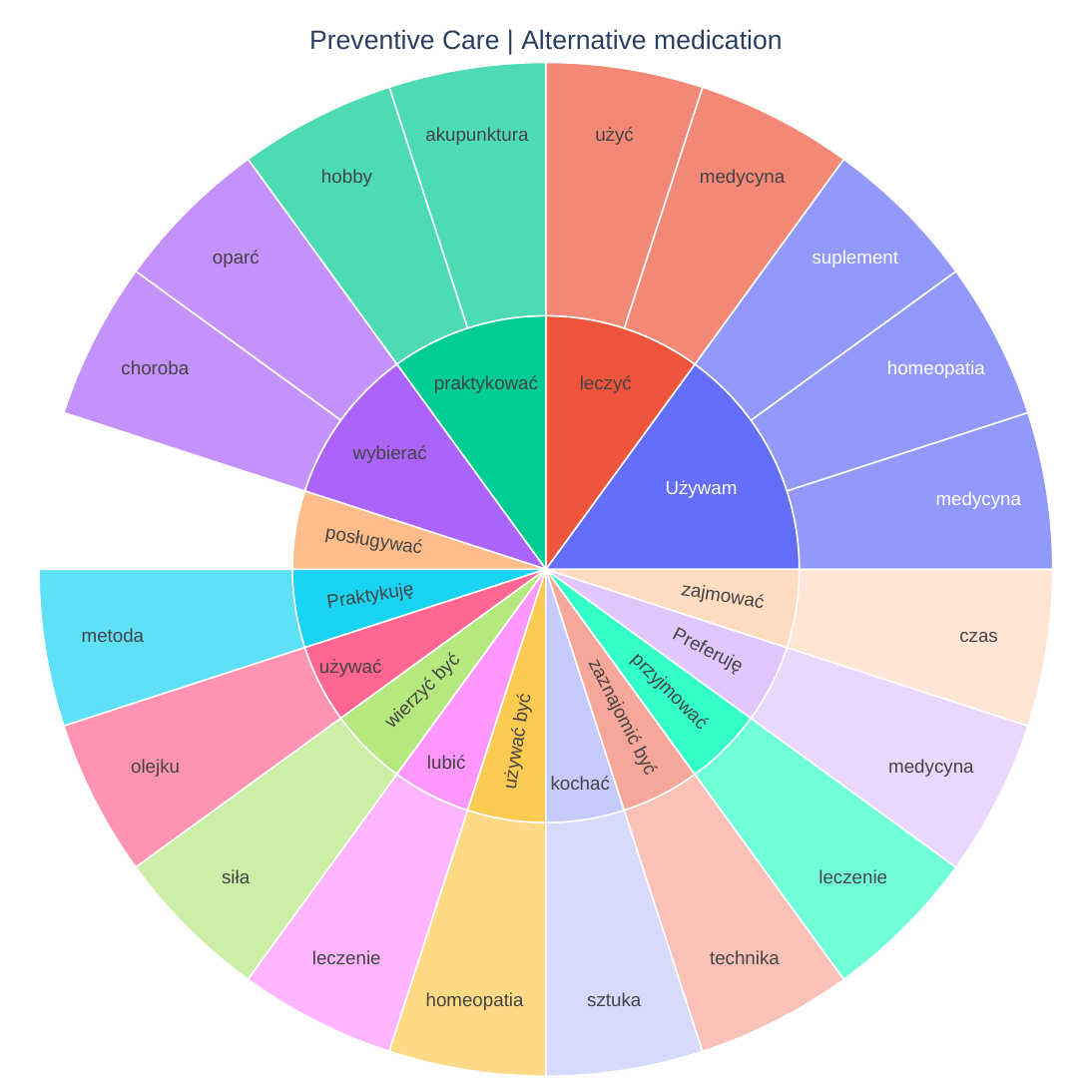}
        \end{subfigure}
    \end{minipage}
    
\caption{Detailed BERTSCORE for Polish Personas in different generation configurations and Sunburst charts of personas taxonomy entities with most root verbs and associated object noun for the different models}    
\end{figure}


\restoregeometry

\newgeometry{top=0.5cm, bottom=1.5cm, left=2.5cm, right=2.5cm}
\subsection{Medium-Resource Languages}

 \subsubsection{\textsc{Vietnamese}}

\begin{figure}[h]
    \centering
    \begin{minipage}{0.45\textwidth}
        \begin{subfigure}{\textwidth}
            \centering
            \includegraphics[height=0.15\textheight]{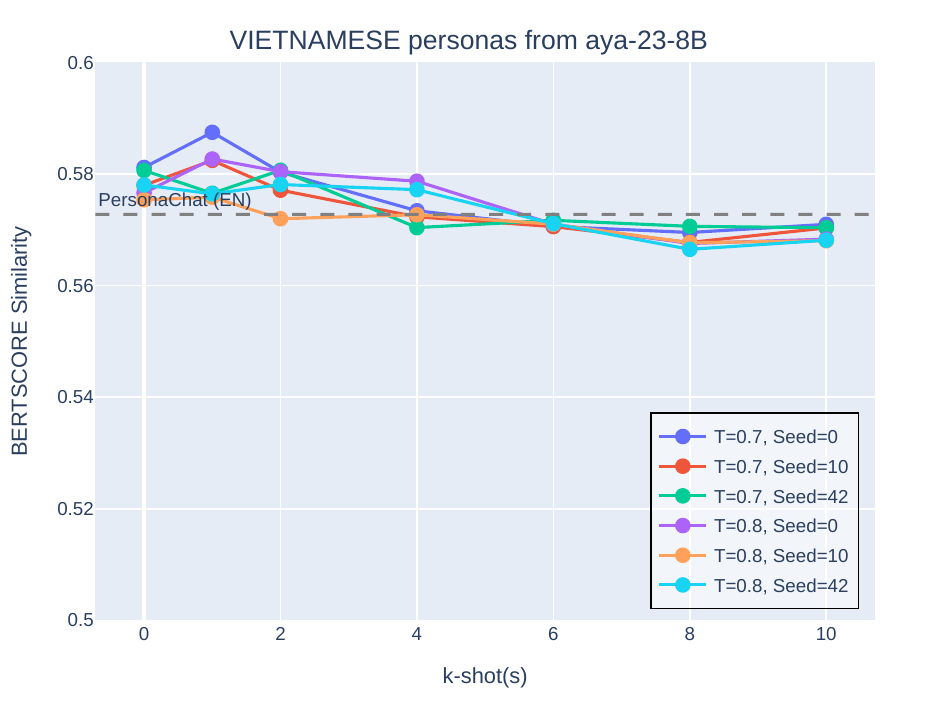}
        \end{subfigure}
        \vskip\baselineskip
        \begin{subfigure}{\textwidth}
            \centering
            \includegraphics[height=0.15\textheight]{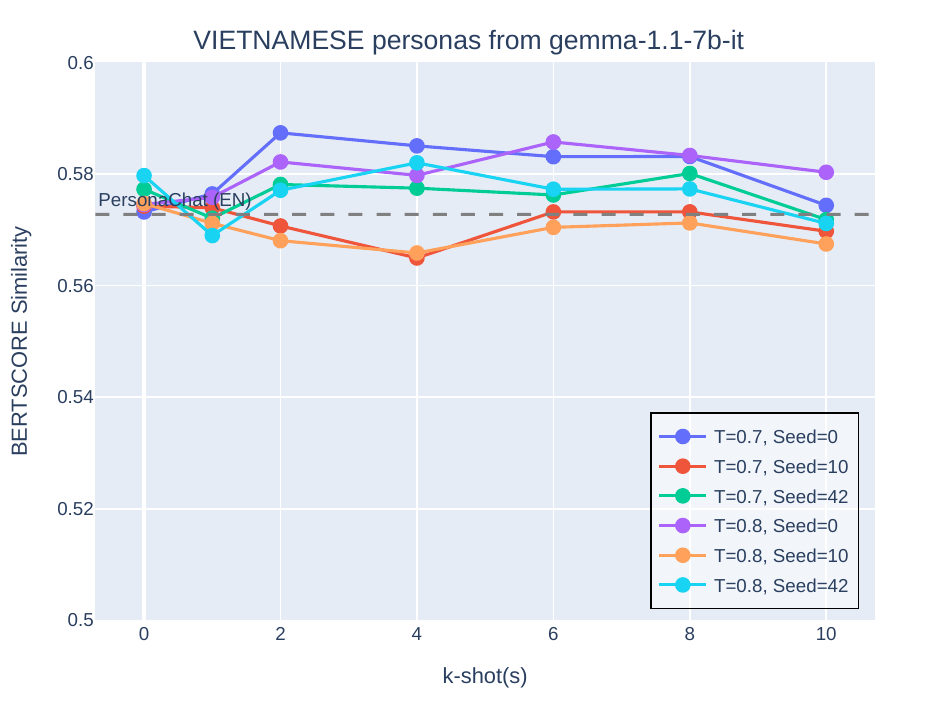}
        \end{subfigure}
    \end{minipage}
    %
    %
    \begin{minipage}{0.45\textwidth}
            \begin{subfigure}{\textwidth}
            \centering
            \includegraphics[height=0.15\textheight]{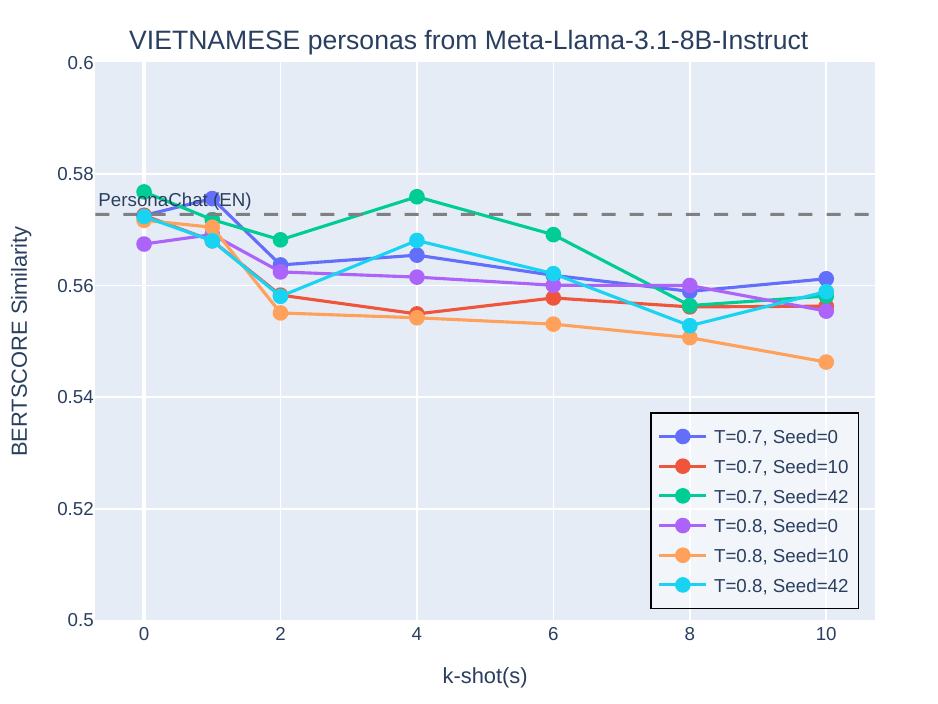}
        \end{subfigure}
        \vskip\baselineskip
        \begin{subfigure}{\textwidth}
            \centering
            \includegraphics[height=0.15\textheight]{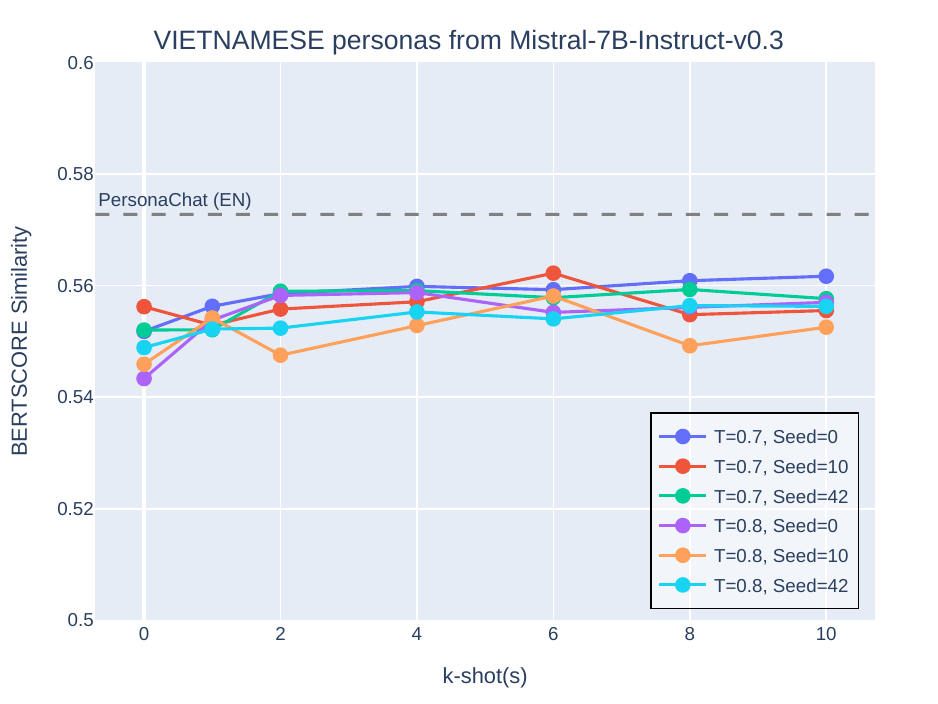}
        \end{subfigure}
    \end{minipage}
    
\caption{Detailed BERTSCORE for Vietnamese Personas in different generation configurations for the different models}    
\end{figure}



\subsubsection{\textsc{Indonesian}}

\begin{figure}[h]
    \centering
    \begin{minipage}{0.45\textwidth}
        \begin{subfigure}{\textwidth}
            \centering
            \includegraphics[height=0.15\textheight]{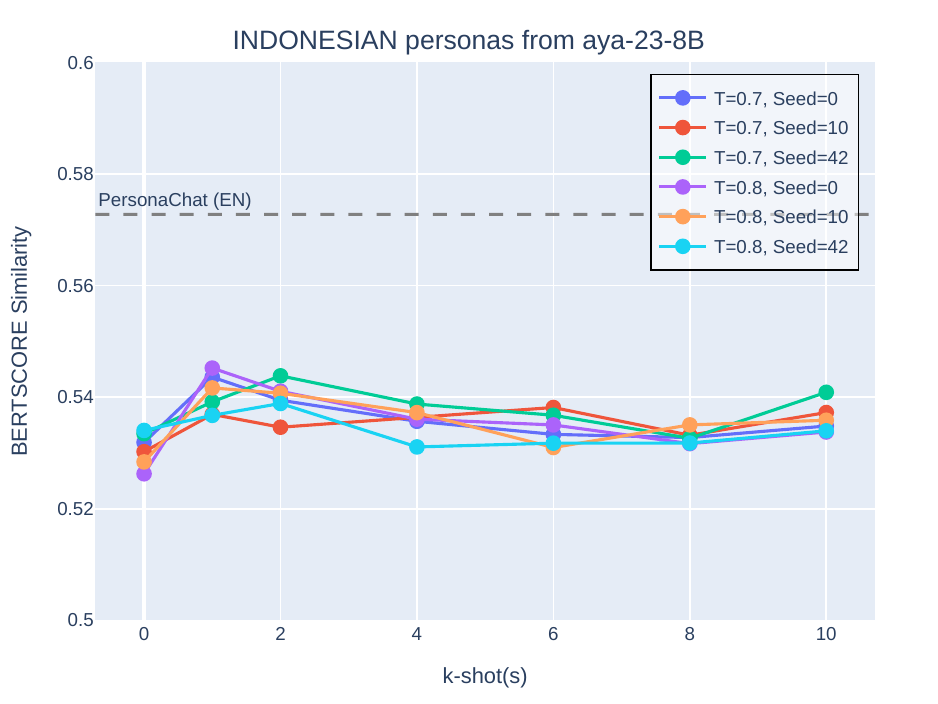}
        \end{subfigure}
        \vskip\baselineskip
        \begin{subfigure}{\textwidth}
            \centering
            \includegraphics[height=0.15\textheight]{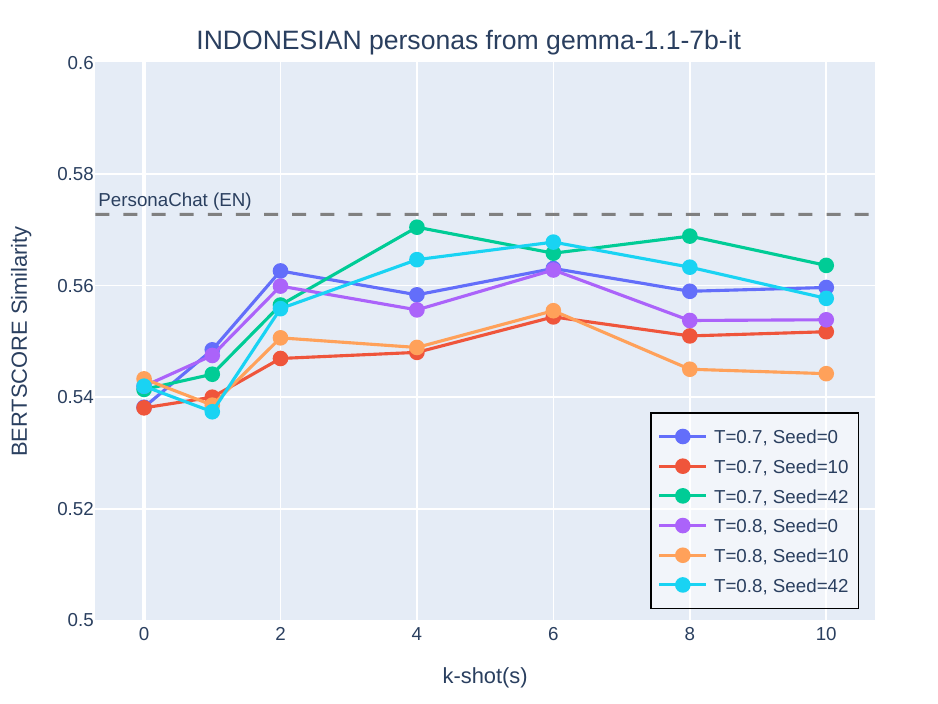}
        \end{subfigure}
    \end{minipage}
    %
    %
    \begin{minipage}{0.45\textwidth}
            \begin{subfigure}{\textwidth}
            \centering
            \includegraphics[height=0.15\textheight]{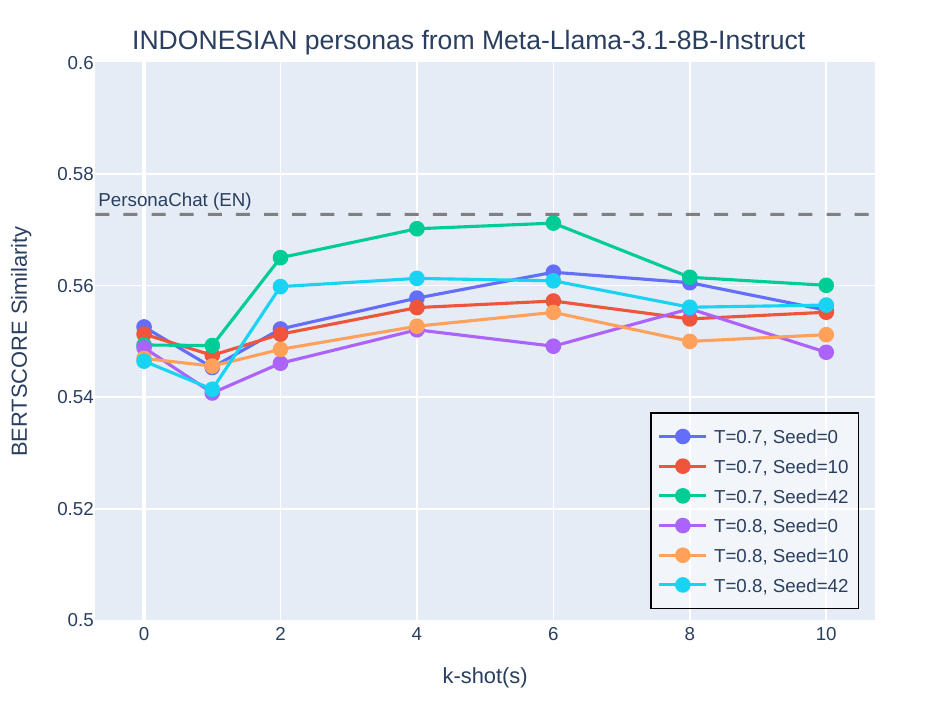}
        \end{subfigure}
        \vskip\baselineskip
        \begin{subfigure}{\textwidth}
            \centering
            \includegraphics[height=0.15\textheight]{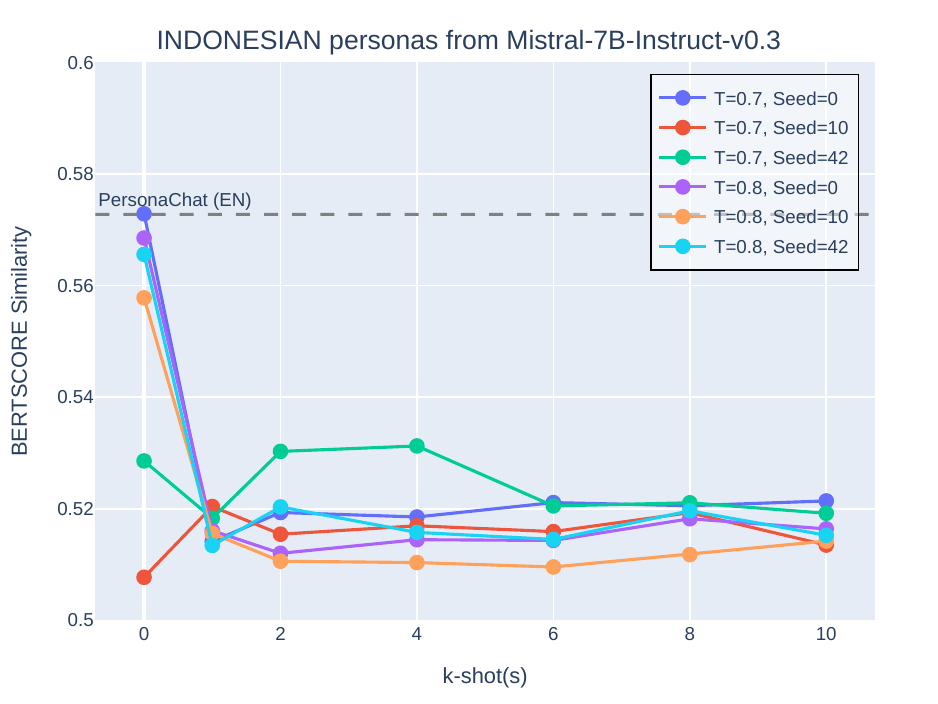}
        \end{subfigure}
    \end{minipage}
    
\caption{Detailed BERTSCORE for Indonesian Personas in different generation configurations for the different models}    
\end{figure}



\newgeometry{top=0.5cm, bottom=1.5cm, left=2.5cm, right=2.5cm}
\subsubsection{\textsc{Korean}}

\begin{figure}[h]
    \centering
    \begin{minipage}{0.45\textwidth}
        \begin{subfigure}{\textwidth}
            \centering
            \includegraphics[height=0.17\textheight]{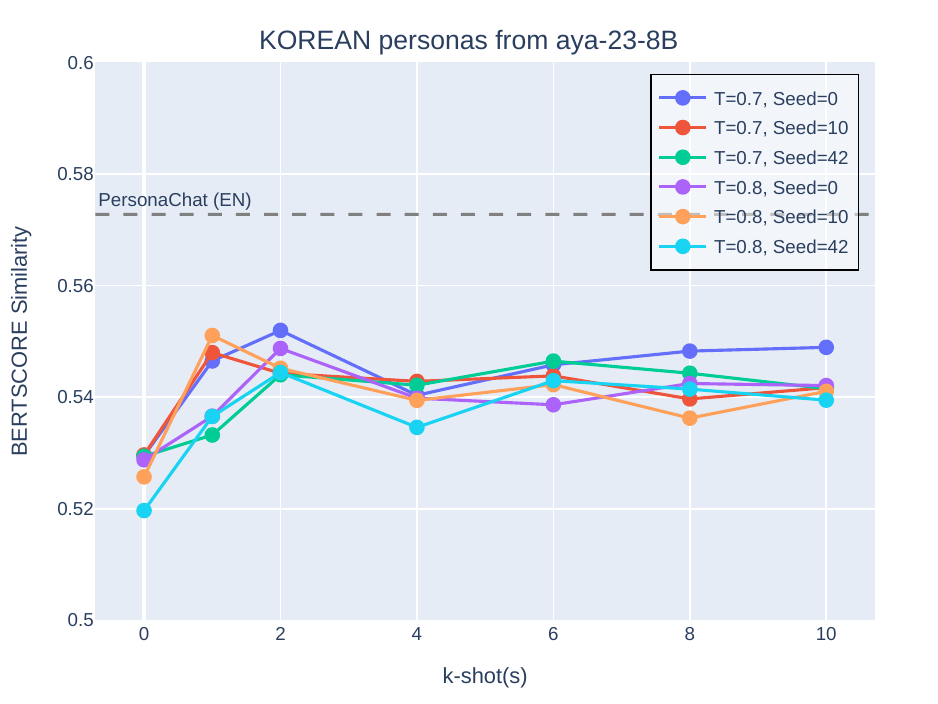}
        \end{subfigure}
        \vskip\baselineskip
        \begin{subfigure}{\textwidth}
            \centering
            \includegraphics[height=0.17\textheight]{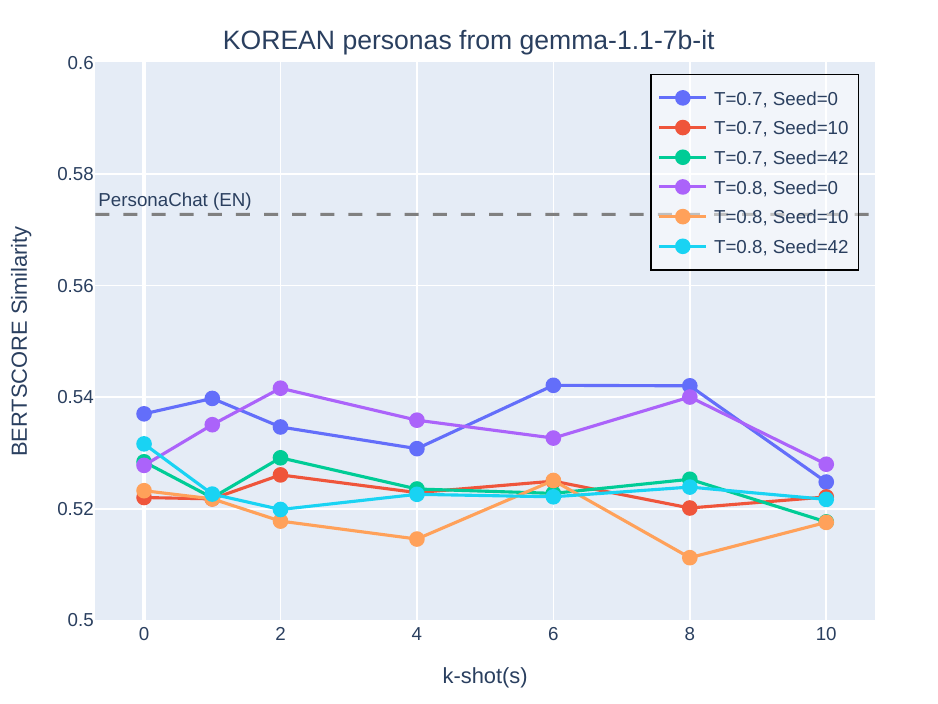}
        \end{subfigure}
        \vskip\baselineskip
        \begin{subfigure}{\textwidth}
            \centering
            \includegraphics[height=0.17\textheight]{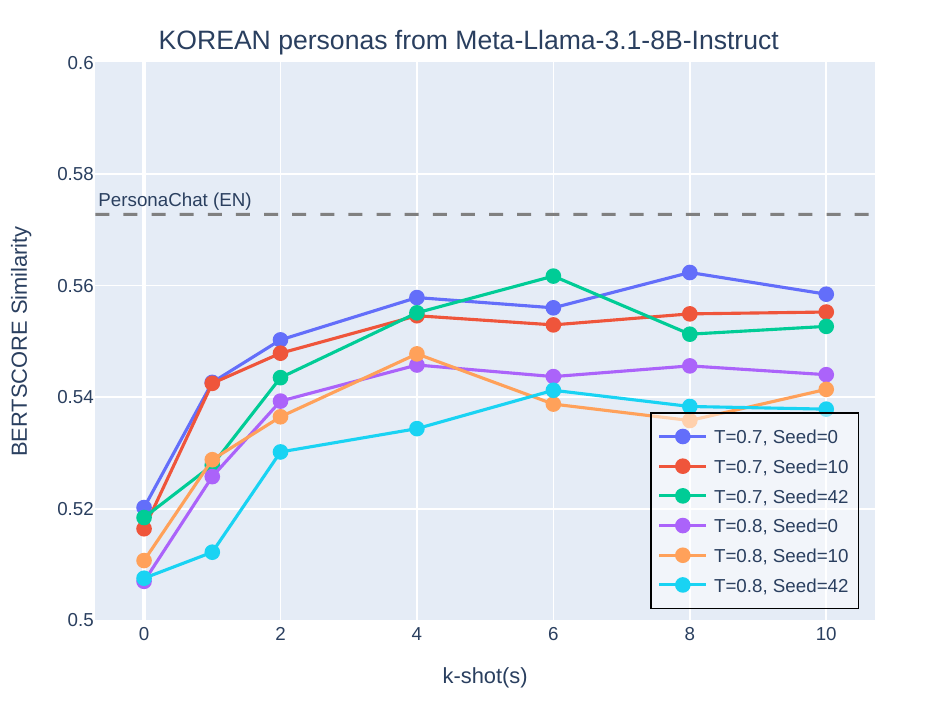}
        \end{subfigure}
        \vskip\baselineskip
        \begin{subfigure}{\textwidth}
            \centering
            \includegraphics[height=0.17\textheight]{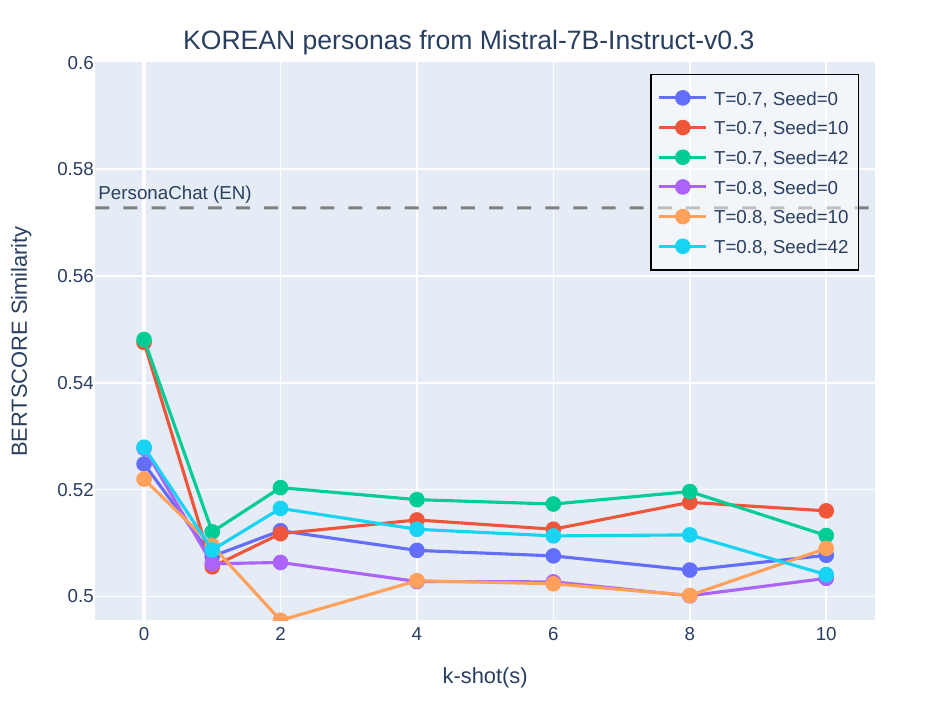}
        \end{subfigure}
    \end{minipage}
    %
    %
    \begin{minipage}{0.45\textwidth}
        \begin{subfigure}{\textwidth}
            \centering
            \includegraphics[height=0.17\textheight]{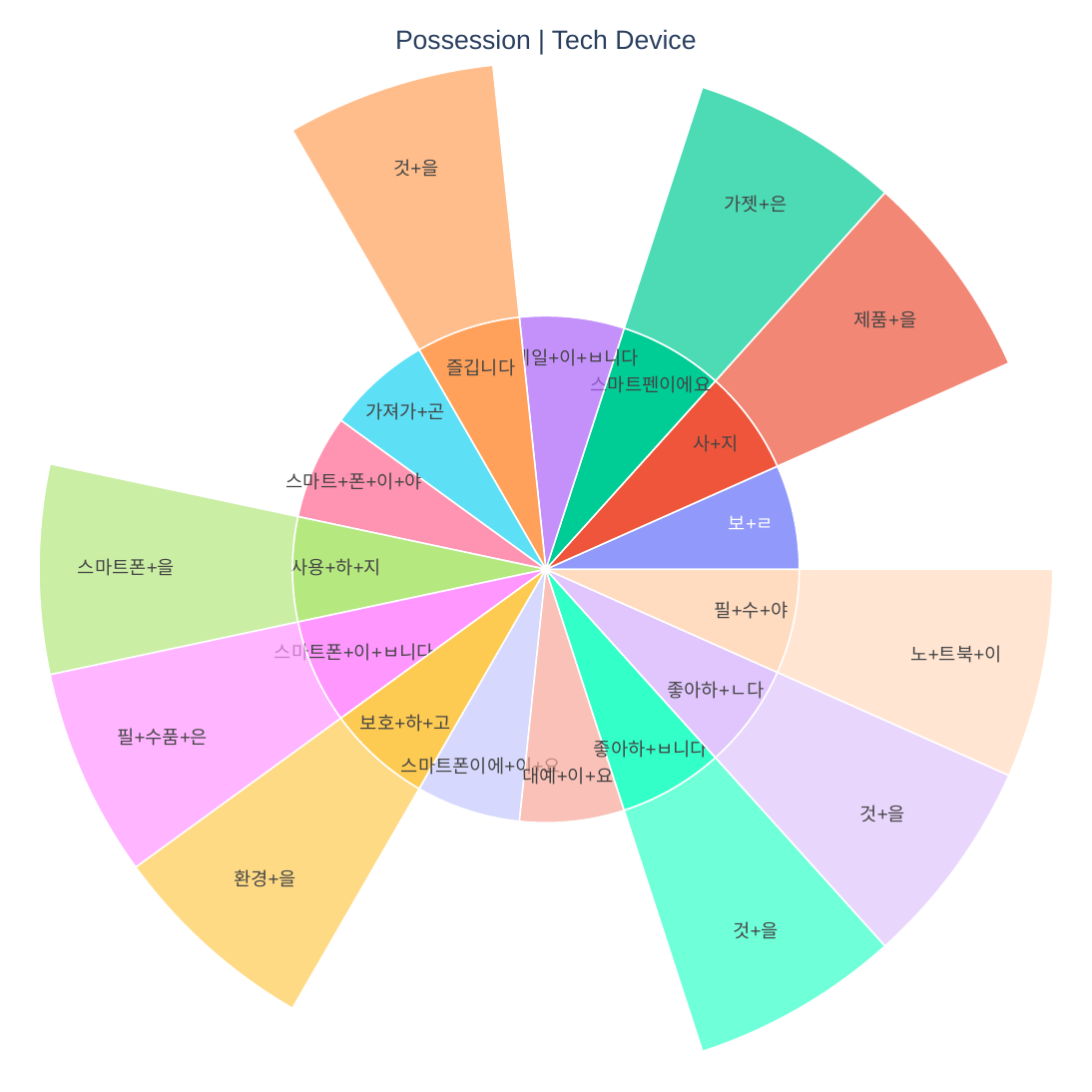}
        \end{subfigure}
        \vskip\baselineskip
        \begin{subfigure}{\textwidth}
            \centering
            \includegraphics[height=0.17\textheight]{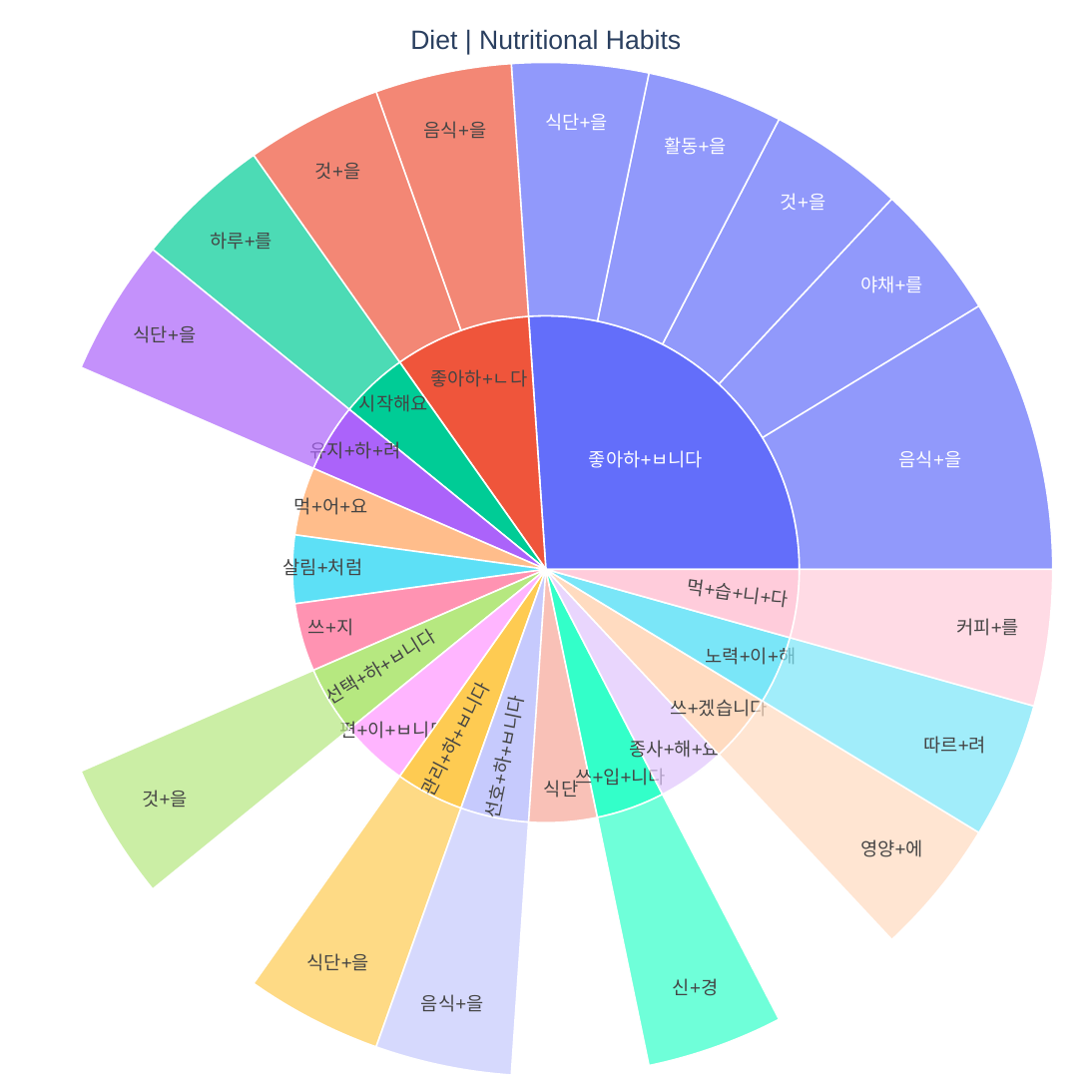}
        \end{subfigure}
        \vskip\baselineskip
        \begin{subfigure}{\textwidth}
            \centering
            \includegraphics[height=0.17\textheight]{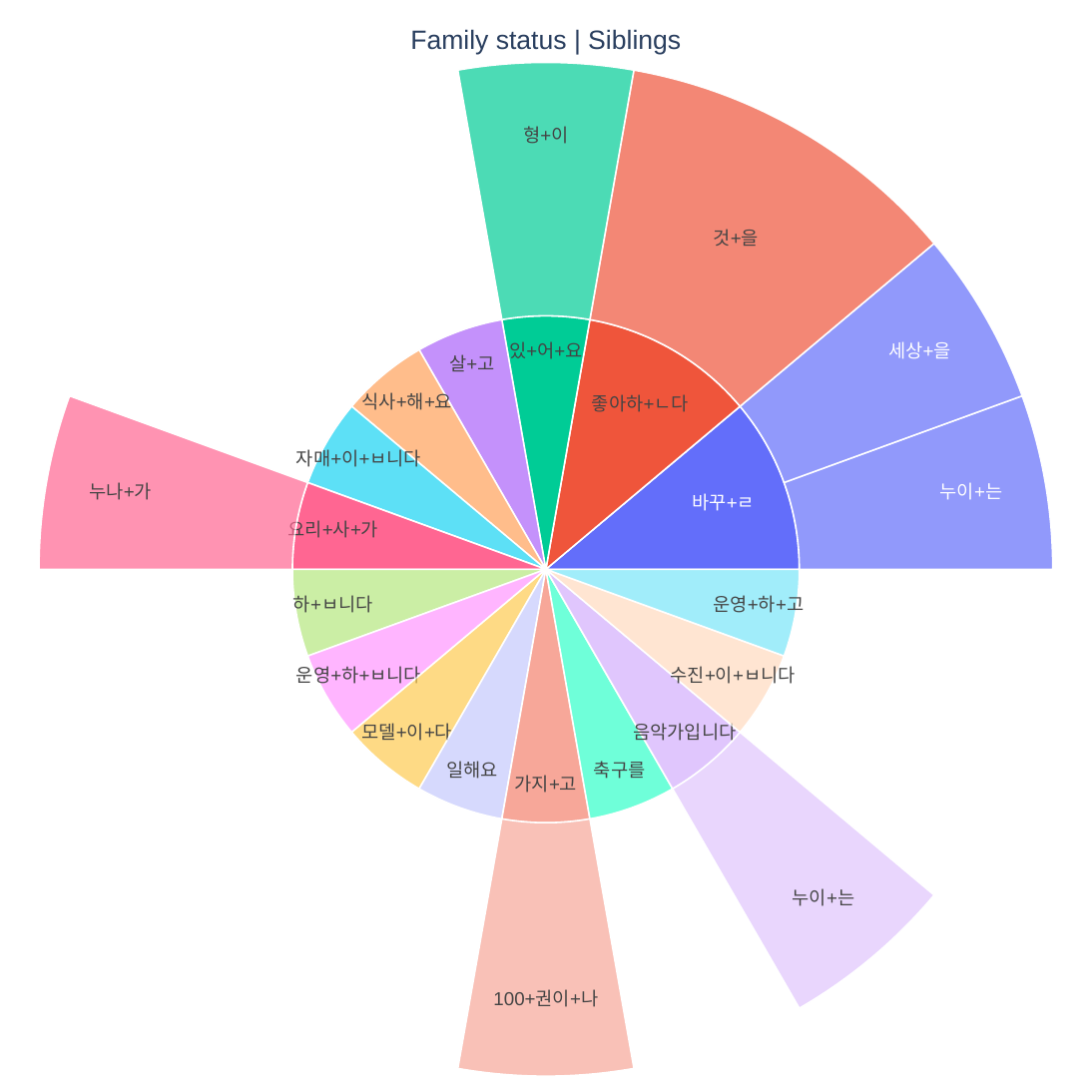}
        \end{subfigure}
        \vskip\baselineskip
        \begin{subfigure}{\textwidth}
            \centering
            \includegraphics[height=0.17\textheight]{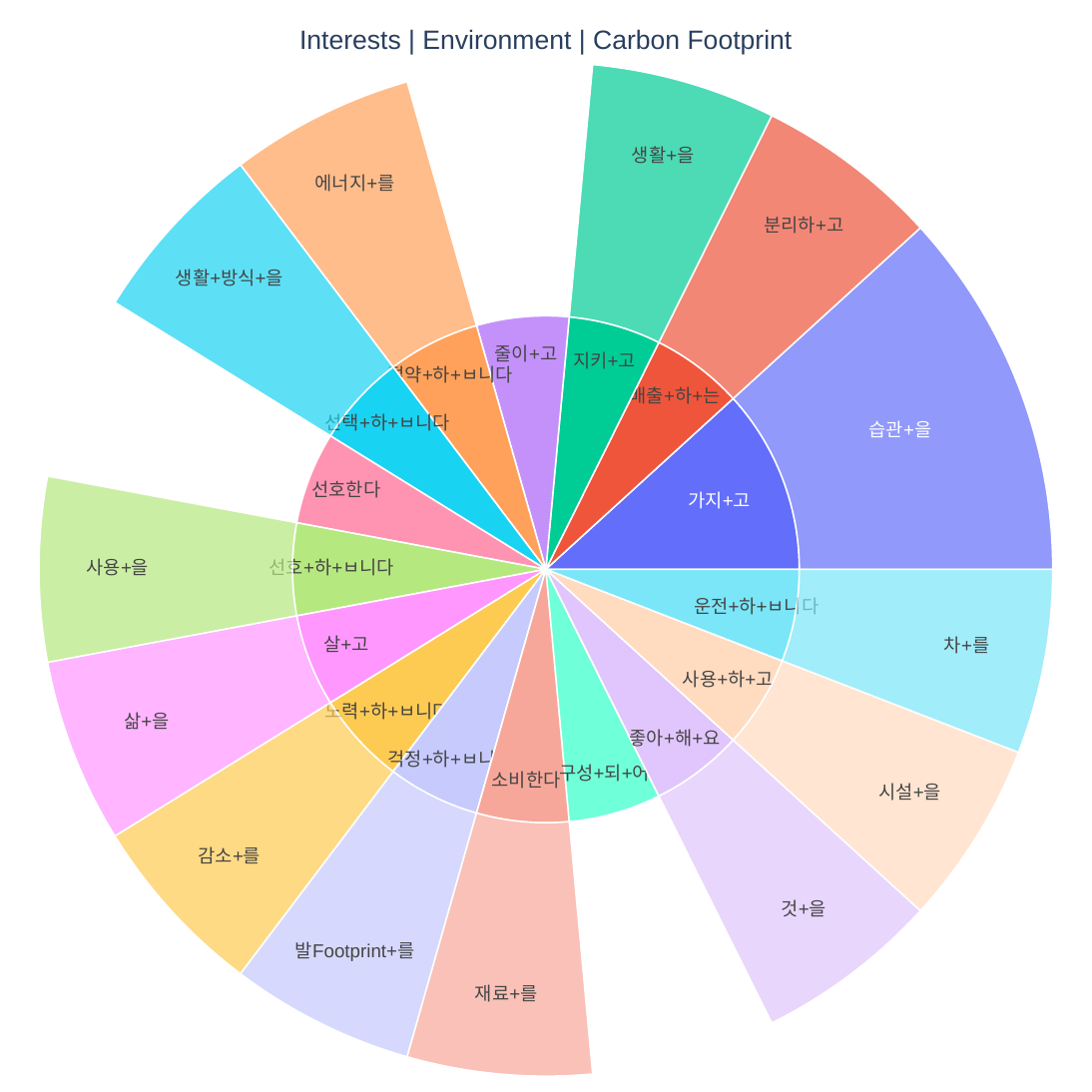}
        \end{subfigure}
    \end{minipage}
    
\caption{Detailed BERTSCORE for Korean Personas in different generation configurations and Sunburst charts of personas taxonomy entities with most root verbs and associated object noun for the different models}    
\end{figure}


\restoregeometry

\newgeometry{top=0.5cm, bottom=1.5cm, left=2.5cm, right=2.5cm}
\subsubsection{\textsc{Swedish}}

\begin{figure}[h]
    \centering
    \begin{minipage}{0.45\textwidth}
        \begin{subfigure}{\textwidth}
            \centering
            \includegraphics[height=0.17\textheight]{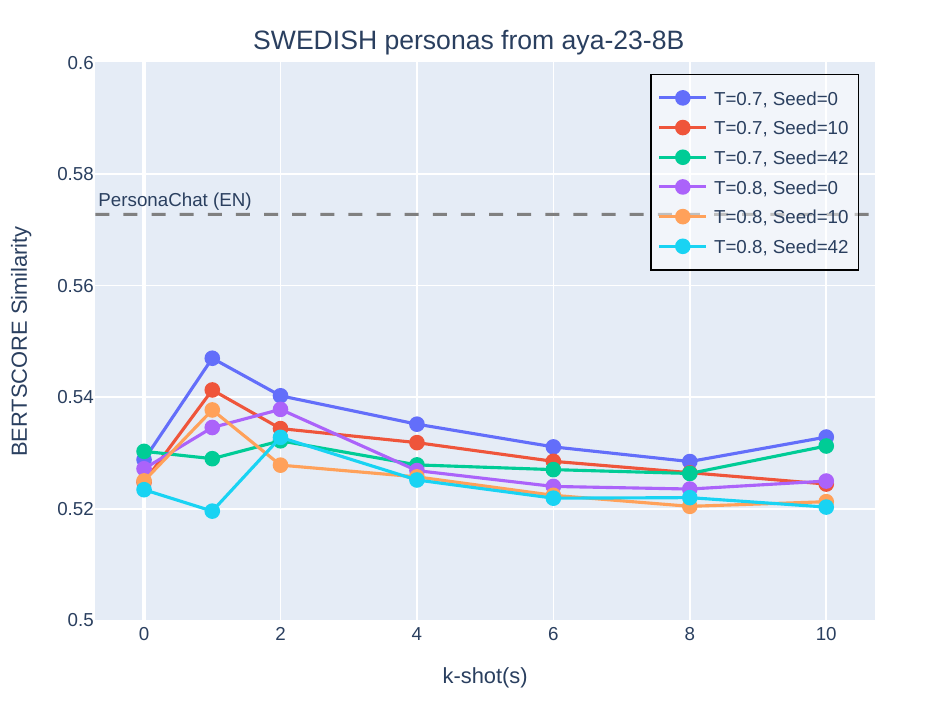}
        \end{subfigure}
        \vskip\baselineskip
        \begin{subfigure}{\textwidth}
            \centering
            \includegraphics[height=0.17\textheight]{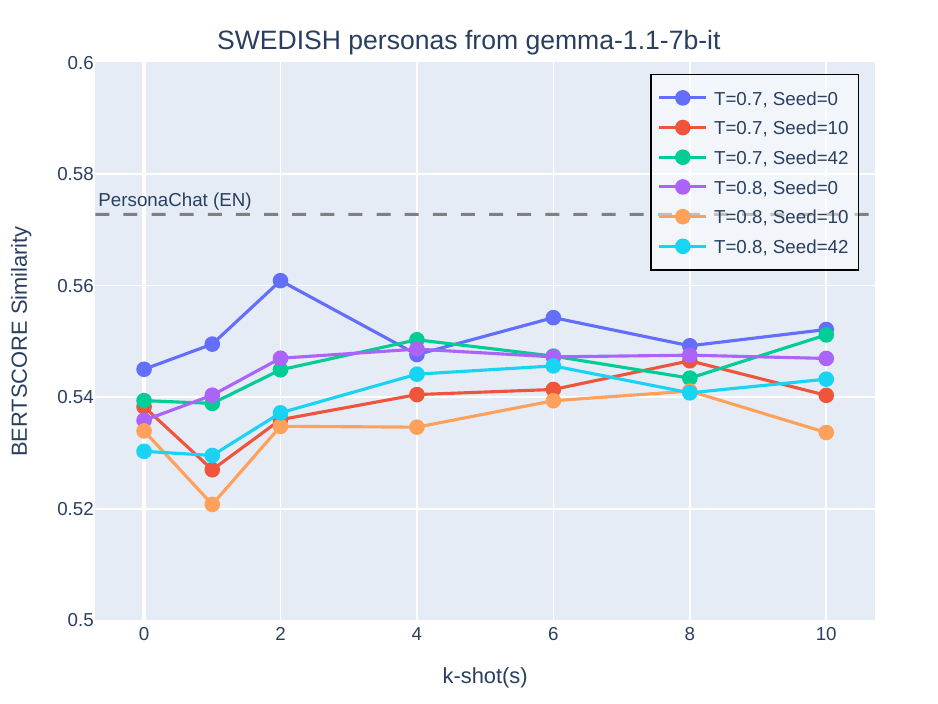}
        \end{subfigure}
        \vskip\baselineskip
        \begin{subfigure}{\textwidth}
            \centering
            \includegraphics[height=0.17\textheight]{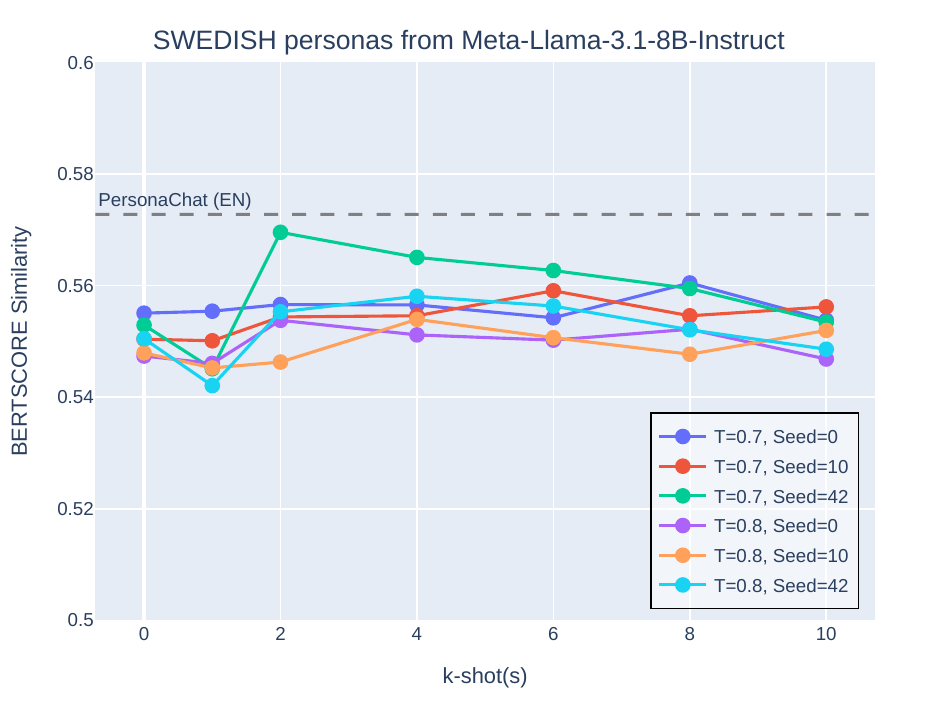}
        \end{subfigure}
        \vskip\baselineskip
        \begin{subfigure}{\textwidth}
            \centering
            \includegraphics[height=0.17\textheight]{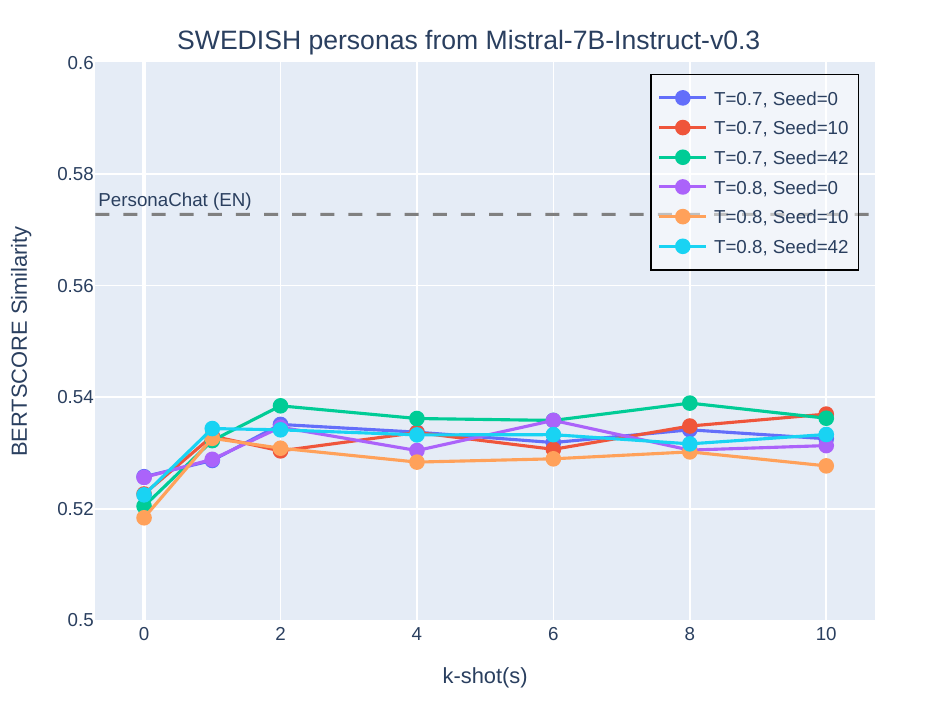}
        \end{subfigure}
    \end{minipage}
    %
    %
    \begin{minipage}{0.45\textwidth}
        \begin{subfigure}{\textwidth}
            \centering
            \includegraphics[height=0.17\textheight]{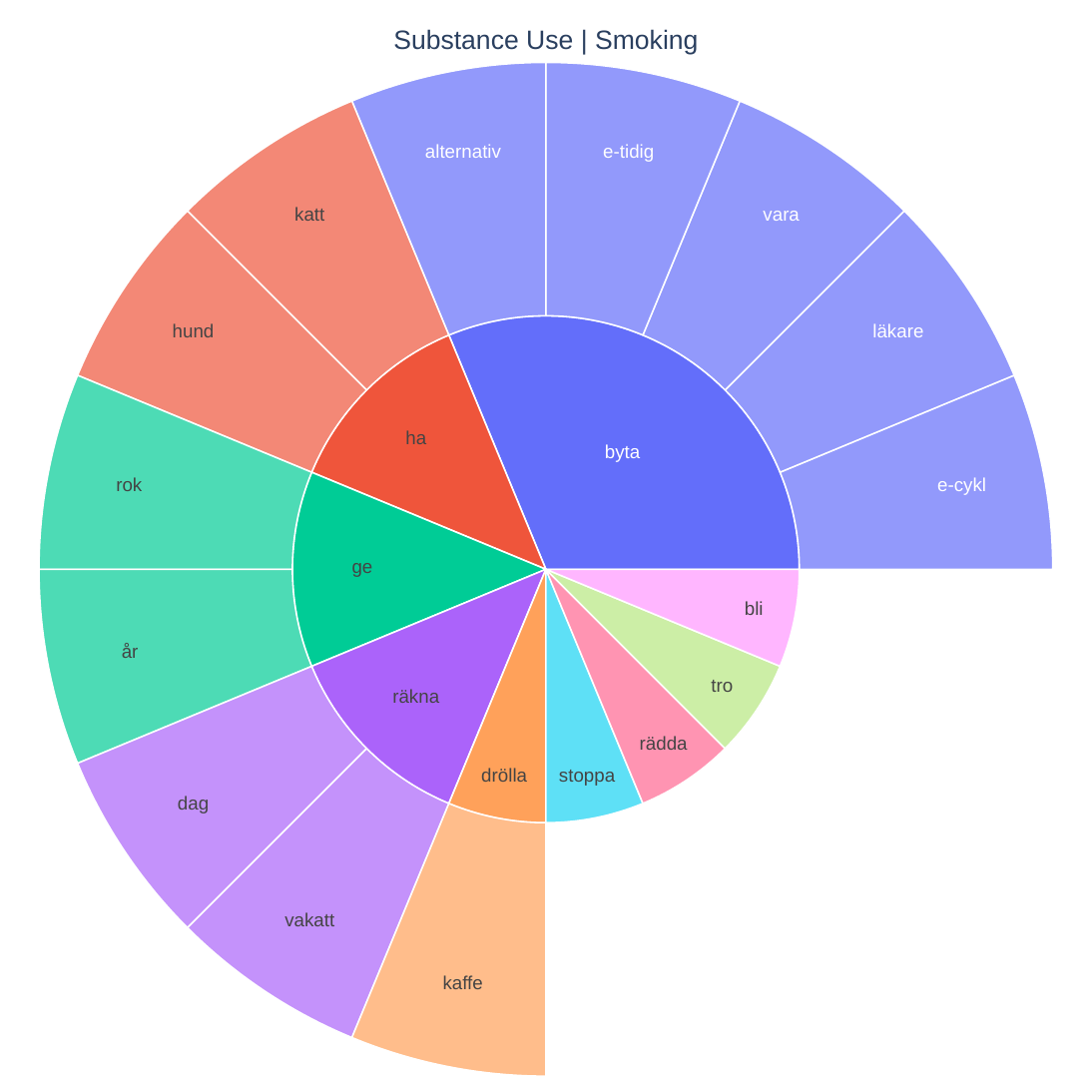}
        \end{subfigure}
        \vskip\baselineskip
        \begin{subfigure}{\textwidth}
            \centering
            \includegraphics[height=0.17\textheight]{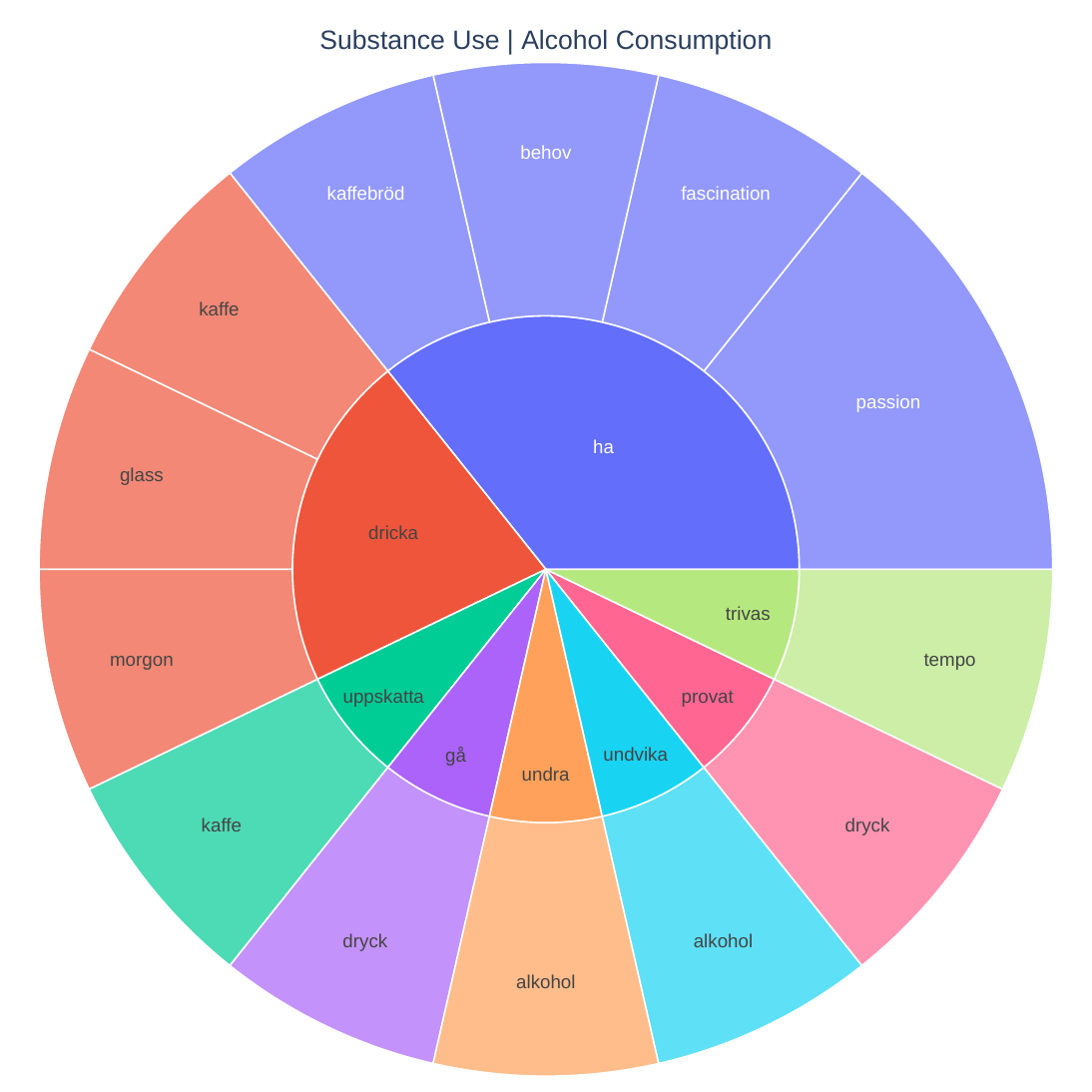}
        \end{subfigure}
        \vskip\baselineskip
        \begin{subfigure}{\textwidth}
            \centering
            \includegraphics[height=0.17\textheight]{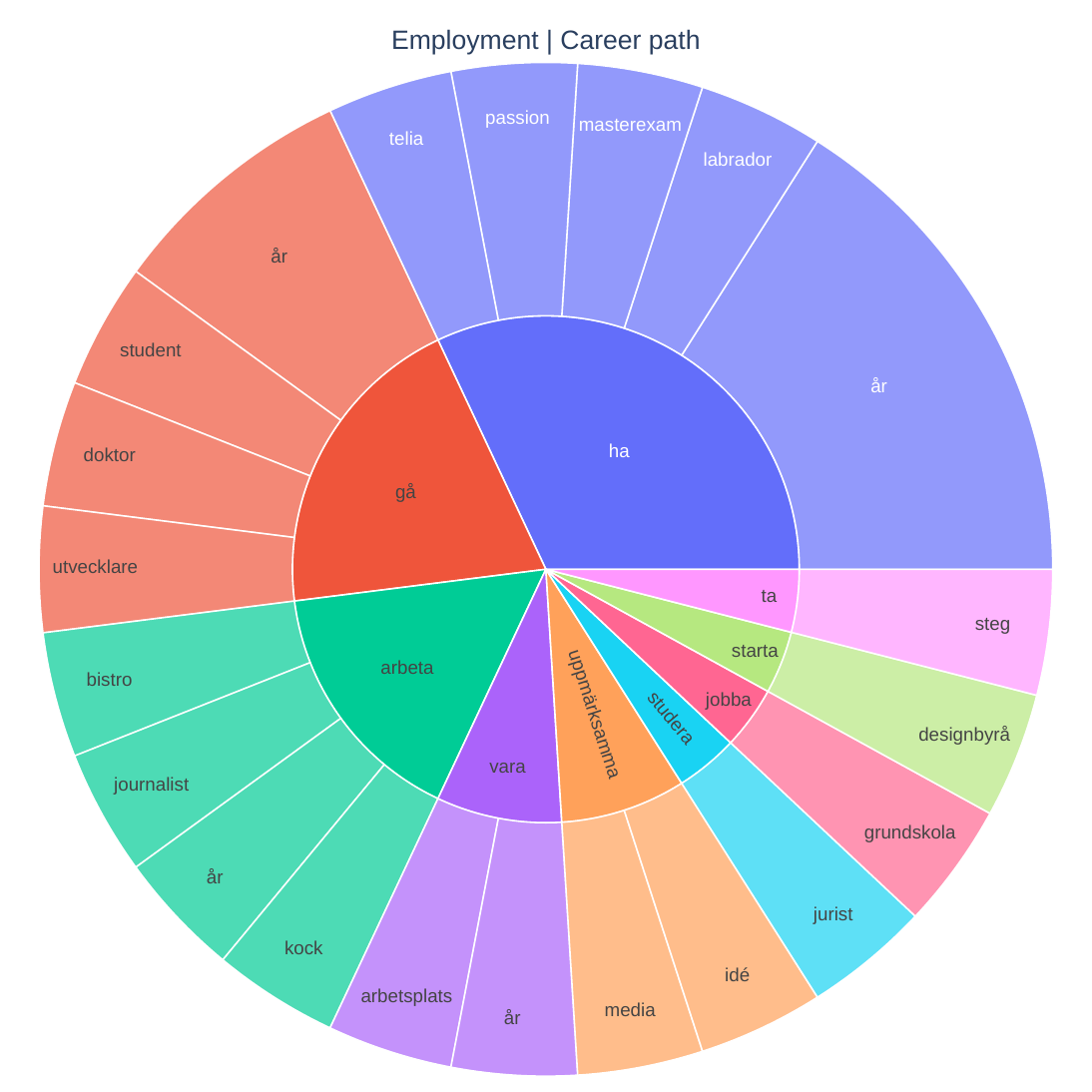}
        \end{subfigure}
        \vskip\baselineskip
        \begin{subfigure}{\textwidth}
            \centering
            \includegraphics[height=0.17\textheight]{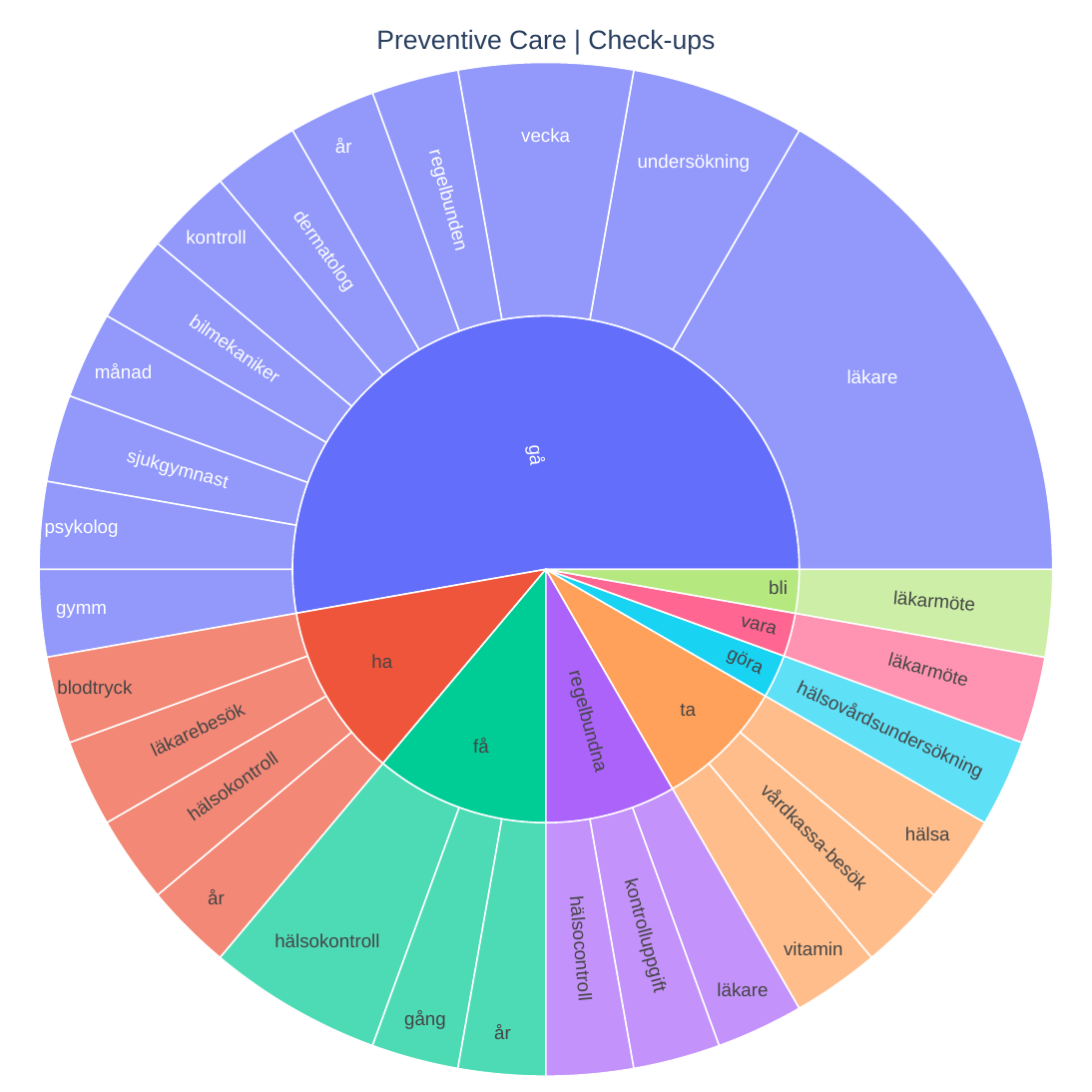}
        \end{subfigure}
    \end{minipage}
    
\caption{Detailed BERTSCORE for Swedish Personas in different generation configurations and Sunburst charts of personas taxonomy entities with most root verbs and associated object noun for the different models}    
\end{figure}


\restoregeometry
\newgeometry{top=0.5cm, bottom=1.5cm, left=2.5cm, right=2.5cm}
 
 \subsubsection{\textsc{Arabic}}

\begin{figure}[h]
    \centering
    \begin{minipage}{0.45\textwidth}
        \begin{subfigure}{\textwidth}
            \centering
            \includegraphics[height=0.15\textheight]{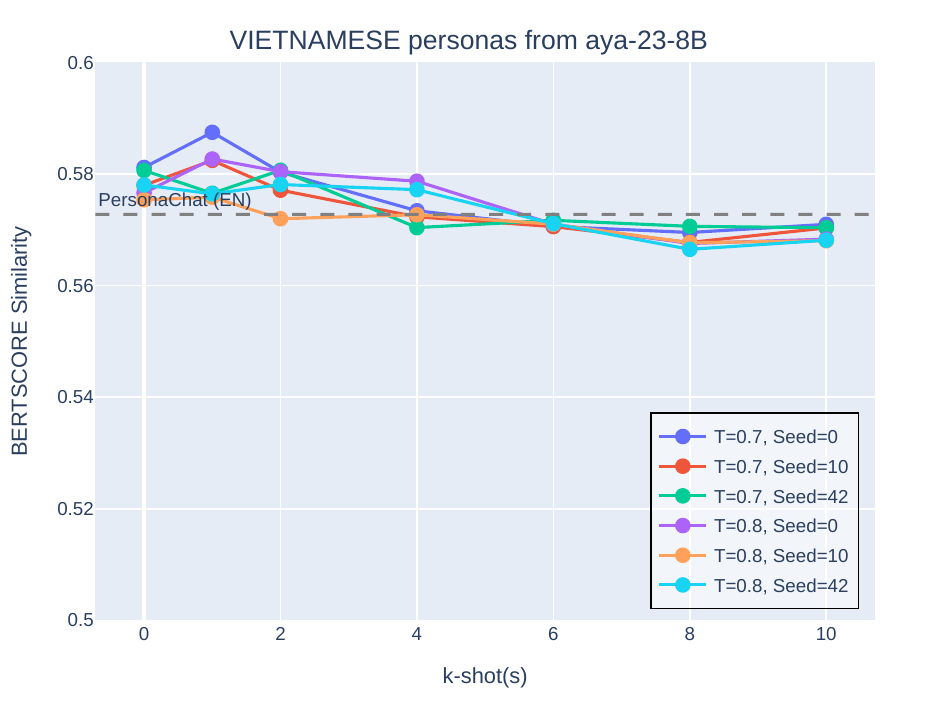}
        \end{subfigure}
        \vskip\baselineskip
        \begin{subfigure}{\textwidth}
            \centering
            \includegraphics[height=0.15\textheight]{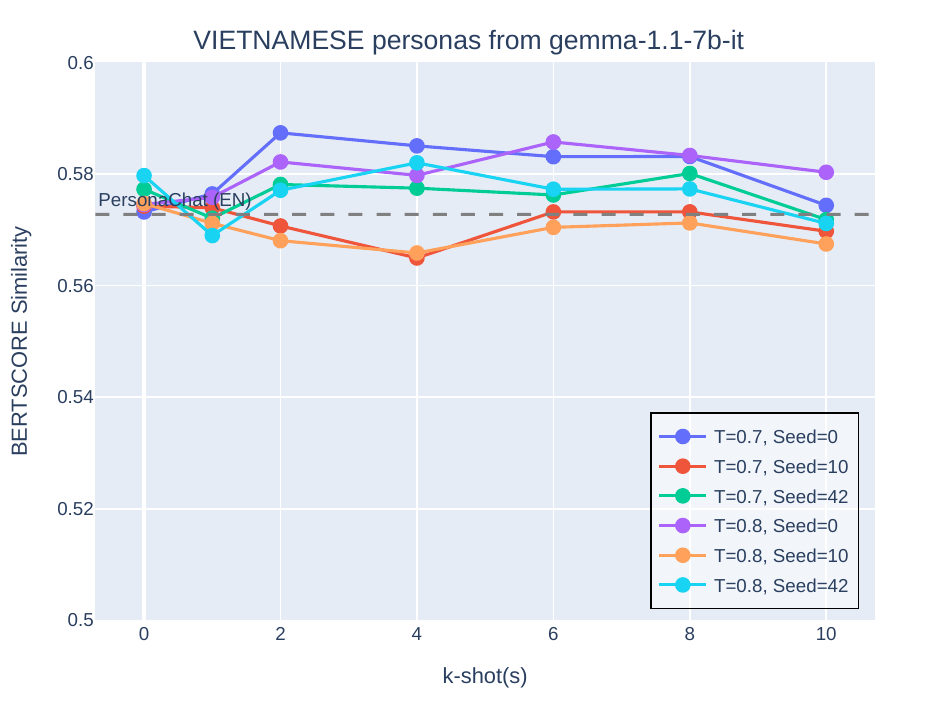}
        \end{subfigure}
    \end{minipage}
    %
    %
    \begin{minipage}{0.45\textwidth}
            \begin{subfigure}{\textwidth}
            \centering
            \includegraphics[height=0.15\textheight]{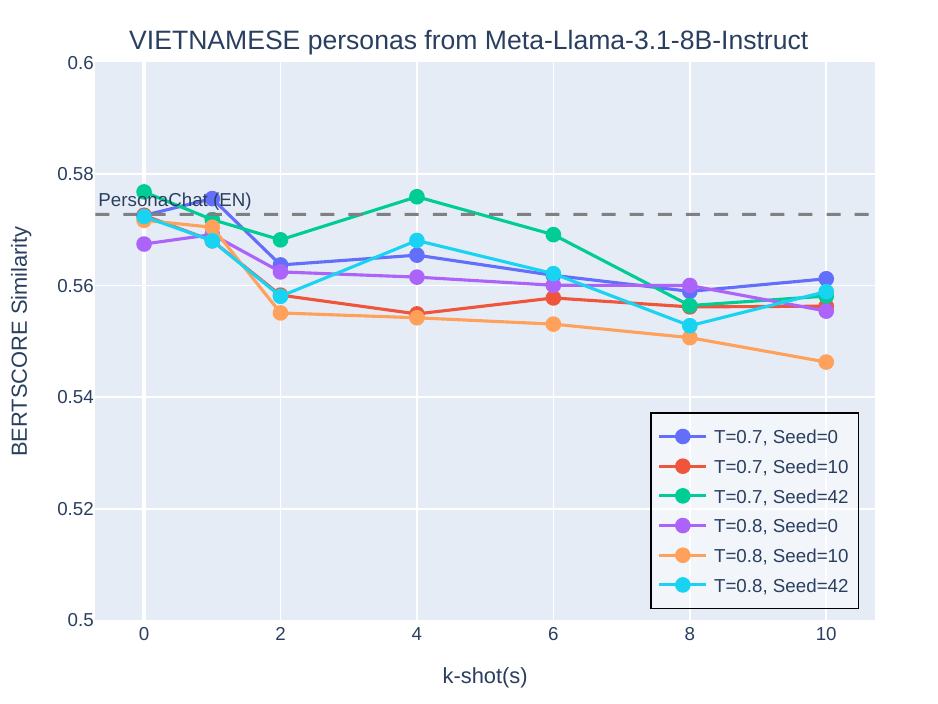}
        \end{subfigure}
        \vskip\baselineskip
        \begin{subfigure}{\textwidth}
            \centering
            \includegraphics[height=0.15\textheight]{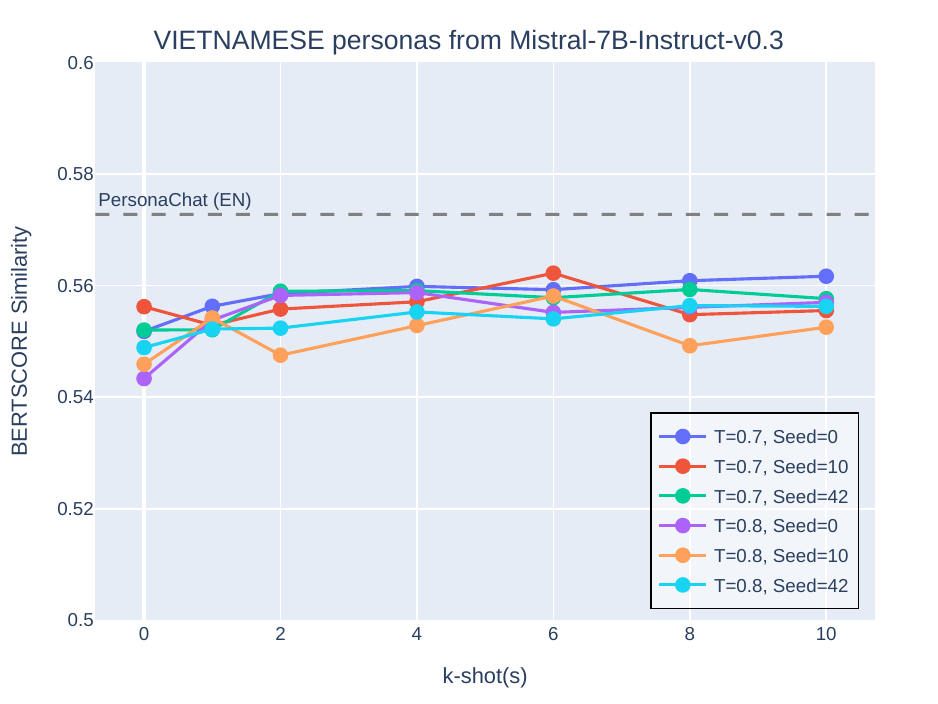}
        \end{subfigure}
    \end{minipage}
    
\caption{Detailed BERTSCORE for Arabic Personas in different generation configurations for the different models}    
\end{figure}



\subsubsection{\textsc{Hungarian}}

\begin{figure}[h]
    \centering
    \begin{minipage}{0.45\textwidth}
        \begin{subfigure}{\textwidth}
            \centering
            \includegraphics[height=0.15\textheight]{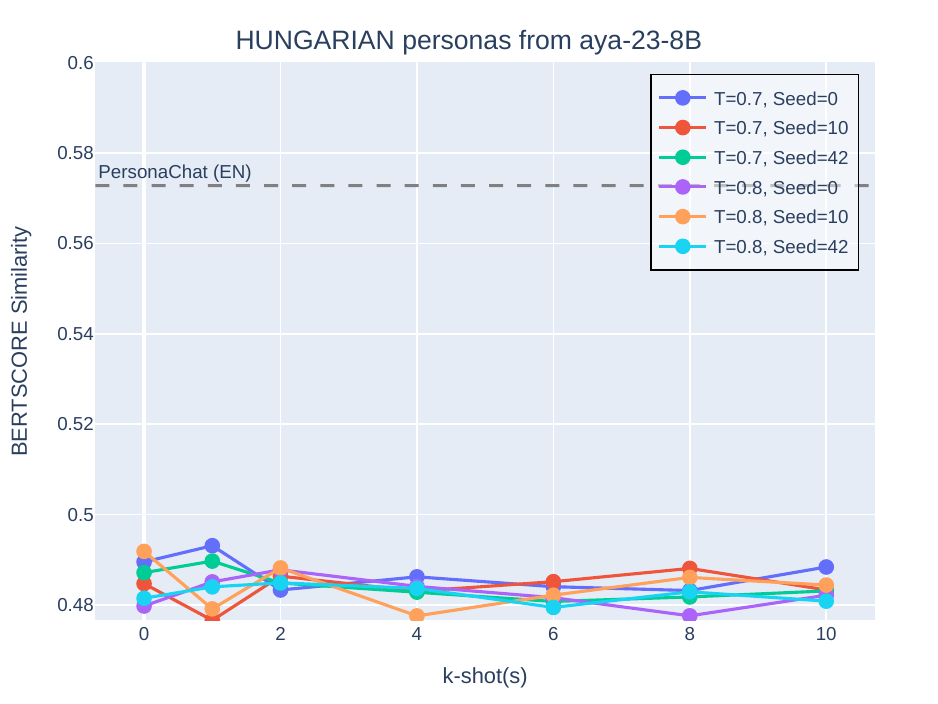}
        \end{subfigure}
        \vskip\baselineskip
        \begin{subfigure}{\textwidth}
            \centering
            \includegraphics[height=0.15\textheight]{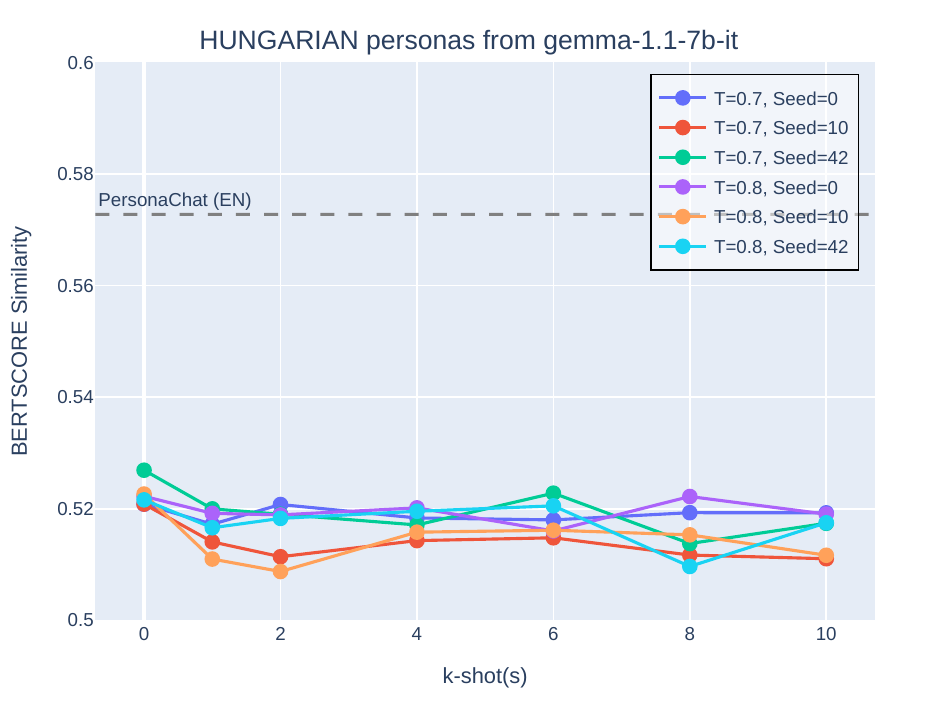}
        \end{subfigure}
    \end{minipage}
    %
    %
    \begin{minipage}{0.45\textwidth}
            \begin{subfigure}{\textwidth}
            \centering
            \includegraphics[height=0.15\textheight]{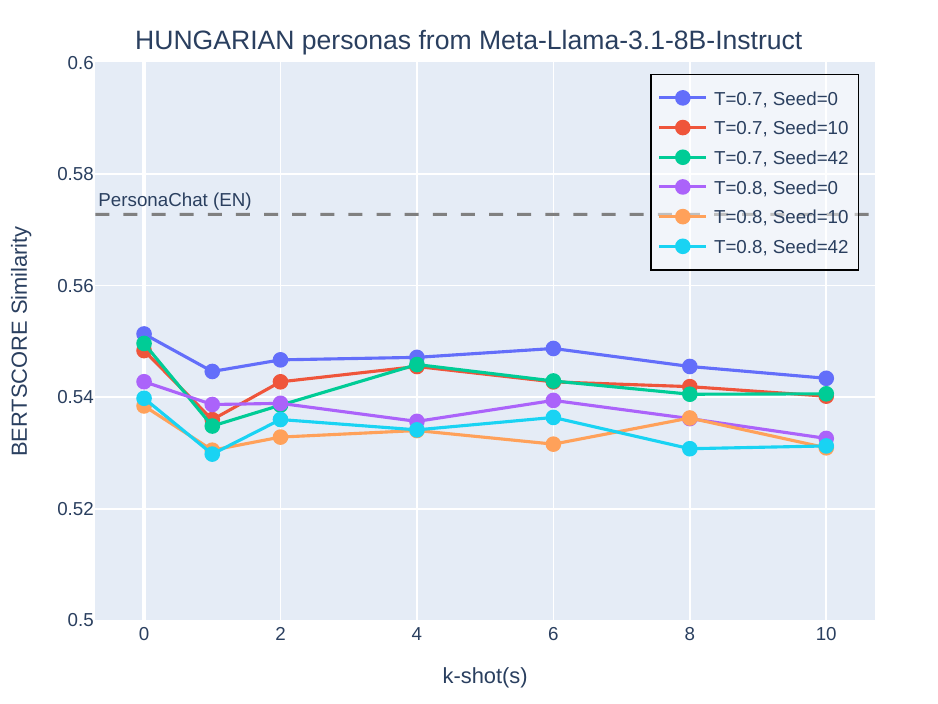}
        \end{subfigure}
        \vskip\baselineskip
        \begin{subfigure}{\textwidth}
            \centering
            \includegraphics[height=0.15\textheight]{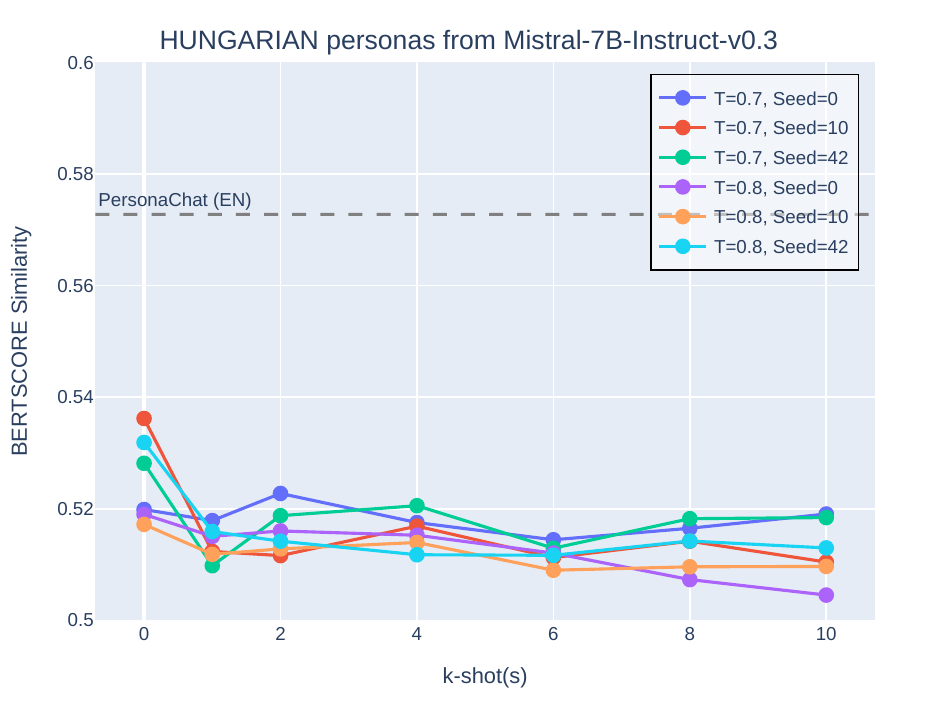}
        \end{subfigure}
    \end{minipage}
    
\caption{Detailed BERTSCORE for Hungarian Personas in different generation configurations for the different models}    
\end{figure}


\restoregeometry


\newgeometry{top=0.5cm, bottom=1.5cm, left=2.5cm, right=2.5cm}
\subsubsection{\textsc{Greek}}

\begin{figure}[h]
    \centering
    \begin{minipage}{0.45\textwidth}
        \begin{subfigure}{\textwidth}
            \centering
            \includegraphics[height=0.17\textheight]{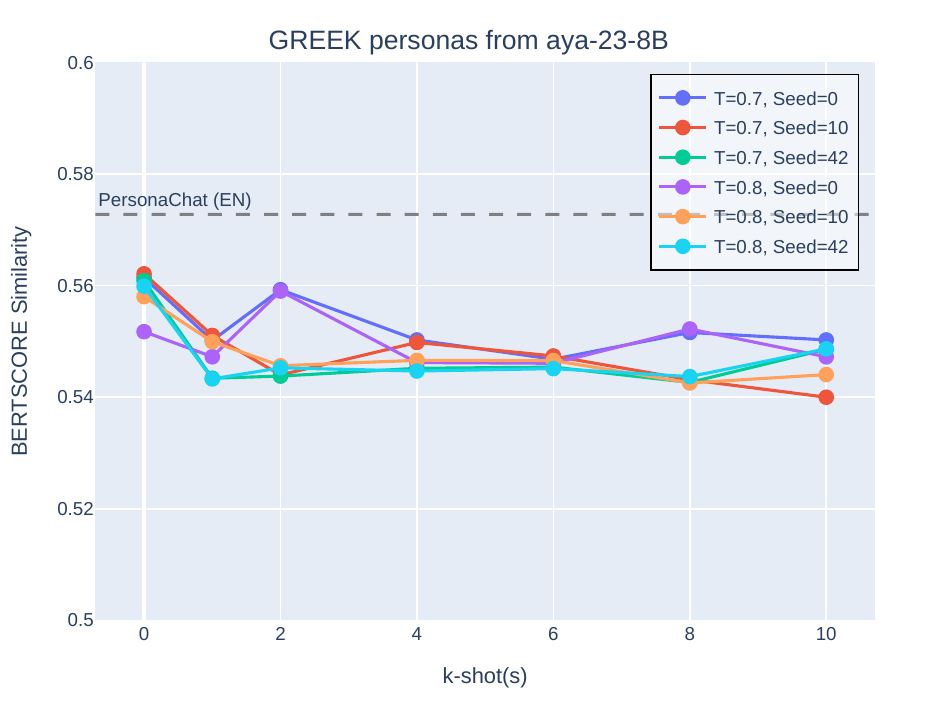}
        \end{subfigure}
        \vskip\baselineskip
        \begin{subfigure}{\textwidth}
            \centering
            \includegraphics[height=0.17\textheight]{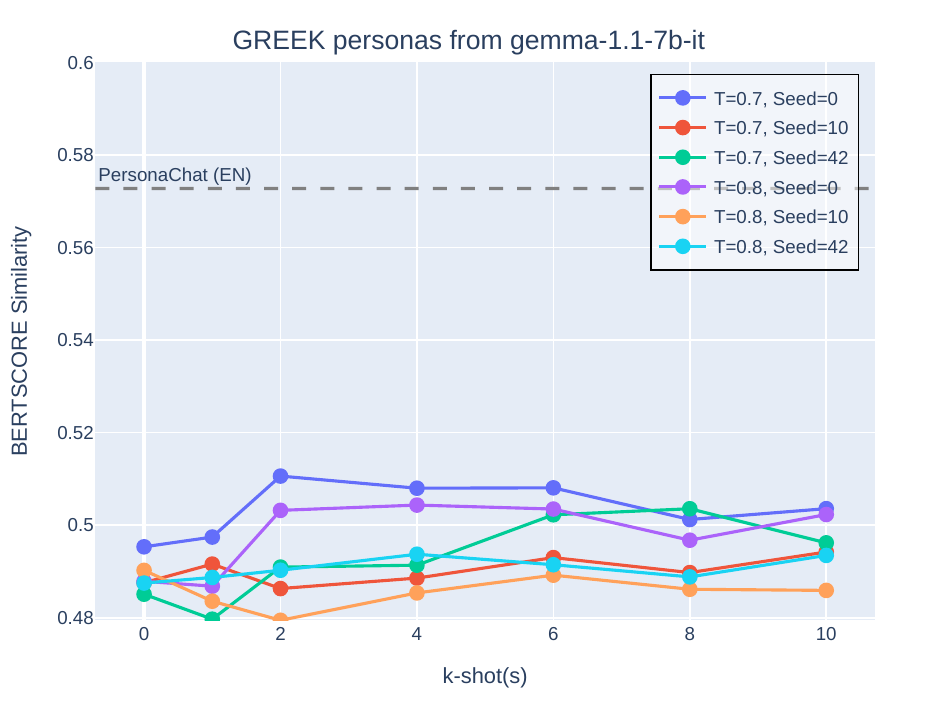}
        \end{subfigure}
        \vskip\baselineskip
        \begin{subfigure}{\textwidth}
            \centering
            \includegraphics[height=0.17\textheight]{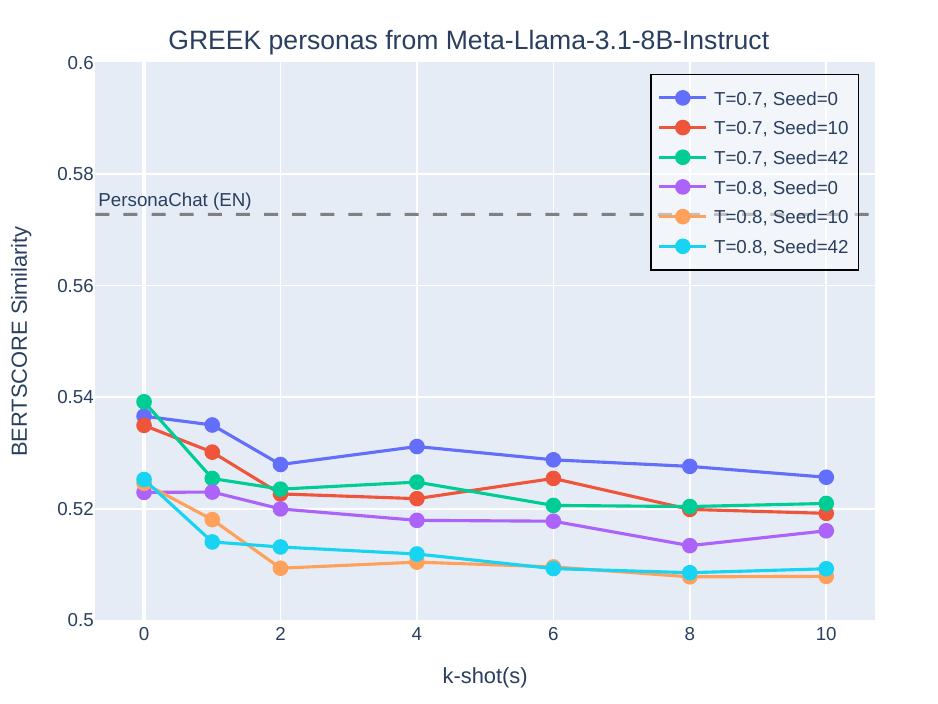}
        \end{subfigure}
        \vskip\baselineskip
        \begin{subfigure}{\textwidth}
            \centering
            \includegraphics[height=0.17\textheight]{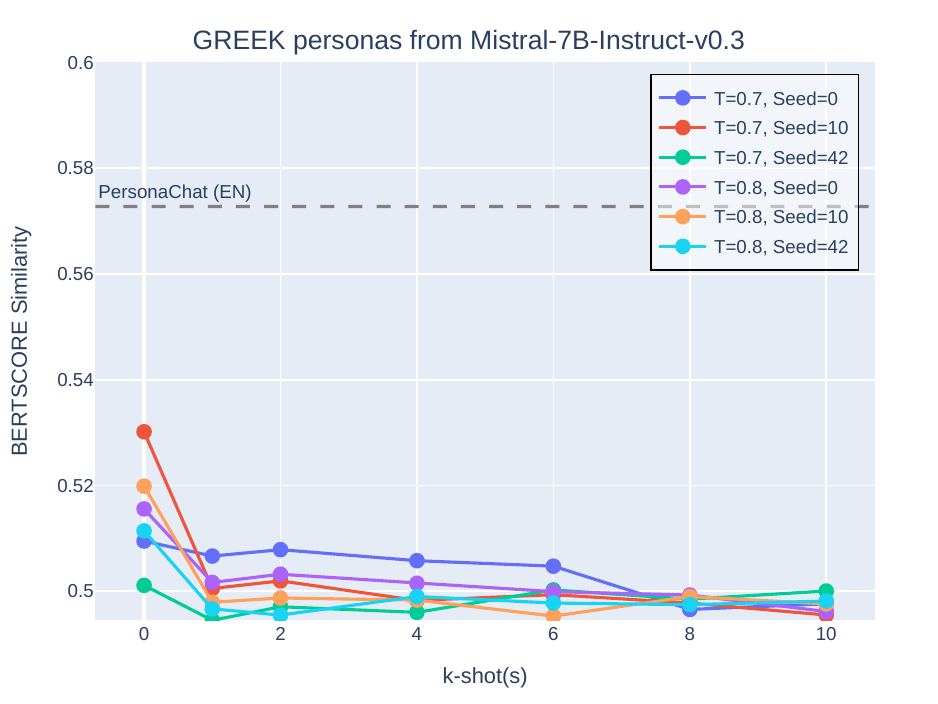}
        \end{subfigure}
    \end{minipage}
    %
    %
    \begin{minipage}{0.45\textwidth}
        \begin{subfigure}{\textwidth}
            \centering
            \includegraphics[height=0.17\textheight]{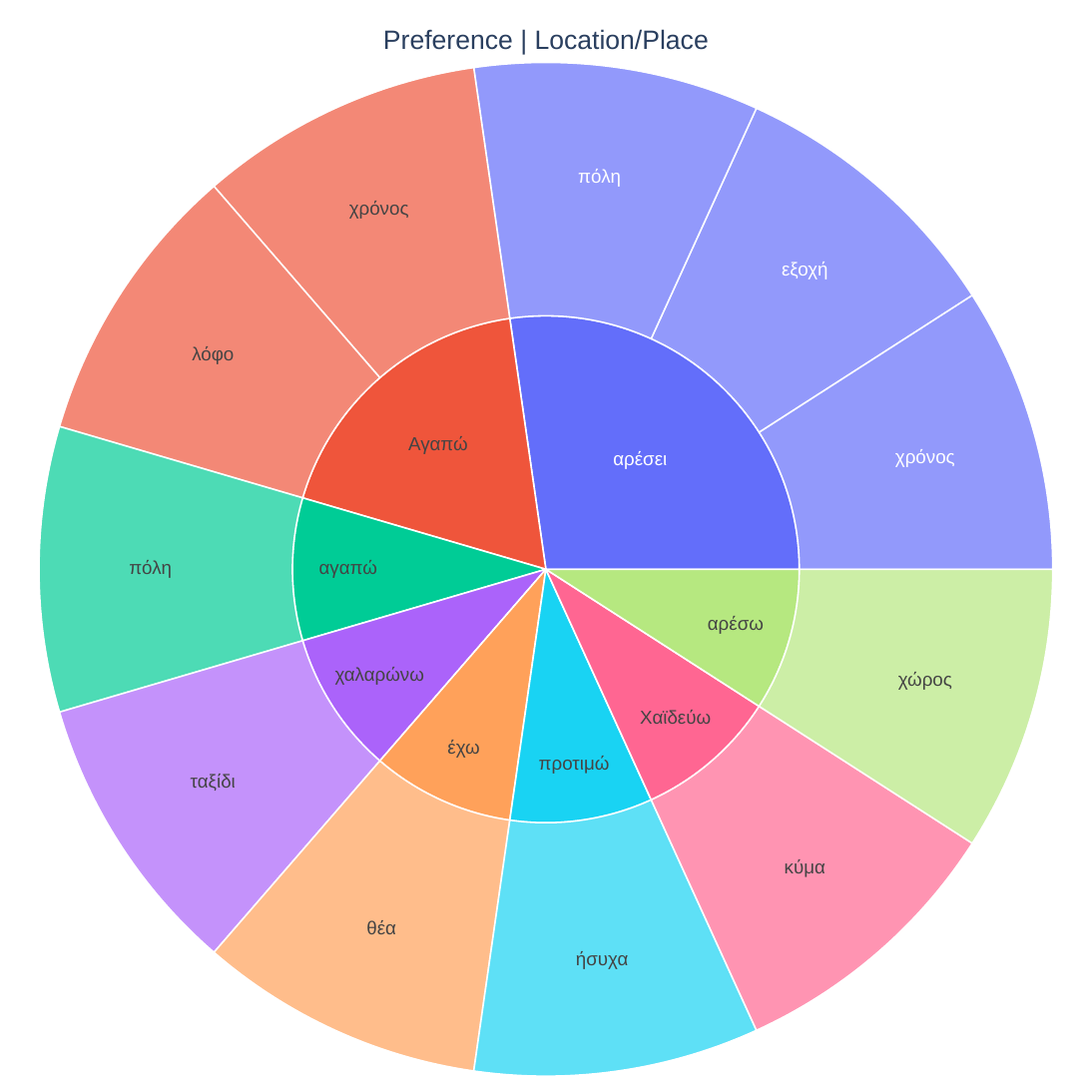}
        \end{subfigure}
        \vskip\baselineskip
        \begin{subfigure}{\textwidth}
            \centering
            \includegraphics[height=0.17\textheight]{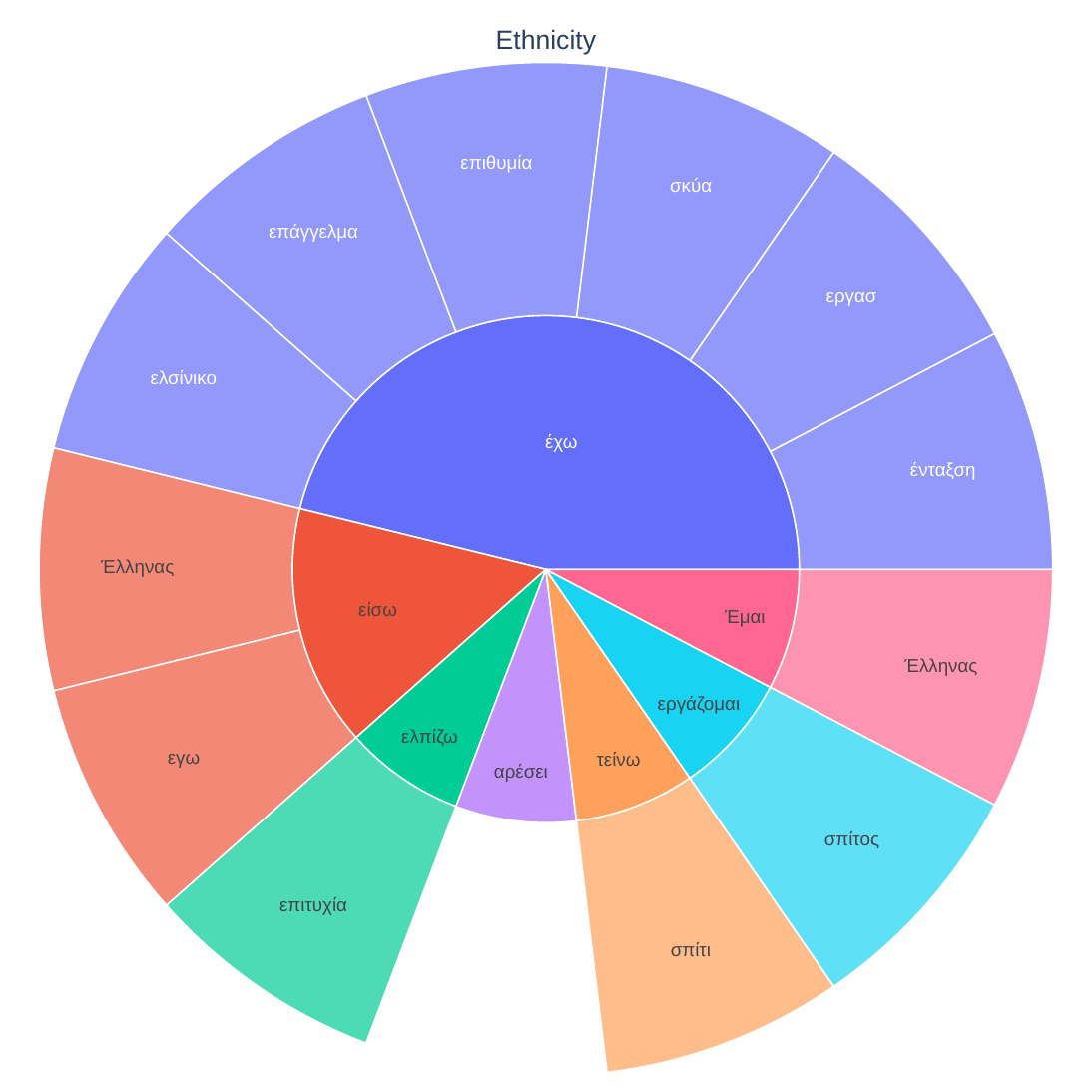}
        \end{subfigure}
        \vskip\baselineskip
        \begin{subfigure}{\textwidth}
            \centering
            \includegraphics[height=0.17\textheight]{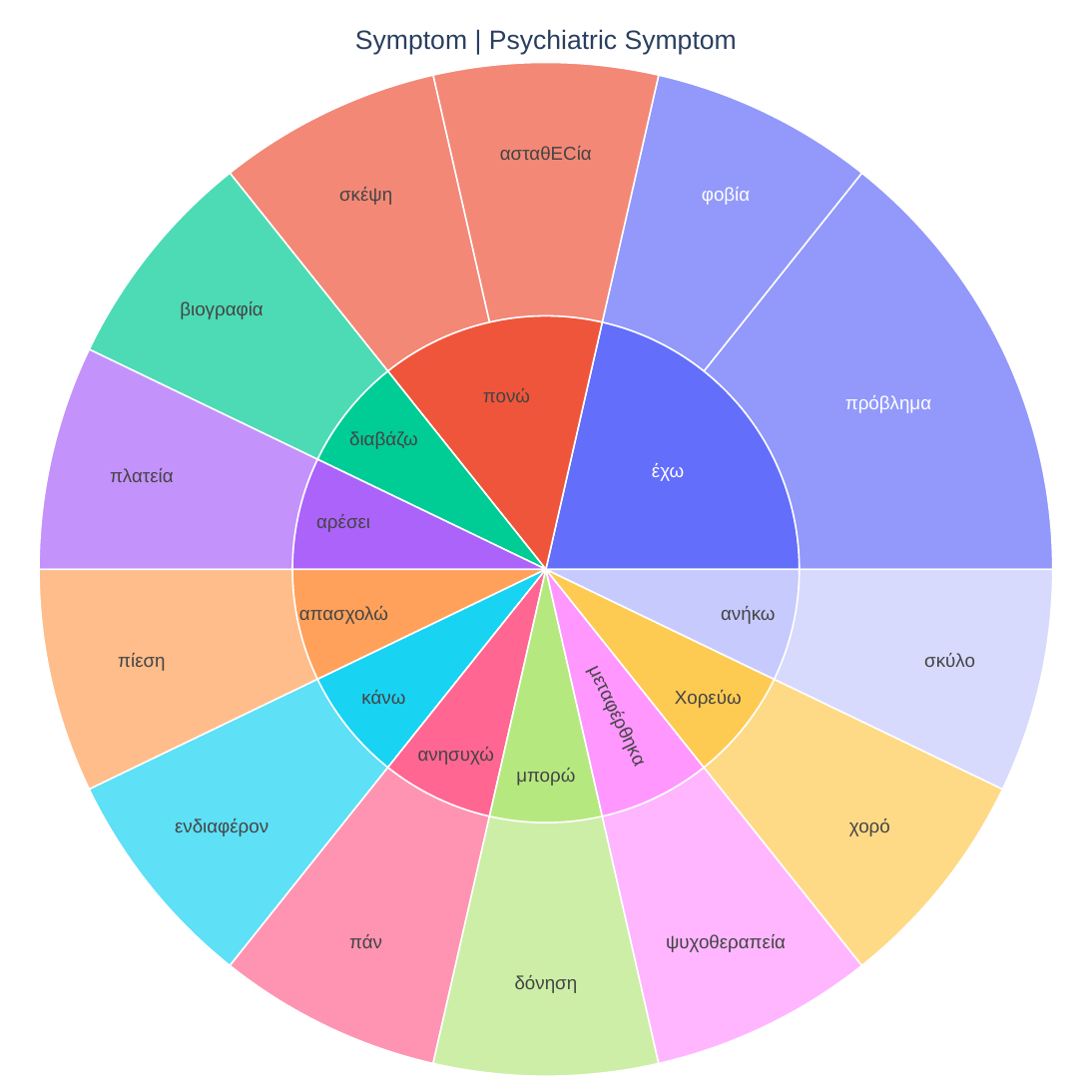}
        \end{subfigure}
        \vskip\baselineskip
        \begin{subfigure}{\textwidth}
            \centering
            \includegraphics[height=0.17\textheight]{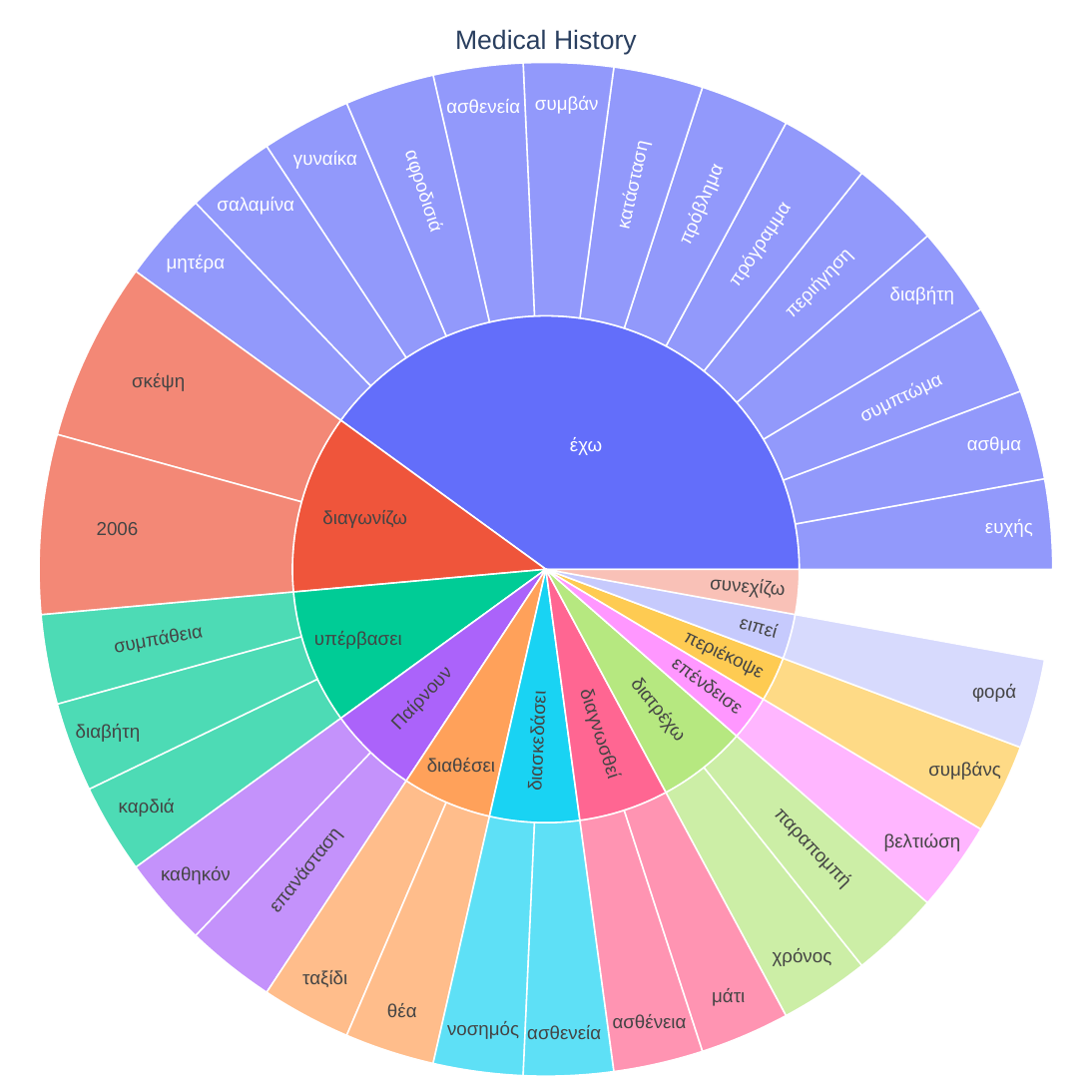}
        \end{subfigure}
    \end{minipage}
    
\caption{Detailed BERTSCORE for Greek Personas in different generation configurations and Sunburst charts of personas taxonomy entities with most root verbs and associated object noun for the different models}    
\end{figure}


\restoregeometry

\newgeometry{top=0.5cm, bottom=1.5cm, left=2.5cm, right=2.5cm}
\subsubsection{\textsc{Ukrainian}}

\begin{figure}[h]
    \centering
    \begin{minipage}{0.45\textwidth}
        \begin{subfigure}{\textwidth}
            \centering
            \includegraphics[height=0.17\textheight]{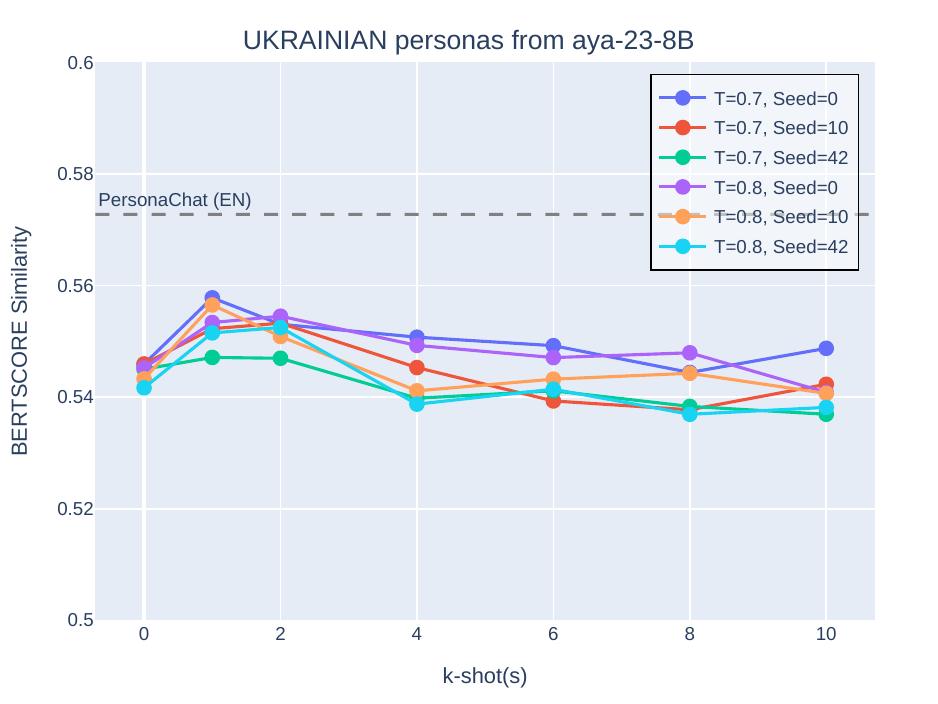}
        \end{subfigure}
        \vskip\baselineskip
        \begin{subfigure}{\textwidth}
            \centering
            \includegraphics[height=0.17\textheight]{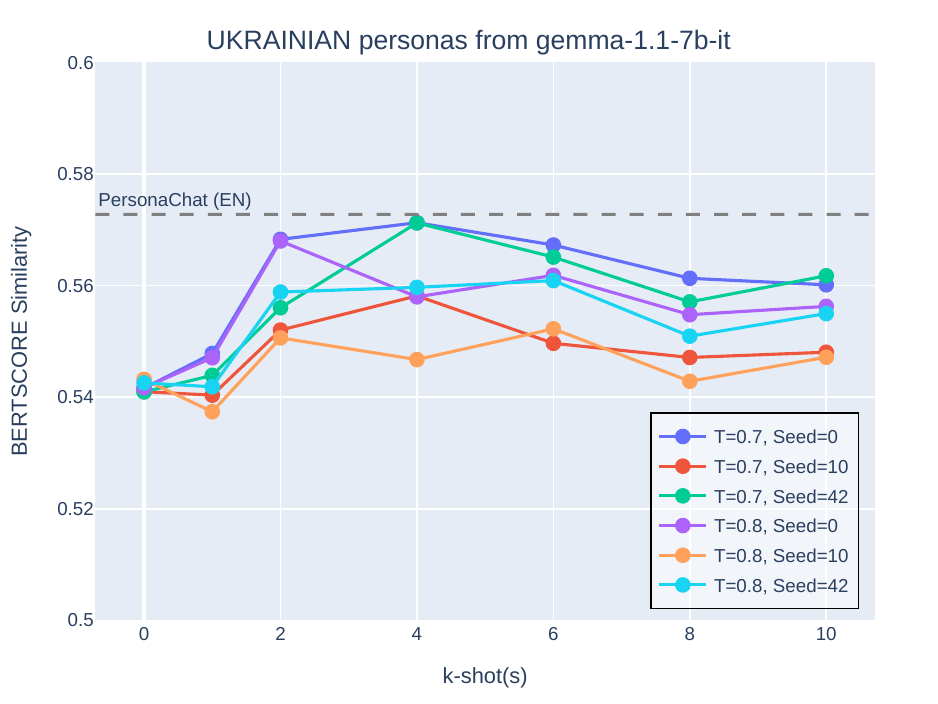}
        \end{subfigure}
        \vskip\baselineskip
        \begin{subfigure}{\textwidth}
            \centering
            \includegraphics[height=0.17\textheight]{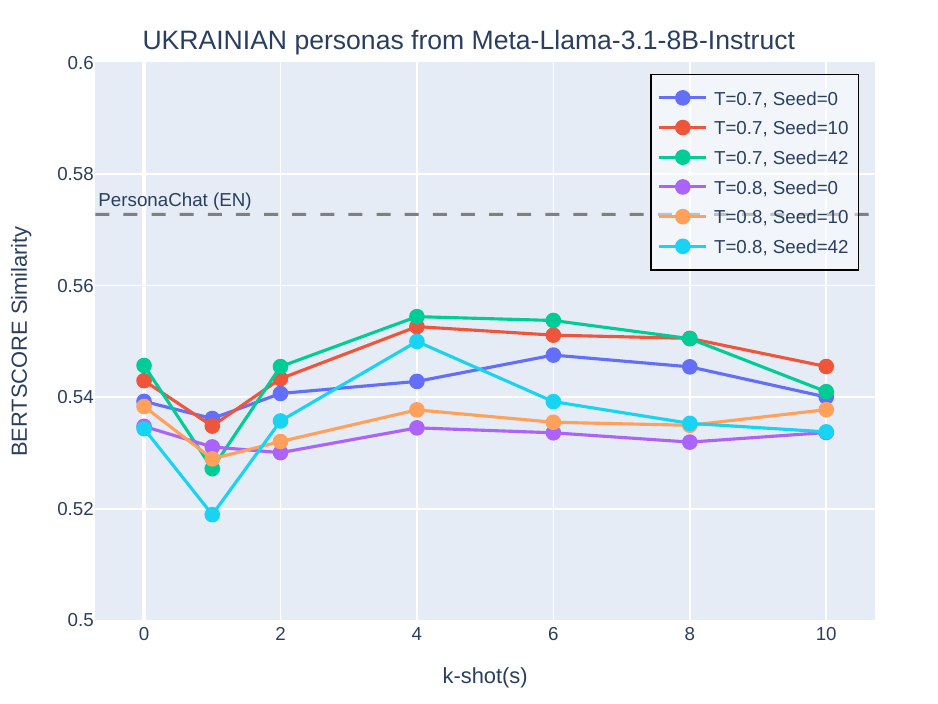}
        \end{subfigure}
        \vskip\baselineskip
        \begin{subfigure}{\textwidth}
            \centering
            \includegraphics[height=0.17\textheight]{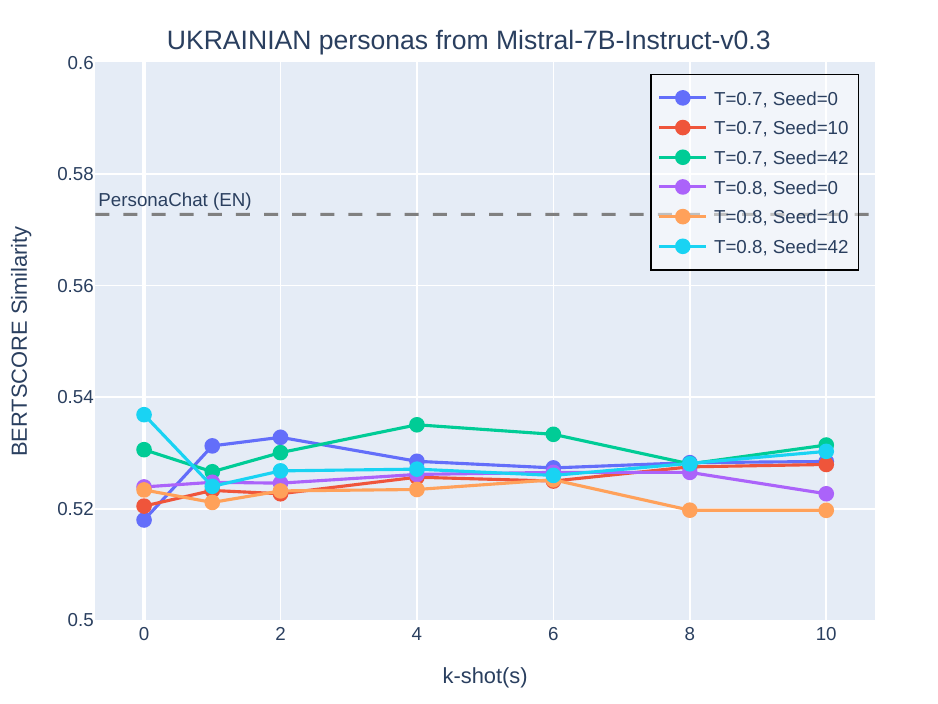}
        \end{subfigure}
    \end{minipage}
    %
    %
    \begin{minipage}{0.45\textwidth}
        \begin{subfigure}{\textwidth}
            \centering
            \includegraphics[height=0.17\textheight]{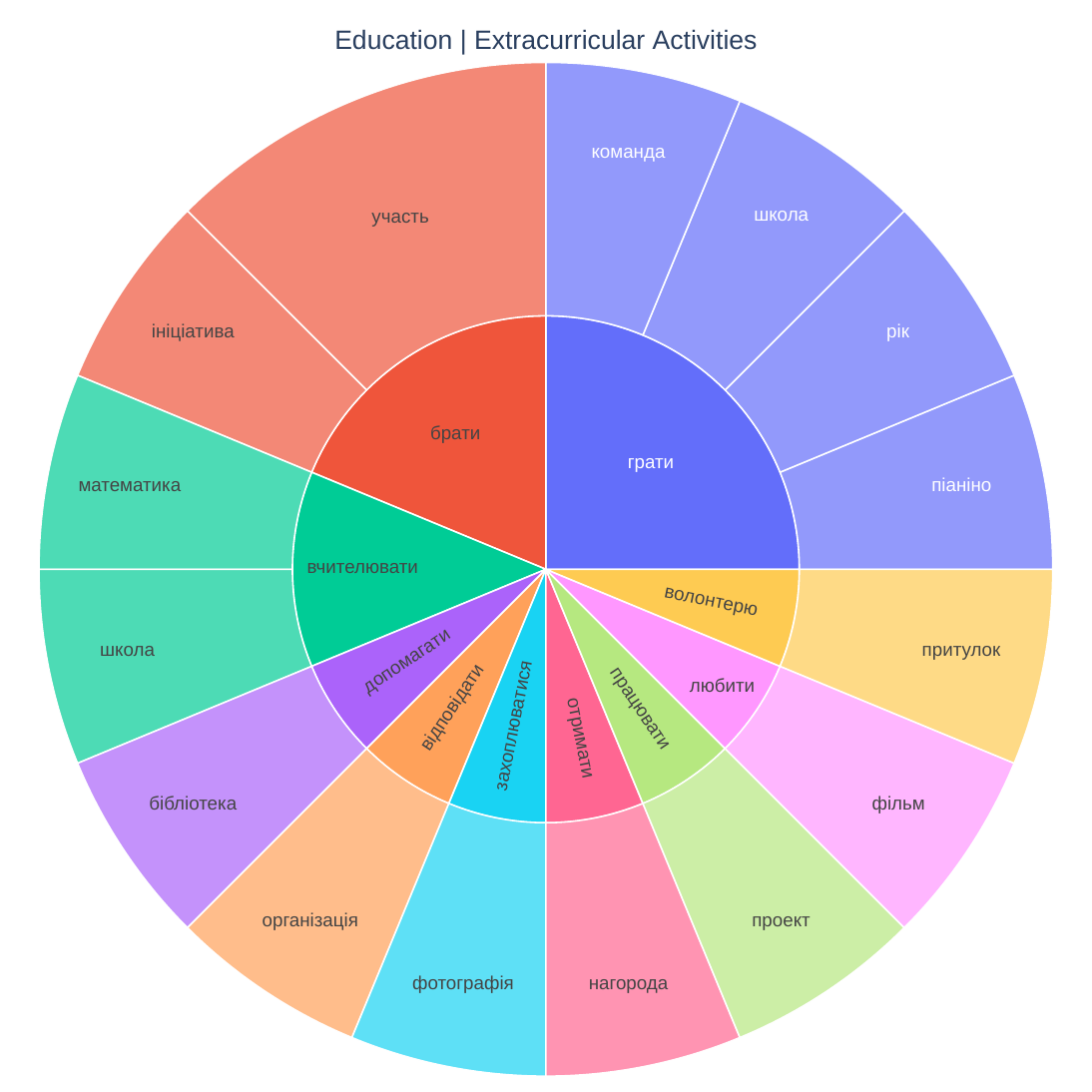}
        \end{subfigure}
        \vskip\baselineskip
        \begin{subfigure}{\textwidth}
            \centering
            \includegraphics[height=0.17\textheight]{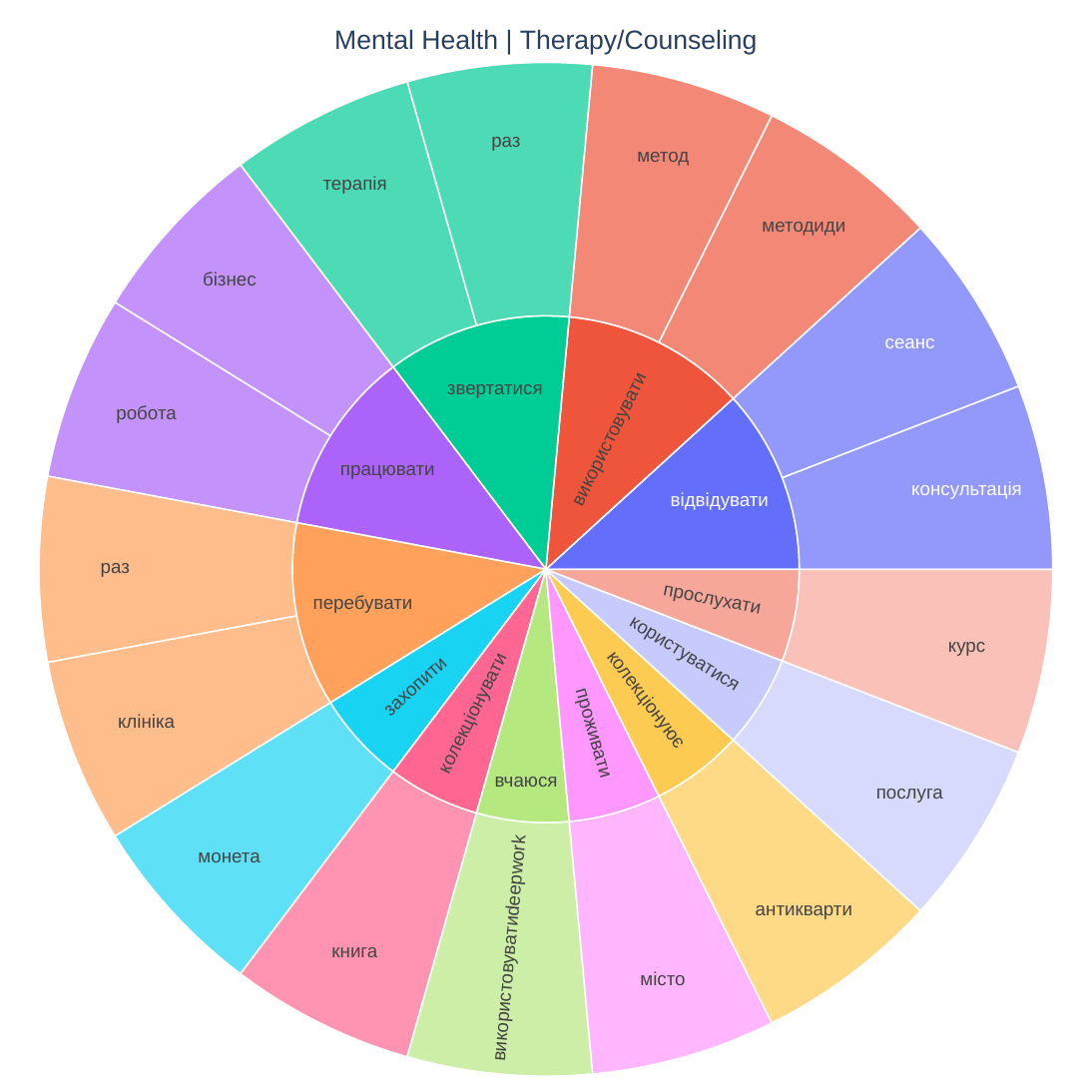}
        \end{subfigure}
        \vskip\baselineskip
        \begin{subfigure}{\textwidth}
            \centering
            \includegraphics[height=0.17\textheight]{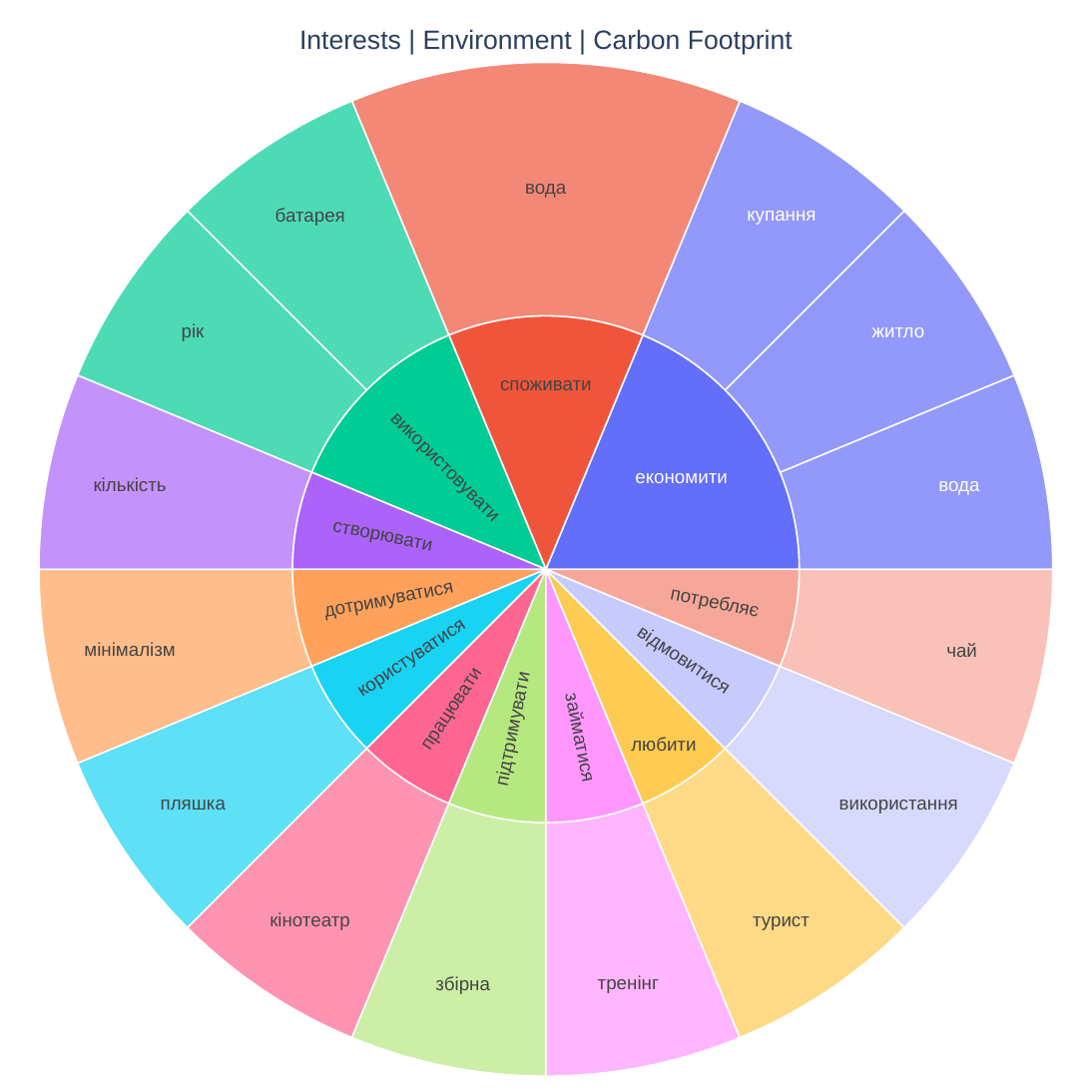}
        \end{subfigure}
        \vskip\baselineskip
        \begin{subfigure}{\textwidth}
            \centering
            \includegraphics[height=0.17\textheight]{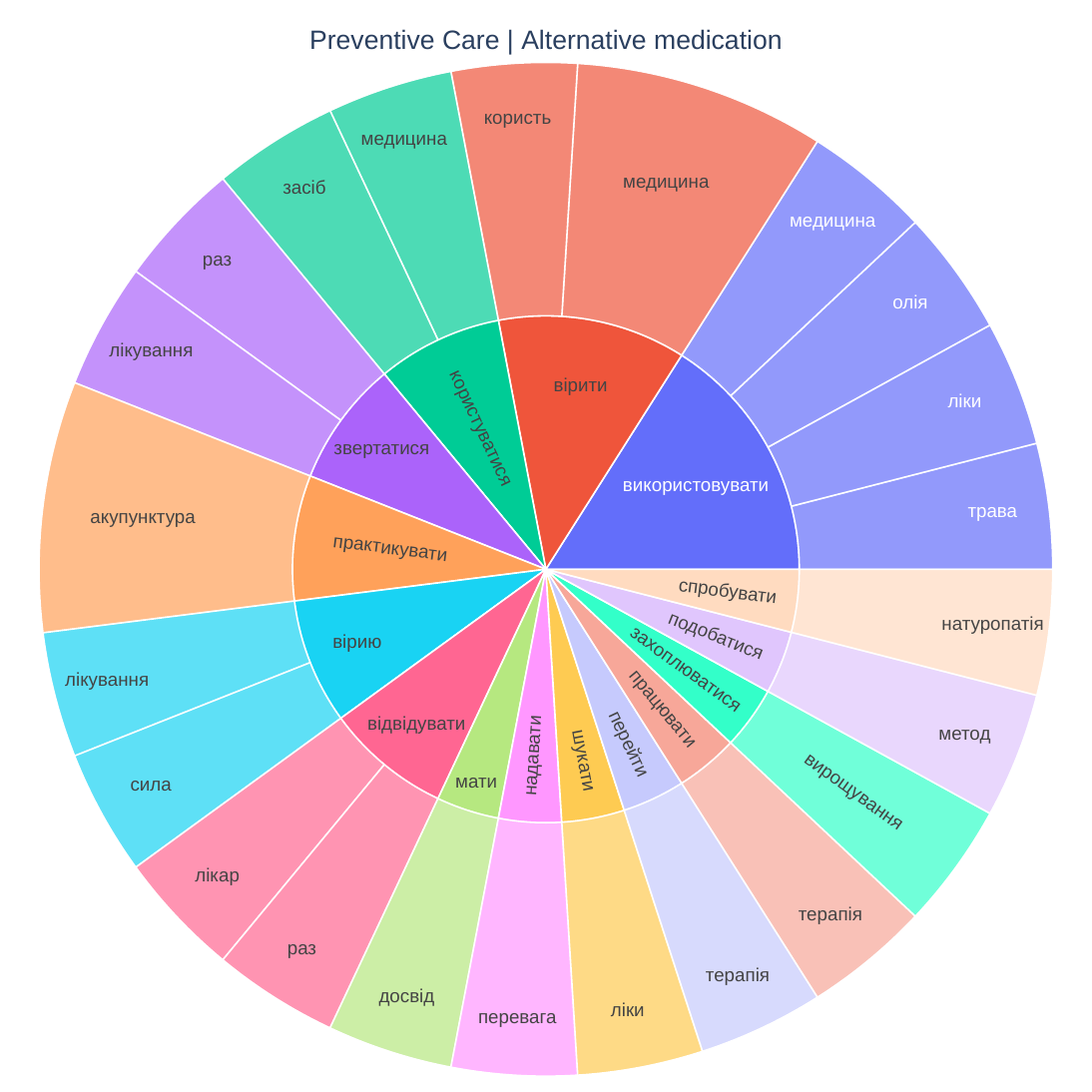}
        \end{subfigure}
    \end{minipage}
    
\caption{Detailed BERTSCORE for Ukrainian Personas in different generation configurations and Sunburst charts of personas taxonomy entities with most root verbs and associated object noun for the different models}    
\end{figure}


\restoregeometry

\newgeometry{top=0.5cm, bottom=1.5cm, left=2.5cm, right=2.5cm}
\subsubsection{\textsc{Danish}}

\begin{figure}[h]
    \centering
    \begin{minipage}{0.45\textwidth}
        \begin{subfigure}{\textwidth}
            \centering
            \includegraphics[height=0.17\textheight]{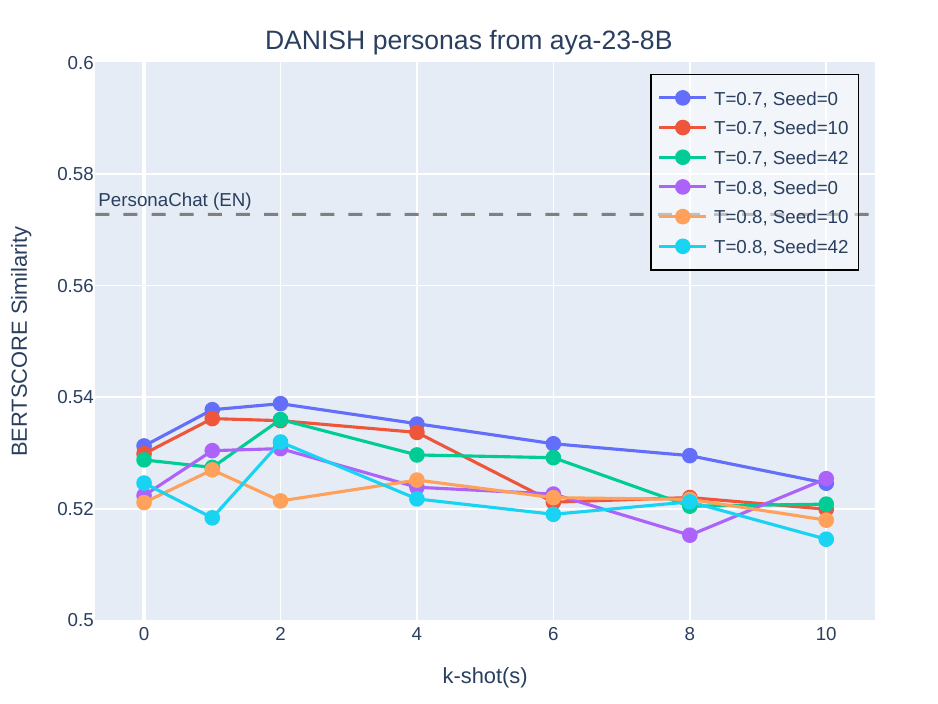}
        \end{subfigure}
        \vskip\baselineskip
        \begin{subfigure}{\textwidth}
            \centering
            \includegraphics[height=0.17\textheight]{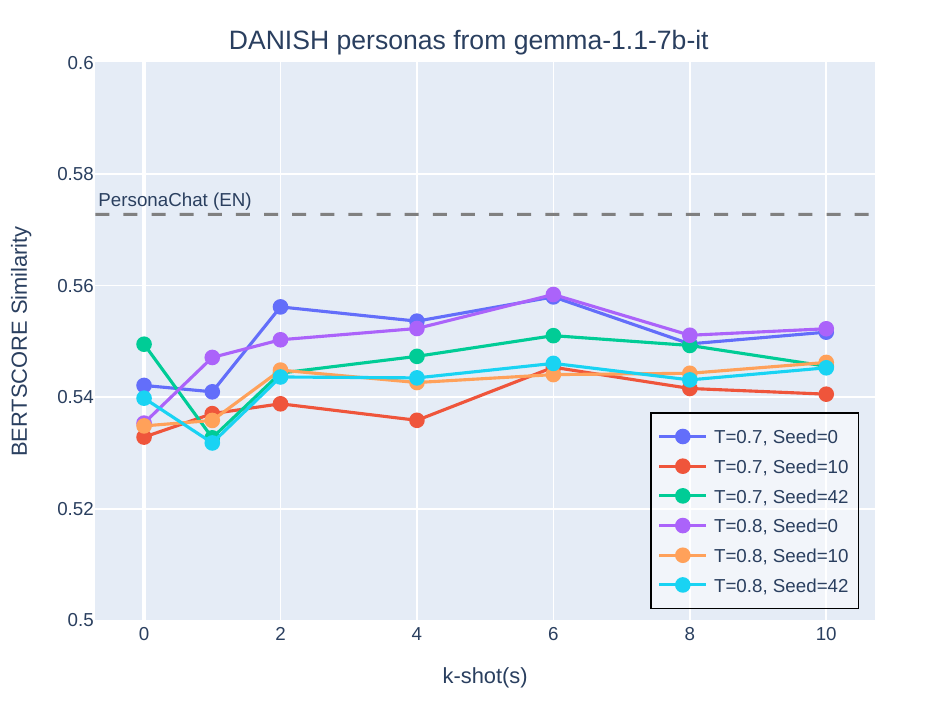}
        \end{subfigure}
        \vskip\baselineskip
        \begin{subfigure}{\textwidth}
            \centering
            \includegraphics[height=0.17\textheight]{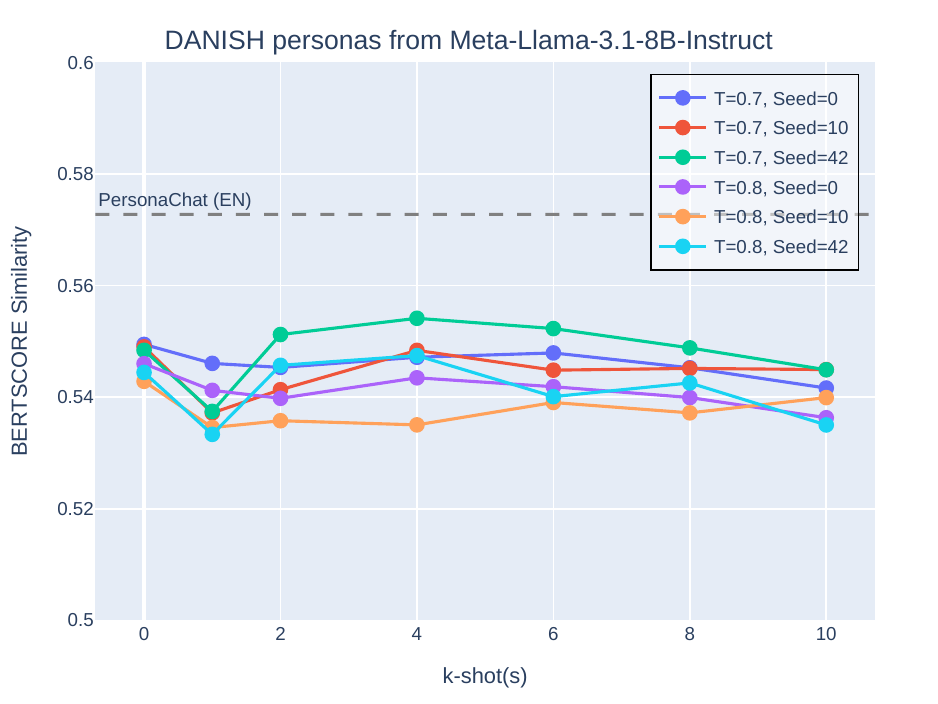}
        \end{subfigure}
        \vskip\baselineskip
        \begin{subfigure}{\textwidth}
            \centering
            \includegraphics[height=0.17\textheight]{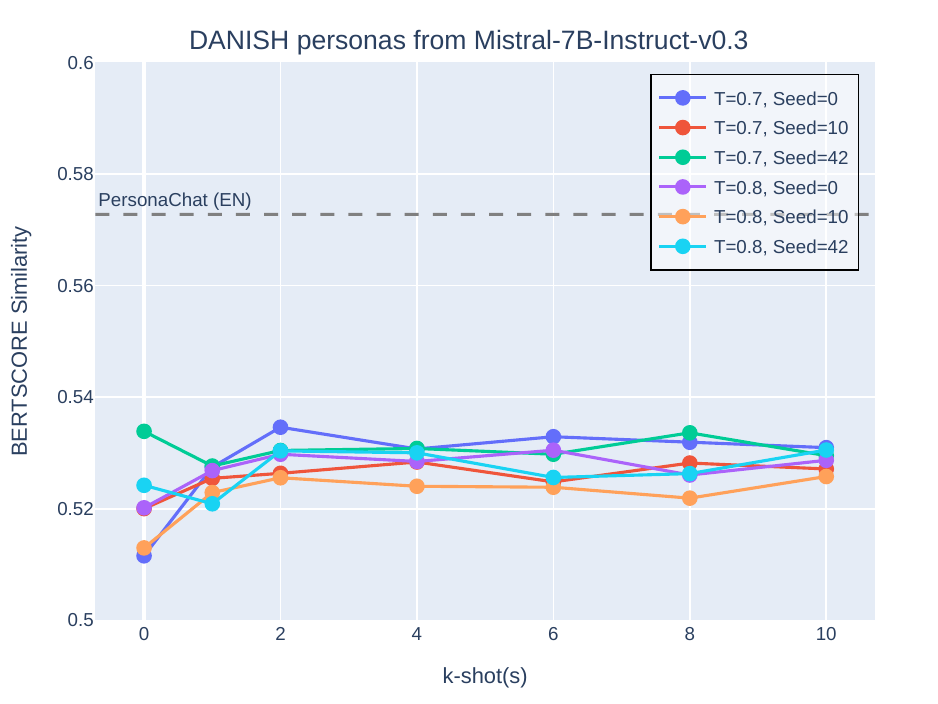}
        \end{subfigure}
    \end{minipage}
    %
    %
    \begin{minipage}{0.45\textwidth}
        \begin{subfigure}{\textwidth}
            \centering
            \includegraphics[height=0.17\textheight]{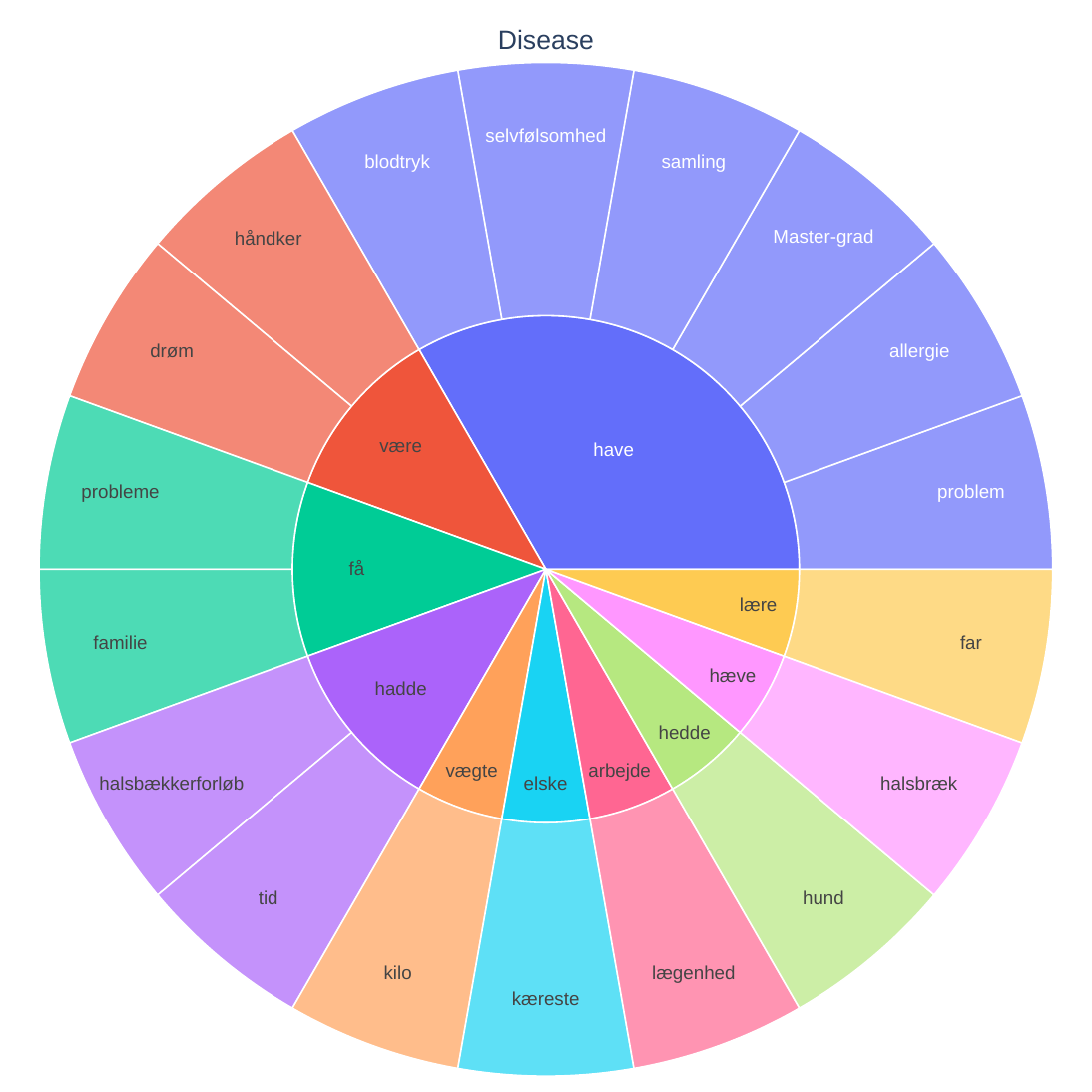}
        \end{subfigure}
        \vskip\baselineskip
        \begin{subfigure}{\textwidth}
            \centering
            \includegraphics[height=0.17\textheight]{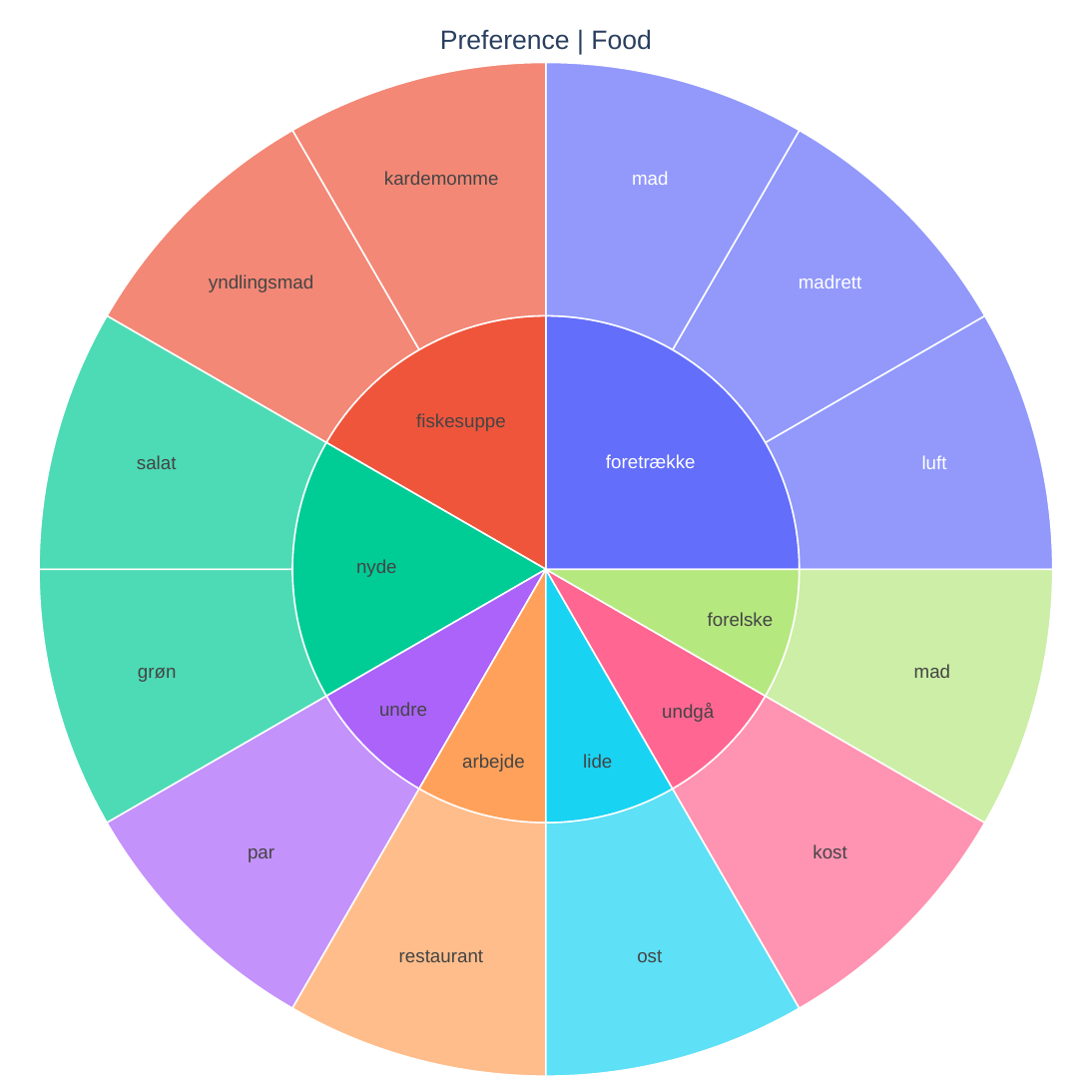}
        \end{subfigure}
        \vskip\baselineskip
        \begin{subfigure}{\textwidth}
            \centering
            \includegraphics[height=0.17\textheight]{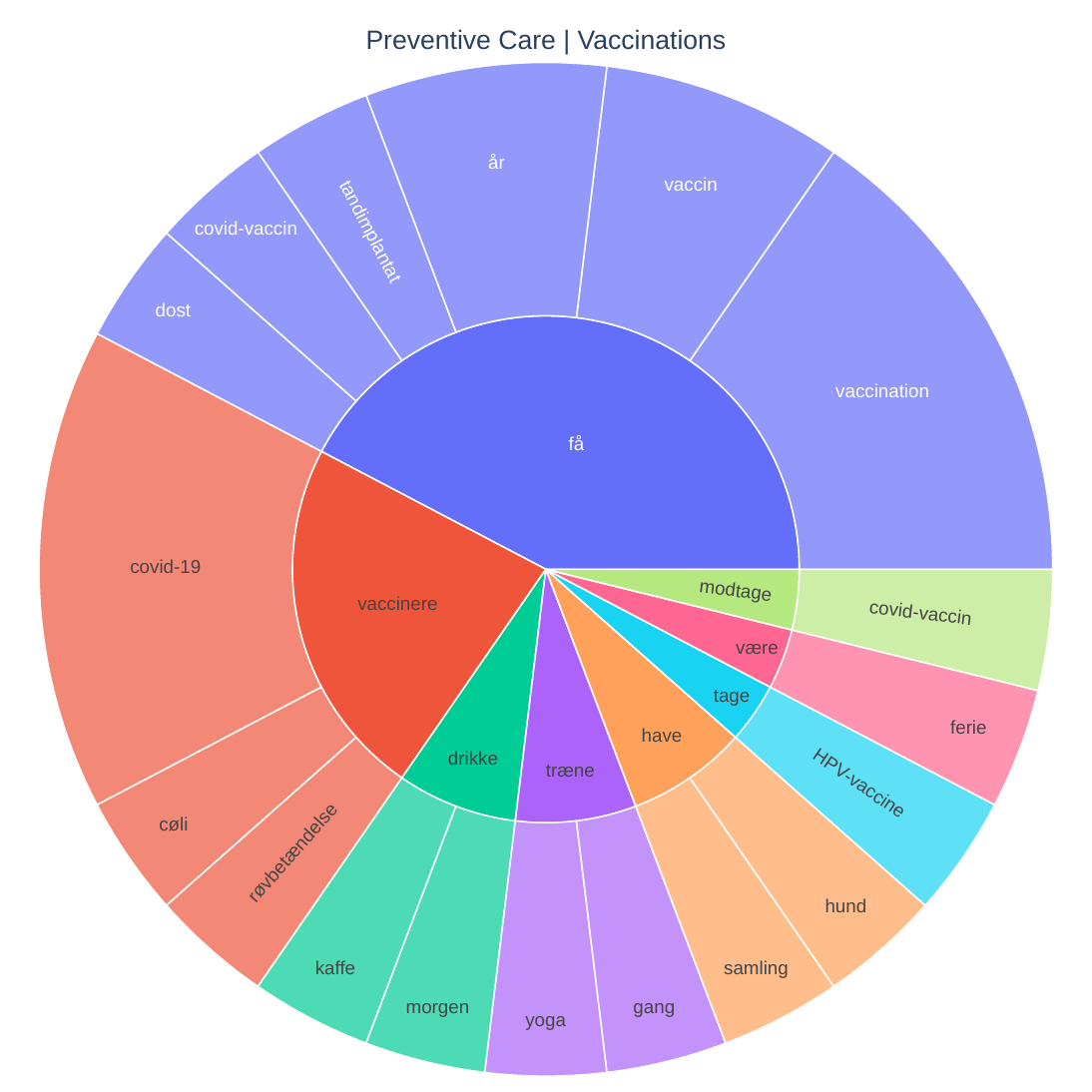}
        \end{subfigure}
        \vskip\baselineskip
        \begin{subfigure}{\textwidth}
            \centering
            \includegraphics[height=0.17\textheight]{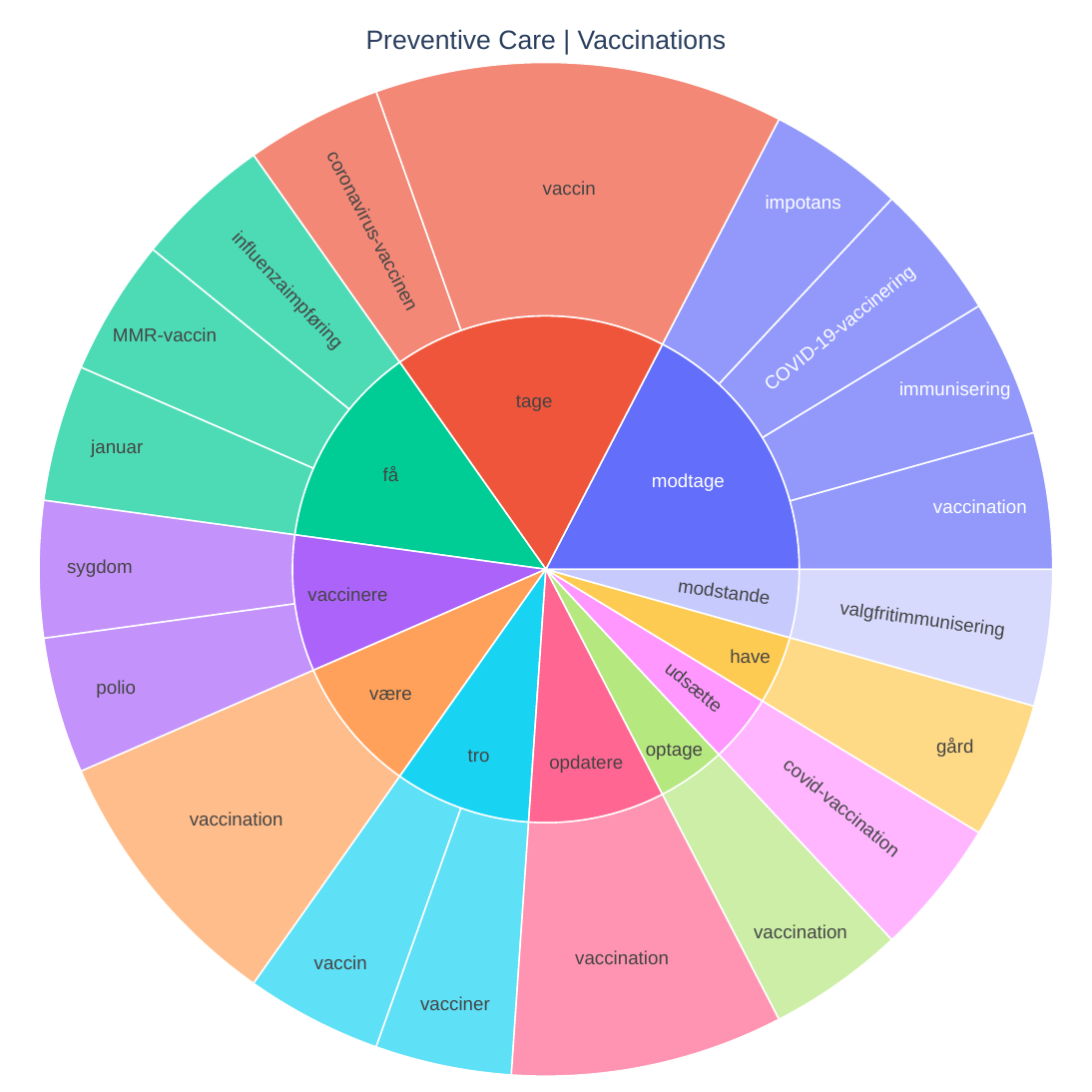}
        \end{subfigure}
    \end{minipage}
    
\caption{Detailed BERTSCORE for Danish Personas in different generation configurations and Sunburst charts of personas taxonomy entities with most root verbs and associated object noun for the different models}    
\end{figure}


\restoregeometry

\newgeometry{top=0.5cm, bottom=1.5cm, left=2.5cm, right=2.5cm}
\subsubsection{\textsc{Finnish}}

\begin{figure}[h]
    \centering
    \begin{minipage}{0.45\textwidth}
        \begin{subfigure}{\textwidth}
            \centering
            \includegraphics[height=0.17\textheight]{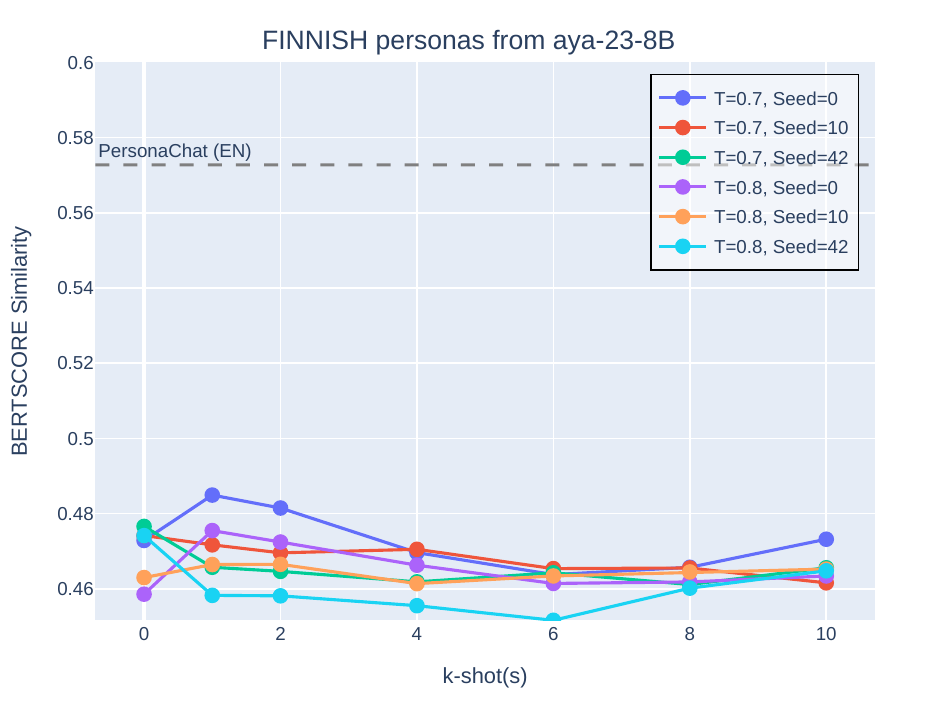}
        \end{subfigure}
        \vskip\baselineskip
        \begin{subfigure}{\textwidth}
            \centering
            \includegraphics[height=0.17\textheight]{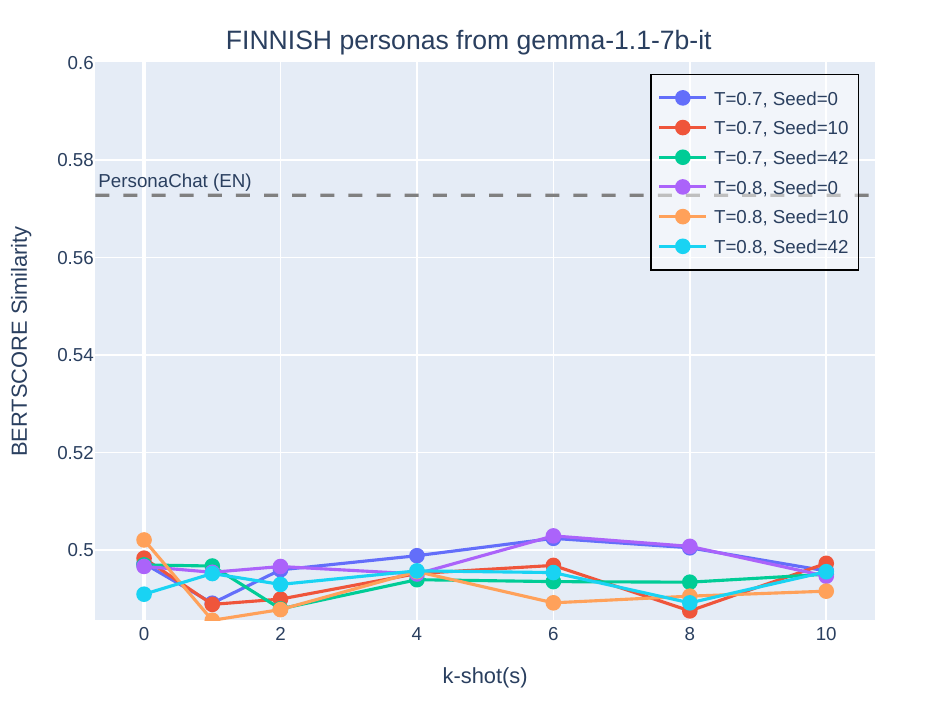}
        \end{subfigure}
        \vskip\baselineskip
        \begin{subfigure}{\textwidth}
            \centering
            \includegraphics[height=0.17\textheight]{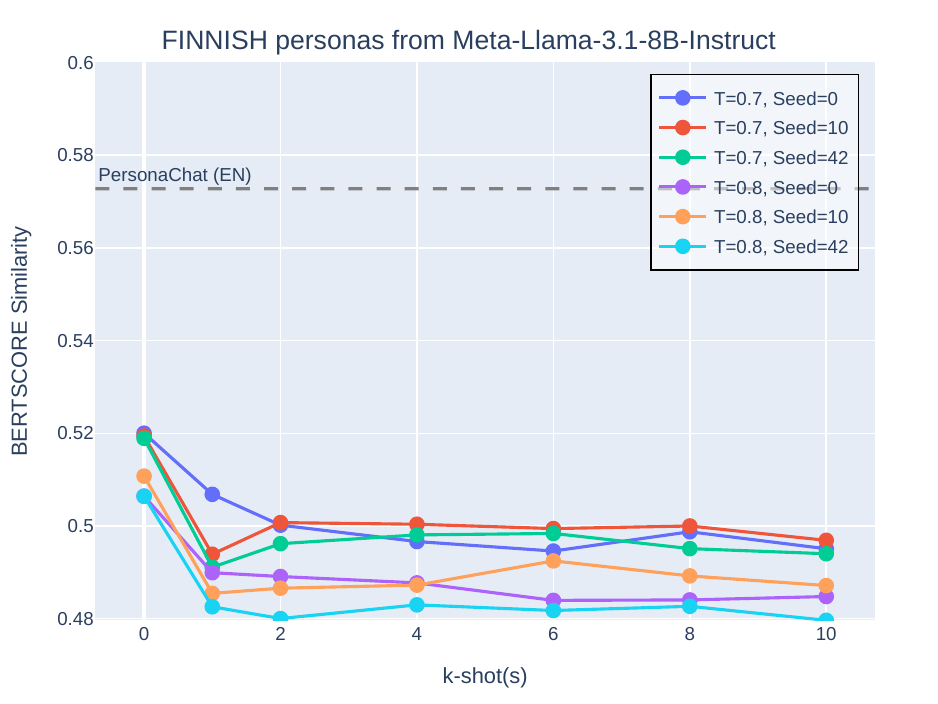}
        \end{subfigure}
        \vskip\baselineskip
        \begin{subfigure}{\textwidth}
            \centering
            \includegraphics[height=0.17\textheight]{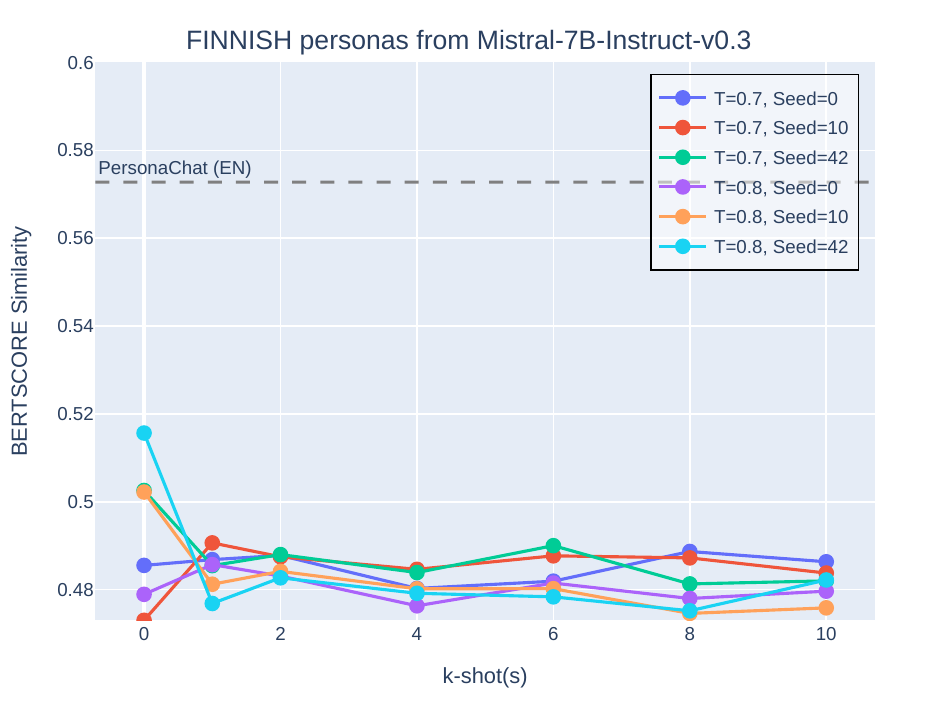}
        \end{subfigure}
    \end{minipage}
    %
    %
    \begin{minipage}{0.45\textwidth}
        \begin{subfigure}{\textwidth}
            \centering
            \includegraphics[height=0.17\textheight]{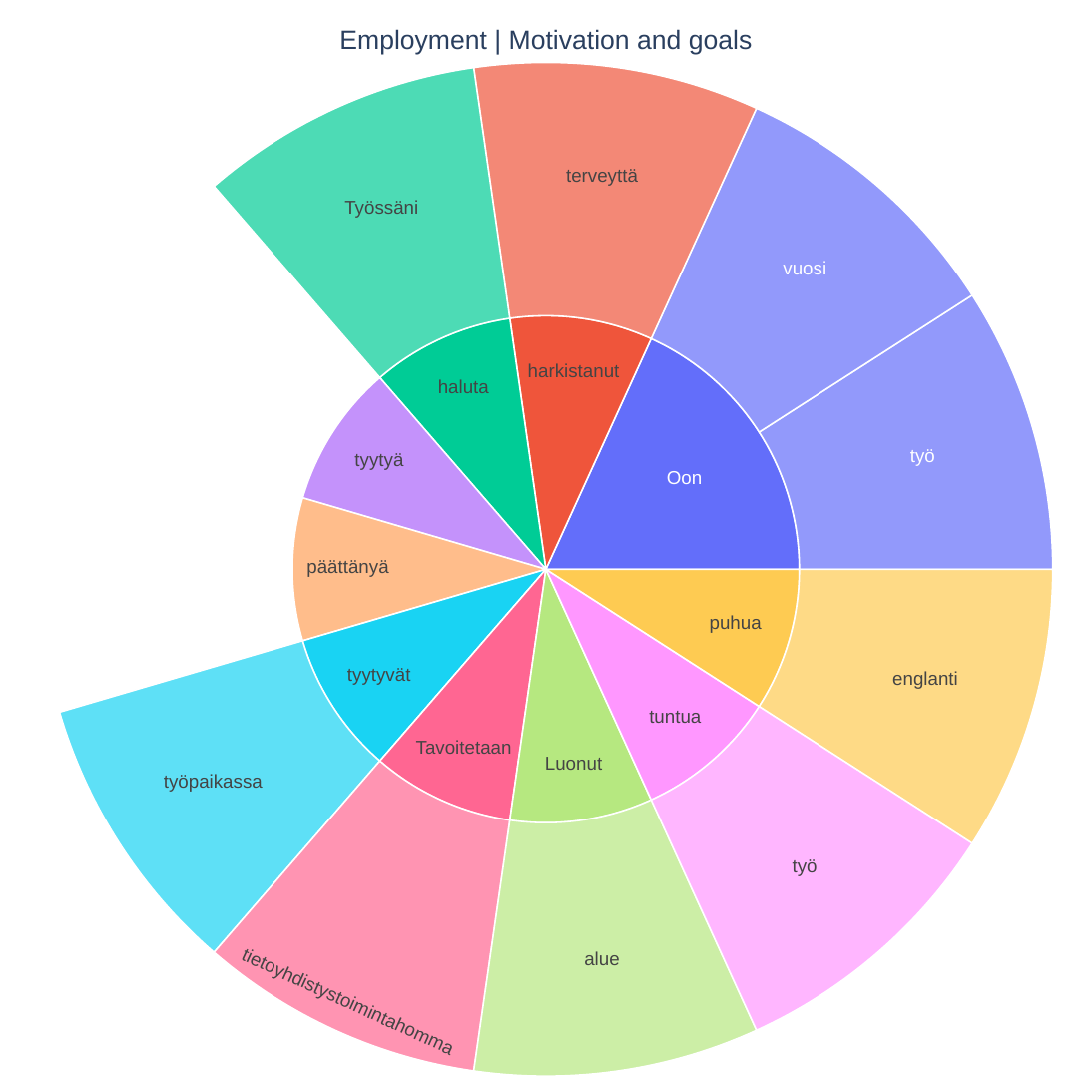}
        \end{subfigure}
        \vskip\baselineskip
        \begin{subfigure}{\textwidth}
            \centering
            \includegraphics[height=0.17\textheight]{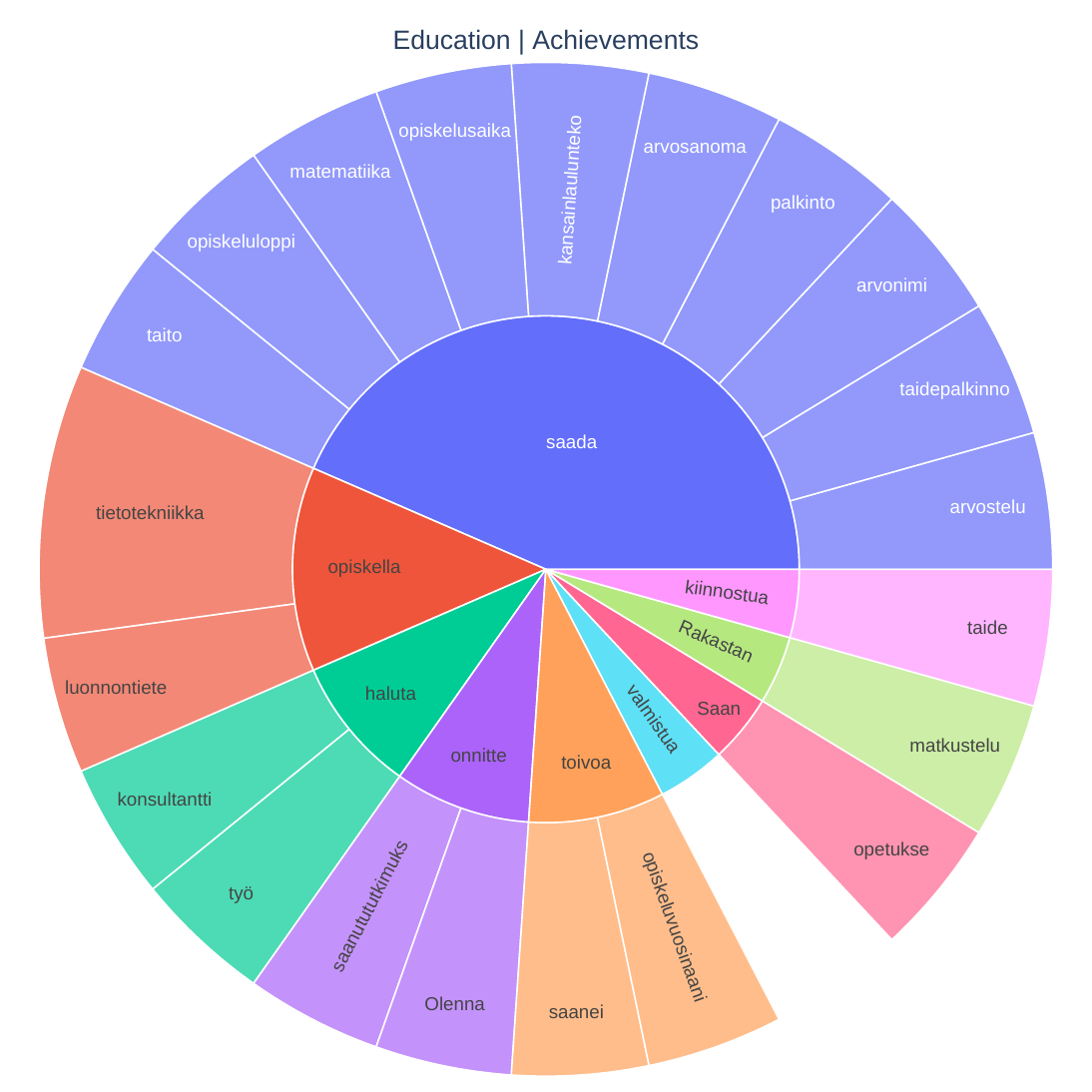}
        \end{subfigure}
        \vskip\baselineskip
        \begin{subfigure}{\textwidth}
            \centering
            \includegraphics[height=0.17\textheight]{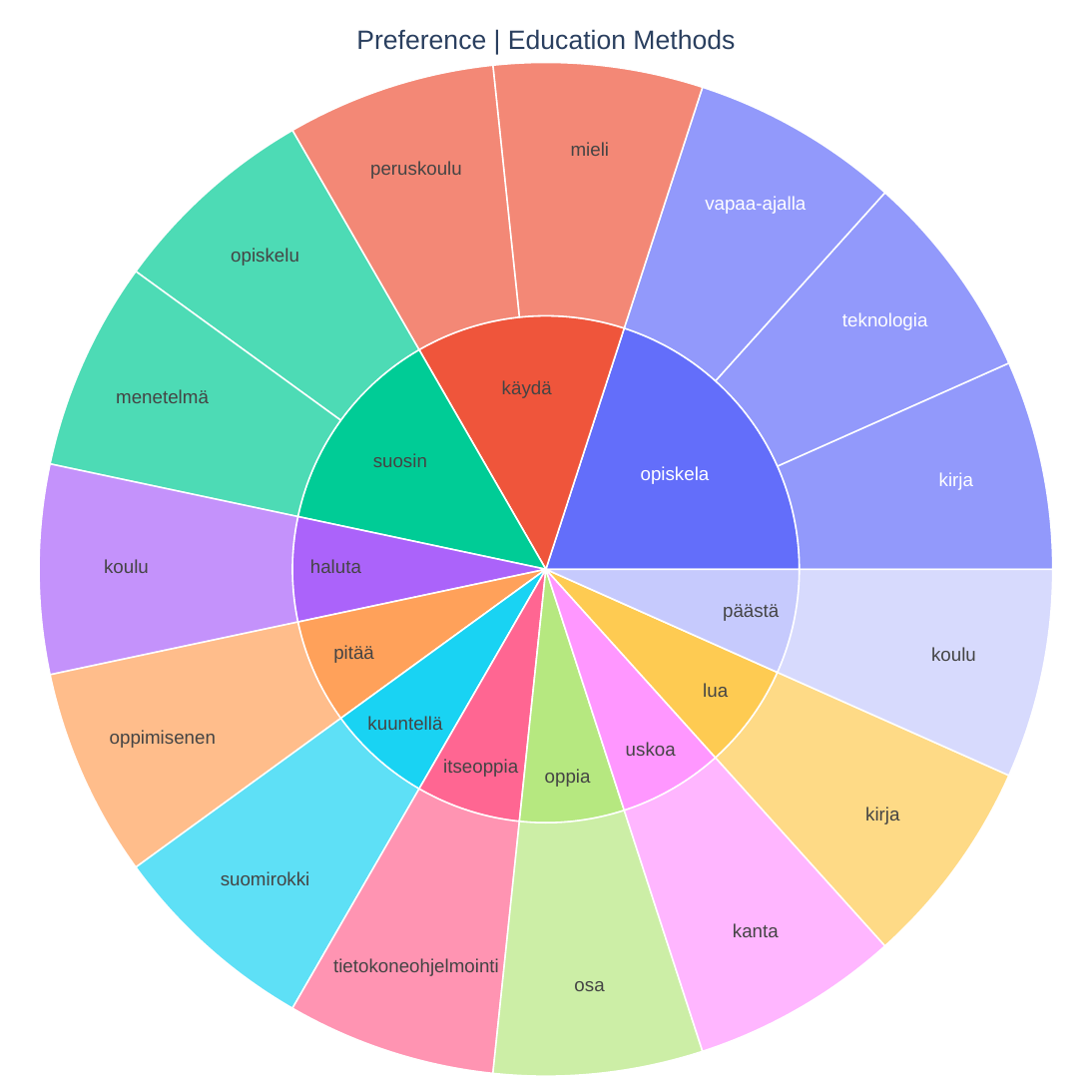}
        \end{subfigure}
        \vskip\baselineskip
        \begin{subfigure}{\textwidth}
            \centering
            \includegraphics[height=0.17\textheight]{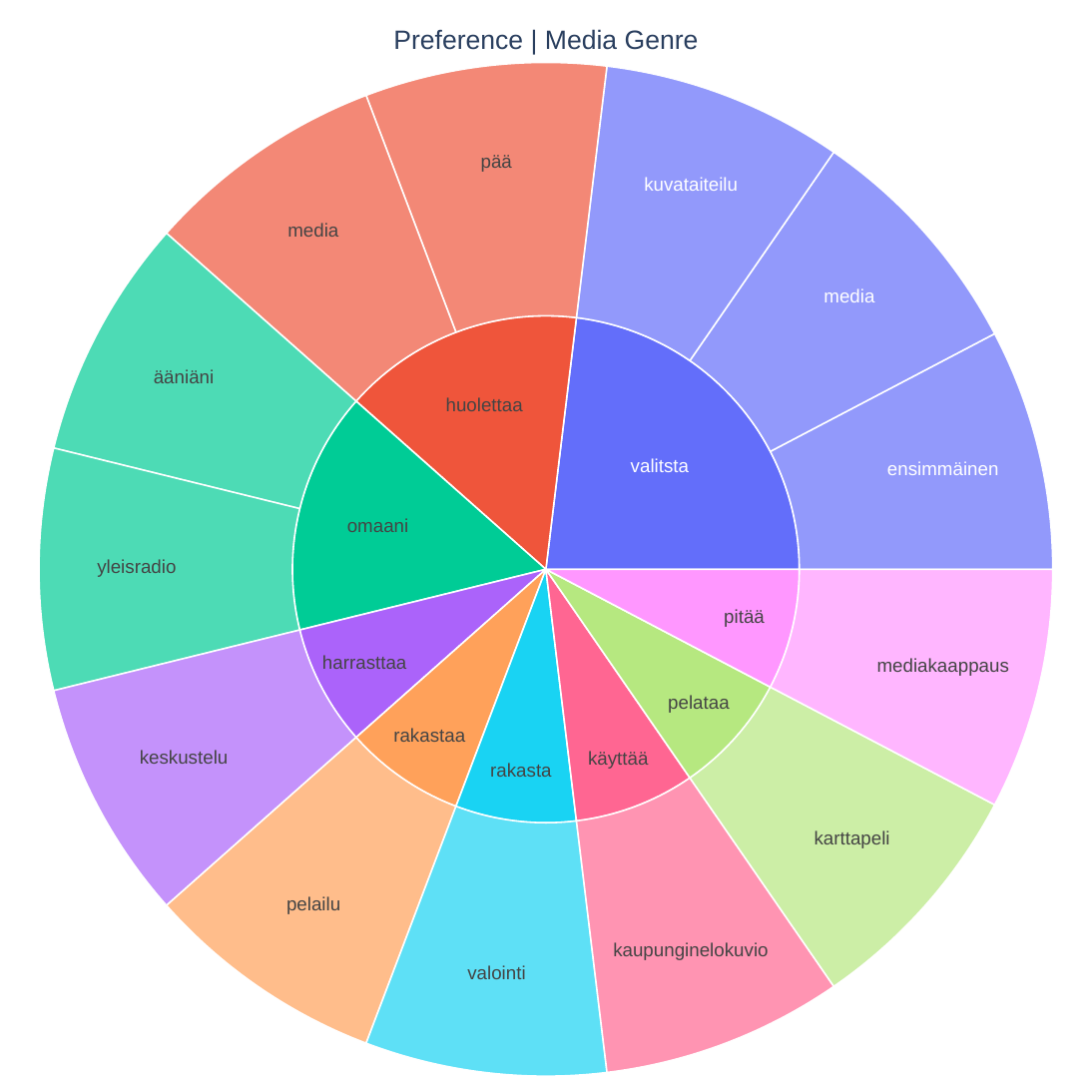}
        \end{subfigure}
    \end{minipage}
    
\caption{Detailed BERTSCORE for Finnish Personas in different generation configurations and Sunburst charts of personas taxonomy entities with most root verbs and associated object noun for the different models}    
\end{figure}


\restoregeometry

\newgeometry{top=0.5cm, bottom=1.5cm, left=2.5cm, right=2.5cm}
\subsubsection{\textsc{Croatian}}

\begin{figure}[h]
    \centering
    \begin{minipage}{0.45\textwidth}
        \begin{subfigure}{\textwidth}
            \centering
            \includegraphics[height=0.17\textheight]{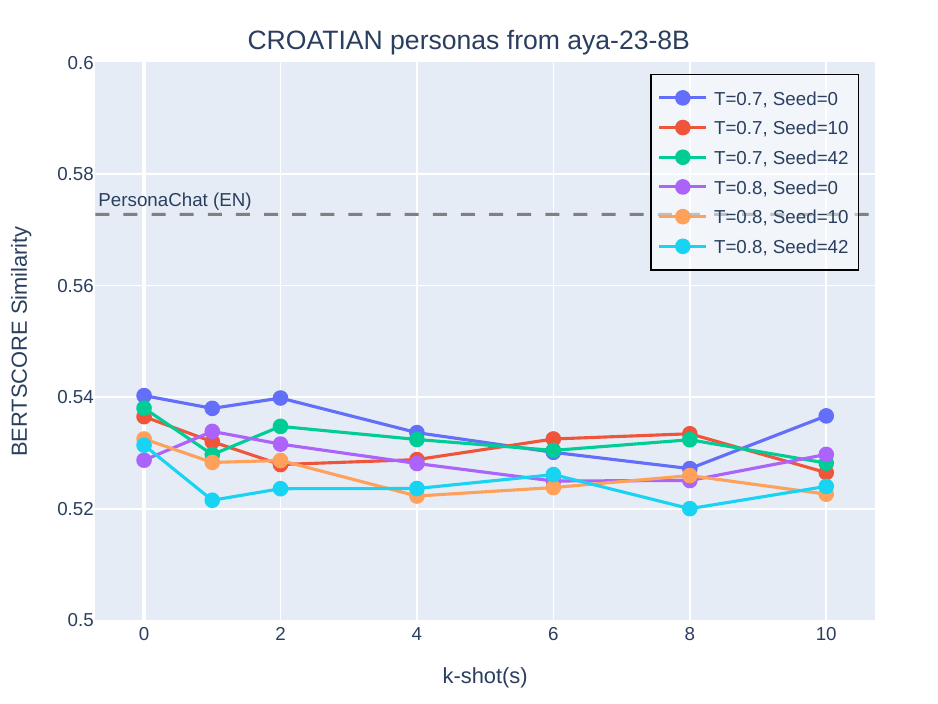}
        \end{subfigure}
        \vskip\baselineskip
        \begin{subfigure}{\textwidth}
            \centering
            \includegraphics[height=0.17\textheight]{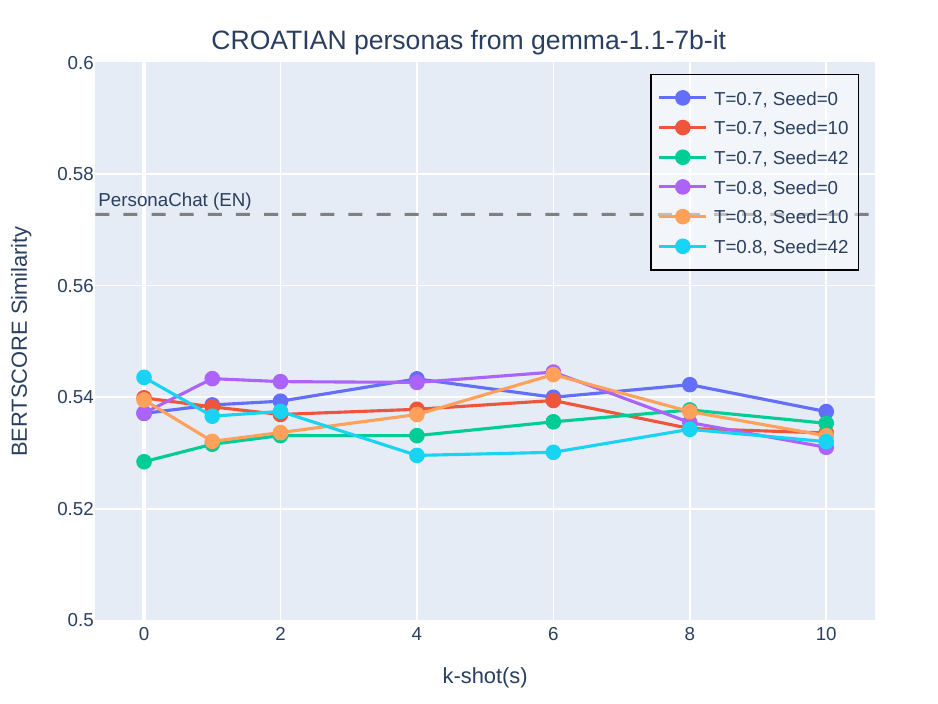}
        \end{subfigure}
        \vskip\baselineskip
        \begin{subfigure}{\textwidth}
            \centering
            \includegraphics[height=0.17\textheight]{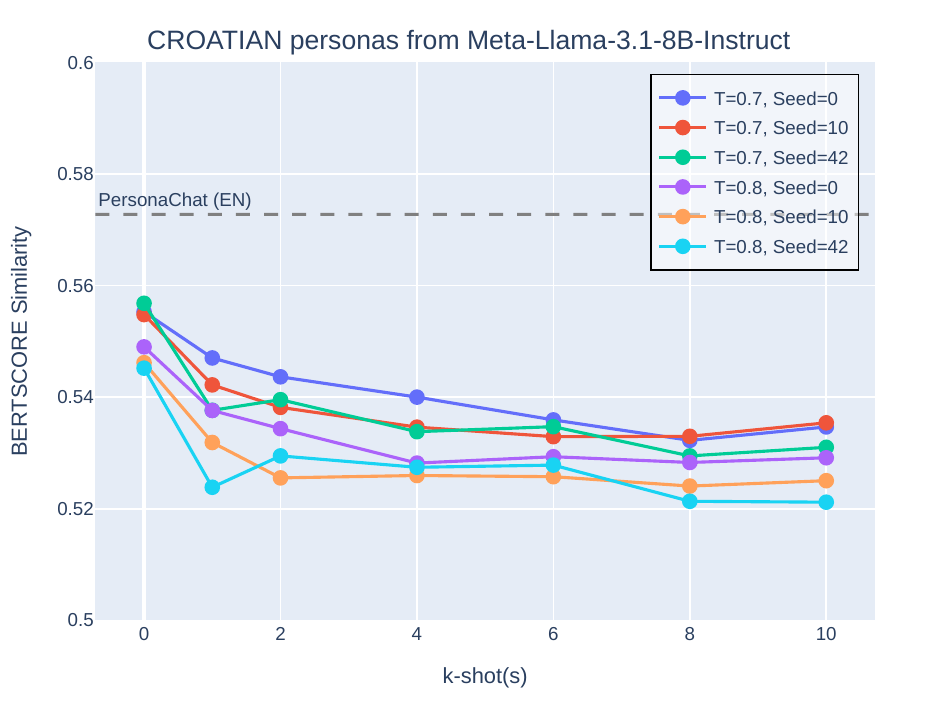}
        \end{subfigure}
        \vskip\baselineskip
        \begin{subfigure}{\textwidth}
            \centering
            \includegraphics[height=0.17\textheight]{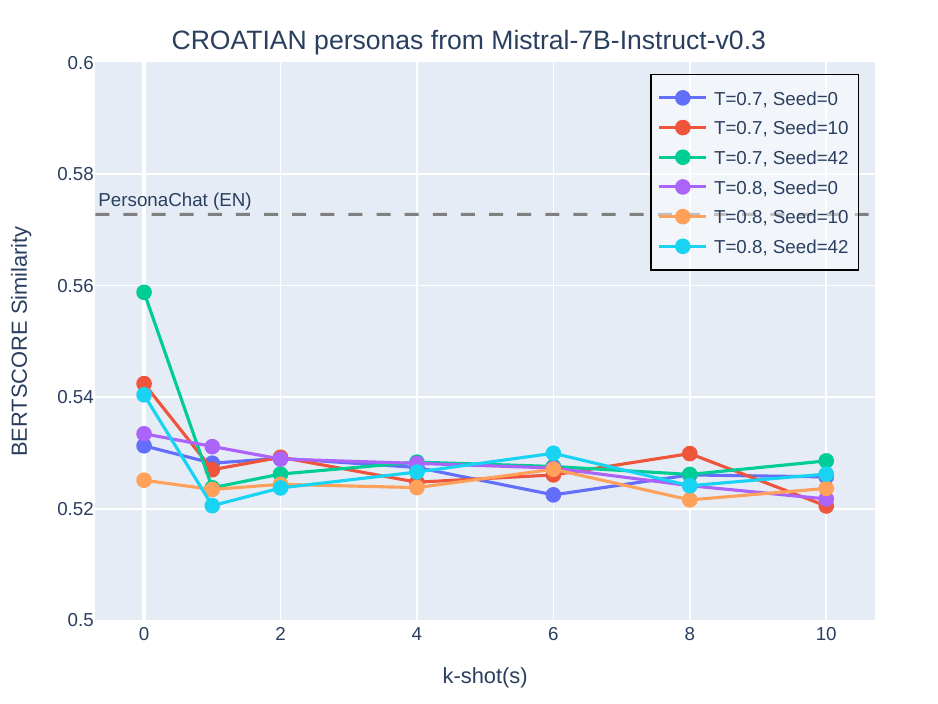}
        \end{subfigure}
    \end{minipage}
    %
    %
    \begin{minipage}{0.45\textwidth}
        \begin{subfigure}{\textwidth}
            \centering
            \includegraphics[height=0.17\textheight]{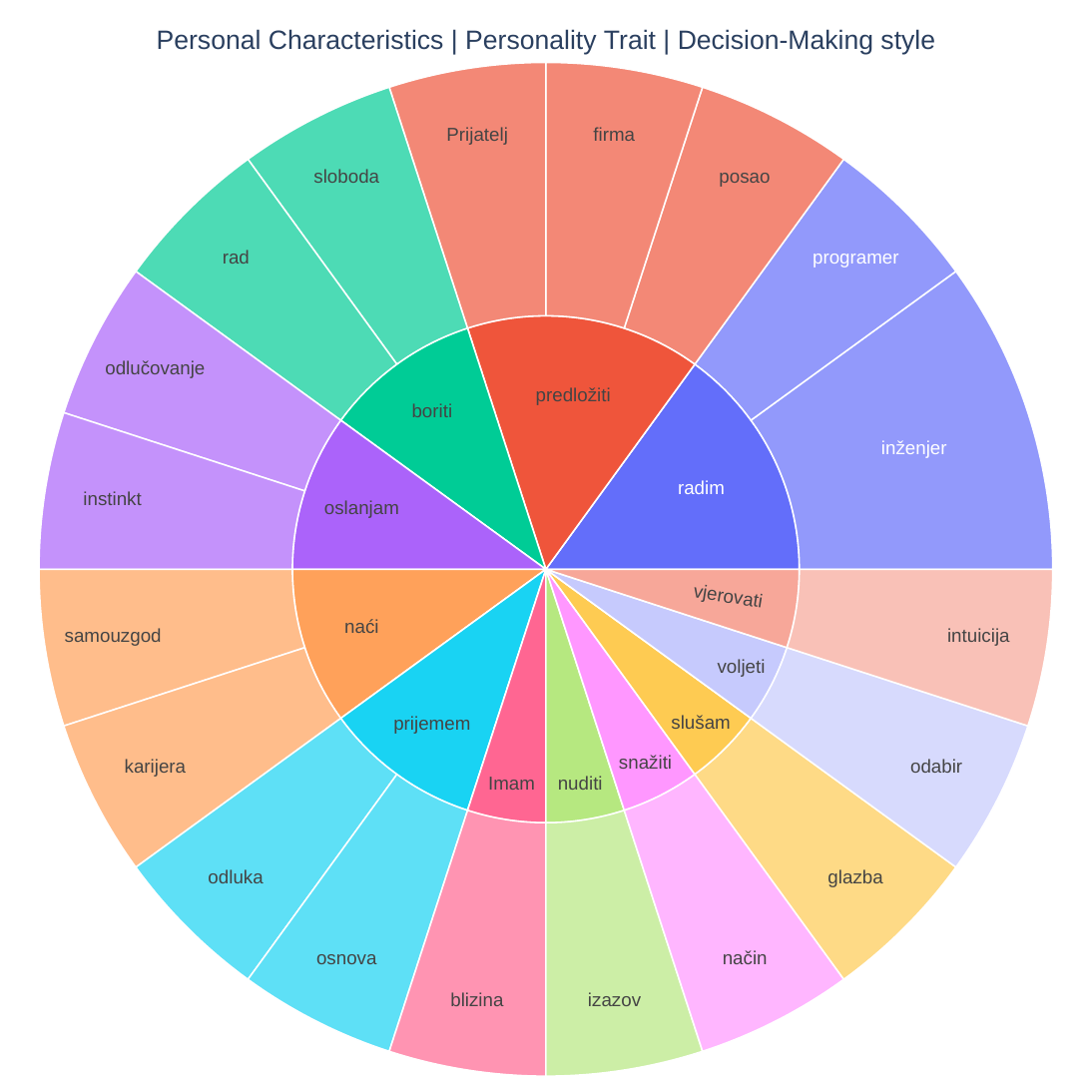}
        \end{subfigure}
        \vskip\baselineskip
        \begin{subfigure}{\textwidth}
            \centering
            \includegraphics[height=0.17\textheight]{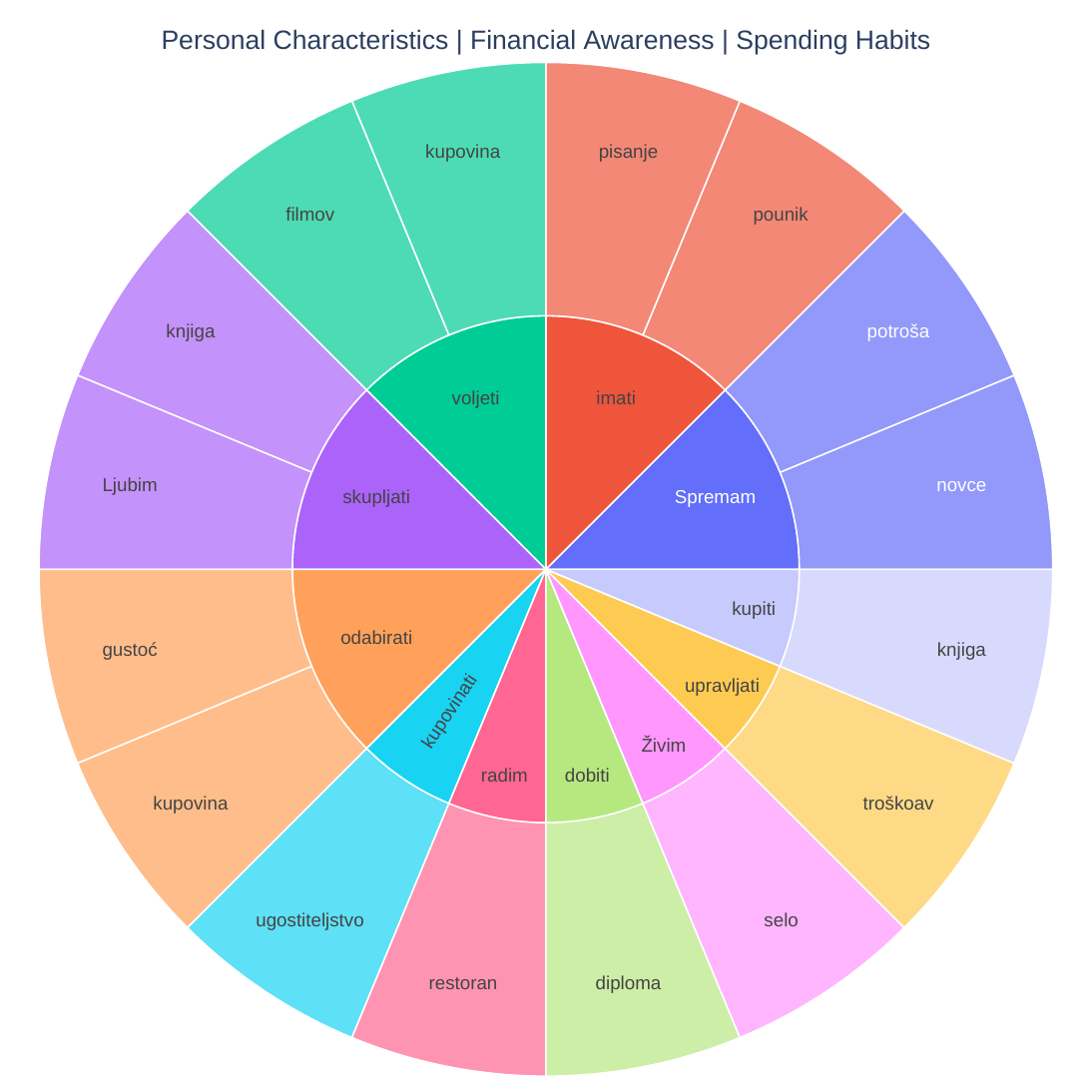}
        \end{subfigure}
        \vskip\baselineskip
        \begin{subfigure}{\textwidth}
            \centering
            \includegraphics[height=0.17\textheight]{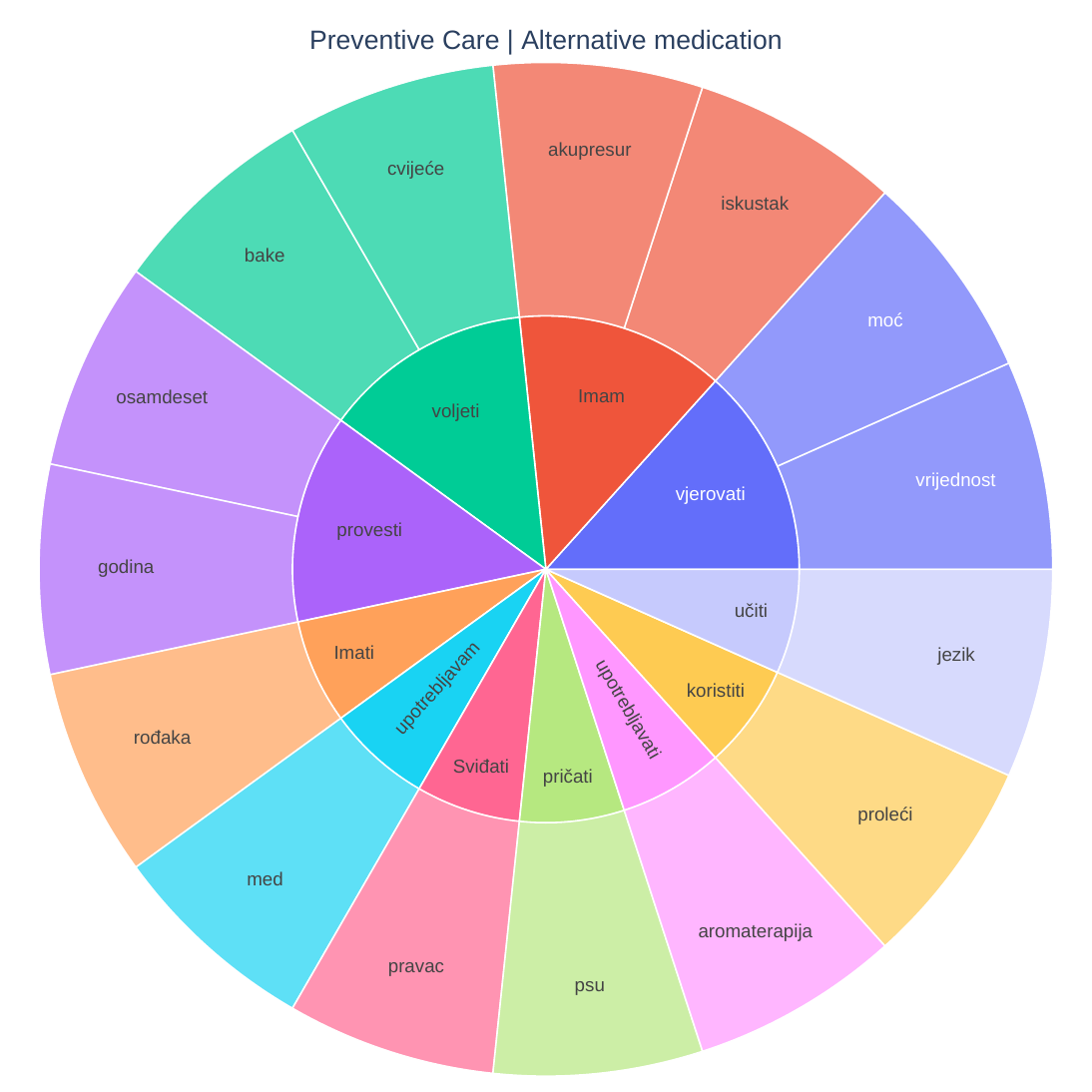}
        \end{subfigure}
        \vskip\baselineskip
        \begin{subfigure}{\textwidth}
            \centering
            \includegraphics[height=0.17\textheight]{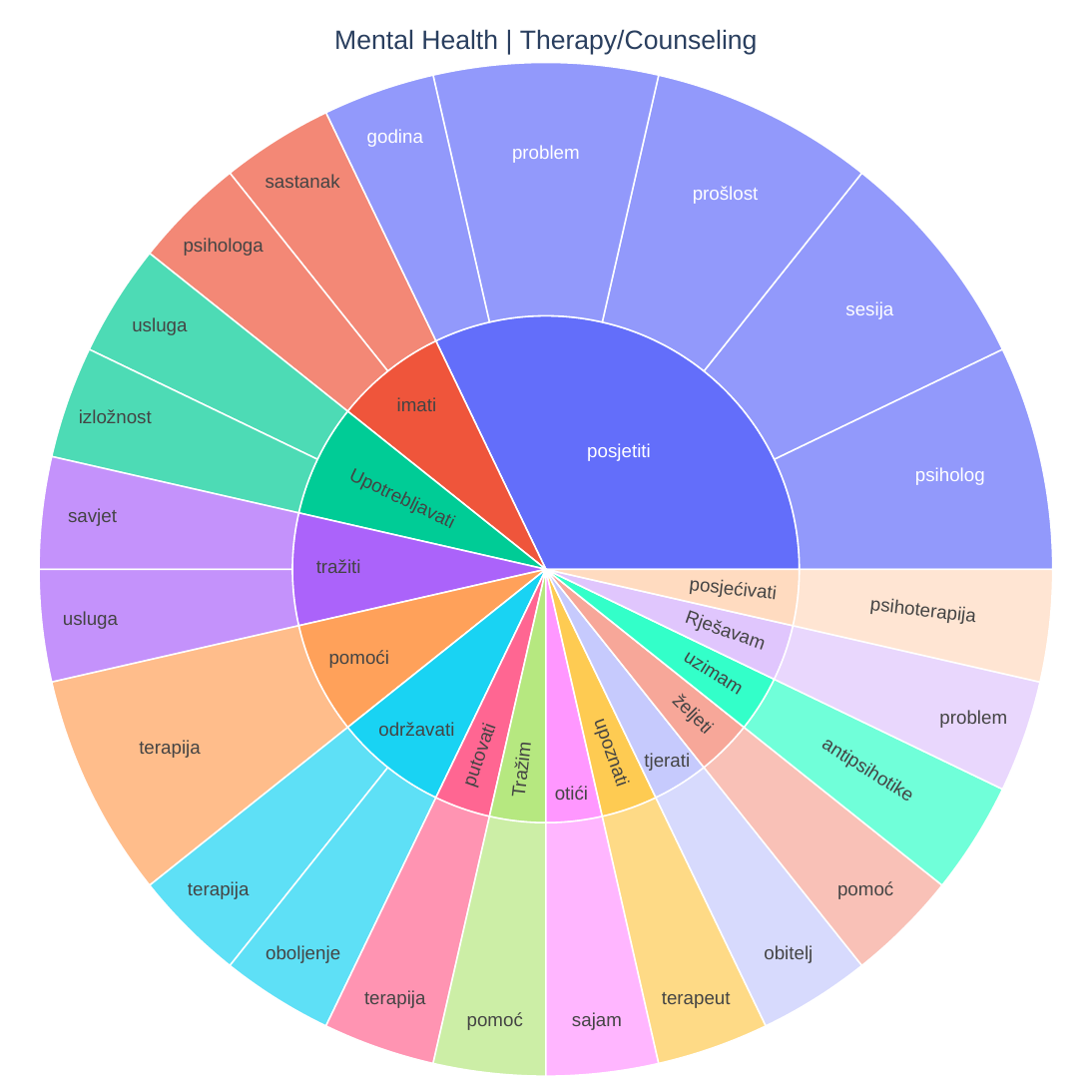}
        \end{subfigure}
    \end{minipage}
    
\caption{Detailed BERTSCORE for Croatian Personas in different generation configurations and Sunburst charts of personas taxonomy entities with most root verbs and associated object noun for the different models}    
\end{figure}


\restoregeometry
\newgeometry{top=0.5cm, bottom=1.5cm, left=2.5cm, right=2.5cm}
 \subsubsection{\textsc{Thai$^*$}}

\begin{figure}[h]
    \centering
    \begin{minipage}{0.45\textwidth}
        \begin{subfigure}{\textwidth}
            \centering
            \includegraphics[height=0.15\textheight]{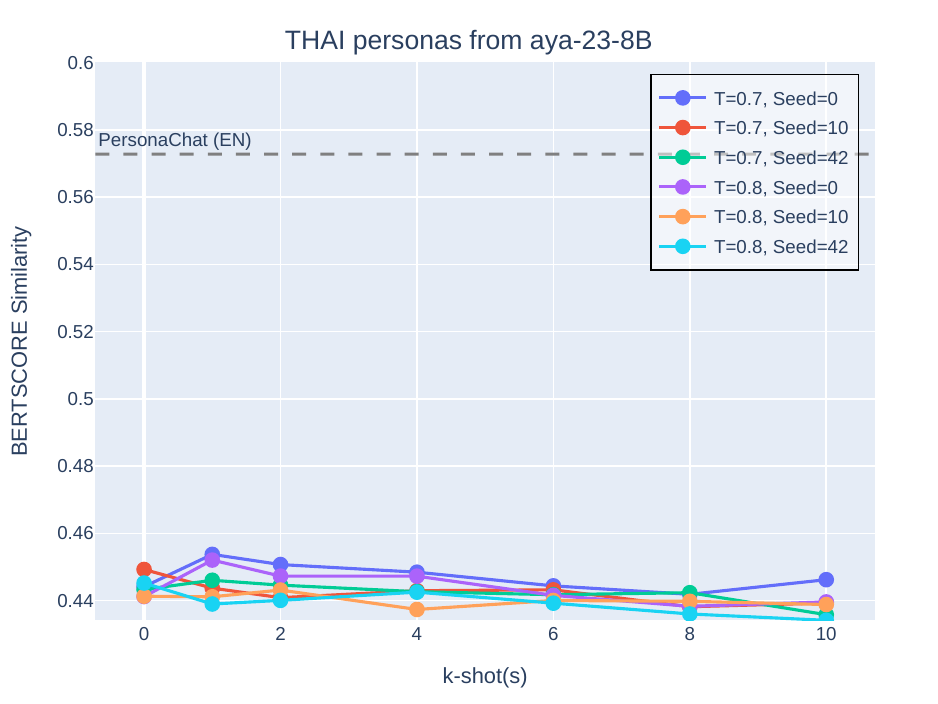}
        \end{subfigure}
        \vskip\baselineskip
        \begin{subfigure}{\textwidth}
            \centering
            \includegraphics[height=0.15\textheight]{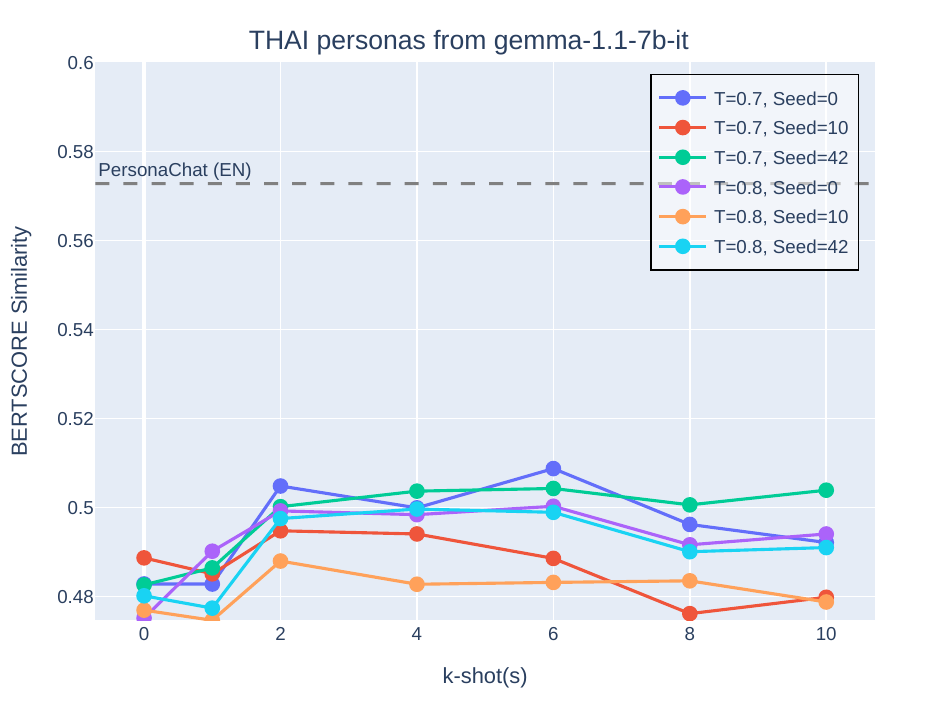}
        \end{subfigure}
    \end{minipage}
    %
    %
    \begin{minipage}{0.45\textwidth}
            \begin{subfigure}{\textwidth}
            \centering
            \includegraphics[height=0.15\textheight]{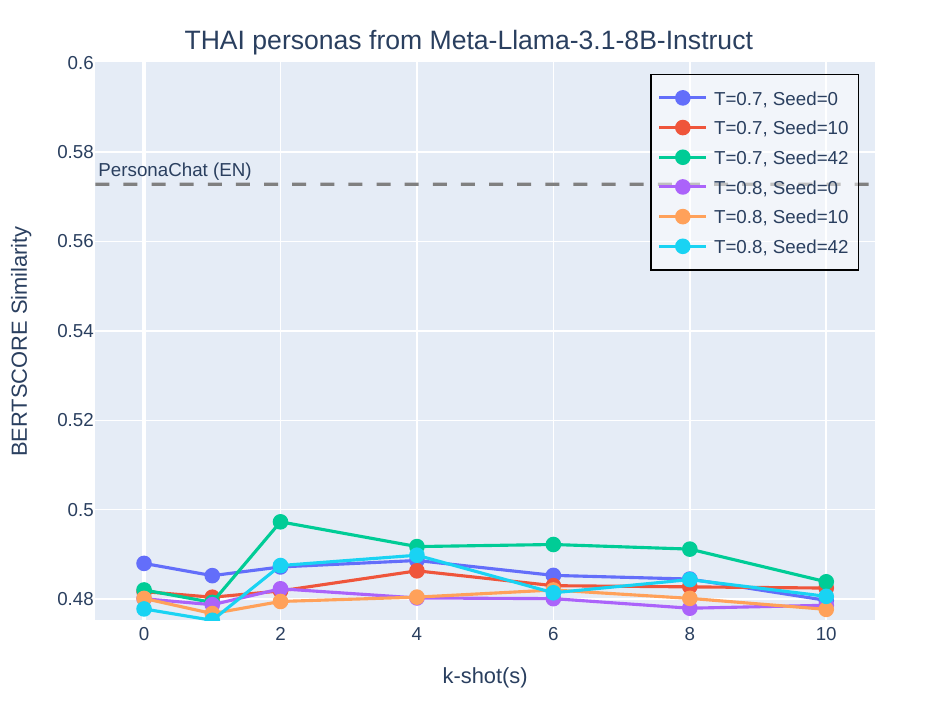}
        \end{subfigure}
        \vskip\baselineskip
        \begin{subfigure}{\textwidth}
            \centering
            \includegraphics[height=0.15\textheight]{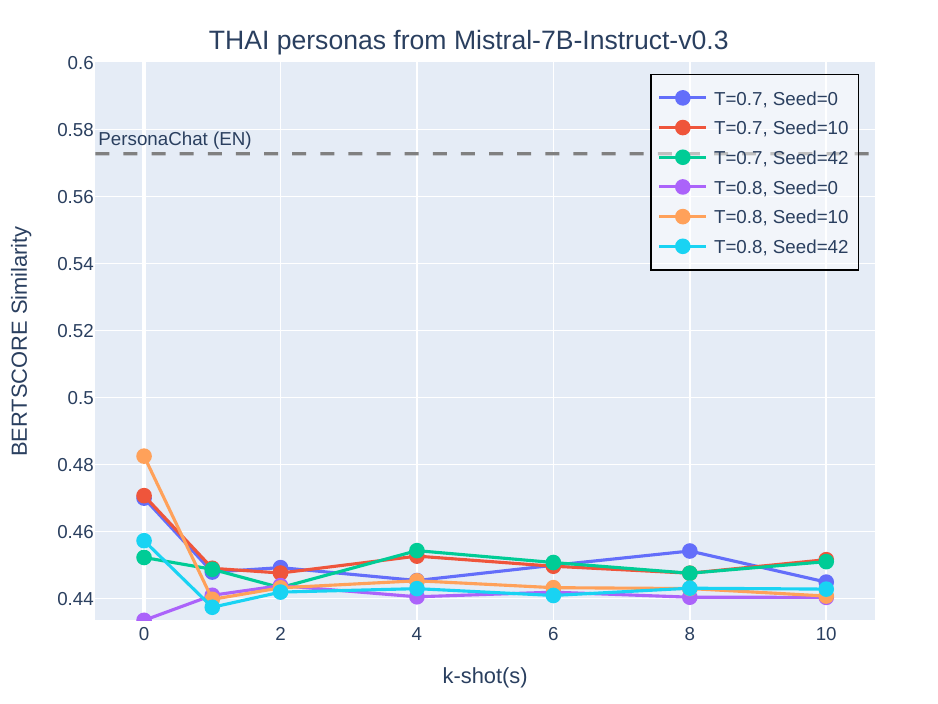}
        \end{subfigure}
    \end{minipage}
    
\caption{Detailed BERTSCORE for Thai Personas in different generation configurations for the different models}    
\end{figure}



\subsubsection{\textsc{Hindi}}

\begin{figure}[h]
    \centering
    \begin{minipage}{0.45\textwidth}
        \begin{subfigure}{\textwidth}
            \centering
            \includegraphics[height=0.15\textheight]{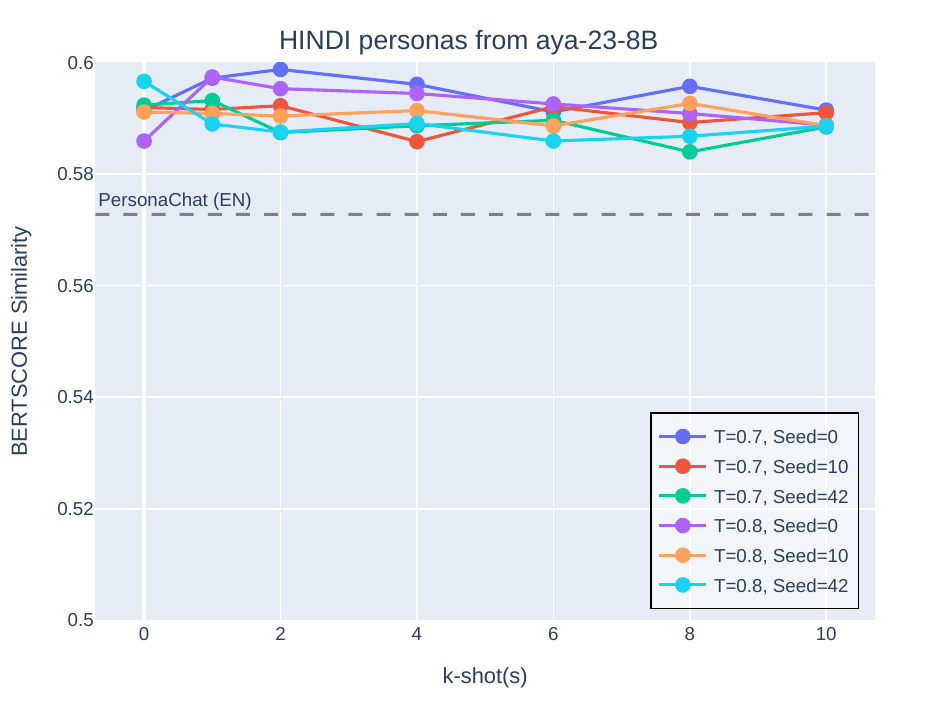}
        \end{subfigure}
        \vskip\baselineskip
        \begin{subfigure}{\textwidth}
            \centering
            \includegraphics[height=0.15\textheight]{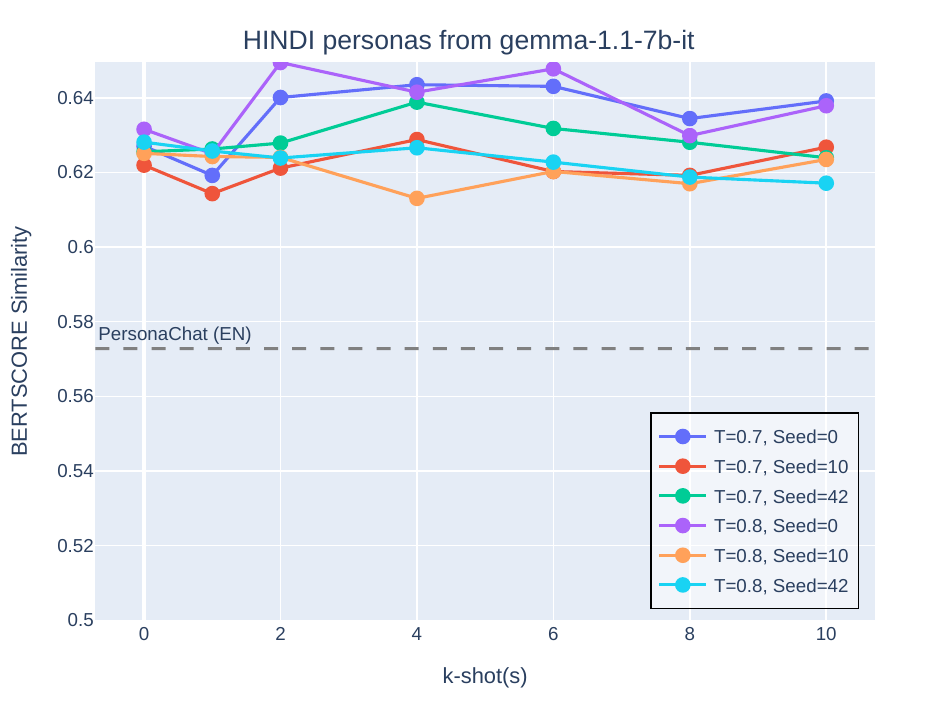}
        \end{subfigure}
    \end{minipage}
    %
    %
    \begin{minipage}{0.45\textwidth}
            \begin{subfigure}{\textwidth}
            \centering
            \includegraphics[height=0.15\textheight]{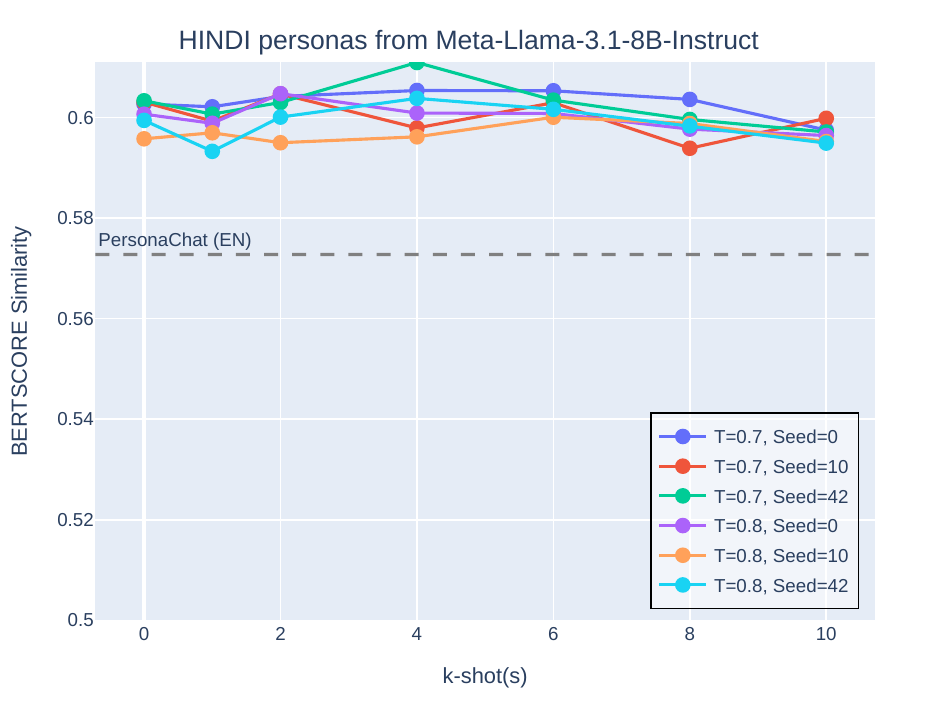}
        \end{subfigure}
        \vskip\baselineskip
        \begin{subfigure}{\textwidth}
            \centering
            \includegraphics[height=0.15\textheight]{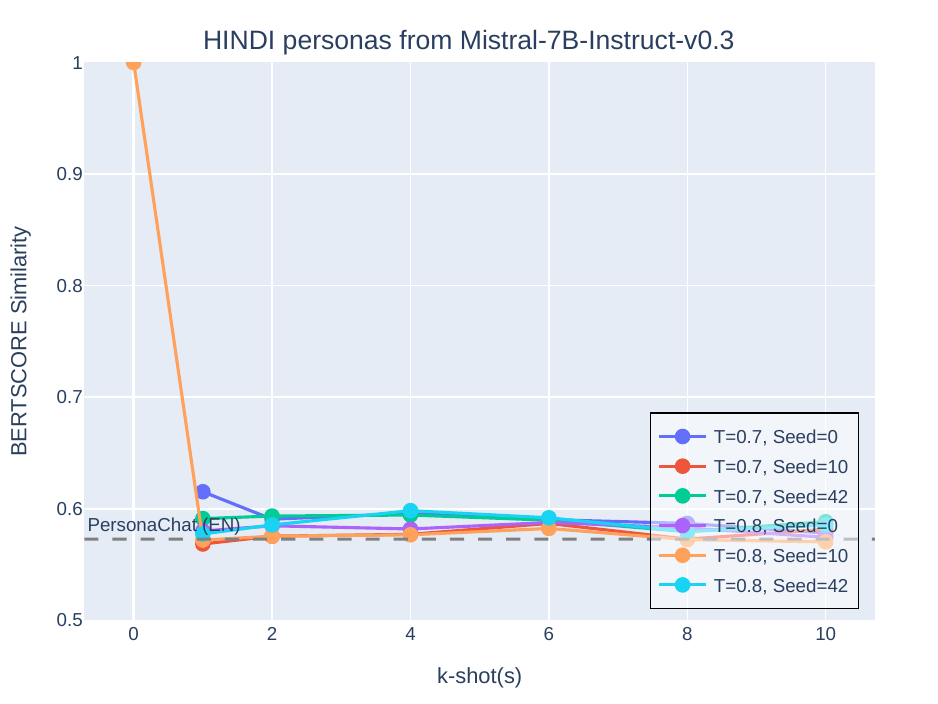}
        \end{subfigure}
    \end{minipage}
    
\caption{Detailed BERTSCORE for Hindi Personas in different generation configurations for the different models}    
\end{figure}


\newgeometry{top=0.5cm, bottom=1.5cm, left=2.5cm, right=2.5cm}
 \subsubsection{\textsc{Bengali}}

\begin{figure}[h]
    \centering
    \begin{minipage}{0.45\textwidth}
        \begin{subfigure}{\textwidth}
            \centering
            \includegraphics[height=0.15\textheight]{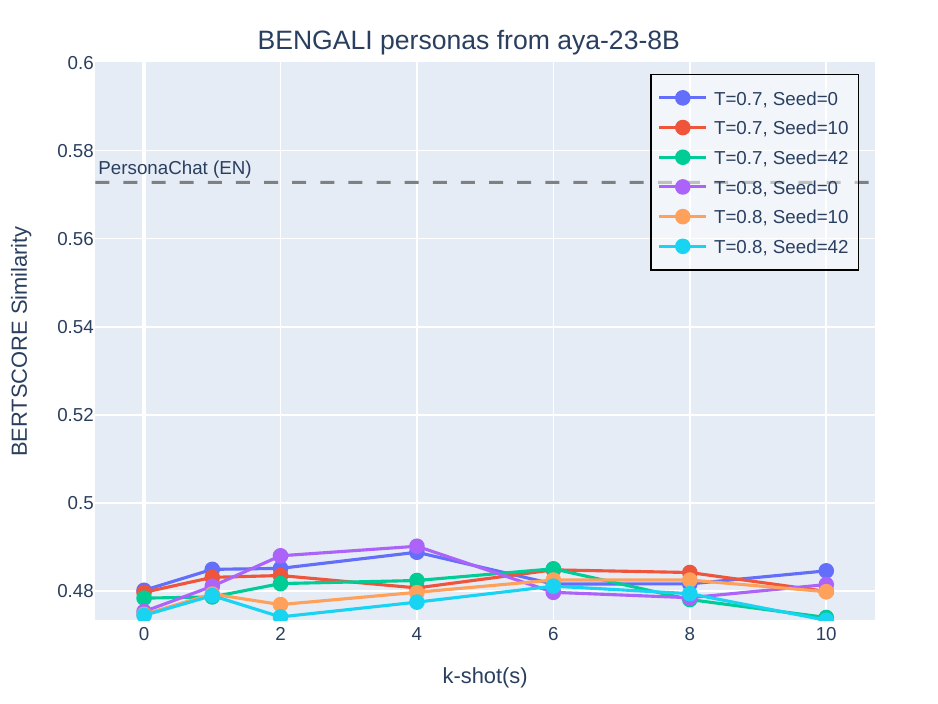}
        \end{subfigure}
        \vskip\baselineskip
        \begin{subfigure}{\textwidth}
            \centering
            \includegraphics[height=0.15\textheight]{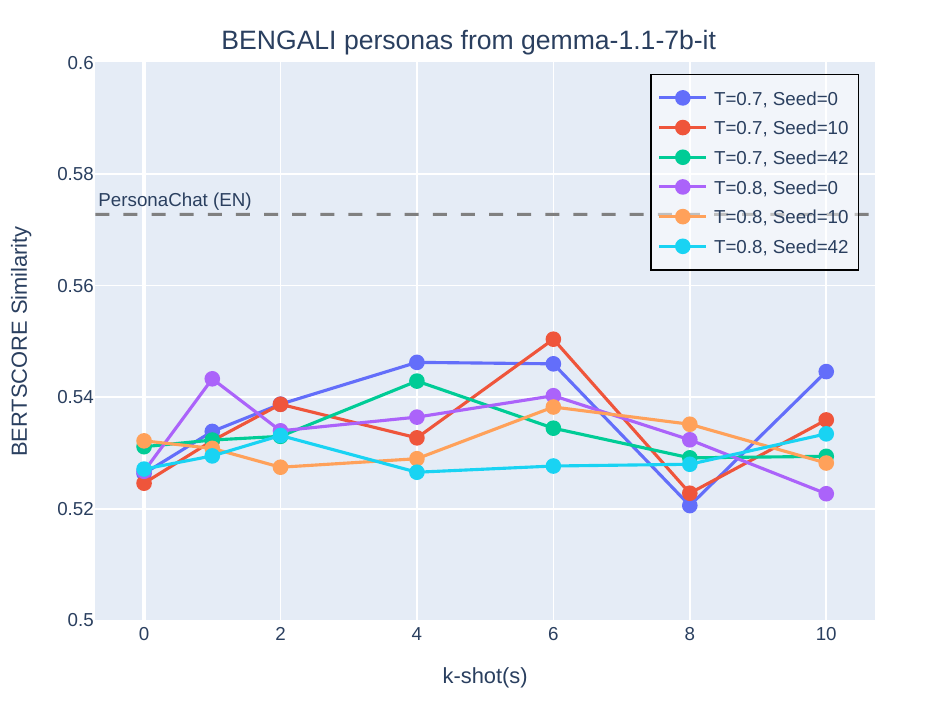}
        \end{subfigure}
    \end{minipage}
    %
    %
    \begin{minipage}{0.45\textwidth}
            \begin{subfigure}{\textwidth}
            \centering
            \includegraphics[height=0.15\textheight]{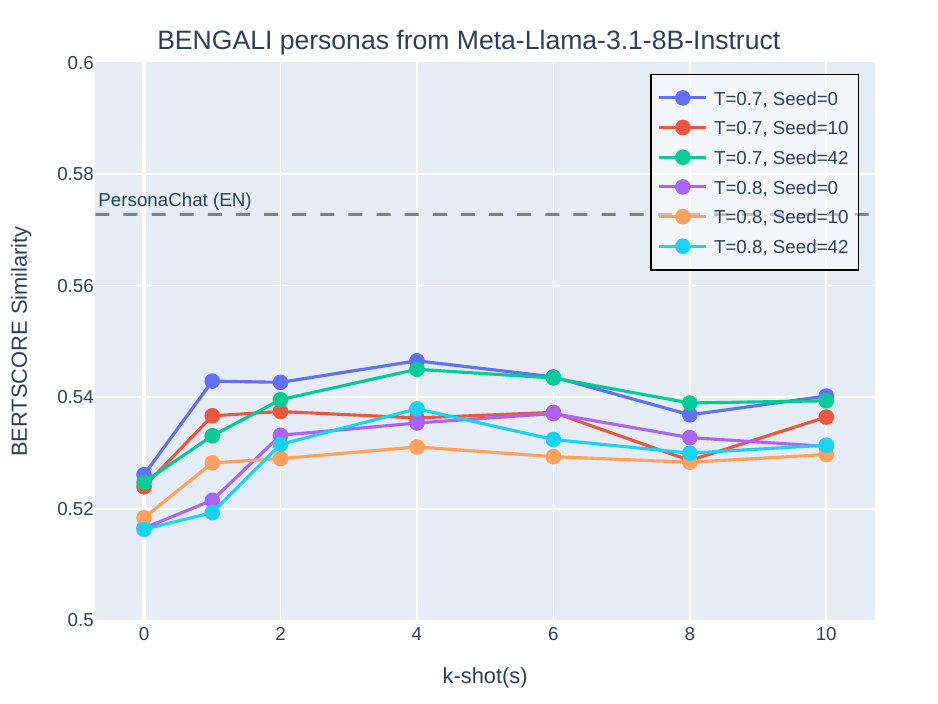}
        \end{subfigure}
        \vskip\baselineskip
        \begin{subfigure}{\textwidth}
            \centering
            \includegraphics[height=0.15\textheight]{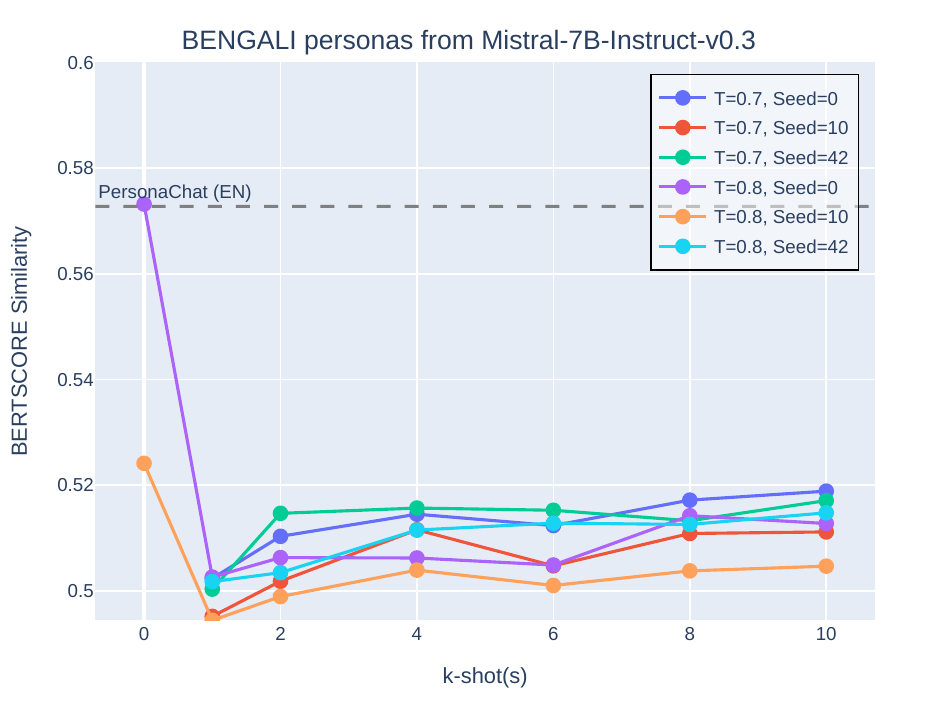}
        \end{subfigure}
    \end{minipage}
    
\caption{Detailed BERTSCORE for Bengali Personas in different generation configurations for the different models}    
\end{figure}



\subsection{Low-Resource Languages}

\subsubsection{\textsc{Afrikaans}}

\begin{figure}[h]
    \centering
    \begin{minipage}{0.45\textwidth}
        \begin{subfigure}{\textwidth}
            \centering
            \includegraphics[height=0.15\textheight]{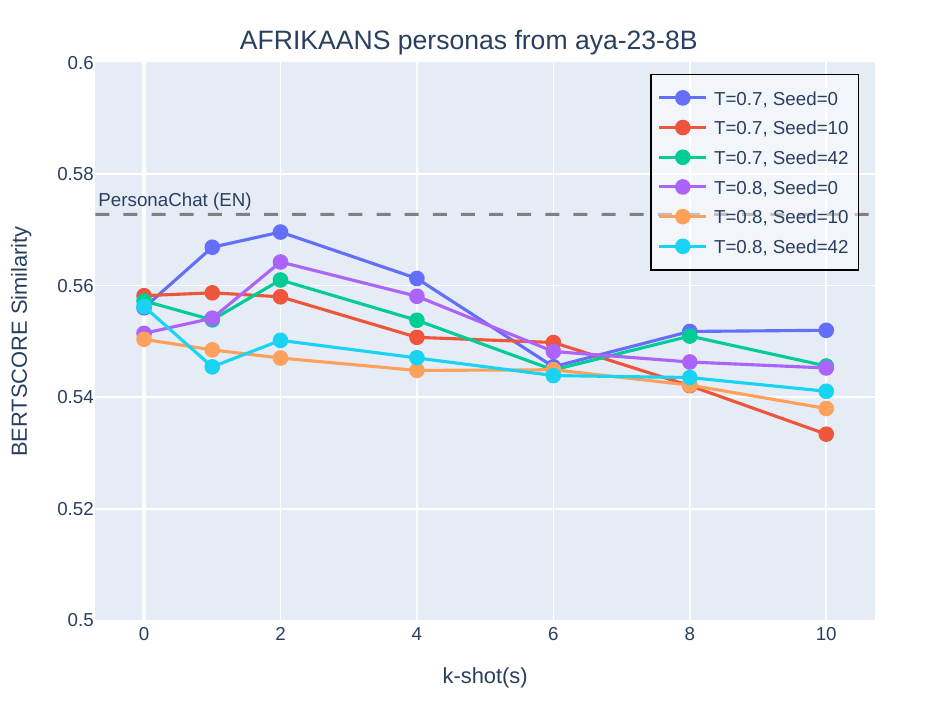}
        \end{subfigure}
        \vskip\baselineskip
        \begin{subfigure}{\textwidth}
            \centering
            \includegraphics[height=0.15\textheight]{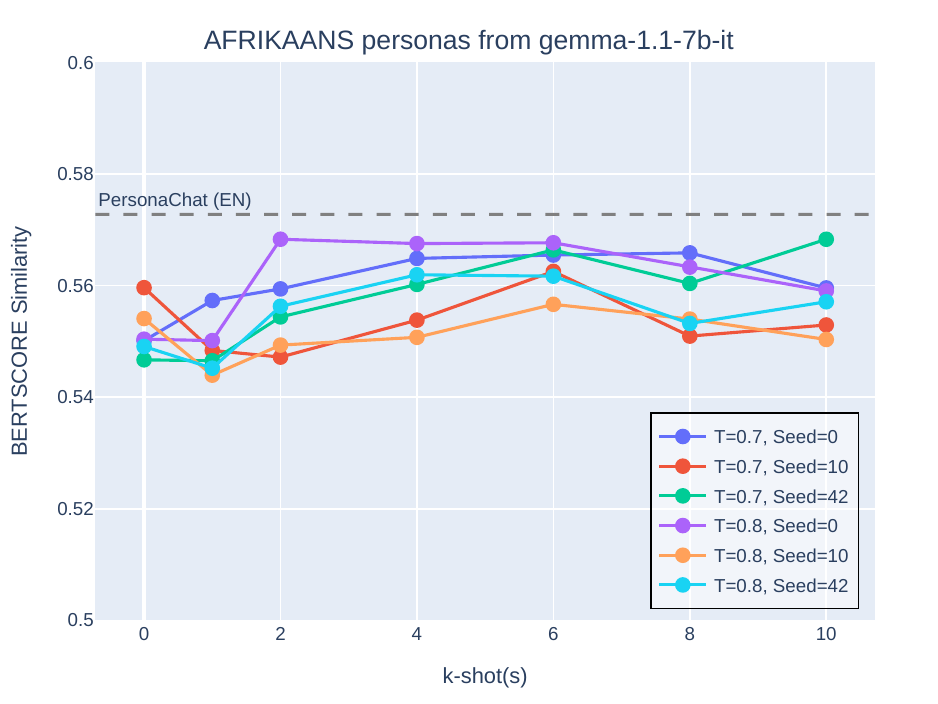}
        \end{subfigure}
    \end{minipage}
    %
    %
    \begin{minipage}{0.45\textwidth}
            \begin{subfigure}{\textwidth}
            \centering
            \includegraphics[height=0.15\textheight]{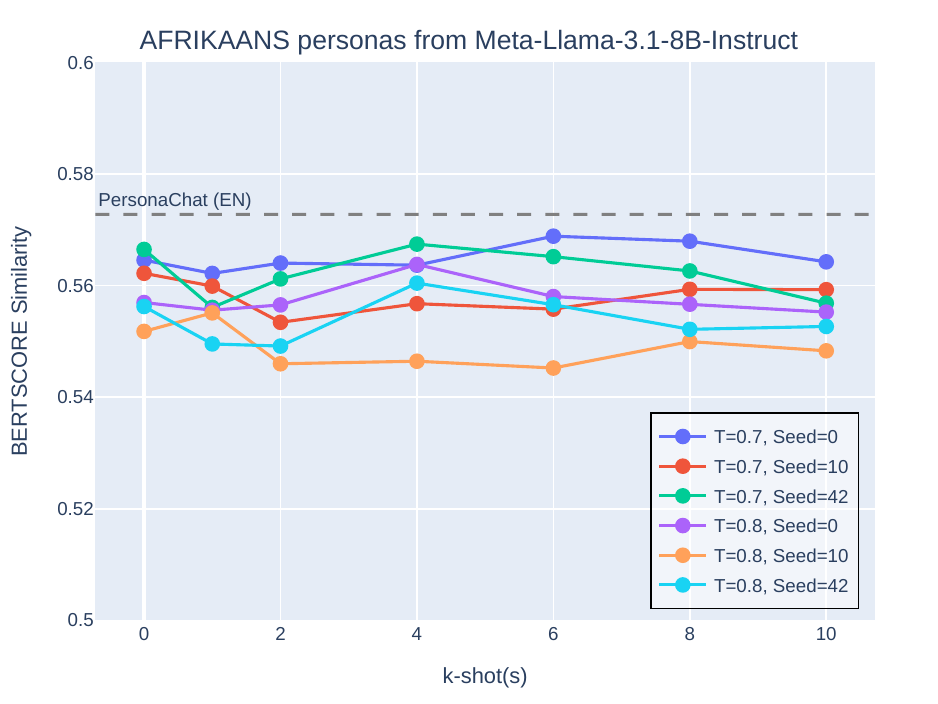}
        \end{subfigure}
        \vskip\baselineskip
        \begin{subfigure}{\textwidth}
            \centering
            \includegraphics[height=0.15\textheight]{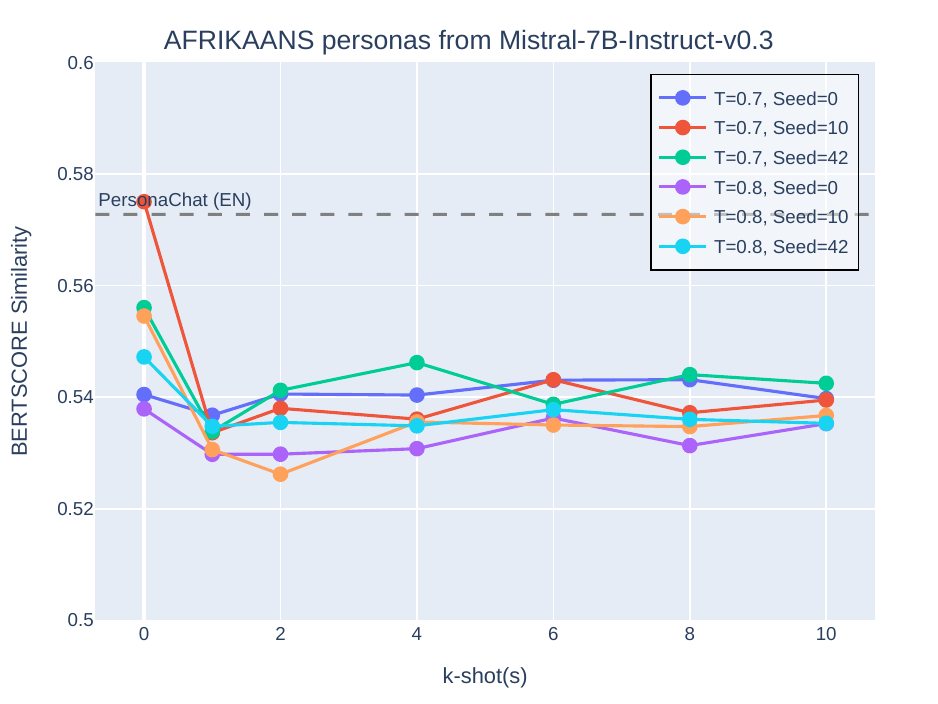}
        \end{subfigure}
    \end{minipage}
    
\caption{Detailed BERTSCORE for Afrikaans Personas in different generation configurations for the different models}    
\end{figure}



\newgeometry{top=0.5cm, bottom=1.5cm, left=2.5cm, right=2.5cm}
\subsubsection{\textsc{Swahili}}

\begin{figure}[h]
    \centering
    \begin{minipage}{0.45\textwidth}
        \begin{subfigure}{\textwidth}
            \centering
            \includegraphics[height=0.15\textheight]{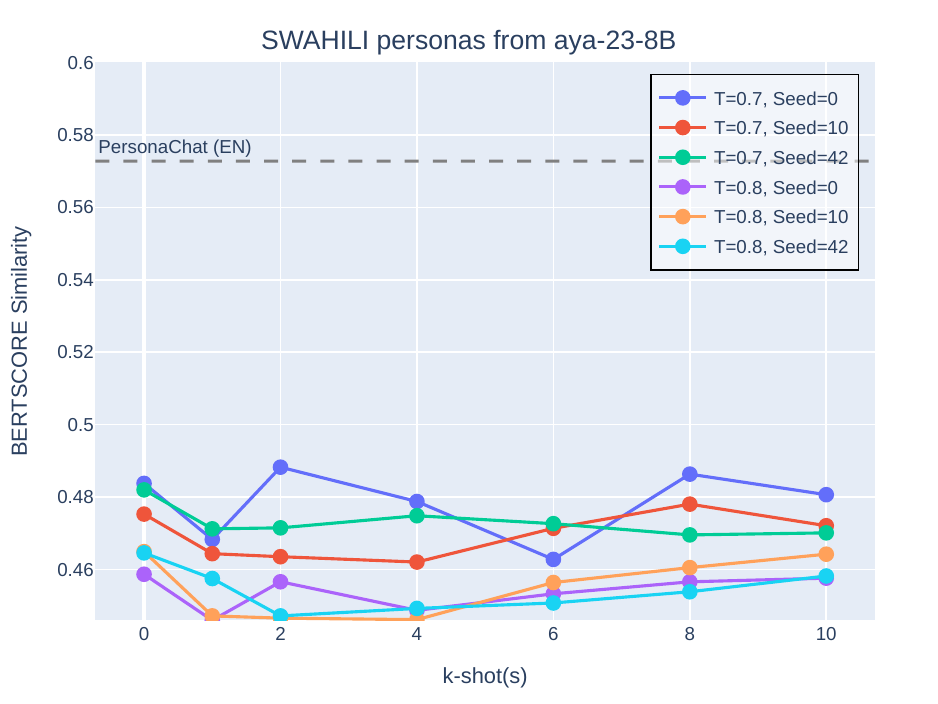}
        \end{subfigure}
        \vskip\baselineskip
        \begin{subfigure}{\textwidth}
            \centering
            \includegraphics[height=0.15\textheight]{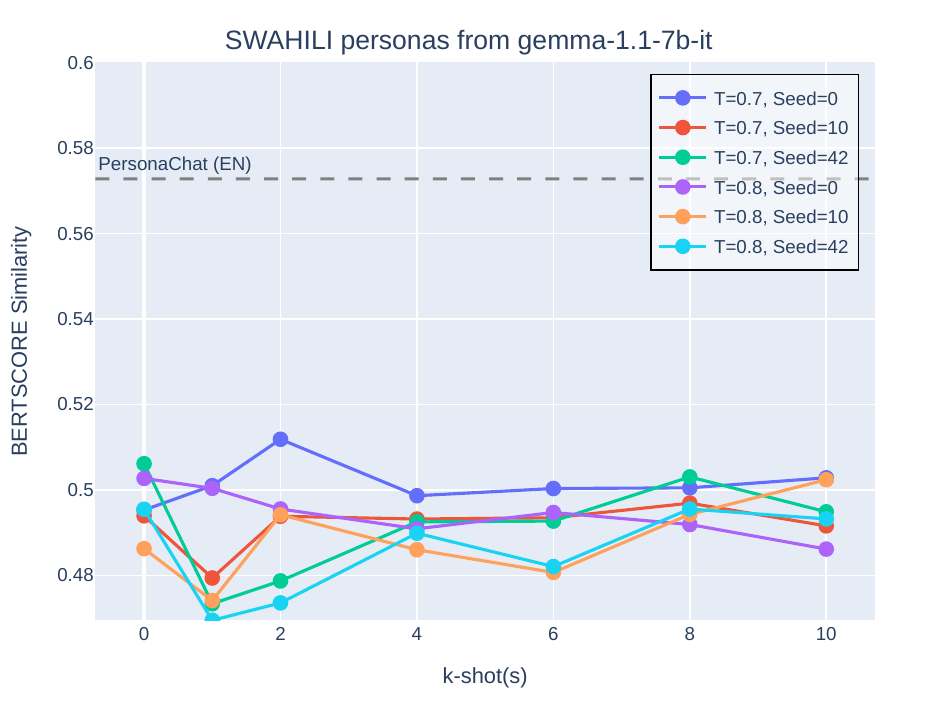}
        \end{subfigure}
    \end{minipage}
    %
    %
    \begin{minipage}{0.45\textwidth}
            \begin{subfigure}{\textwidth}
            \centering
            \includegraphics[height=0.15\textheight]{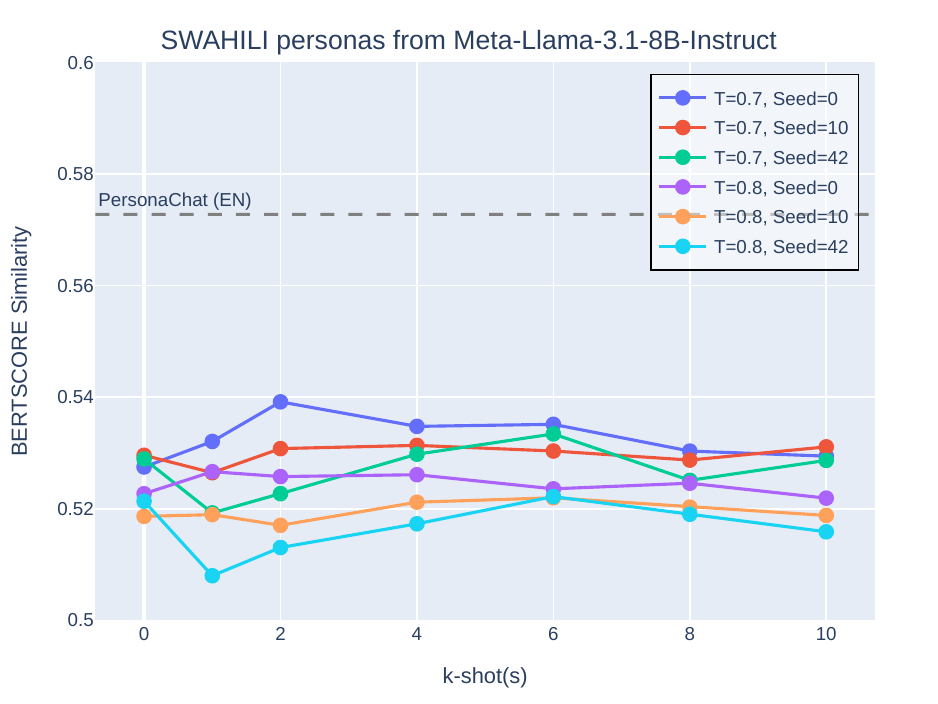}
        \end{subfigure}
        \vskip\baselineskip
        \begin{subfigure}{\textwidth}
            \centering
            \includegraphics[height=0.15\textheight]{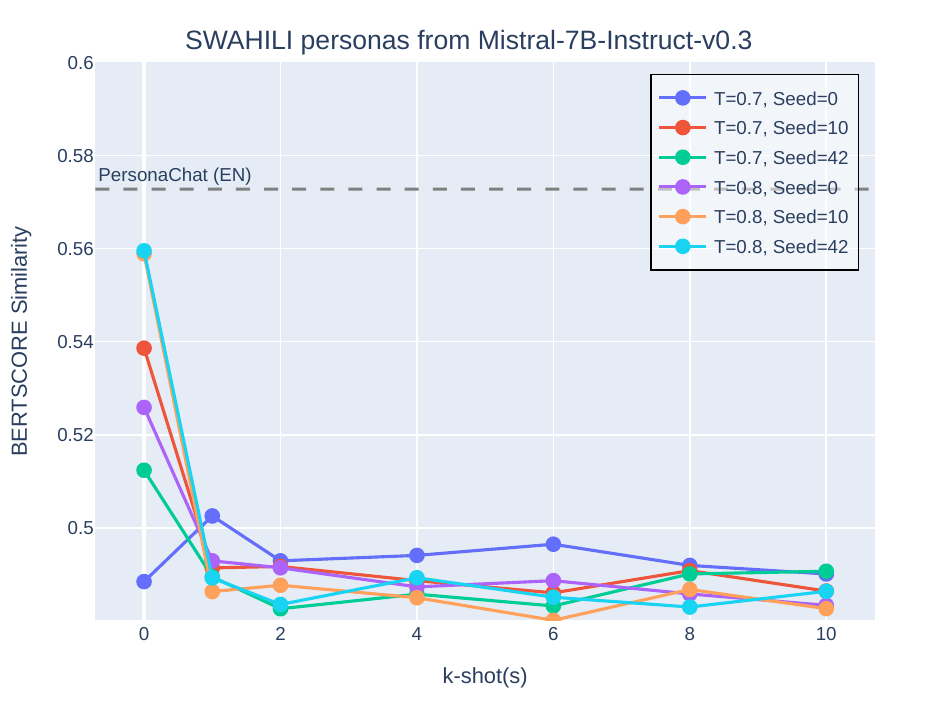}
        \end{subfigure}
    \end{minipage}
    
\caption{Detailed BERTSCORE for Swahili Personas in different generation configurations for the different models}    
\end{figure}



\subsubsection{\textsc{Yoruba}}

\begin{figure}[h!]
    \centering
    \begin{minipage}{0.45\textwidth}
        \begin{subfigure}{\textwidth}
            \centering
            \includegraphics[height=0.15\textheight]{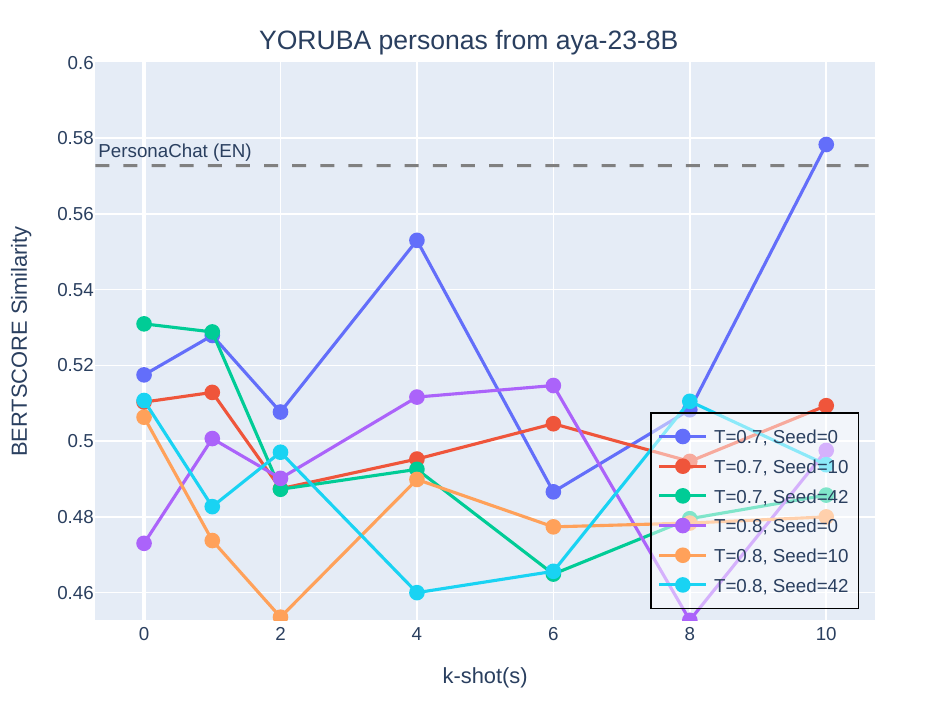}
        \end{subfigure}
        \vskip\baselineskip
        \begin{subfigure}{\textwidth}
            \centering
            \includegraphics[height=0.15\textheight]{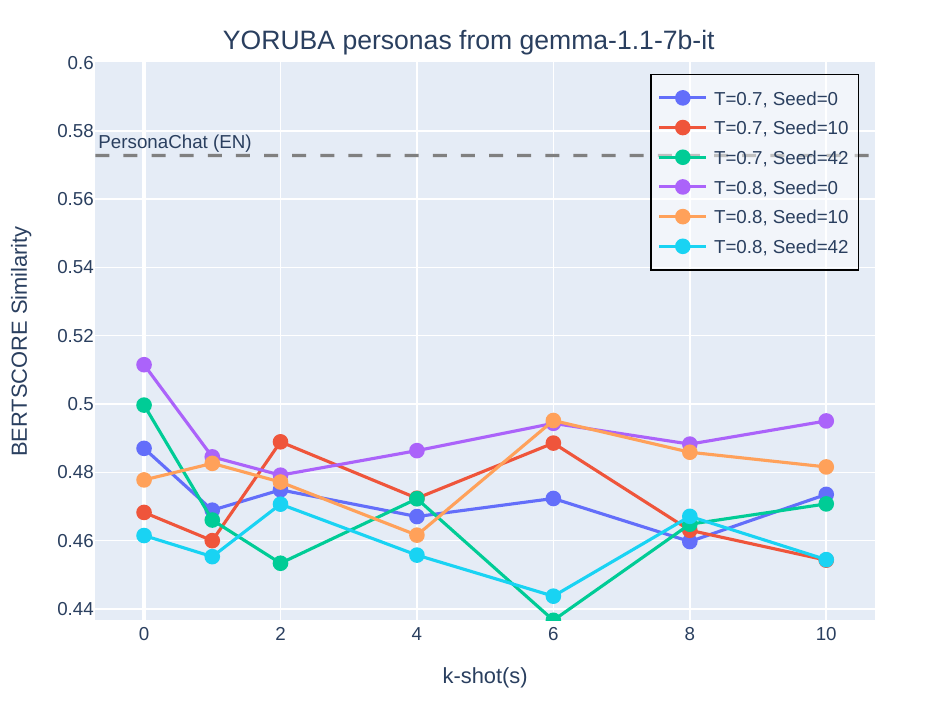}
        \end{subfigure}
    \end{minipage}
    %
    %
    \begin{minipage}{0.45\textwidth}
            \begin{subfigure}{\textwidth}
            \centering
            \includegraphics[height=0.15\textheight]{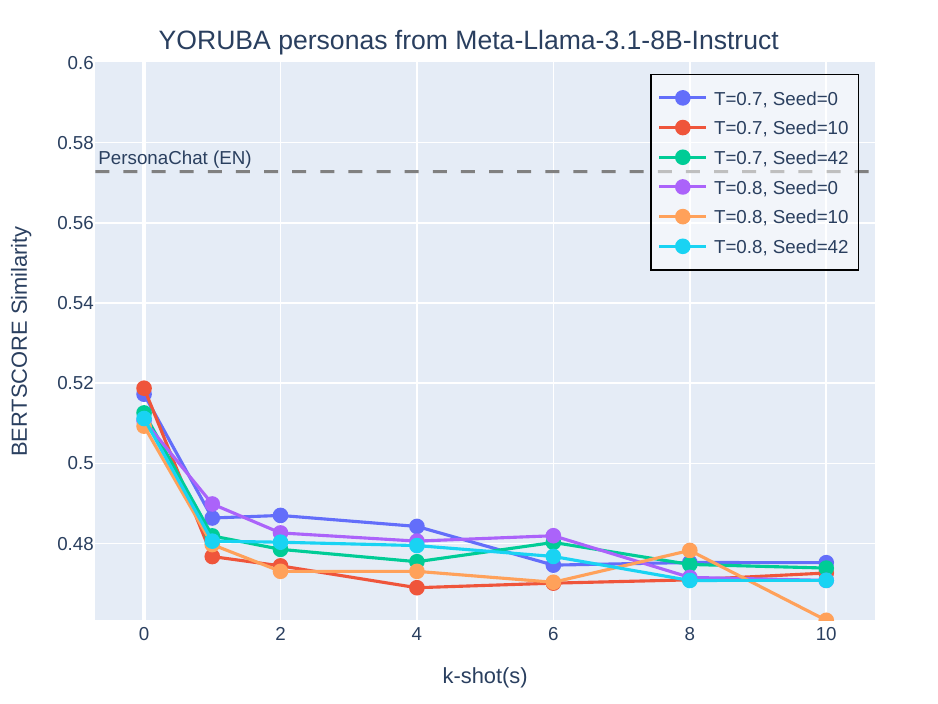}
        \end{subfigure}
        \vskip\baselineskip
        \begin{subfigure}{\textwidth}
            \centering
            \includegraphics[height=0.15\textheight]{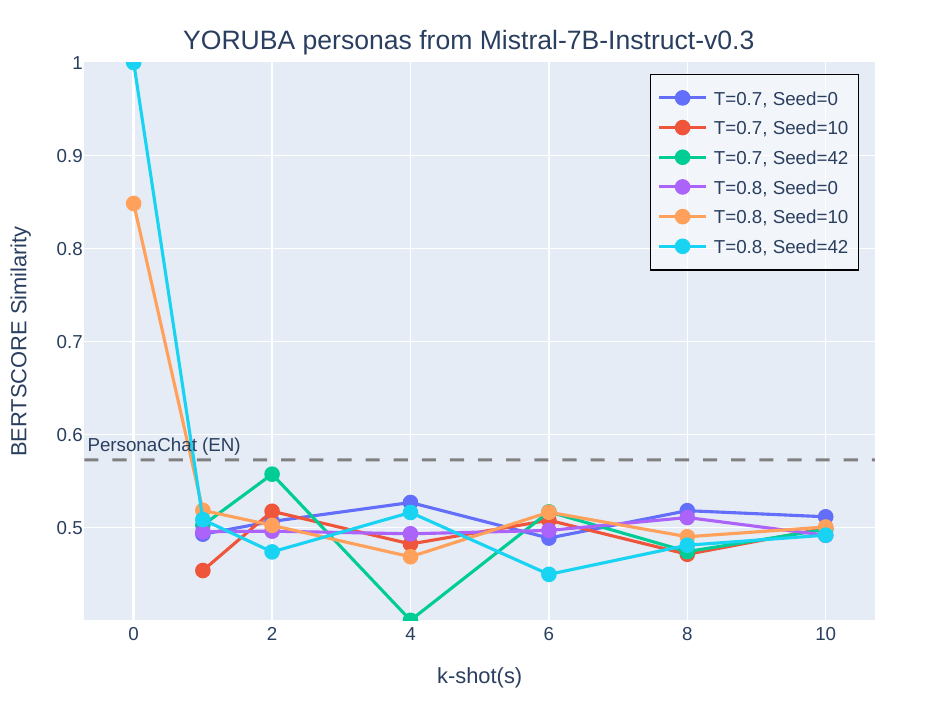}
        \end{subfigure}
    \end{minipage}
    
\caption{Detailed BERTSCORE for Yoruba Personas in different generation configurations for the different models}    
\end{figure}


\restoregeometry


\end{document}